\documentclass[lettersize,journal]{IEEEtran}
\usepackage{amsmath,amsfonts}
\usepackage{algorithm}
\usepackage{algorithmic}
\usepackage{array}
\usepackage[caption=false,font=footnotesize,labelformat=empty,labelfont=sf,textfont=sf]{subfig}
\usepackage{textcomp}
\usepackage{stfloats}
\usepackage{url}
\usepackage{verbatim}
\usepackage{graphicx}
\usepackage{stmaryrd}
\usepackage{cite}
\usepackage{makecell}
\usepackage{booktabs}
\usepackage[table]{xcolor}
\usepackage{multirow}
\usepackage{hyperref}
\begin{document}

\title{3DTTNet: Multimodal Fusion-Based 3D Traversable Terrain Modeling for Off-Road Environments}

\author{Zitong Chen, Chao Sun, Shida Nie, Chen Min, Changjiu Ning, Haoyu Li, and Bo Wang
\thanks{This work was supported by Key-Area Research and Development Program of Guangdong Province (2023B0909040001), Shenzhen Science and Technology Program (No. KJZD20231023100304010), National Key Research and Development Program (2022YFB2503203). (Corresponding author: Chao Sun.)}
\thanks{Zitong Chen, Chao Sun, Changjiu Ning, Haoyu Li, and Bo Wang are with the Shenzhen Automotive Research Institute, Beijing Institute of Technology, Shenzhen, 518118, China, and also with the School of Mechanical Engineering, Beijing Institute of Technology, Beijing, 100081, China (e-mail: zitongchen2000@163.com; chaosun@bit.edu.cn; ning18500681112@163.com; rabbit.yujixyz@gmail.com; 3120205224@bit.edu.cn).}
\thanks{Shida Nie is with the School of Mechanical Engineering, Beijing Institute of Technology, Beijing, 100081, China, and also with the Advanced Technology Research Institute (Jinan), Beijing Institute of Technology, Jinan, 250307, China (email:nieshida@bit.edu.cn).}
\thanks{Chen Min is with the Research Center for Intelligent Computing Systems, Institute of Computing Technology, Chinese Academy of Sciences, Beijing, 100190, China (e-mail: mincheng@ict.ac.cn).}}

% The paper headers
\markboth{Journal of \LaTeX\ Class Files,~Vol.~14, No.~8, August~2021}
{Shell \MakeLowercase{\textit{et al.}}: A Sample Article Using IEEEtran.cls for IEEE Journals}

\maketitle

\begin{abstract}
Off-road environments remain significant challenges for autonomous ground vehicles, due to the lack of structured roads and the presence of complex obstacles, such as uneven terrain, vegetation, and occlusions. Traditional perception algorithms, primarily designed for structured environments, often fail in unstructured scenarios. In this paper, traversable area recognition is achieved through semantic scene completion. A novel multimodal method, 3DTTNet, is proposed to generate dense traversable terrain estimations by integrating LiDAR point clouds with monocular images from a forward-facing perspective. By integrating multimodal data, environmental feature extraction is strengthened, which is crucial for accurate terrain modeling in complex terrains. Furthermore, RELLIS-OCC, a dataset with 3D traversable annotations, is introduced, incorporating geometric features such as step height, slope, and unevenness. Through a comprehensive analysis of vehicle obstacle-crossing conditions and the incorporation of vehicle body structure constraints, four traversability cost labels are generated: lethal, medium-cost, low-cost, and free. Experimental results demonstrate that 3DTTNet outperforms the comparison approaches in 3D traversable area recognition, particularly in off-road environments with irregular geometries and partial occlusions. Specifically, 3DTTNet achieves a 42\% improvement in scene completion IoU compared to other models. The proposed framework is scalable and adaptable to various vehicle platforms, allowing for adjustments to occupancy grid parameters and the integration of advanced dynamic models for traversability cost estimation.
\end{abstract}


\section{Introduction}
\begin{figure}[t]
    \centering
    \subfloat[{\scriptsize (a)}]{%
        \includegraphics[width=0.48\columnwidth]{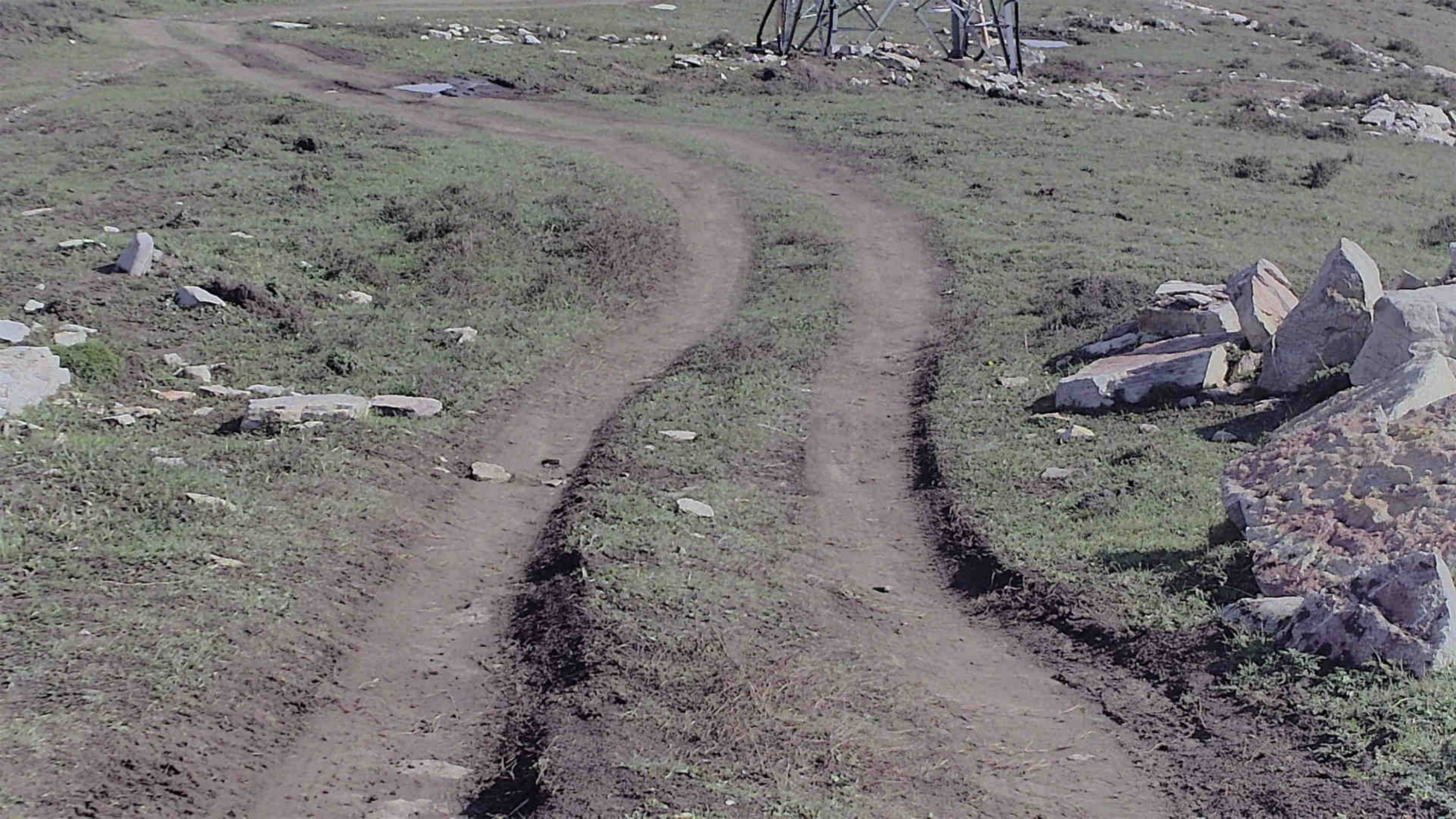}%
        \label{fig:irregular_geometries}
    }
    \hfill
    \subfloat[{\scriptsize (b)}]{%
        \includegraphics[width=0.48\columnwidth]{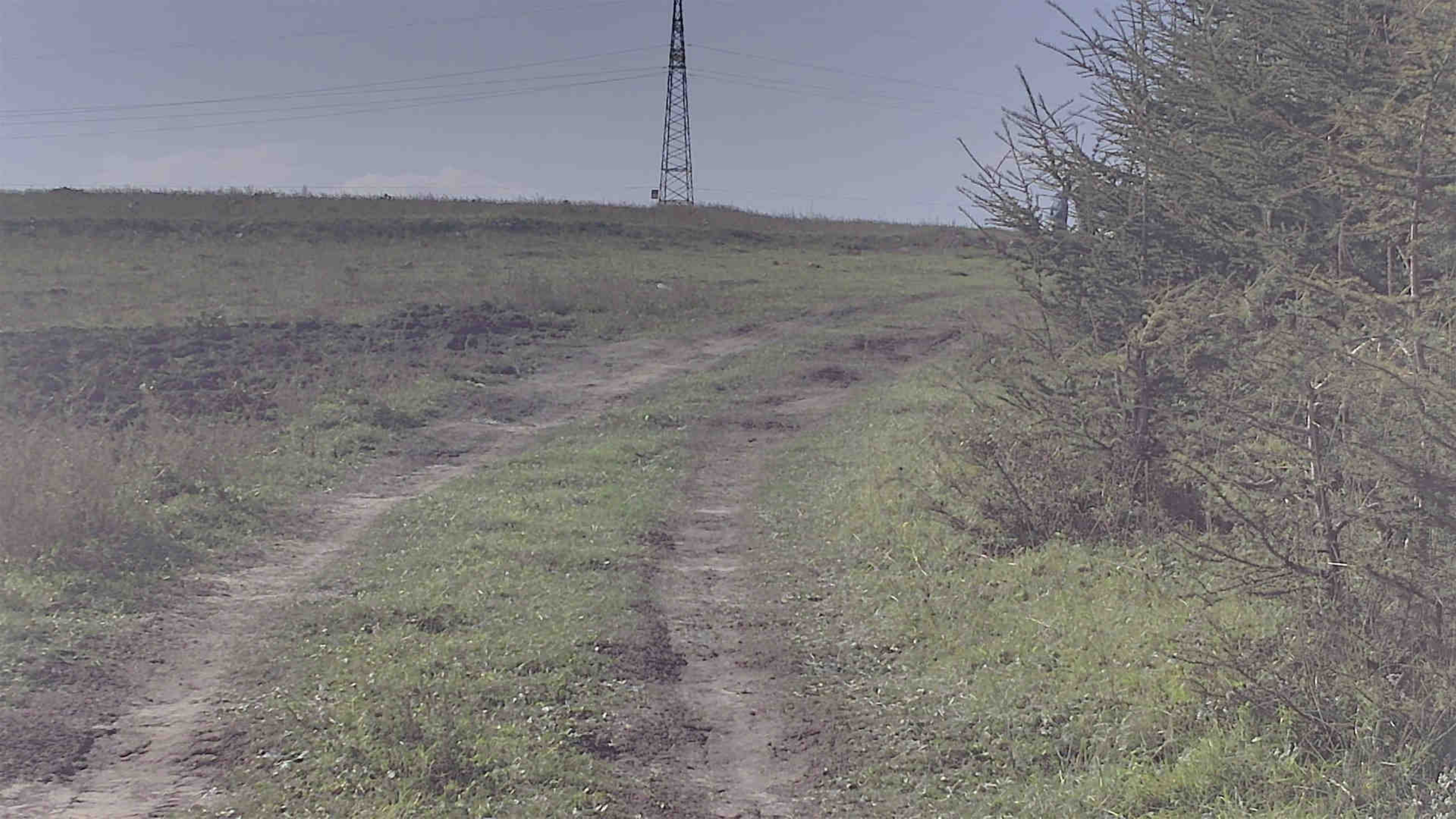}%
        \label{fig:partial_occlusion}
    }
    \caption{Common challenges in off-road scenarios: (a) Irregular geometries formed by rocky piles, and (b) Partial occlusions caused by vegetation.}
    \label{fig_0}
\end{figure}

\IEEEPARstart{A}{utonomous} vehicles are increasingly deployed to various environments, including urban, suburban, and off-road settings. While significant advancements have been made in structured environments with well-defined roads and traffic regulations, navigating unstructured off-road terrains remains a substantial challenge. Off-road environments lack clear pathways and usually come with irregular obstacles, as well as diverse terrain features such as uneven ground, vegetation, and occlusions caused by rocky piles or dense foliage, as shown in Fig.~\ref{fig_0}. These complexities require advanced perception capabilities to ensure safe and efficient navigation.

In urban environments, datasets such as KITTI \cite{ref_add1}, nuScenes \cite{ref_add2}, and Waymo Open \cite{ref_add3} provide valuable data for obstacle detection and localization. Traditional perception methods \cite{ref_add5},\cite{ref_add6},\cite{ref_add8} which are designed for 3D object detection tasks, detect and localize obstacles, thereby representing them with bounding boxes. However, these methods struggle to identify irregular and continuously distributed objects in off-road conditions. Recent approaches \cite{ref_add9},\cite{ref_add10},\cite{ref_add12} for 3D occupancy prediction focus on capturing fine-grained spatial structures and semantic information through voxel-based representations. While these methods offer more accurate modeling of obstacles, off-road scenarios also require evaluating the traversability of the environment.

To our knowledge, no work has yet employed semantic scene completion for 3D traversable terrain modeling in off-road scenarios. This gap motivates our novel approach to traversable area recognition. To address these challenges, 3DTTNet, a novel multimodal method for off-road traversability estimation, is proposed. This approach integrates LiDAR point clouds with monocular camera images to generate dense traversable predictions from a forward-facing perspective. By leveraging the complementary advantages of LiDAR's geometric precision and the rich semantic information provided by camera sensors, 3DTTNet enhances the extraction of environmental features, which is essential for accurate terrain modeling in complex terrains.

Furthermore, RELLIS-OCC is introduced, featuring 3D traversable annotations. This dataset encompasses geometric features such as step height, slope, and unevenness, crucial for evaluating the relationship between terrain traversability and vehicle dynamics. Through comprehensive analysis of vehicle obstacle-crossing conditions and incorporation of vehicle body structure constraints, a mapping from voxel semantic categories to traversability cost is established, resulting in four traversability cost labels: lethal, medium-cost, low-cost, and free. This mapping provides a detailed understanding of terrain difficulty, thus offering precise information for autonomous navigation decision-making. 

Building upon this foundation, extensive experiments demonstrate that 3DTTNet significantly outperforms existing approaches in 3D traversable area recognition, especially in off-road environments where obstacles exhibit irregular geometries and are often partially obstructed. Specifically, 3DTTNet achieves a 42\% improvement in the scene completion IoU metric compared to other models. This performance boost underscores the effectiveness of the multimodal fusion and advanced feature extraction techniques.

Moreover, the proposed framework is highly scalable and adaptable to various vehicle platforms. It offers a flexibility to adjust grid parameters and can incorporate more sophisticated dynamic models for traversability cost estimation. This adaptability ensures that the method can be tailored to the specific geometric and dynamic characteristics of different autonomous vehicles, further enhancing their ability to navigate complex off-road environments safely and efficiently.

\begin{table*}[t] 
\small
    \centering
    \caption{Comparison of Different Off-Road Environment Datasets.(Pro.: Proprioceptive, Seg.: Segmentation, Det.: Detection, Exc.: Excitation, SSD: Self-supervised Driving, Ann.: Annotation)}
    \setlength{\tabcolsep}{3pt}
    \resizebox{\textwidth}{!}{
    \begin{tabular}{>{\centering\arraybackslash}p{2.6cm}|c|cccc
        >{\centering\arraybackslash}p{1.1cm}
        >{\centering\arraybackslash}p{0.9cm}
        >{\centering\arraybackslash}p{0.6cm}ccc|
        >{\centering\arraybackslash}p{0.6cm}
        >{\centering\arraybackslash}p{0.6cm}
        >{\centering\arraybackslash}p{0.6cm}ccc|c|
        >{\centering\arraybackslash}p{1.1cm}|
        >{\centering\arraybackslash}p{2cm}}
        \toprule
            \multirow{3}{*}{\textbf{Dataset}}& \multirow{3}{*}{\textbf{Year}}& \multicolumn{10}{c|}{\textbf{Sensors}}& \multicolumn{6}{c|}{\textbf{Tasks}}& \multirow{3}{*}{\textbf{Size}}& \multirow{3}{*}{\textbf{Ann.}}& \multirow{3}{*}{\textbf{Ann. Type}} \\
            
            && \textbf{RGB}& \textbf{Stereo}& \textbf{Depth}& \textbf{NIR}& \textbf{Laser /Lidar}& \textbf{4D Radar}& \textbf{IMU /INS}& \textbf{GPS}& \textbf{Action}& \textbf{Pro.}& \textbf{2D Seg.}& \textbf{3D Seg.}& \textbf{2D Det.}& \textbf{SLAM}& \textbf{Exc.}& \textbf{SSD}&&& \\
        \midrule	
            \textbf{Tong~\cite{ref1}}& 2013&-&-&-&-& Laser&-& $\checkmark$& $\checkmark$&-&-&-&-&-& $\checkmark$&-&-& 384&-& -\\
            
            \textbf{DeepScene~\cite{ref2}}& 2016& $1024\times768$&-& $\checkmark$& $\checkmark$&-&-&-&-&-&-& $\checkmark$&-&-&-&-&-& 15,000& 366& pixel-wise \\
            
            \textbf{YCOR~\cite{ref3}}& 2018& $1024\times544$&-&-&-&-&-&-&-&-&-&-&-&-&-&-&-& 1,076& 1,076& pixel-wise \\
            
            \textbf{RUGD~\cite{ref4}}& 2019& $1376\times1110$&-&-&-&-&-&-&-&-&-& $\checkmark$&-&-&-&-&-& 37,000& 7,546& pixel-wise \\
            
            \textbf{Gresenz~\cite{ref5}}& 2021& $3840\times2160$&-&-&-&-&-& $\checkmark$& $\checkmark$&-& $\checkmark$&-&-&-&-& $\checkmark$&-& 12,982&-& - \\
            
             \multirow{2}{*}{\textbf{RELLIS-3D~\cite{ref6}}}& \multirow{2}{*}{2021}& \multirow{2}{*}{$1920\times1200$}&  \multirow{2}{*}{$\checkmark$}& \multirow{2}{*}{-}& \multirow{2}{*}{-}& 32-line/ 64-line& \multirow{2}{*}{-}&  \multirow{2}{*}{$\checkmark$}& \multirow{2}{*}{-}& \multirow{2}{*}{-}& \multirow{2}{*}{-}&  \multirow{2}{*}{$\checkmark$}&  \multirow{2}{*}{$\checkmark$}&\multirow{2}{*}{-}&  \multirow{2}{*}{$\checkmark$}&\multirow{2}{*}{-}&\multirow{2}{*}{-}& \multirow{2}{*}{6,235}& 6,235 /13,556& pixel-wise /point-wise \\
            
            \textbf{ROOAD~\cite{ref7}}& 2021& $1920\times1200$&-&-&-&-&-& $\checkmark$& $\checkmark$&-&-&-&-&-& $\checkmark$&-&-& 20,000&-& - \\
            
            \textbf{CaT~\cite{ref8}}& 2022& $1928\times1208$&-&-&-&-&-&-&-&-&-& $\checkmark$&-&-&-&-&-& 12,300& 1,812& traversability \\
            
            \textbf{ORFD~\cite{ref9}}& 2022& $1280\times720$&-&-&-& 40-line&-&-&-&-&-& $\checkmark$&-&-&-&-&-& 12,198& 12,198& traversability \\
            
            \textbf{SORT~\cite{ref10}}& 2022& $1280\times720$&-&-&-&-&-&-&-&-&-& $\checkmark$& -&-& -&-&-& 20,834& 20,834& pixel-wise \\
            
            \textbf{TartanDrive~\cite{ref11}}& 2022& $1024\times512$& $\checkmark$& $\checkmark$&-&-&-& $\checkmark$& $\checkmark$& $\checkmark$& $\checkmark$&-&-&-& $\checkmark$&-& $\checkmark$& 5 hours&-& - \\
            
            \multirow{3}{*}{\textbf{WaterScenes~\cite{ref13}}}&  \multirow{3}{*}{2023}&  \multirow{3}{*}{$1920\times1080$}& \multirow{3}{*}{-}& \multirow{3}{*}{-}& \multirow{3}{*}{-}& \multirow{3}{*}{-}&  \multirow{3}{*}{$\checkmark$}&  \multirow{3}{*}{$\checkmark$}&  \multirow{3}{*}{$\checkmark$}& \multirow{3}{*}{-}& \multirow{3}{*}{-}&  \multirow{3}{*}{$\checkmark$}&  \multirow{3}{*}{$\checkmark$}&  \multirow{3}{*}{$\checkmark$}& \multirow{3}{*}{-}& \multirow{3}{*}{-}& \multirow{3}{*}{-}&  \multirow{3}{*}{54,120}& 54,120/ 202,807 /54,120& pixel-wise/ bounding box /point-wise \\
            
            \textbf{Wild-Places~\cite{ref14}}& 2023&-&-&-&-& 16-line&-& $\checkmark$&-&-&-&-&-&-& $\checkmark$&-&-& 66,863&-& - \\
            
            \multirow{2}{*}{\textbf{GOOSE~\cite{ref15}}}& \multirow{2}{*}{2024}& \multirow{2}{*}{$2448\times2048$}& \multirow{2}{*}{-}& \multirow{2}{*}{-}&  \multirow{2}{*}{$\checkmark$}& 32-line/ 128-line & \multirow{2}{*}{-}&  \multirow{2}{*}{$\checkmark$}& \multirow{2}{*}{$\checkmark$}& \multirow{2}{*}{-}& \multirow{2}{*}{-}&  \multirow{2}{*}{$\checkmark$}& \multirow{2}{*}{$\checkmark$}& \multirow{2}{*}{-}&  \multirow{2}{*}{$\checkmark$}& \multirow{2}{*}{-}& \multirow{2}{*}{-}& \multirow{2}{*}{15,000}& 10,000 /10,000& pixel-wise /point-wise \\
            
            \textbf{TartanDrive v2.0~\cite{ref12}}& \multirow{2}{*}{2024}& \multirow{2}{*}{$1024\times512$}& \multirow{2}{*}{$\checkmark$}& \multirow{2}{*}{$\checkmark$}&\multirow{2}{*}{-}& solid/ 32-line& \multirow{2}{*}{-}& \multirow{2}{*}{$\checkmark$}& \multirow{2}{*}{$\checkmark$}& \multirow{2}{*}{$\checkmark$}& \multirow{2}{*}{$\checkmark$}& \multirow{2}{*}{-}& \multirow{2}{*}{-}& \multirow{2}{*}{-}& \multirow{2}{*}{$\checkmark$}& \multirow{2}{*}{$\checkmark$}& \multirow{2}{*}{$\checkmark$}& \multirow{2}{*}{7 hours}& \multirow{2}{*}{-}& \multirow{2}{*}{-} \\

        \bottomrule
    \end{tabular}}
    \label{tab:table_1}
\end{table*}

In summary, the main contributions are as follows:
\begin{itemize}
    \item[$\bullet$]An innovative framework is proposed for 3D traversable terrain modeling in off-road terrains based on semantic scene completion. This framework enables the completion and inference of occluded regions, a capability that traditional traversability estimation methods cannot achieve.
    
    \item[$\bullet$]3DTTNet, a transformer-based model that integrates multi-sensor features for traversability prediction. By leveraging a deformable attention mechanism, 3DTTNet enhances the interaction between multimodal features.
    
    \item[$\bullet$]A novel pipeline for generating 3D traversable annotations is proposed, integrating semantic accessibility analysis and geometric passability considerations. This method effectively addresses occlusion challenges and facilitates end-to-end model training with quantifiable traversability labels.
    
\end{itemize}

The remainder of the paper is organized as follows. Section \ref{sec:rela} provides a brief review of the relevant literature. Subsequently, the proposed network and dataset is detailed in Section \ref{sec:meth}. Experimental results and discussion are demonstrated in Section \ref{sec:expe}. Finally, a conclusion is drawn in Section \ref{sec:conc}.

\section{Related works}
\label{sec:rela}
\subsection{Off-road Environment Datasets}

Perception technology in autonomous driving largely relies on large-scale training datasets; however, existing datasets often do not clearly distinguish between off-road and unstructured road environments. In view of this situation, datasets designed for unstructured roads have been compiled and summarized, as presented in Table~\ref{tab:table_1}. Public datasets can be categorized based on their data sources into real-vehicle datasets \cite{ref1},\cite{ref2},\cite{ref3},\cite{ref4},\cite{ref5},\cite{ref6},\cite{ref7},\cite{ref8},\cite{ref9},\cite{ref11},\cite{ref12},\cite{ref13},\cite{ref14},\cite{ref15} and simulation datasets \cite{ref10}. Due to differences in data acquisition platform configurations, these datasets typically contain image \cite{ref2},\cite{ref3},\cite{ref4}, LiDAR point clouds \cite{ref1},\cite{ref14}, positioning data \cite{ref7}, or multimodal data \cite{ref6},\cite{ref13}. Notably, the TartanDrive \cite{ref11} dataset, in addition to multimodal data, is comprised of vehicle behavior data (pedal positions, steering) and vehicle proprioceptive data (suspension travel, wheel speed). Some datasets of unstructured roads include annotation information, which can be divided into semantic annotations \cite{ref4},\cite{ref6} and traversable area annotations \cite{ref8},\cite{ref9}. The RUGD \cite{ref4} and RELLIS-3D \cite{ref6} datasets employ the same labeling rules to annotate the semantic categories of objects in the environment. The CaT \cite{ref8} and ORFD \cite{ref9} datasets feature traversable area annotations made by experienced drivers, who subjectively judge based on the type of platform. However, despite these advances, there is a lack of 3D traversable datasets specifically tailored for off-road scenarios.

\subsection{Traversability Estimation Scheme}
Dima et al. \cite{ref16} defined obstacles as "areas that a vehicle cannot or should not traverse." The initiation of the DARPA Challenge in 2004 advanced research on traversable regions in off-road environments \cite{ref16},\cite{ref17},\cite{ref21}. Early studies frequently employed Support Vector Machines (SVMs) to recognize vehicle traversable areas using various perception schemes, including LiDAR and vision fusion \cite{ref22}, pure LiDAR methods \cite{ref23}, and pure vision approaches \cite{ref24}. With the development of deep learning, these techniques have been widely applied to off-road research, encompassing traversable area image segmentation \cite{ref25},\cite{ref26}, road detection and classification \cite{ref29}, LiDAR point cloud obstacle recognition \cite{ref30},\cite{ref31}, multi-sensor fusion \cite{ref33},\cite{ref35}, and weakly supervised traversable area recognition \cite{ref36}. Traversable area recognition results are typically represented in various forms: images \cite{ref37}, 2D grid maps \cite{ref38}, 2.5D grid maps \cite{ref32}, and 3D grid maps \cite{ref42},\cite{ref55}. Among these, 3D grid maps are capable of more precisely characterizing the shape and occupancy information of obstacles, making them a more ideal representation form.

\subsection{Traversability Quantification}
Goodin et al. \cite{ref39} defines "traversability" as a quantification of the drivability of vehicles with certain characteristics on terrain. In off-road scenarios, layered cost maps are composed of multiple sub-maps to address different terrain features, which are widely used for traversable area analysis \cite{ref40},\cite{ref54}. These layers typically consist of ground semantic and geometric information, making the selection of appropriate cost functions crucial for accurate traversability estimation.

Cameras capture rich semantic scene data, while LiDAR provides precise geometric information; thus, fusing these modalities is a common approach \cite{ref41},\cite{ref42},\cite{ref56}. Zhao et al. \cite{ref41} integrated images and LiDAR point clouds to construct an elevation-semantic 2D grid map, assigning traversability probabilities to various semantic categories to generate a traversability probability grid. Leung et al. \cite{ref43} developed a traversability cost map by combining image-based semantic segmentation with geometric attributes such as slope, step height, and roughness. Shaban et al. \cite{ref31} classified point clouds into four traversability levels: free passage, low cost, medium cost, and obstacles, and used deep learning networks to produce a Bird's Eye View (BEV) traversability map. However, these methods exhibit limited capability in 3D environmental representation and are unable to effectively infer and reconstruct occluded regions.

\section{Methods}
\label{sec:meth}
The objective is to predict a 3D traversable grid map within the camera view in front of the platform by leveraging monocular images and LiDAR point clouds as input, thereby enabling accurate and reliable traversability estimation in off-road environments. Specifically, the image $I_t$ and LiDAR point cloud $L_t$ are taken as inputs, and the output is a voxel grid $Y_t \in \{c_0, c_1, c_2, \cdots, c_M\}^{H \times W \times Z}$, where each voxel is classified as either empty $\{c_0\}$ or occupied by a specific traversability class within $\{c_1, c_2\cdots, c_M\}$. In this context, $M$ denotes the total number of traversability classes, while $H$, $W$, and $Z$ represent the length, width, and height of the voxel grid, respectively. To facilitate this process, RELLIS-OCC is introduced as the first occupancy dataset with traversability cost annotations specifically designed for off-road scenarios.

\subsection{Architecture of 3DTTNet}
\begin{figure}[t]
\centering
\includegraphics[width=\columnwidth]{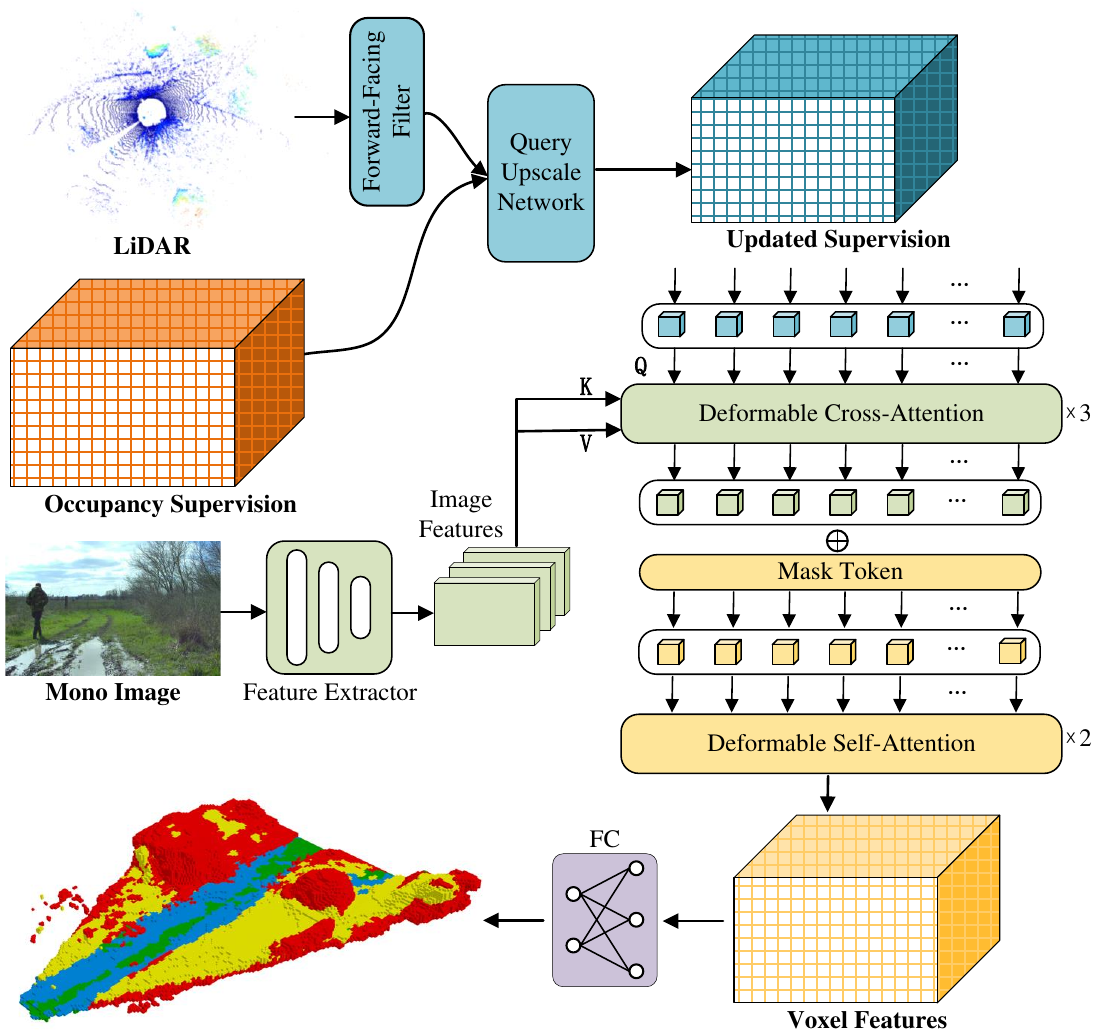}
\caption{Overall framework of 3DTTNet. The process starts by extracting multimodal features from LiDAR point clouds and monocular images, with LiDAR providing geometric details and images contributing semantic information. These features are then fused using deformable attention mechanisms to enhance spatial and semantic interactions. Following fusion, the refined features are processed through deformable attention layers to generate a detailed 3D traversability map that highlights traversable areas.}
\label{fig_8}
\end{figure}

Off-road environments lack structured roads and traffic regulation information, therefore the traversability of such terrains is closely linked to the geometric constraints of the platform and necessitates detailed geometric ground analysis. Consequently, there is an increased demand for advanced 3D traversability prediction capabilities in perception algorithms. To address these challenges, a multimodal semantic scene completion method specifically designed for off-road conditions is proposed. This method takes advantages of LiDAR point clouds and monocular images to generate dense 3D traversable area predictions from a forward-facing perspective. The overall pipeline of 3DTTNet is illustrated in Fig.~\ref{fig_8}. 

The model begins by extracting features from both modalities: the LiDAR data which provides precise geometric information, and images which contribute rich semantic details. These features are then fused using deformable attention mechanisms to effectively integrate spatial and semantic information. Subsequently, the fused features are processed through deformable self-attention layers, on purpose to enhance the 3D occupancy representation. Eventually, it produces a dense 3D grid map highlighting traversable areas.

\subsubsection{Feature Extraction}
To achieve more accurate terrain modeling, multimodal information is utilized to extract environmental features. The model employs a ResNet-50 backbone to extract image features $F_t^{2D} \in \mathbb{R}^{l \times b \times d}$ from the monocular camera image $I_t$, where $l \times b$ represents the spatial resolution and $d$ denotes the feature dimension. Concurrently, the sparse point clouds $L_t$ from the LiDAR input are voxelized to generate sparse voxel supervision $S_{\text{spa}} \in \{0, 1\}^{H \times W \times Z}$. Subsequently, a query upscale network is utilized to produce lower spatial resolution voxel supervision $S_{\text{den}} \in \{0, 1\}^{H/2 \times W/2 \times Z/2}$, thereby enhancing the robustness of the queries.

\textbf{Mask Token.} The mask token module proposed by Y. Li \cite{ref45} is incorporated to represent empty voxel spaces within the environment. When combined with voxel supervision, this approach will form a more comprehensive 3D voxel feature, facilitating improved occupancy predictions.

\subsubsection{Feature Fusion}
In the model, deformable attention (DA) mechanisms \cite{ref50} are adopted to propel the interaction between local regions of interest in the 3D and 2D feature spaces. Deformable attention dynamically samples $N_s$ points around a reference point, and the attention weights are adaptively learned based on the query vector, while the spatial offsets for sampled points are predicted relative to the reference point, enabling flexible and efficient feature aggregation. The attention results are computed based on these samples using the following equation:
\begin{equation}
\text{DA}\left(q, p, F\right) = \sum_{s=1}^{N_s}{A_s W_s F(p + \delta p_s)},
\label{eqs1}
\end{equation}
where $q$ is the query vector, $p$ denotes the reference point associated with $q$, $F$ represents the input feature map, $W_s \in \mathbb{R}^{d \times d}$ are learnable weight matrices for value generation at each sampled point, $A_s \in [0, 1]$ are attention weights for each sampled point, dynamically learned based on the query, $\delta p_s \in \mathbb{R}^2$ are the predicted offsets for the sampled points relative to the reference point $p$, and $F(p + \delta p_s)$ denotes the feature vector at the location $p + \delta p_s$, extracted via bilinear interpolation from the input features.

\textbf{Deformable Cross-Attention (DCA).} Multiple DCA layers are employed to enhance the interaction between the occupancy supervision queries $S_{\text{den}}$ and the image features $F_t^{2D}$, thereby reinforcing the integration of multimodal information. The 3D grids are projected onto the 2D image feature space $F_t^{2D}$ using projection matrices. The resulting 2D points serve as reference points for the queries $S_{\text{den}}$, from which surrounding features are sampled. A weighted sum of the sampled features is computed to produce the output of the DCA layers. Consequently, refined queries $\hat{S}_{\text{den}}$ are obtained, fusing environmental semantic features with geometric features.
 
\begin{equation}
\text{DCA}\left(s_{\text{den}}, F_t^{2D}\right) = \text{DA}(s_{\text{den}}, \mathcal{P}(p, t), F_t^{2D}),
\label{eqs2}
\end{equation}
for each query $s_{\text{den}}$ at position $p = (x, y, z)$, the camera projection function $\mathcal{P}(p, t)$ is applied to map the query to a corresponding reference point on the image $I_t$.

\textbf{Deformable Self-Attention (DSA).} The refined queries $\hat{S}_{\text{den}}$ are concatenated with mask tokens to form the initial voxel features $F^{3D} \in \mathbb{R}^{H/2 \times W/2 \times Z/2 \times d}$. Subsequently, these features are processed through multiple DSA layers to obtain the enhanced voxel features $\hat{F}^{3D} \in \mathbb{R}^{H/2 \times W/2 \times Z/2 \times d}$. The DSA layers enable the model to learn spatial 3D occupancy features more efficiently and accurately. In the equations, the query $q_p$ can either be a mask token or a refined query located at position $p = (x, y, z)$.

\begin{equation}
\text{DSA}\left(F^{3D}, F^{3D}\right) = \text{DA}(q_p, p, F^{3D}).
\label{eqs3}
\end{equation}

\subsubsection{Output}
The enhanced voxel features $\hat{F}^{3D}$ are input to fully connected (FC) layers and passed through a softmax function to produce dense traversability predictions $Y_t \in \{c_0, c_1, c_2, \cdots, c_M\}^{H \times W \times Z}$, where each voxel is classified as either empty $\{c_0\}$ or occupied by a specific traversability class within $\{c_1, c_2\cdots, c_M\}$. Mapping these predictions back to the 3D grid yields a detailed occupancy map that highlights traversable areas and provides traversability labels essential for autonomous off-road navigation.

\subsubsection{Training Loss}
3DTTNet is trained end-to-end from scratch by minimizing three primary loss components: the semantic scale loss $\mathcal{L}_{\text{scal}}^{\text{sem}}$, the geometric scale loss $\mathcal{L}_{\text{scal}}^{\text{geo}}$, and the weighted cross-entropy loss $\mathcal{L}_{\text{ce}}$. The total training loss $\mathcal{L}_{\text{total}}$ is defined as the sum of these individual losses:
\begin{equation}
\mathcal{L}_{\text{total}} = \mathcal{L}_{\text{ce}} + \mathcal{L}_{\text{scal}}^{\text{sem}} + \mathcal{L}_{\text{scal}}^{\text{geo}}.
\label{eqs4}
\end{equation}

The weighted cross-entropy loss is computed as follows:
\begin{equation}
\mathcal{L}_{\text{ce}} = \frac{1}{\Omega} \sum_{i \in \Omega} \omega_{\text{tar}_i} \left( y_{i,\text{tar}_i} - \log{\left(\sum_{c=1}^{M} e^{y_{i,c}}\right)} \right),
\label{eqs5}
\end{equation}
where $\Omega$ represents the set of valid voxels, $M$ is the number of classes, $\text{tar}_i$ denotes the true class of voxel $i$, $\omega_{\text{tar}_i}$ is the class weight for the true class $\text{tar}_i$ of voxel $i$, $y_{i,\text{tar}_i}$ is the logit output of the model for the true class $\text{tar}_i$, and $y_{i,c}$ is the logit output of the model predicting voxel $i$ as class $c$.

In the context of semantic scene completion, model performance is evaluated via several metrics, including Precision $P_c$, Recall $R_c$, and Specificity $S_c$. These metrics assess the ability of the model to correctly classify occupied and non-occupied voxels. The metrics are formulated as follows:
\begin{equation}
P_c\left(\text{tar}, p\right) = \log\left( \frac{\sum_{i \in \Omega} p_{i, c} \cdot \llbracket \text{tar}_i = c \rrbracket }{\sum_{i \in \Omega} p_{i, c}} \right),
\label{eqs6}
\end{equation}

\begin{equation}
R_c\left(\text{tar}, p\right) = \log\left( \frac{\sum_{i \in \Omega} p_{i, c} \cdot \llbracket \text{tar}_i = c \rrbracket}{\sum_{i \in \Omega} \llbracket \text{tar}_i = c \rrbracket} \right),
\label{eqs7}
\end{equation}

\begin{equation}
S_c\left(\text{tar}, p\right) = \log\left( \frac{\sum_{i \in \Omega} (1 - p_{i, c}) \cdot \llbracket \text{tar}_i \neq c \rrbracket}{\sum_{i \in \Omega} \llbracket \text{tar}_i \neq c \rrbracket} \right),
\label{eqs8}
\end{equation}
where $p_{i, c}$ represents the predicted probability that voxel $i$ belongs to class $c$, and $\llbracket \cdot \rrbracket$ denotes the Iverson bracket.

For greater generality, the scale loss $\mathcal{L}_{\text{scal}}$ maximizes the above class-wise metrics:
\begin{equation}
\mathcal{L}_{\text{scal}}(\text{tar}, p) = -\frac{1}{M} \sum_{c=1}^{M} \left(P_c(\text{tar}, p) + R_c(\text{tar}, p) + S_c(\text{tar}, p)\right).
\label{eqs9}
\end{equation}

Based on the formulation, we compute the semantic loss 
$\mathcal{L}_{\text{scal}}^{\text{sem}} = \mathcal{L}_{\text{scal}}(\text{tar}_{\text{sem}}, p_{\text{sem}})$ and the geometric loss $\mathcal{L}_{\text{scal}}^{\text{geo}} = \mathcal{L}_{\text{scal}}(\text{tar}_{\text{geo}}, p_{\text{geo}})$, where $\text{tar}_{\text{sem}}$ and $\text{tar}_{\text{geo}}$ represent the ground truth labels for semantics and geometry, respectively, with $p_{\text{sem}}$ and $p_{\text{geo}}$ denoting the corresponding model predictions.

\subsection{3D Traversable Annotations}
We propose a labeling pipeline that integrates semantic accessibility analysis with geometric passability considerations, resulting in the generation of the RELLIS-OCC dataset. This dataset, derived from the semantically annotated point clouds of the RELLIS-3D dataset \cite{ref6}, comprises five sequences and a total of 11,201 frames. Sequences 00000 to 00003 are designated as the training set, while sequence 00004 is used for validation.

\subsubsection{Dense Semantic Annotations}
In the experiments, networks trained with sparse LiDAR points fail to predict sufficiently dense occupancy, highlighting the need for dense occupancy labels. Dense forward-facing labels are generated from the semantically annotated point clouds of the RELLIS-3D dataset. We transform multi-frame LiDAR scans into a unified coordinate system and voxelize the dense points into grids, including dynamic objects. This process captures surface information since LiDAR measures points on object surfaces.

\begin{figure}[t]
\centering
\includegraphics[width=\columnwidth]{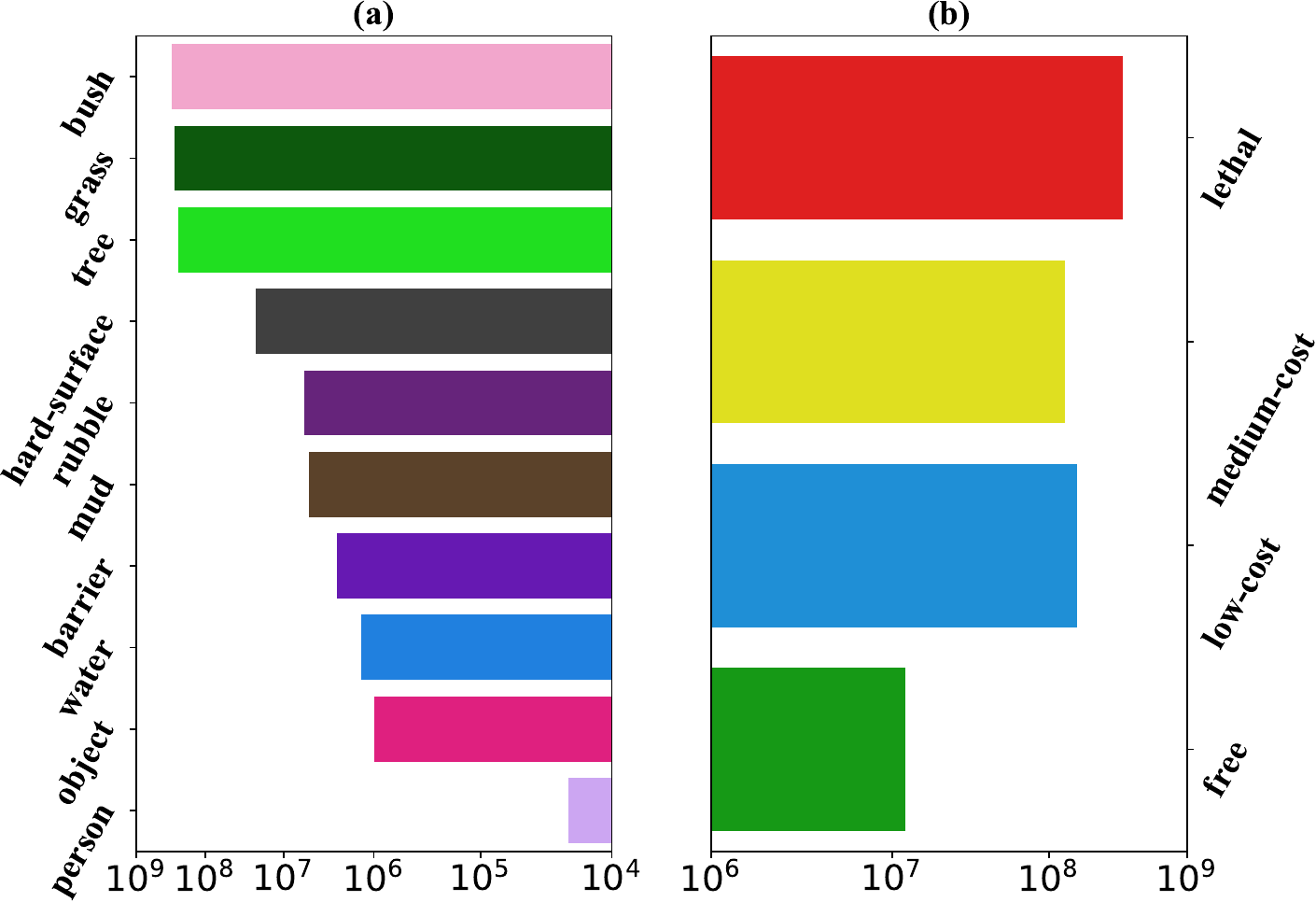}
\caption{Voxel Label Distribution: (a) the distribution of semantic annotations, and (b) the distribution of traversability cost annotations. Each category's color corresponds to its label.}
\label{fig_2}
\end{figure}

Due to the complex and occluded nature of off-road obstacles, ground truth within a range of $[38.4\,\mathrm{m}, 51.2\,\mathrm{m}, 8\,\mathrm{m}]$ is selected. The prediction range is set to $[0\,\mathrm{m}, 38.4\,\mathrm{m}]$ along the X-axis, $[-25.6\,\mathrm{m}, 25.6\,\mathrm{m}]$ along the Y-axis, and $[-2\,\mathrm{m}, 6\,\mathrm{m}]$ along the Z-axis. The ground truth is voxelized into a $192 \times 256 \times 40$ grid with a voxel size of $0.2\,\mathrm{m}$ and data is filtered to retain voxels within the camera's field of view. A remapping of the original categories is then performed to assess their impact on traversability. Voxel data statistics of the RELLIS-OCC dataset, comprising 11 semantic category labels (including void), are presented in Fig.~\ref{fig_2}(a).

\subsubsection{Geometric Feature Parameter Calculation}A 3D occupancy voxel is constructed by integrating registered dense point clouds. Each voxel $G_j = \{x_j, y_j, z_j\}$ is centered at coordinates ($x_j$, $y_j$) in the ground map system and has an elevation $z_j$. The neighborhood of each voxel is defined as a circular region $\Omega$ arround $G_j$, where the point distribution approximates a circle. Each point from the registered dense point cloud is assigned to a voxel, and all points within a voxel form a matrix $\mathbf{g}$.

\begin{equation}
\mathbf{g} = 
\begin{bmatrix}
x_1 & x_2 & x_3 & \cdots & x_n \\
y_1 & y_2 & y_3 & \cdots & y_n \\
z_1 & z_2 & z_3 & \cdots & z_n
\end{bmatrix}.
\label{eqs10}
\end{equation}

\textbf{Step Height $h$:} The maximum elevation difference between the central voxel $G$ and other voxels $G_j$ within its neighborhood $\mathrm{\Omega}$ is calculated, in order to assess the variation in terrain height around the central voxel.

\begin{equation}
h = \max\left(\left|G^z - G_j^z\right|\right), \quad G_j \in \mathrm{\Omega},
\label{eqs11}
\end{equation}
where $G^z$ and $G_j^z$ represent the elevations of the central voxel $G$ and the neighboring voxel $G_j$, respectively.

\textbf{Slope $s$:} The slope is calculated from the elevation values within the neighborhood. For each voxel $G$, the centers of all voxels in its neighborhood $\mathrm{\Omega}$ are fitted to a plane. The angle between the normal vector $\mathbf{n}$ of this plane and the unit vector $\mathbf{Z} = (0, 0, 1)$ along the Z-axis of the map coordinate system is defined as the slope.

\begin{equation}
s = \arccos \frac{\mathbf{n} \cdot \mathbf{Z}}{\|\mathbf{n}\| \|\mathbf{Z}\|},
\label{eqs12}
\end{equation}
where the value of $\mathbf{n}$ is obtained by calculating the covariance matrix using Principal Component Analysis (PCA).

\textbf{Unevenness $u$:} The unevenness is estimated by calculating the logarithmic mean squared error (MSE) between the actual elevation of each voxel within the neighborhood $\mathrm{\Omega}$ and the fitted plane. The unevenness is expressed as:
\begin{equation}
u = \log \left(\frac{1}{m} \sum_{i=1}^{m} [z_{\text{actual}} - (a_0 \cdot x + a_1 \cdot y + c)]_i^2\right),\quad m \in \mathrm{\Omega},
\label{eqs13}
\end{equation}
where $a_0$ and $a_1$ are the coefficients representing the slope in the $x$ and $y$ directions, respectively, $c$ is the intercept, and $z_{\text{actual}}$ represents the actual elevation of the current voxel.

\subsubsection{Vehicle Obstacle Crossing Condition Analysis}
In off-road conditions, the traversable area of a vehicle is strongly coupled with its geometric passability, as the difficulty of traversing the ground relates to various road geometric features. By analyzing the failure conditions during vehicle obstacle negotiation, we introduce the Geometric Passability Assessment Mask (GPAM).

\textbf{GPAM:} The failure conditions for vehicle obstacle crossing are composed of body suspension, front-end collision, vertical obstacles, ditches, and lateral tipping. Since the dataset does not encompass particularly challenging scenarios, the analysis is focused on four specific cases: overcoming vertical obstacles, crossing ditches, crossing overhang obstacles and traversing longitudinal slopes \cite{ref44}. 

\paragraph{Vehicle Vertical Obstacle Overcoming Conditions}The performance of wheeled vehicles in overcoming step-like obstacles and vertical protrusions is primarily evaluated based on the maximum height of vertical obstacles they can surmount. The dimensionless expression for the maximum obstacle height $h_{\text{max}}$ that the front wheels of a 4×4 wheeled vehicle can overcome is:
\begin{equation}
\frac{h_{\text{max}}}{r} = \frac{1 - \mu \frac{r}{l} + \eta^2 - \eta \sqrt{1 - 2\mu \frac{r}{l} + \eta^2}}{\left(1 + \mu \frac{r}{l}\right)^2 + \eta^2},
\label{eqs14}
\end{equation}

\begin{equation}
\eta = \frac{1 - \mu \frac{r}{l} - (1 + \mu^2) \frac{a}{l}}{\mu},
\label{eqs15}
\end{equation}
where $r$ denotes the wheel radius, $\mu$ signifies the coefficient of friction, $l$ refers to the wheelbase of the vehicle, and $a$ corresponds to the horizontal distance from the front axle to the center of gravity of the vehicle. Thus, the condition for a vehicle to overcome a vertical obstacle is expressed as follows:
\begin{equation} 
\Psi_{\text{ver}}(G_j) = \left \llbracket h \leq h_{\text{max}} \right \rrbracket.
\label{eqs16} 
\end{equation}

\paragraph{Vehicle Traversing Trenches Conditions}The width of the trenches that a wheeled vehicle can overcome is evaluated using the ratio of the trench width $l_d$ to the wheel diameter $D$, expressed as $\frac{l_d}{D}$. The ratio of the height of vertical obstacles $h_d$ that can be overcome to the wheel diameter, expressed as $\frac{h_d}{D}$, can be converted to $\frac{l_d}{D}$, as shown in Equation\ref{eqs17}. Given a specific obstacle height, the corresponding trench width ratio $\frac{l_d}{D}$ can be determined.

\begin{equation}
\frac{l_{d}}{D} = 2\sqrt{\frac{h_d}{D} - \left(\frac{h_d}{D}\right)^2}.
\label{eqs17}
\end{equation}

Therefore, the condition for a vehicle to overcome a trench, where $l_{d{\text{max}}}$ represents the maximum trench width the vehicle can cross, calculated when $h_d$ equals $h_{\text{max}}$ using Equation\ref{eqs17}, is expressed as:
\begin{equation} 
\Psi_{\text{tre}}(G_j) = \left \llbracket l_d \leq l_{d_{\max}} \right \rrbracket.
\label{eqs18} 
\end{equation}

\paragraph{Vehicle Crossing Overhang Obstacles Conditions}To ensure a vehicle successfully traversing overhang obstacles, the height $h_{\text{obj}}$ of the lowest point of the obstacle above the ground must exceed the sum of $h_{\text{pc}}$ (which is the height of the obstacle point cloud relative to the LiDAR sensor), and $h_{\text{Lid}}$ (which is the height of the LiDAR sensor above the ground). Once the vehicle encounters an overhang obstacle, the following conditions must be met:
\begin{equation} 
\Psi_{\text{ove}}(G_j) = \left \llbracket h_{\text{obj}} > h_{\text{pc}} + h_{\text{Lid}} \right \rrbracket.
\label{eqs19} 
\end{equation}

\paragraph{Vehicle Traversing Longitudinal Slopes Conditions}
In off-road conditions, the maximum slope angle $\emptyset_m$ that a vehicle can traverse is governed by the terrain's friction-related angle $\emptyset_{\mu}$ and the vehicle's approach angle $\emptyset_a$. The friction-related angle $\emptyset_{\mu}$ can be calculated using:
\begin{equation}
\emptyset_{\mu} = \sin^{-1}\left( \frac{F_{\text{max}}}{mg\sqrt{\mu_s^2 + 1}} \right) - \tan^{-1}(\mu_s),
\label{eqs20}
\end{equation}
where $F_{\text{max}}$ is the vehicle's maximum propulsion force, $m$ is the vehicle mass, $g$ is gravitational acceleration, and $\mu_s$ is the terrain's unevenness parameter. This formula accounts for both surface roughness and traction limitations \cite{ref57}. The ultimate traversable slope angle $\emptyset_m$ is then determined by the limiting factor between terrain friction and vehicle geometry:
\begin{equation}
\emptyset_m = \min(\emptyset_{\mu}, \emptyset_a),
\label{eqs21}
\end{equation}

Therefore, the condition for a vehicle to overcome a slope is expressed as:
\begin{equation} 
\Psi_{\text{slo}}(G_j) = \left \llbracket s \leq \emptyset_m \right \rrbracket.
\label{eqs22} 
\end{equation}

Above, the expression of GPAM is:
\begin{equation}
\text{GPAM}(G_j) = \left\llbracket \Psi_{\text{ver}}(G_j) \land \Psi_{\text{tre}}(G_j) \land \Psi_{\text{ove}}(G_j) \land \Psi_{\text{slo}}(G_j) \right\rrbracket.
\label{eqs23}
\end{equation}

\subsubsection{Traversability Cost Annotations}
To establish an initial mapping from voxel semantic categories to traversability cost, a set of rules is defined, incorporating adjustments based on the previously computed GPAM, which evaluates the vehicle's geometric passability. Voxels failing to meet the required passability criteria are assigned a lethal cost. This results in four distinct traversability cost labels: lethal, medium-cost(M-cost), low-cost(L-cost), and free, which represent a spectrum of traversability difficulty, from challenging to easy. The statistical distribution of these labels is shown in Fig.~\ref{fig_2}(b).

The traversability cost labeling function \( F(G_j) \) is formalized as follows:
\begin{equation}
F(G_j) =
\begin{cases}
\text{Lethal}, & \text{if } S(G_j) \in S_{\text{lethal}} \ \lor \ \text{GPAM}(G_j) = 0 \\
\text{M-cost}, & \text{if } S(G_j) \in S_{\text{traversable}} \ \land \\
& \phi_{\text{geo}}(G_j) \geq \tau_{\text{medium}}(V) \\
\text{L-cost}, & \text{if } S(G_j) \in S_{\text{traversable}} \ \land \\ 
& \tau_{\text{low}}(V) \leq \phi_{\text{geo}}(G_j) < \tau_{\text{medium}}(V) \\
\text{Free}, & \text{if } S(G_j) \in S_{\text{free}} \ \land \ \phi_{\text{geo}}(G_j) < \tau_{\text{low}}(V) 
\end{cases}
\label{eqs24}
\end{equation}

\paragraph{Semantic Mapping}
The semantic categories are divided into three types: lethal obstacles that the vehicle cannot traverse (\( S_{\text{lethal}} = \{ \text{person}, \text{object}, \text{barrier}, \text{tree}\} \)), traversable terrains that require verification (\( S_{\text{traversable}} = \{ \text{water}, \text{soil}, \text{bush}, \text{grass}\} \)), and free surfaces considered inherently traversable (\( S_{\text{free}} = \{ \text{hard-surface}, \text{mud}, \text{rubble}\} \)).

\paragraph{Geometric Risk Evaluation Function}
The geometric risk evaluation function \( \phi_{\text{geo}}(G_j) \) quantifies the passability of each voxel based on several terrain characteristics:
\begin{equation}
\phi_{\text{geo}}(G_j) = \frac{h}{h_{\max}(V)} + \frac{s}{s_{\max}(V)} + \frac{u}{u_{\text{critic}}(V)},
\label{eqs25}
\end{equation}
where \( h_{\max}(V) \) is the maximum vertical obstacle height that the vehicle can overcome (Equation \ref{eqs14}), \( s_{\max}(V) \) represents the maximum slope angle the vehicle can traverse (Equation \ref{eqs21}), and \( u_{\text{critic}}(V) = \log \left( \min \left( z_{\text{susp}}, {z_{\text{clearance}}} \right) \right) \) is the critical unevenness, which depends on the suspension displacement \( z_{\text{susp}} \) and the minimum ground clearance \( z_{\text{clearance}} \) of the vehicle chassis.

\paragraph{Threshold Calibration}
Thresholds for classifying terrain into medium-cost and low-cost categories are dynamically derived based on vehicle parameters:
\begin{equation}
\tau_{\text{medium}}(V) = \eta_{\text{medium}} \times \phi_{\text{geo, max}}(V),
\label{eqs26}
\end{equation}

\begin{equation}
\tau_{\text{low}}(V) = \eta_{\text{low}} \times \phi_{\text{geo, max}}(V),
\label{eqs27}
\end{equation}
where \( \eta_{\text{medium}} \) and \( \eta_{\text{low}} \) are scaling factors that adjust the thresholds for medium and low-cost classifications, respectively. \( \phi_{\text{geo, max}}(V) \) represents the theoretical maximum geometric risk when \( h = h_{\max}, s = s_{\max}, u = u_{\text{critic}} \).

In conclusion, the pipeline for generating traversability cost annotations is outlined in Algorithm \ref{algorithm 1}.

\begin{algorithm}[htbp]
\caption{Labeling Pipeline}
\label{algorithm 1}
\begin{algorithmic}[1]
\REQUIRE Point clouds $g$, Voxel grid $G$, Vehicle parameters $V$
\ENSURE Traversability labels $F(G_j)$

\STATE Initialize semantic mapping:
\STATE $S_{\text{lethal}} \gets \{\text{person}, \text{object}, \text{barrier}, \text{tree}\}$
\STATE $S_{\text{traversable}} \gets \{\text{water}, \text{soil}, \text{bush}, \text{grass}\}$
\STATE $S_{\text{free}} \gets \{\text{hard-surface}, \text{mud}, \text{rubble}\}$

\FOR{each voxel $G_j$}
    \IF{$S(G_j) \in S_{\text{lethal}}$}
        \STATE $F(G_j) \gets \text{Lethal}$
    \ELSIF{$S(G_j) \in S_{\text{free}}$}
        \STATE $F(G_j) \gets \text{Free}$
    \ELSE
        \STATE Compute geometric features:
        \STATE $h \gets \max(|G_j^z - G_k^z|)$ \COMMENT{Step height}
        \STATE $s \gets \arccos(\mathbf{n} \cdot \mathbf{Z})$ \COMMENT{Slope}
        \STATE $u \gets \log(\text{MSE}(\text{plane}, z_{\text{actual}}))$ \COMMENT{Unevenness}
        
        \STATE Evaluate GPAM:
        \STATE $\Psi_{\text{ver}} \gets \llbracket h \leq h_{\max}(V) \rrbracket$
        \STATE $\Psi_{\text{tre}} \gets \llbracket l_d \leq l_{d_{\max}}(V) \rrbracket$
        \STATE $\Psi_{\text{ove}} \gets \llbracket h_{\text{obj}} > h_{\text{pc}} + h_{\text{Lid}} \rrbracket$
        \STATE $\Psi_{\text{slo}} \gets \llbracket s \leq \emptyset_m(V) \rrbracket$
        \STATE $\text{GPAM} \gets \llbracket \Psi_{\text{ver}} \land \Psi_{\text{tre}} \land \Psi_{\text{ove}} \land \Psi_{\text{slo}} \rrbracket$
        
        \IF{$\text{GPAM} = 0$}
            \STATE $F(G_j) \gets \text{Lethal}$
        \ELSE
            \STATE Compute geometric risk:
            \STATE $\phi_{\text{geo}} \gets \frac{h}{h_{\max}} + \frac{s}{s_{\max}} + \frac{u}{u_{\text{critic}}}$
            
            \STATE Assign cost label:
            \IF{$\phi_{\text{geo}} \geq \tau_{\text{medium}}(V)$}
                \STATE $F(G_j) \gets \text{M-cost}$
            \ELSIF{$\phi_{\text{geo}} \geq \tau_{\text{low}}(V)$}
                \STATE $F(G_j) \gets \text{L-cost}$
            \ELSE
                \STATE $F(G_j) \gets \text{Free}$
            \ENDIF
        \ENDIF
    \ENDIF
\ENDFOR
\end{algorithmic}
\end{algorithm}

\subsection{Architecture of ORD-BKI}

\begin{figure}[htbp]
\centering
\includegraphics[width=\columnwidth]{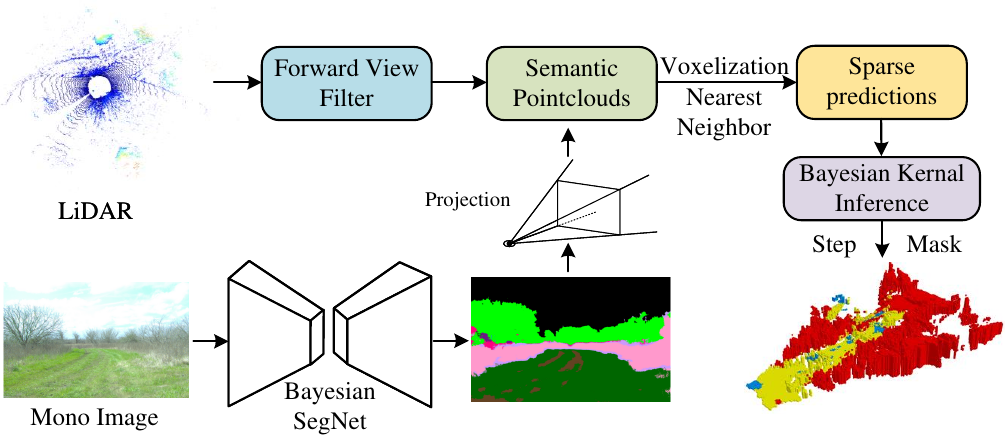}
\caption{Overall framework of ORD-BKI.}
\label{fig_9}
\end{figure}

The research presented in \cite{ref48} introduces the S-BKI model, which infers grid attributes for discrete semantic grid maps to achieve continuous and dense semantic mapping. To ensure a fair comparison, the S-BKI model is constructed using LiDAR point clouds as input. Based on the vehicle's geometric passability, a feasible domain recognition model, ORD-BKI, is developed, which employs Bayesian Kernel Inference for semantic scene completion, as illustrated in Fig.~\ref{fig_9}.

\begin{figure}[b]
\centering
\includegraphics[width=\columnwidth]{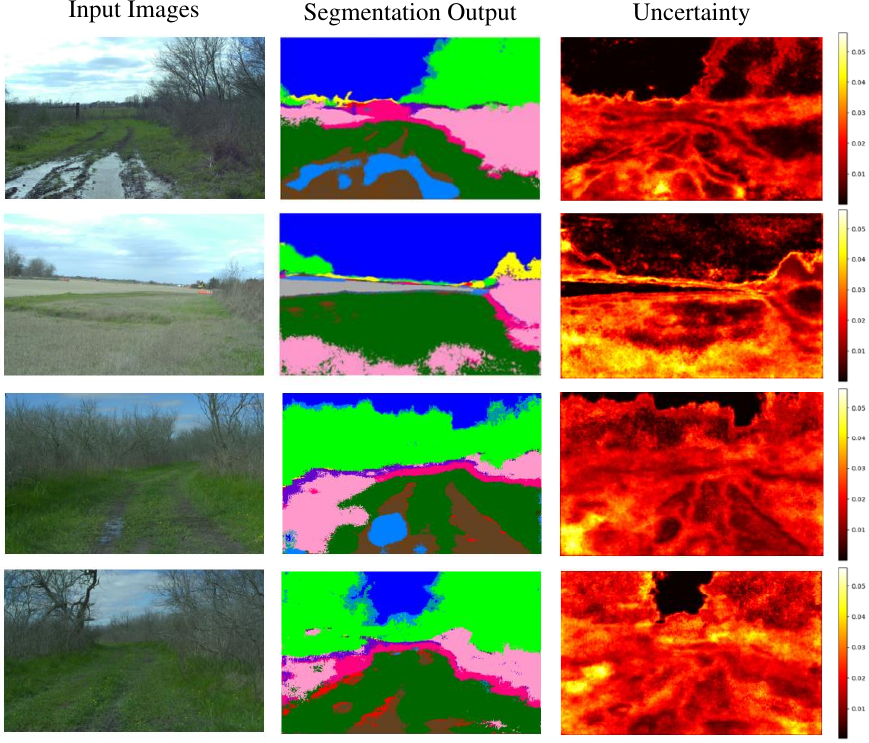}
\caption{Uncertainty estimation results on the RELLIS-3D Dataset.}
\label{fig_10}
\end{figure}

The ORD-BKI model utilizes the Bayesian SegNet \cite{ref46} network to extract semantic information from images, capturing the uncertainty of pixel-wise semantic segmentation. As shown in Fig.~\ref{fig_10}, the boundaries of road edges and vegetation with varying traversability in off-road environments are often ambiguous, prompting the choice of a network with uncertainty extraction capabilities to enhance the robustness of environmental recognition.

The image semantic segmentation results are projected onto the point clouds to obtain semantically enriched point clouds. Due to the discrete and sparse nature of LiDAR point clouds, it is essential to model the distribution of grid semantic attributes during voxelization to infer the semantic attributes of neighboring grids \cite{ref48}. The ORD-BKI model exploits Bayesian Kernel Inference to achieve semantic scene completion, approximating a Gaussian process, which reduces the time complexity from cubic to linear complexity \cite{ref47}. The Nearest Neighbor algorithm is then used to assign semantics, generating dense semantic occupancy results. The model utilizes an anisotropic kernel function for inference and analyzes the model results under historical frame point cloud inputs. The geometric passability analysis method adopted by the model is consistent with the discussions demonstrated earlier in the article.

The anisotropic kernel function is as follows:
\begin{equation}
K(x, \acute{x}) = e^{-\frac{1}{2}d_M^2},
\label{eqs28}
\end{equation}
where $d_M$ represents the Mahalanobis distance, calculated using the formula:
\begin{equation}
d_M = \sqrt{(x - \acute{x})^T S^{-1} (x - \acute{x})}.
\label{eqs29}
\end{equation}

\section{Experiments}
\label{sec:expe}
\subsection{Quantitative Analysis}
Experiments are conducted on the RELLIS-OCC dataset for the tasks of 3D traversability cost estimation, of which the results are compared to those obtained by the algorithms presented in \cite{ref51},\cite{ref52},\cite{ref53},\cite{ref48}. Due to GPU memory constraints, the MonoScene model is trained on $4 \times V100 \ GPUs$, while the other models are trained on $4 \times 3090 \ GPUs$. A consistent batch size of 4 is maintained, and each model is trained for 20 epochs.

\textbf{Evaluation Metrics:} The RELLIS-OCC dataset comprises both semantic labels and traversability cost annotations. The traversability cost estimation task is framed as a semantic occupancy prediction task, encompassing four categories: lethal, medium-cost, low-cost, and free. Consequently, traditional metrics employed in 3D semantic occupancy prediction tasks are utilized for evaluation. As regards the scene completion (SC) task, the Intersection over Union (IoU) of occupied voxels, disregarding their semantic classes, is used as the evaluation metric. Correspondingly, with respect to the semantic scene completion (SSC) task, the IoU of each semantic class and the mean IoU (mIoU) across all semantic classes are adopted as evaluation metrics.

\textbf{3D Traversability Cost Estimation:}The results of the 3D Traversability Cost Estimation task are illustrated in Table~\ref{tab:table_2}. Of all the methods, those on basis of semantic scene completion outperform the ORD-BKI method, which relies on Bayesian Kernel Inference. This emphasizes the efficiency of semantic scene completion approaches in enhancing off-road traversable area recognition. Notably, methods using only LiDAR point clouds as input, such as LMSCNet and SSCNet, exhibit inferior performance in the traversability cost estimation task compared to models that incorporate camera input. This suggests that image input is indispensable for accurate traversability cost estimation.

\begin{table*}[ht]
    \centering
    \caption{Traversability Estimate Results on the RELLIS-OCC Dataset.}

        \begin{tabular}{c|cccc|c|c}
            \toprule
            \multirow{2}*{\textbf{Method}} &\multicolumn{4}{c|}{\textbf{SSC}} &{\textbf{SSC}} &{\textbf{SC}} \\
            %\cline{2-13}
            &\textbf{Lethal} &\textbf{Medium-cost} &\textbf{Low-cost} &\textbf{Free} &\textbf{mIoU} &\textbf{IoU} \\
            \midrule	
                \textbf{MonoScene\cite{ref51}}& 12.07& 13.37& \textbf{27.30}& \textbf{5.51}& 14.59& 34.92\\		
            \textbf{SSCNet\cite{ref52}}& 11.58& 8.29& 13.95& 0.004& 8.56& 23.34\\	
                \textbf{SSCNet-full\cite{ref52}}& 11.35& 8.34& 15.60& 0.002& 8.86& 24.70\\
                \textbf{LMSCNet\cite{ref53}}& 2.87& 7.32& 7.62& 0.000& 4.45& 9.59\\
                \textbf{LMSCNet-SS\cite{ref53}}& 1.04& 5.87& 10.64& 0.000& 4.39& 10.09\\
                \textbf{ORD-BKI\cite{ref48}}& 6.29& 0.76& 0.85& 0.000& 1.98& 17.47\\
            \midrule
                \rowcolor{gray!10}\textbf{3DTTNet(Ours)}& \textbf{19.98}& \textbf{13.89}& 23.50& 5.41& \textbf{15.70}& \textbf{49.78}\\			
            \bottomrule
        \end{tabular}
    
    \label{tab:table_2}
\end{table*}

Our investigation reveals that while methods employing image-to-point cloud semantic mapping are widely used for traversable area recognition tasks \cite{ref33},\cite{ref42},\cite{ref48}, there is a lack of quantitative evaluation of their environmental perception results. Taking the ORD-BKI method as an example, a quantitative analysis of its perception performance is conducted by setting the parameter $k$ in the Bayesian Kernel Inference to 8. To assess the effect of the Bayesian Kernel Inference module, point clouds are voxelized using only semantic mapping, and evaluations are performed. The comparative results, presented in Table~\ref{tab:table_3}, demonstrate that Bayesian Kernel Inference can greatly strengthen voxel occupancy prediction. Nevertheless, since the inference relies solely on voxelized grids of sparse point clouds, all metrics of the ORD-BKI model remain substantially lower than those of the semantic scene completion-based approach.

\begin{table}[htbp] 
	\centering
        \caption{Impact of Bayesian Kernel Inference on Semantic Mapping.}
	%\setlength{\tabcolsep}{0.5mm}
	% \resizebox{0.3\textwidth}{!}
	{
		\begin{tabular}{c|c|c}
			\toprule
			\textbf{Method} & \textbf{SSC mIoU} & \textbf{SC IoU} \\
			\midrule
			\textbf{ORD-BKI} & 2.03 & 14.48 \\
			\textbf{Semantic-Projection} & 0.25 & 1.5 \\
			\bottomrule
		\end{tabular}
	}
	\label{tab:table_3}
\end{table}

Notably, 3DTTNet achieves state-of-the-art results in the lethal and medium-cost categories, with IoU scores of 19.98 and 13.89, respectively. Specifically, in the lethal traversability category, 3DTTNet attains an IoU score of 19.98, which is 7.91 IoU higher than that of the second-best model, MonoScene. Furthermore, in the free and low-cost categories, 3DTTNet delivers competitive performance. The overall SSC mIoU and SC IoU scores are 15.70 and 49.78, respectively. These results represent significant improvements of 42\% in SC IoU compared to other methods, indicating the excellent performance of 3DTTNet in occupancy prediction. Importantly, the enhanced geometric perception capabilities of 3DTTNet are critical for identifying complex negative and overhanging obstacles in off-road environments, where accurate 3D traversability estimation is crucial for safe navigation.

\subsection{Qualitative Analysis}

\begin{figure*}[htbp]       
    % 1 row of images with titles on the left
    \begin{minipage}{0.16\textwidth}
        \centering
        {Camera View}
    \end{minipage}%
    \begin{minipage}{0.16\textwidth}
        \centering
        \includegraphics[width=0.95\textwidth]{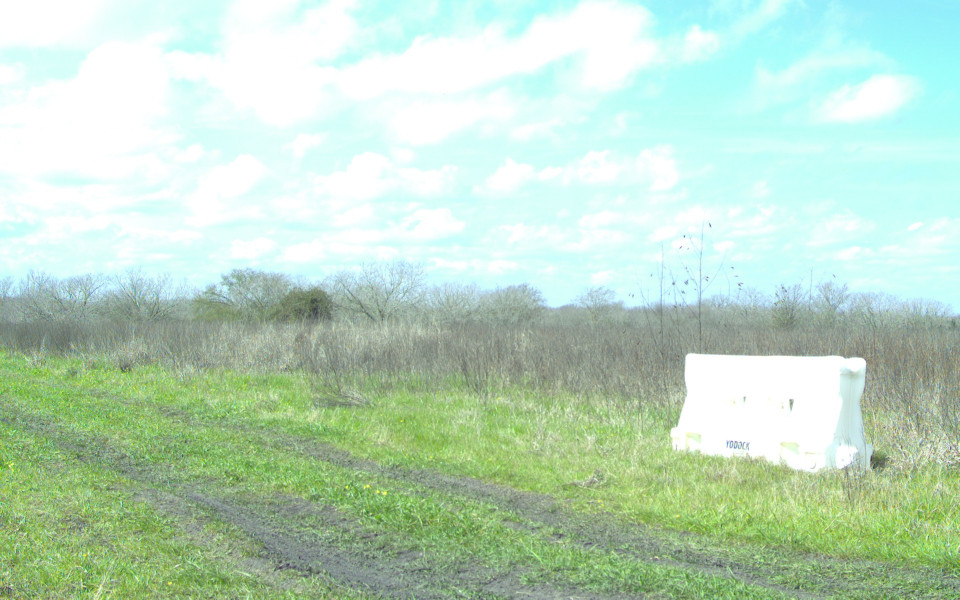}
    \end{minipage}%
    \begin{minipage}{0.16\textwidth}
        \centering
        \includegraphics[width=0.95\textwidth]{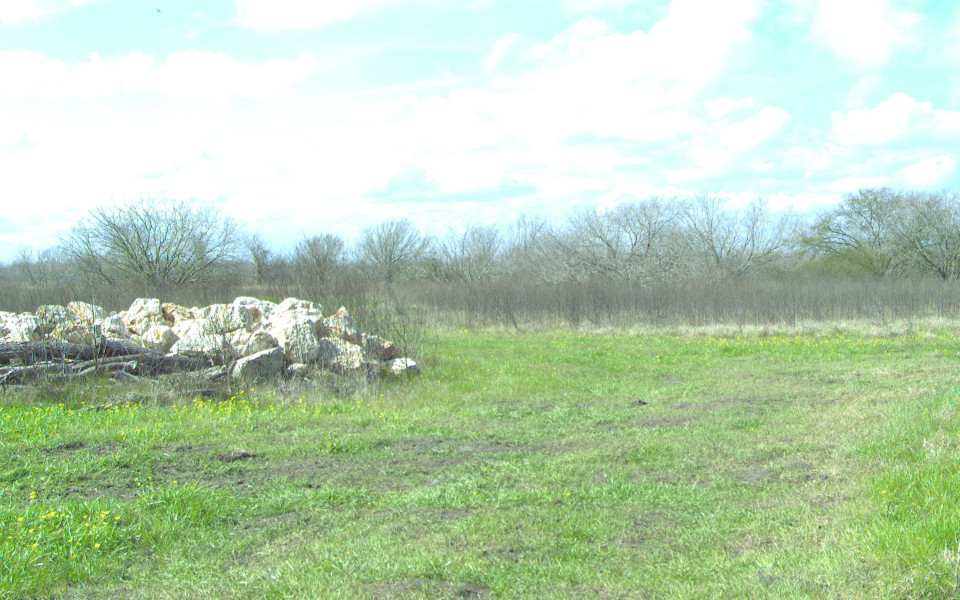}
    \end{minipage}%
    \begin{minipage}{0.16\textwidth}
        \centering
        \includegraphics[width=0.95\textwidth]{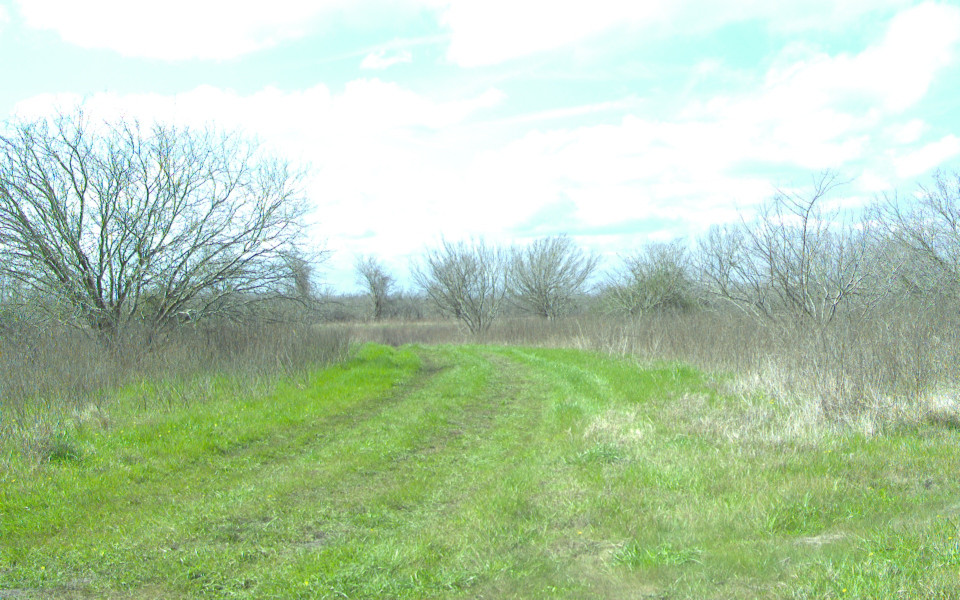}
    \end{minipage}%
    \begin{minipage}{0.16\textwidth}
        \centering
        \includegraphics[width=0.95\textwidth]{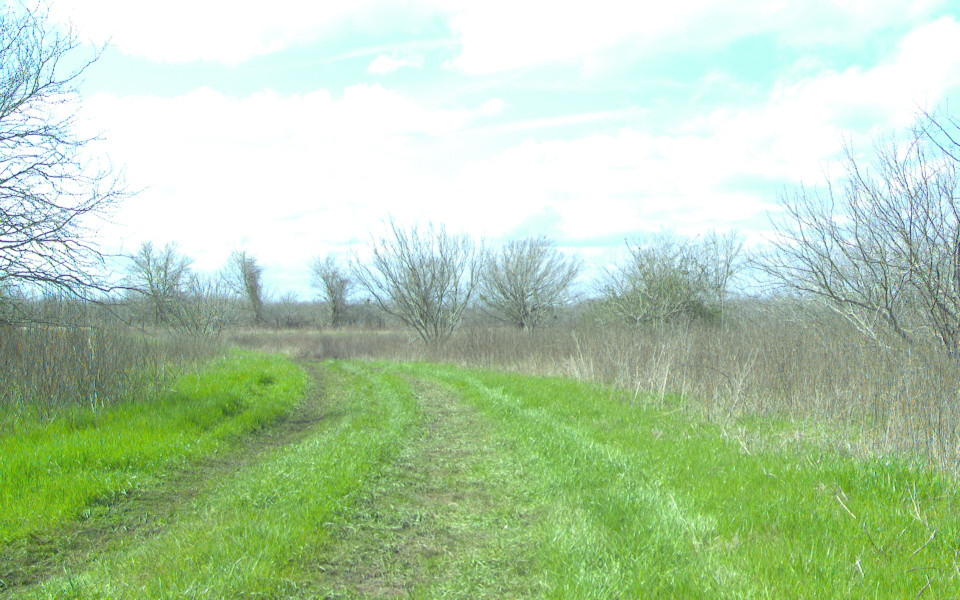}
    \end{minipage}%
    \begin{minipage}{0.16\textwidth}
        \centering
        \includegraphics[width=0.95\textwidth]{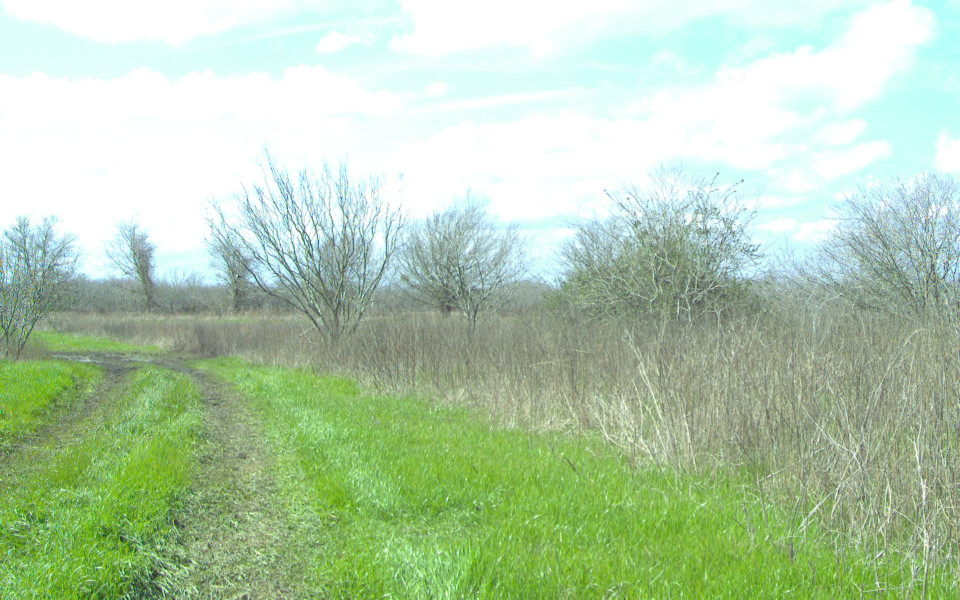}
    \end{minipage}
    \vskip\baselineskip

    % 2 row of images with titles on the left
    \begin{minipage}{0.16\textwidth}
        \centering
        {Semantic Annotations}
    \end{minipage}%
    \begin{minipage}{0.16\textwidth}
        \centering
        \includegraphics[width=0.95\textwidth]{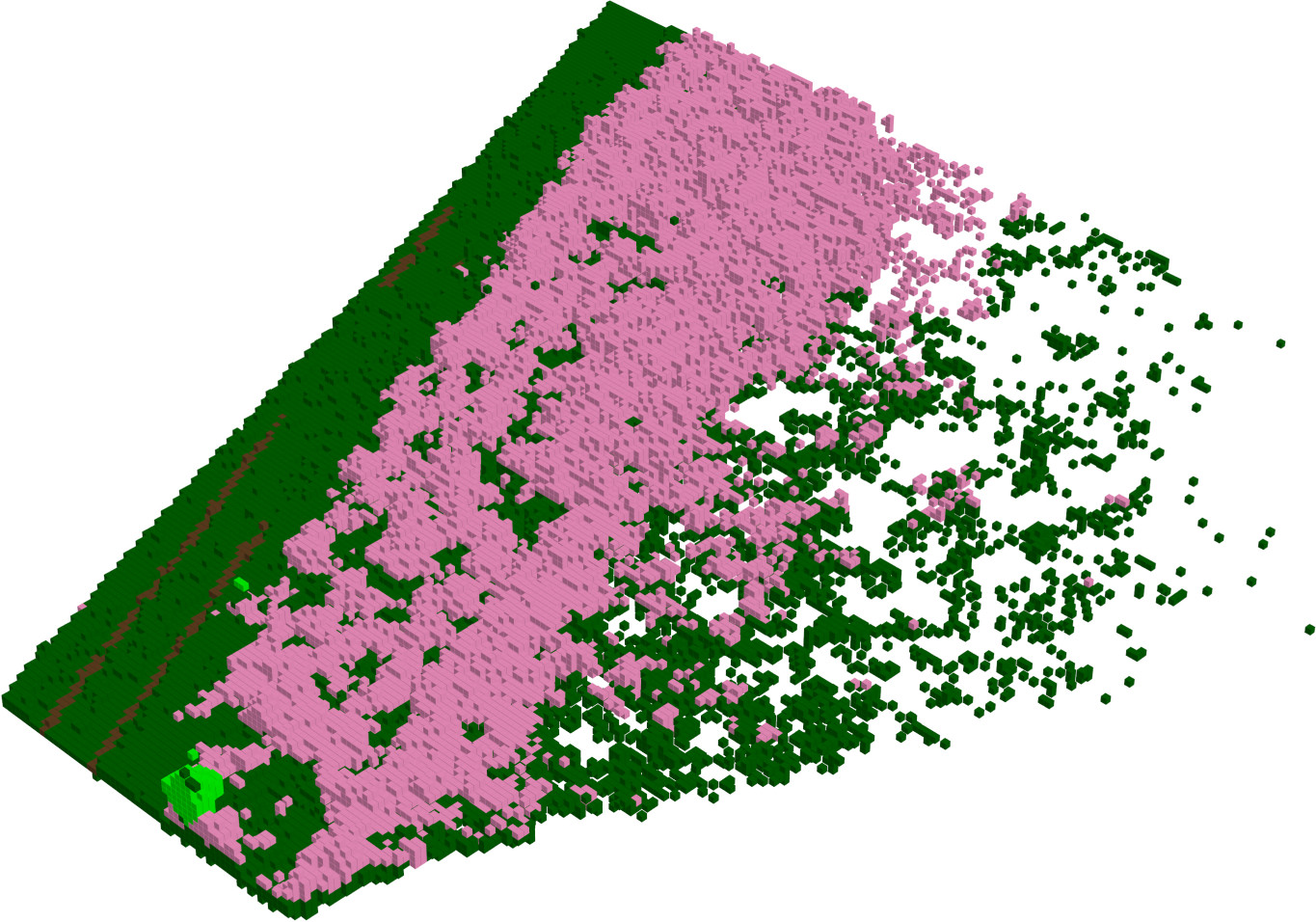}
    \end{minipage}%
    \begin{minipage}{0.16\textwidth}
        \centering
        \includegraphics[width=0.95\textwidth]{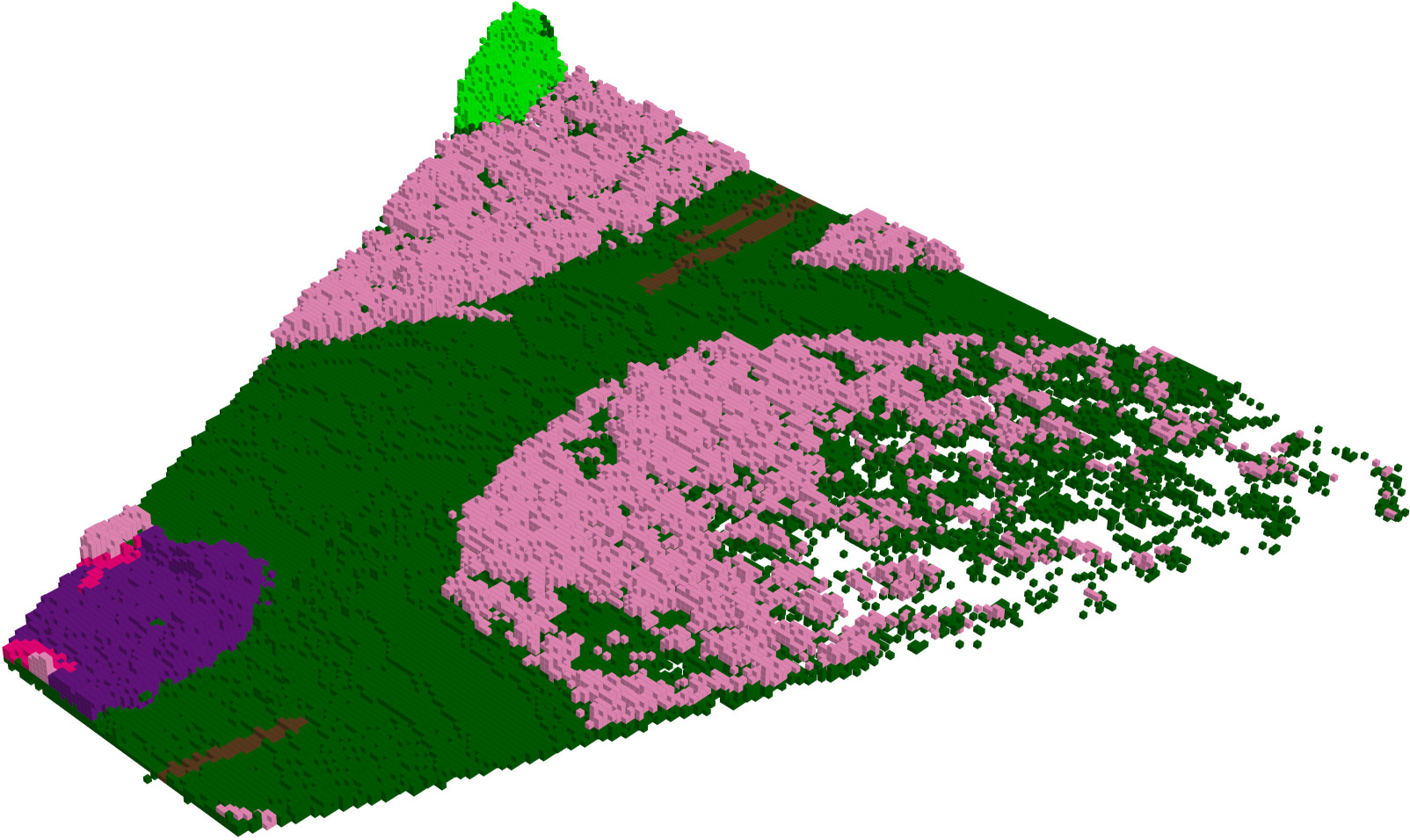}
    \end{minipage}%
    \begin{minipage}{0.16\textwidth}
        \centering
        \includegraphics[width=0.95\textwidth]{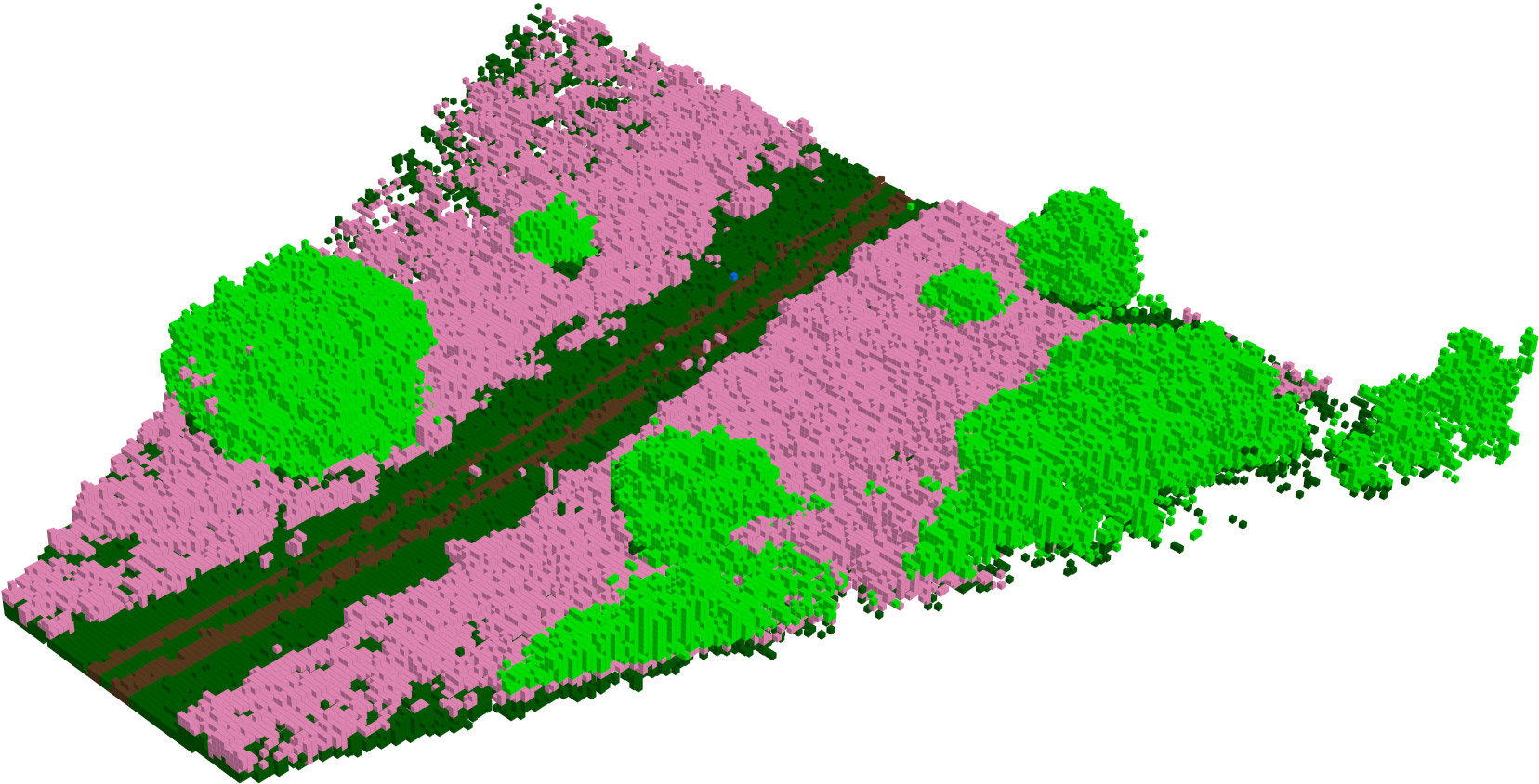}
    \end{minipage}%
    \begin{minipage}{0.16\textwidth}
        \centering
        \includegraphics[width=0.95\textwidth]{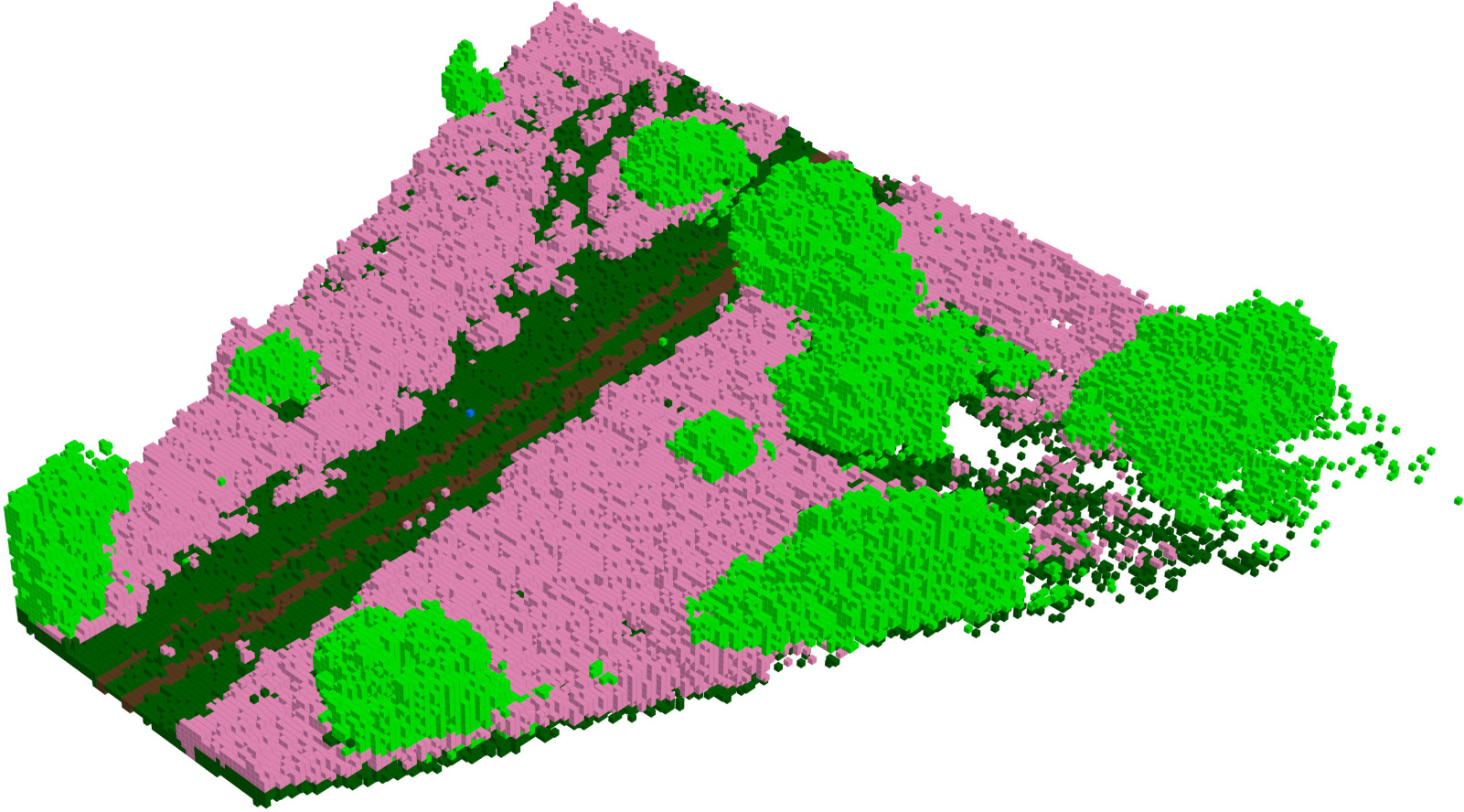}
    \end{minipage}%
    \begin{minipage}{0.16\textwidth}
        \centering
        \includegraphics[width=0.95\textwidth]{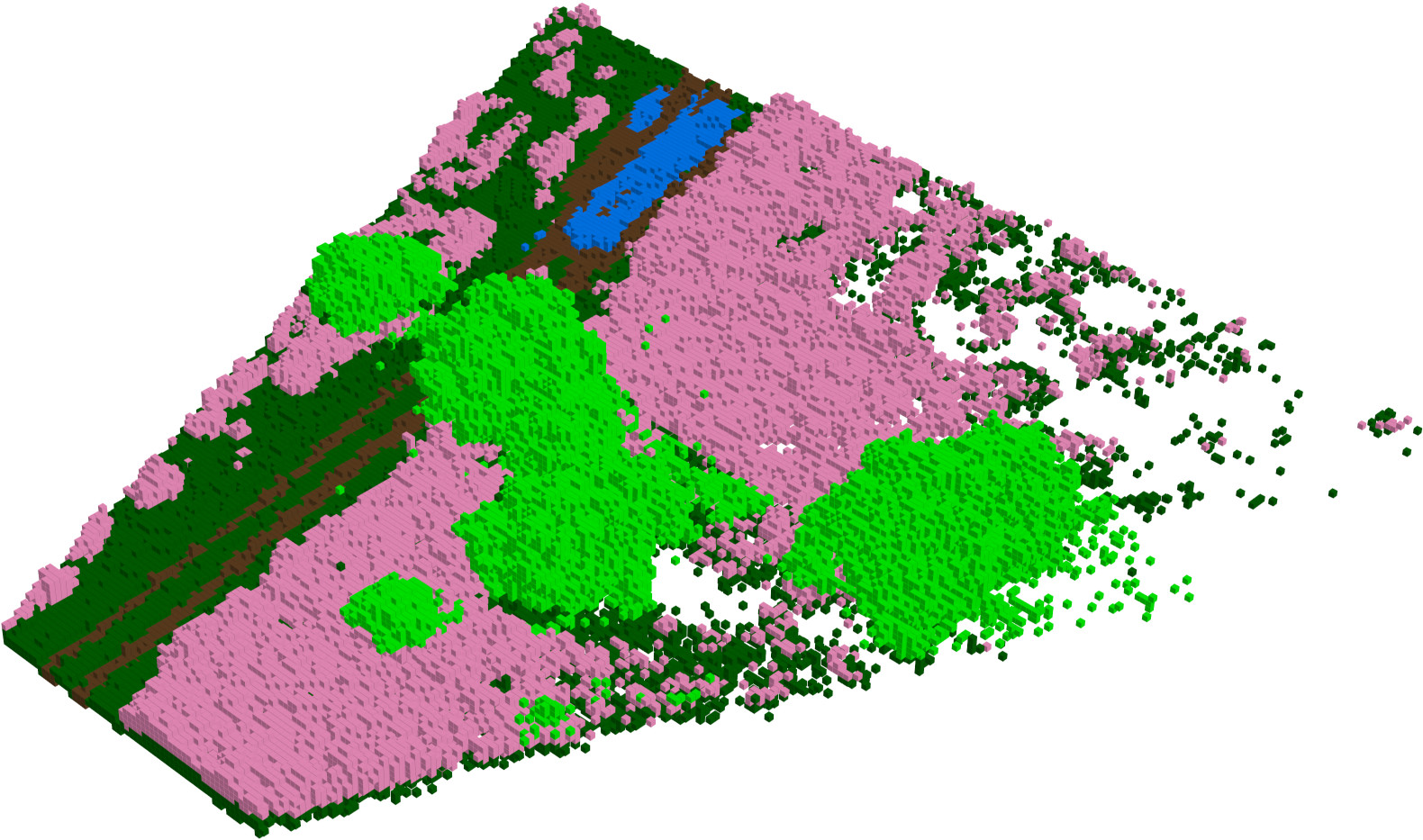}
    \end{minipage}
    \vskip\baselineskip

     % 2 row of images with titles on the left
    \begin{minipage}{0.16\textwidth}
        \centering
        {Traversability Annotations}
    \end{minipage}%
    \begin{minipage}{0.16\textwidth}
        \centering
        \includegraphics[width=0.95\textwidth]{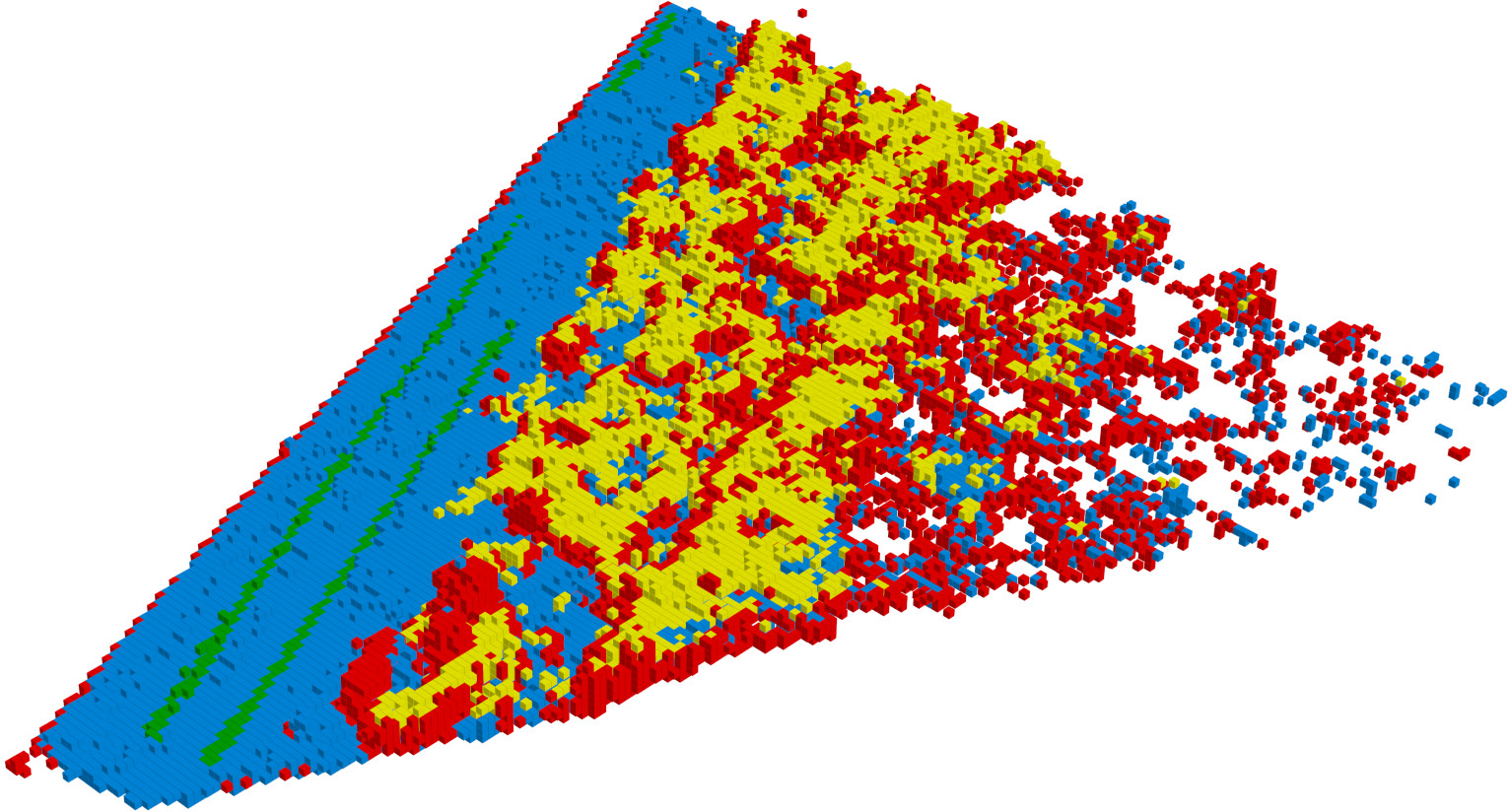}
    \end{minipage}%
    \begin{minipage}{0.16\textwidth}
        \centering
        \includegraphics[width=0.95\textwidth]{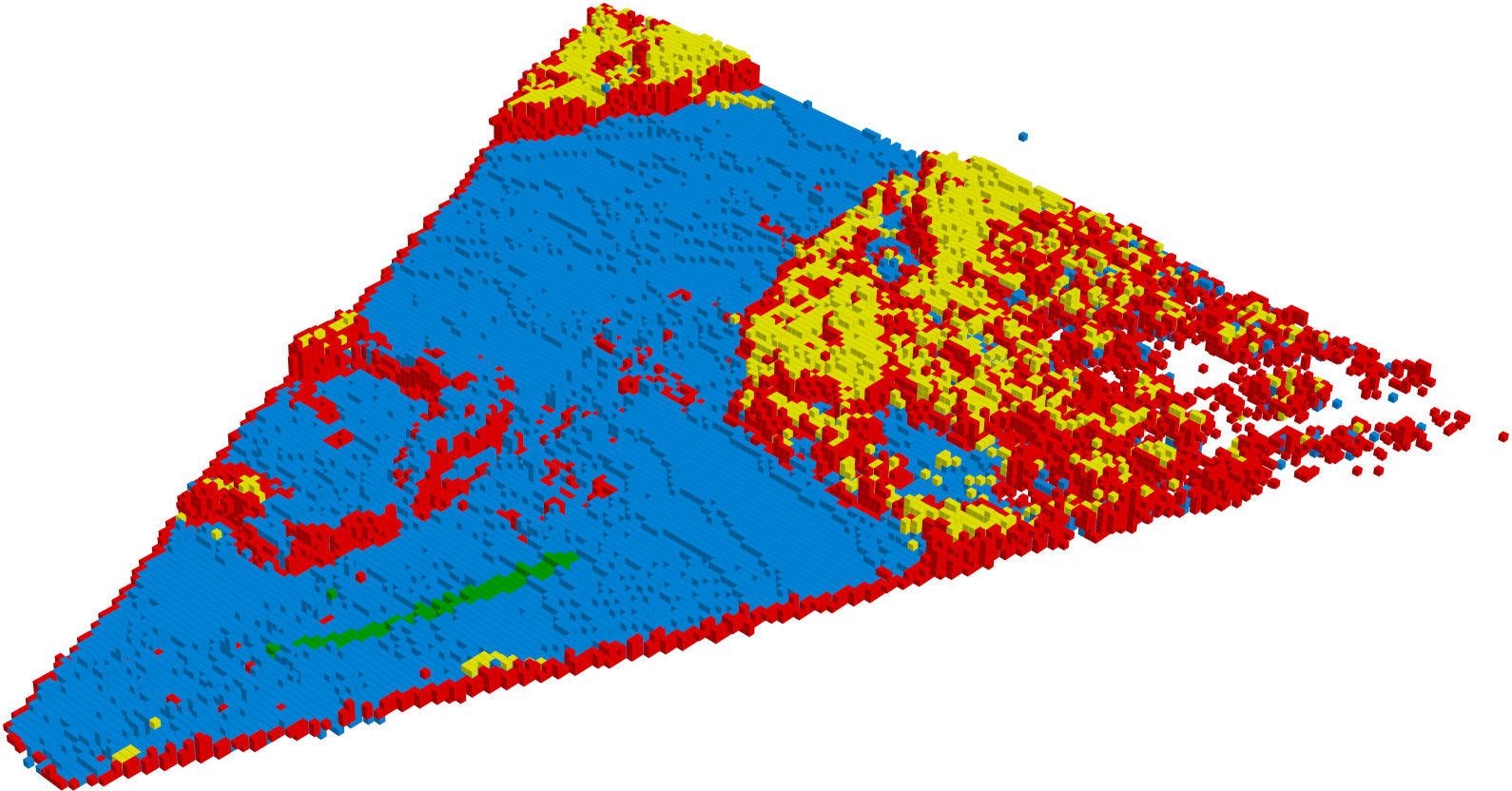}
    \end{minipage}%
    \begin{minipage}{0.16\textwidth}
        \centering
        \includegraphics[width=0.95\textwidth]{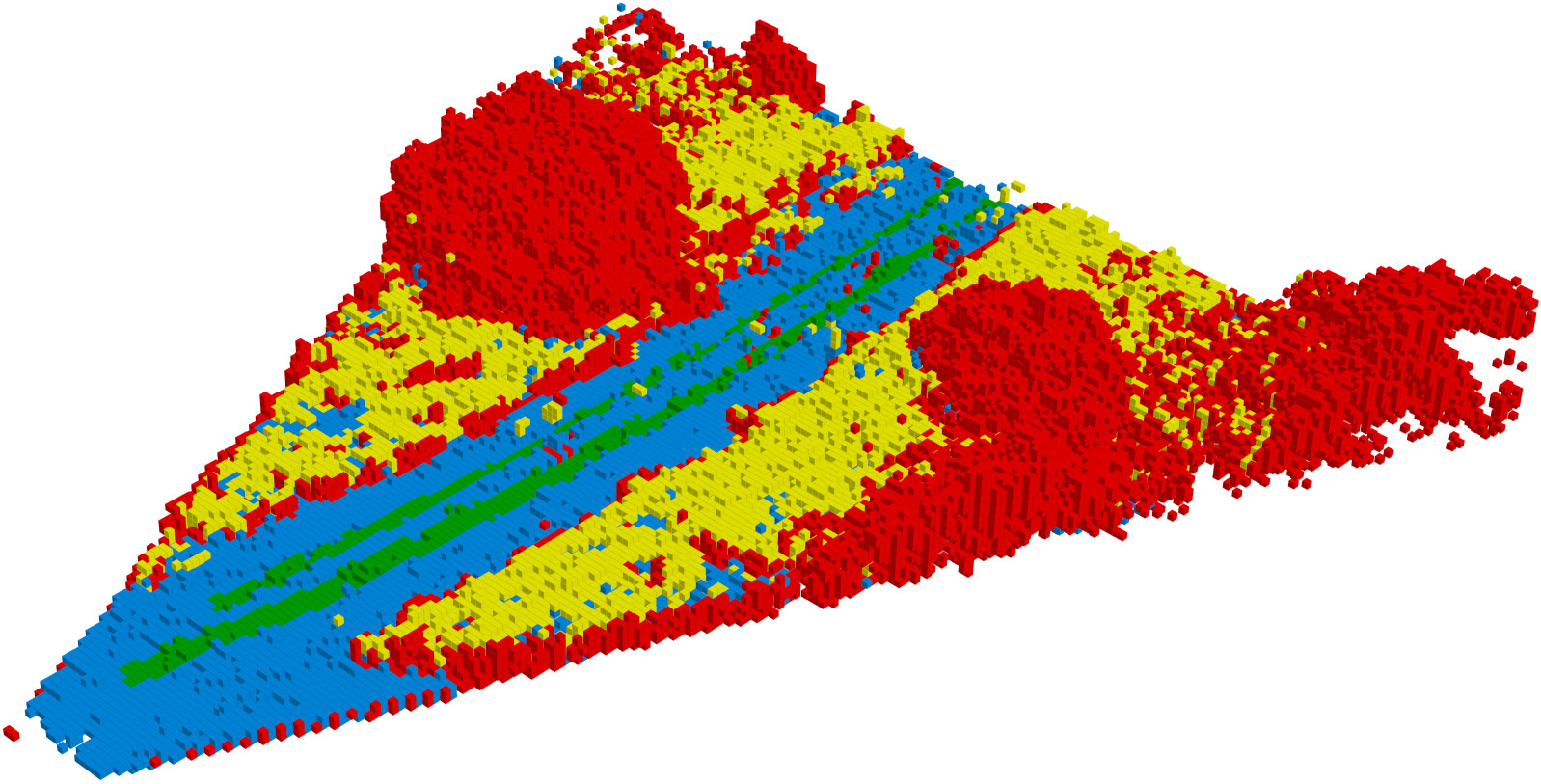}
    \end{minipage}%
    \begin{minipage}{0.16\textwidth}
        \centering
        \includegraphics[width=0.95\textwidth]{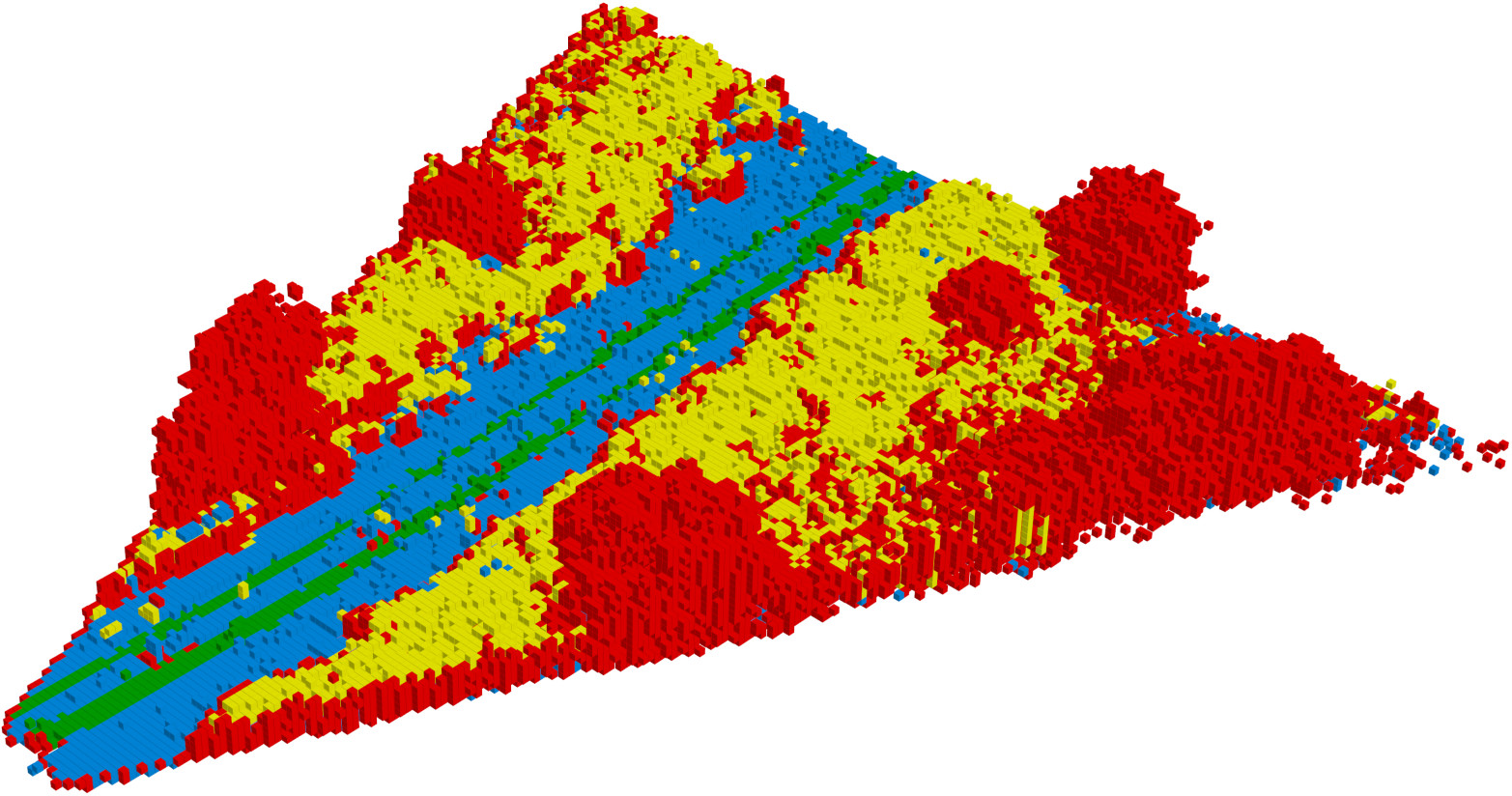}
    \end{minipage}%
    \begin{minipage}{0.16\textwidth}
        \centering
        \includegraphics[width=0.95\textwidth]{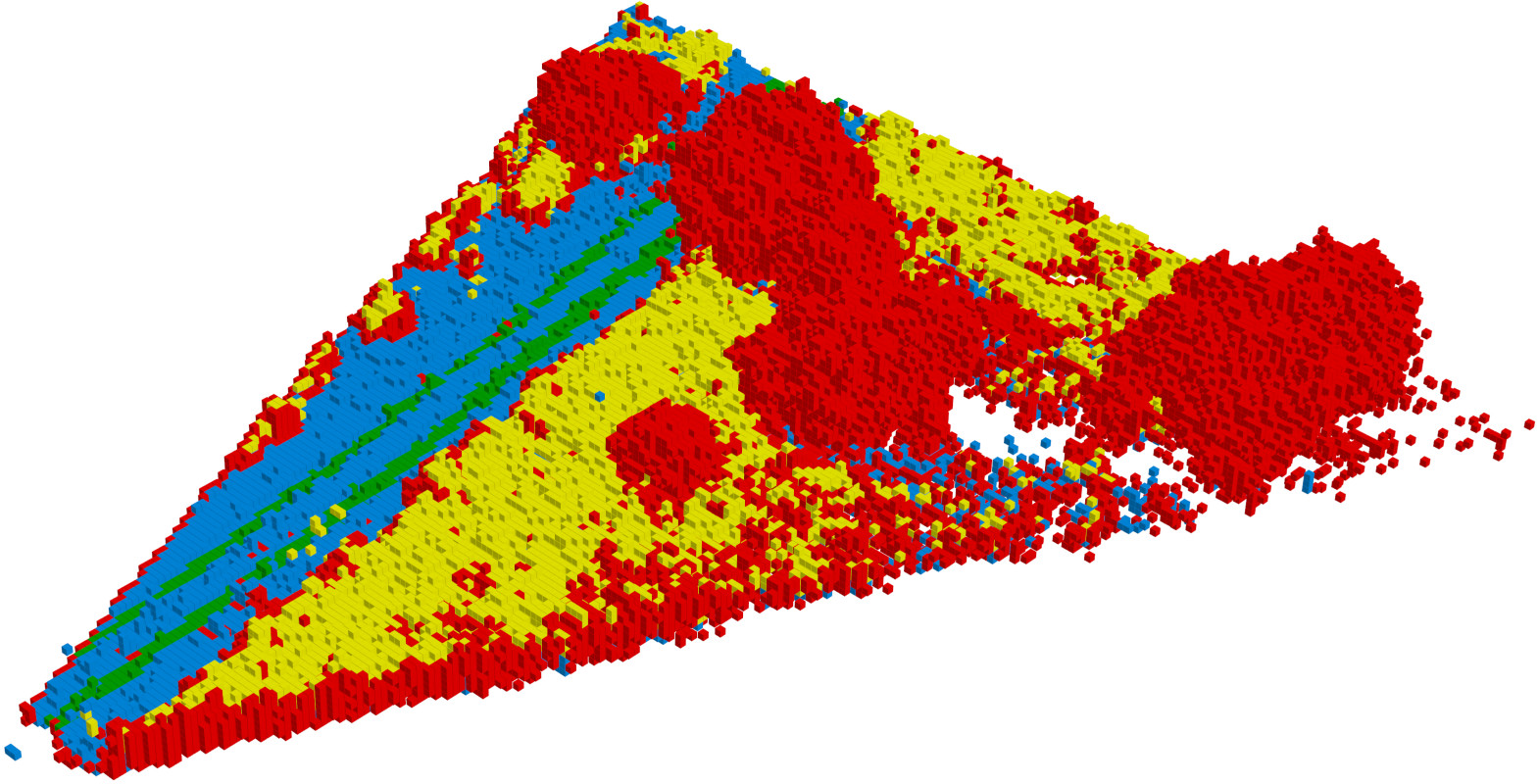}
    \end{minipage}
    \vskip\baselineskip

     % 3 row of images with titles on the left
    \begin{minipage}{0.16\textwidth}
        \centering
        {3DTTNet(Ours)}
    \end{minipage}%
    \begin{minipage}{0.16\textwidth}
        \centering
        \includegraphics[width=0.95\textwidth]{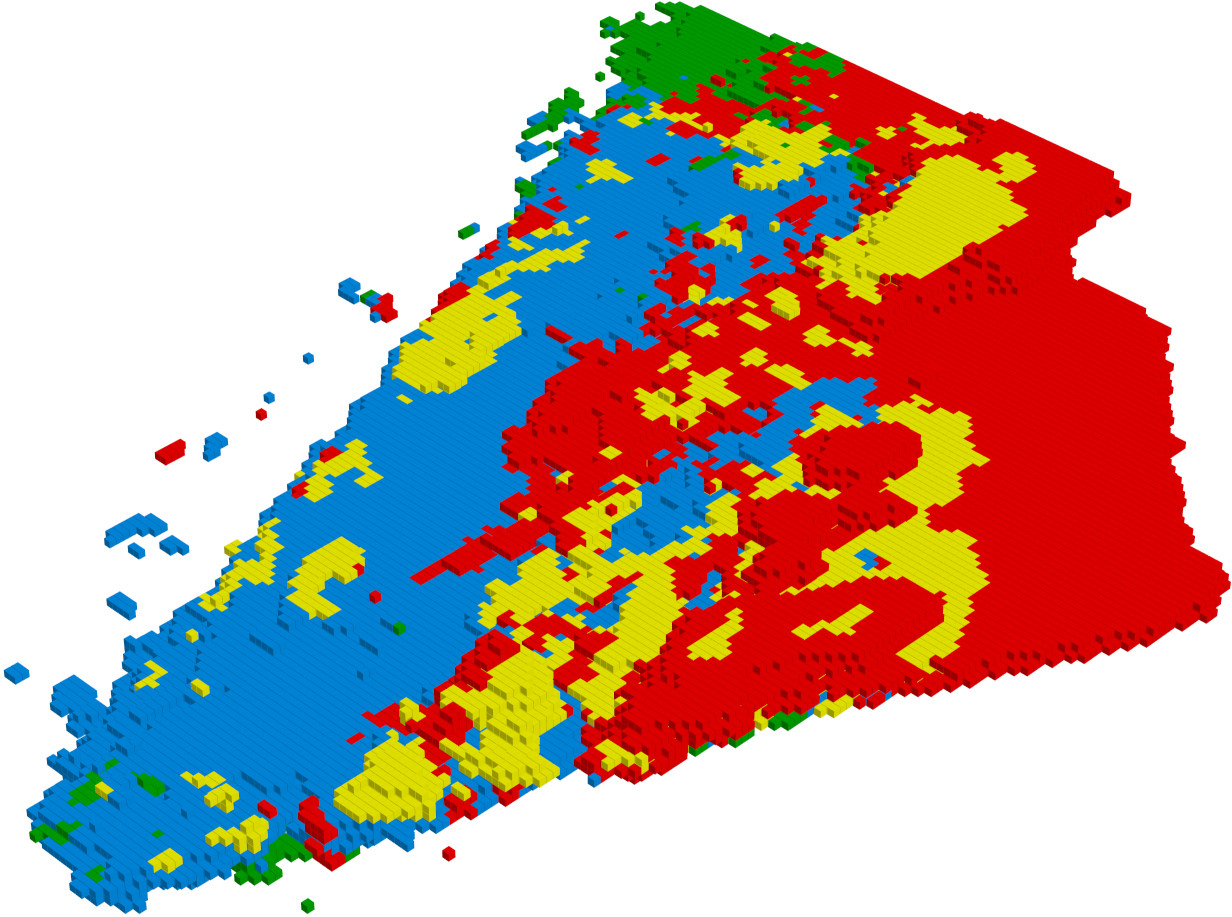}
    \end{minipage}%
    \begin{minipage}{0.16\textwidth}
        \centering
        \includegraphics[width=0.95\textwidth]{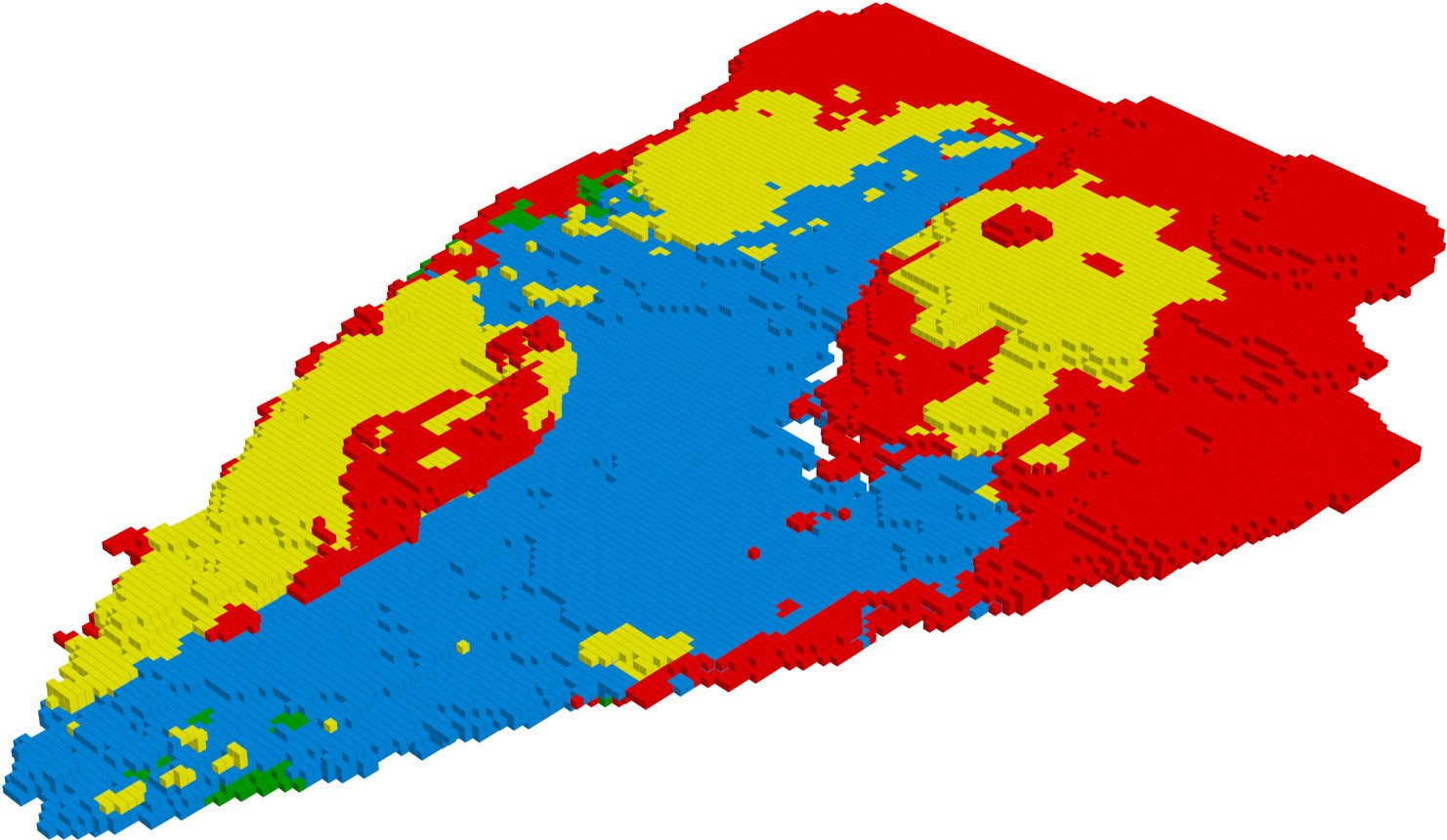}
    \end{minipage}%
    \begin{minipage}{0.16\textwidth}
        \centering
        \includegraphics[width=0.95\textwidth]{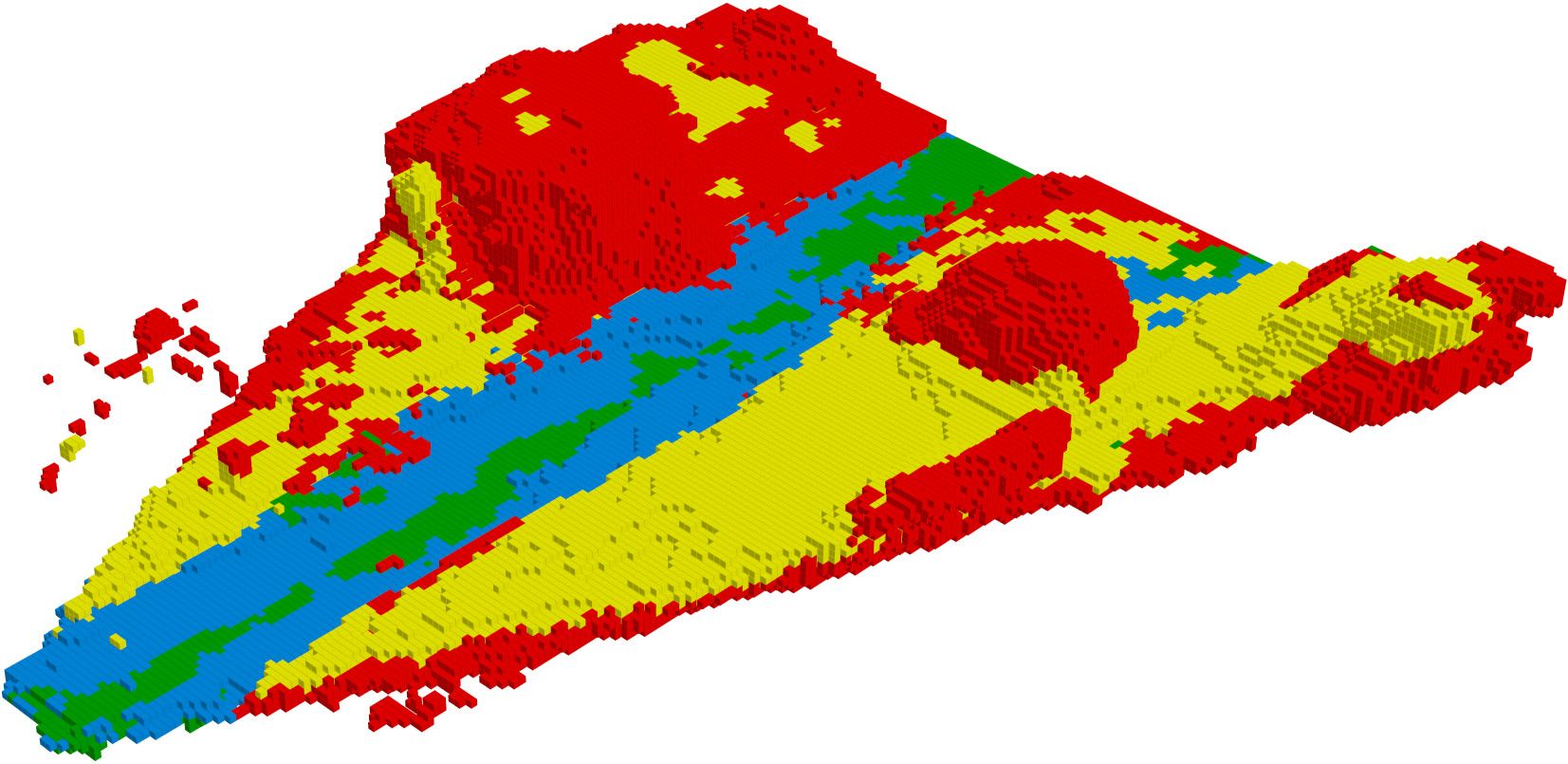}
    \end{minipage}%
    \begin{minipage}{0.16\textwidth}
        \centering
        \includegraphics[width=0.95\textwidth]{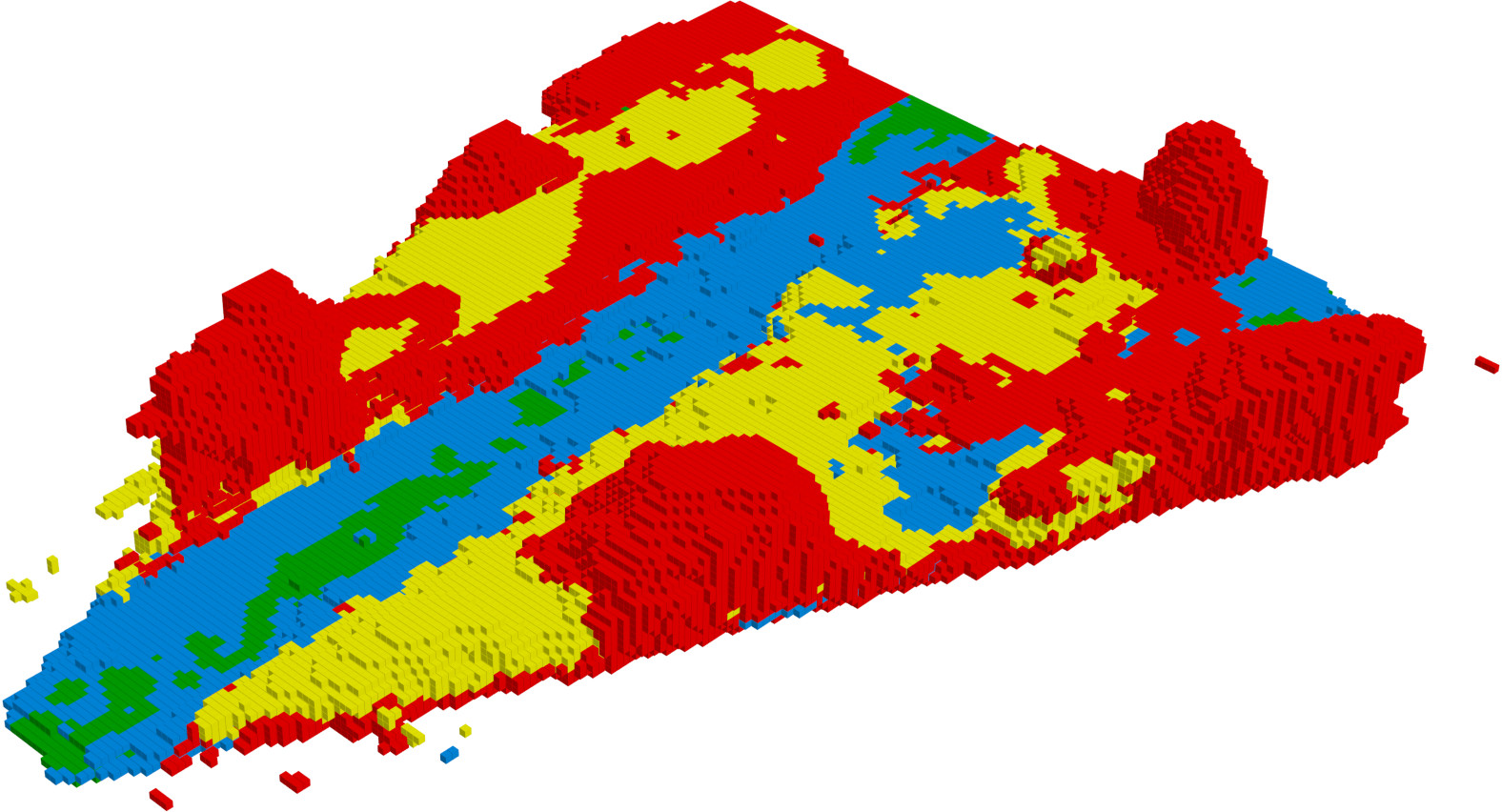}
    \end{minipage}%
    \begin{minipage}{0.16\textwidth}
        \centering
        \includegraphics[width=0.95\textwidth]{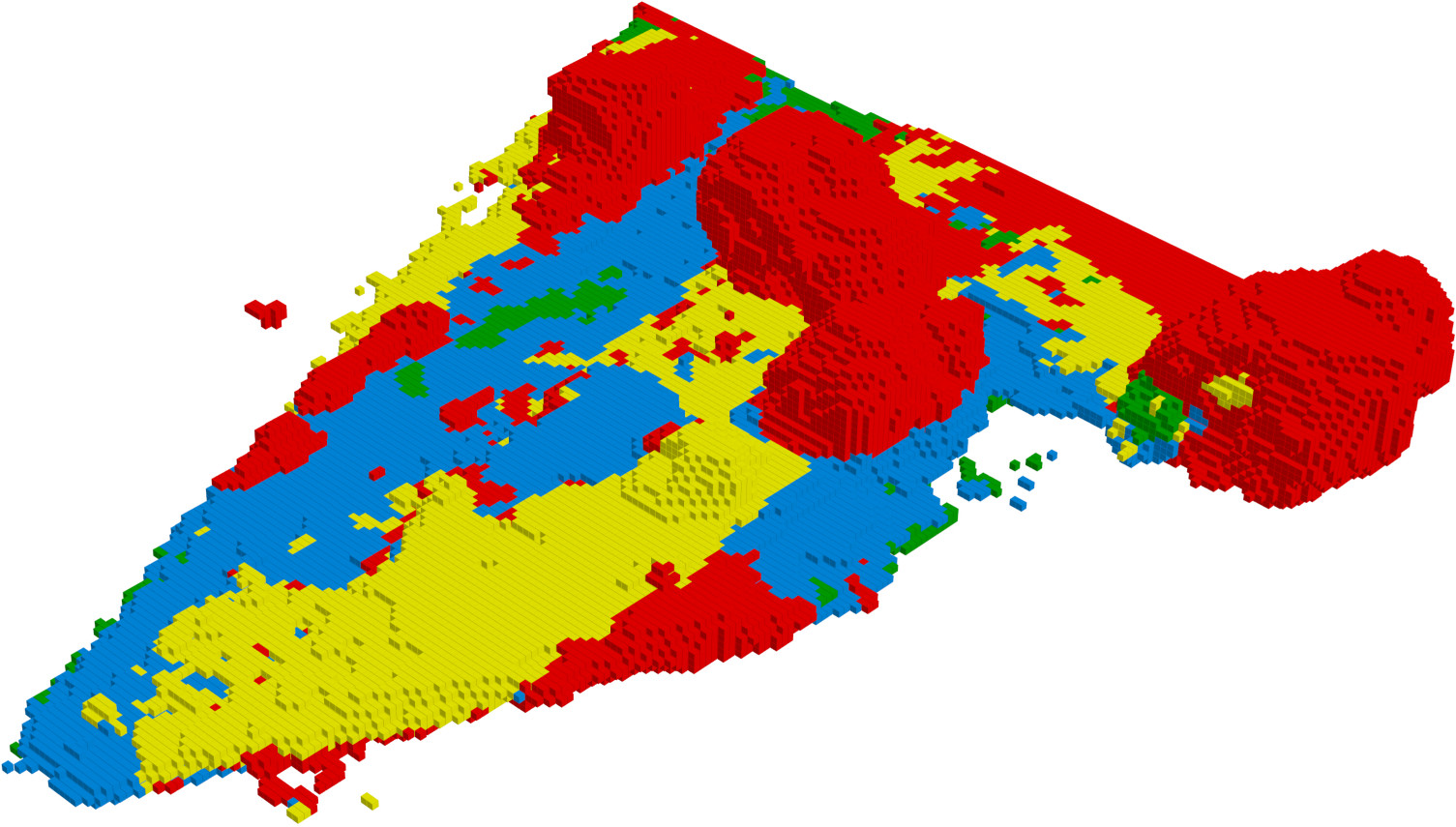}
    \end{minipage}
    \vskip\baselineskip

     % 4 row of images with titles on the left
    \begin{minipage}{0.16\textwidth}
        \centering
        {MonoScene}
    \end{minipage}%
    \begin{minipage}{0.16\textwidth}
        \centering
        \includegraphics[width=0.95\textwidth]{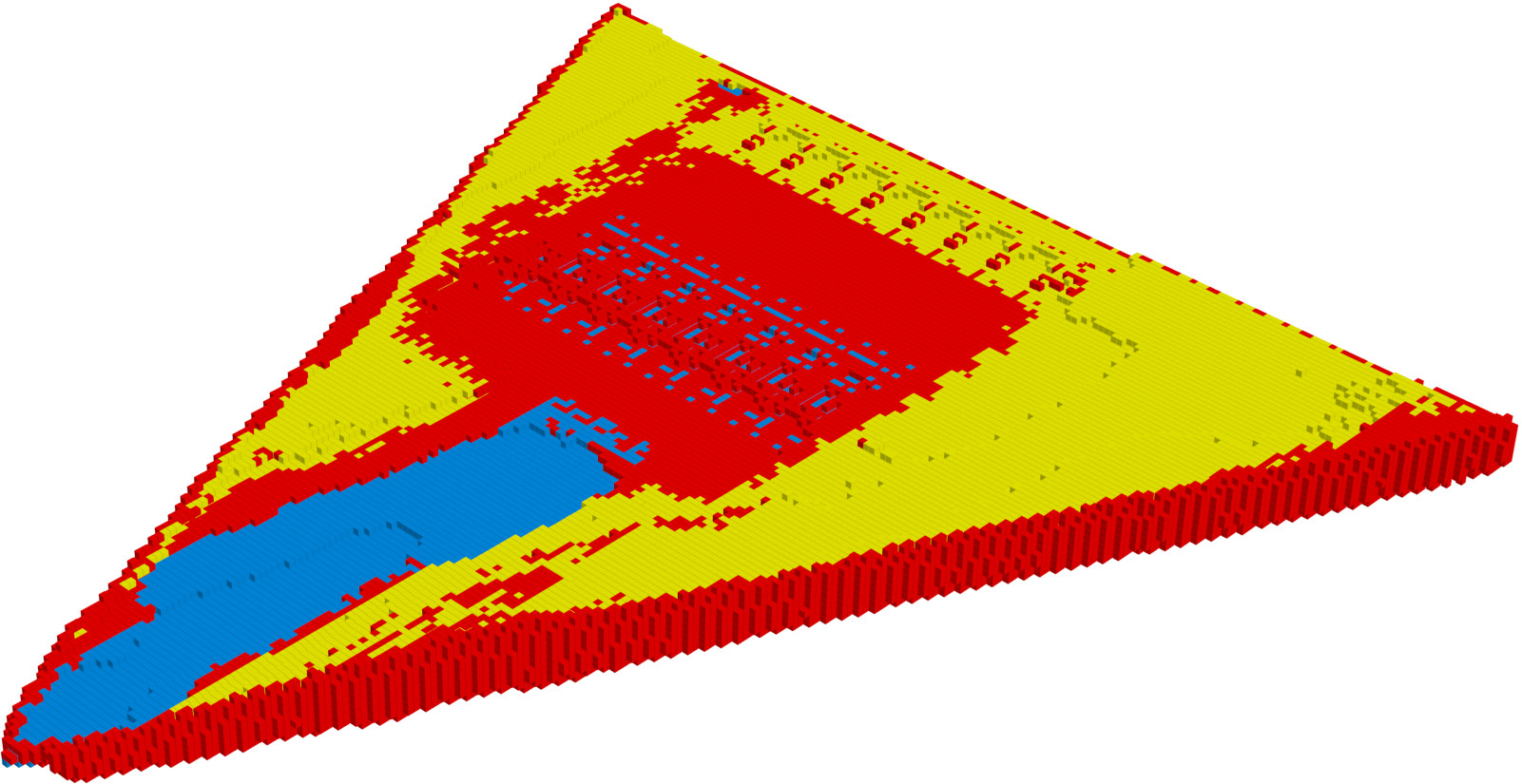}
    \end{minipage}%
    \begin{minipage}{0.16\textwidth}
        \centering
        \includegraphics[width=0.95\textwidth]{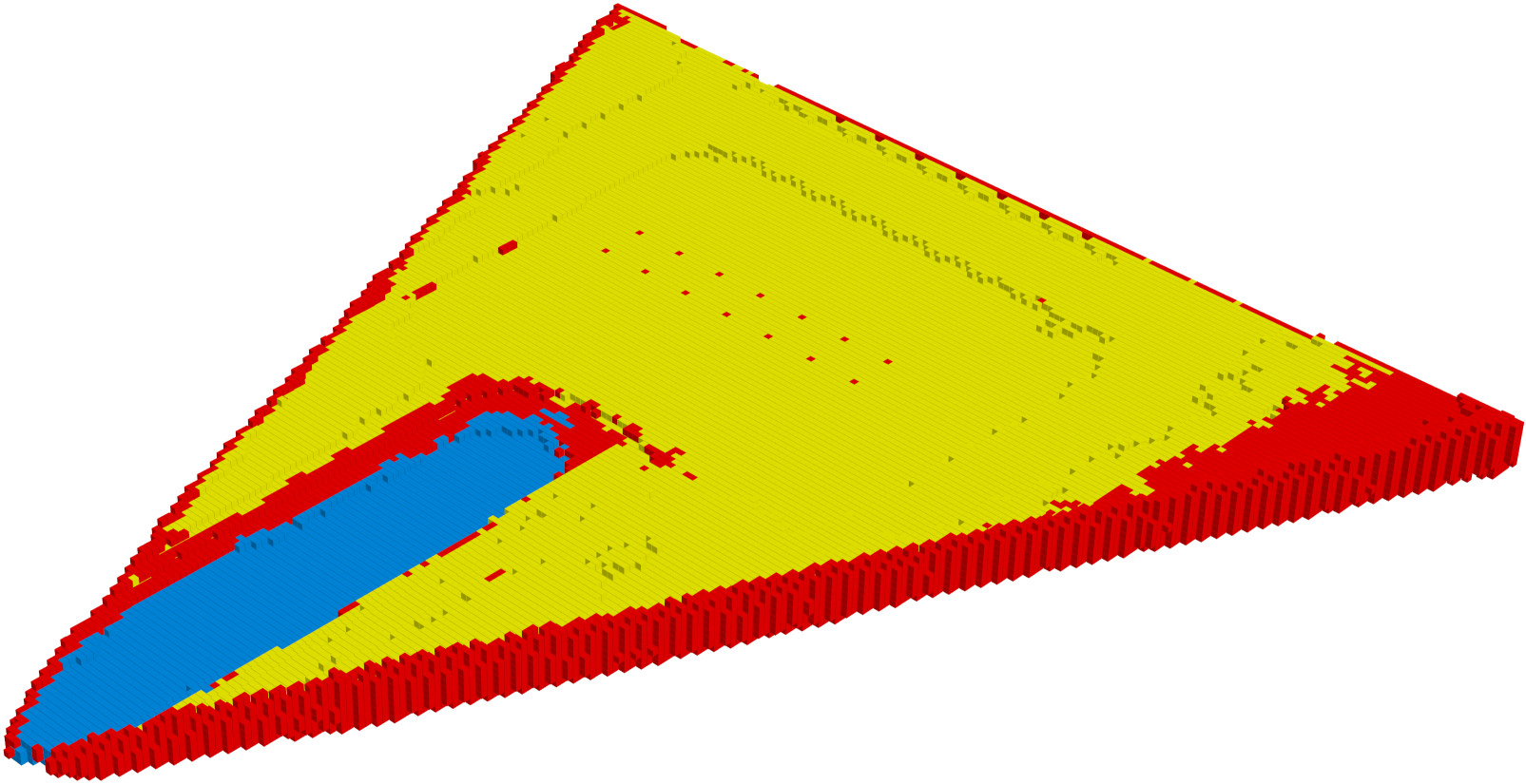}
    \end{minipage}%
    \begin{minipage}{0.16\textwidth}
        \centering
        \includegraphics[width=0.95\textwidth]{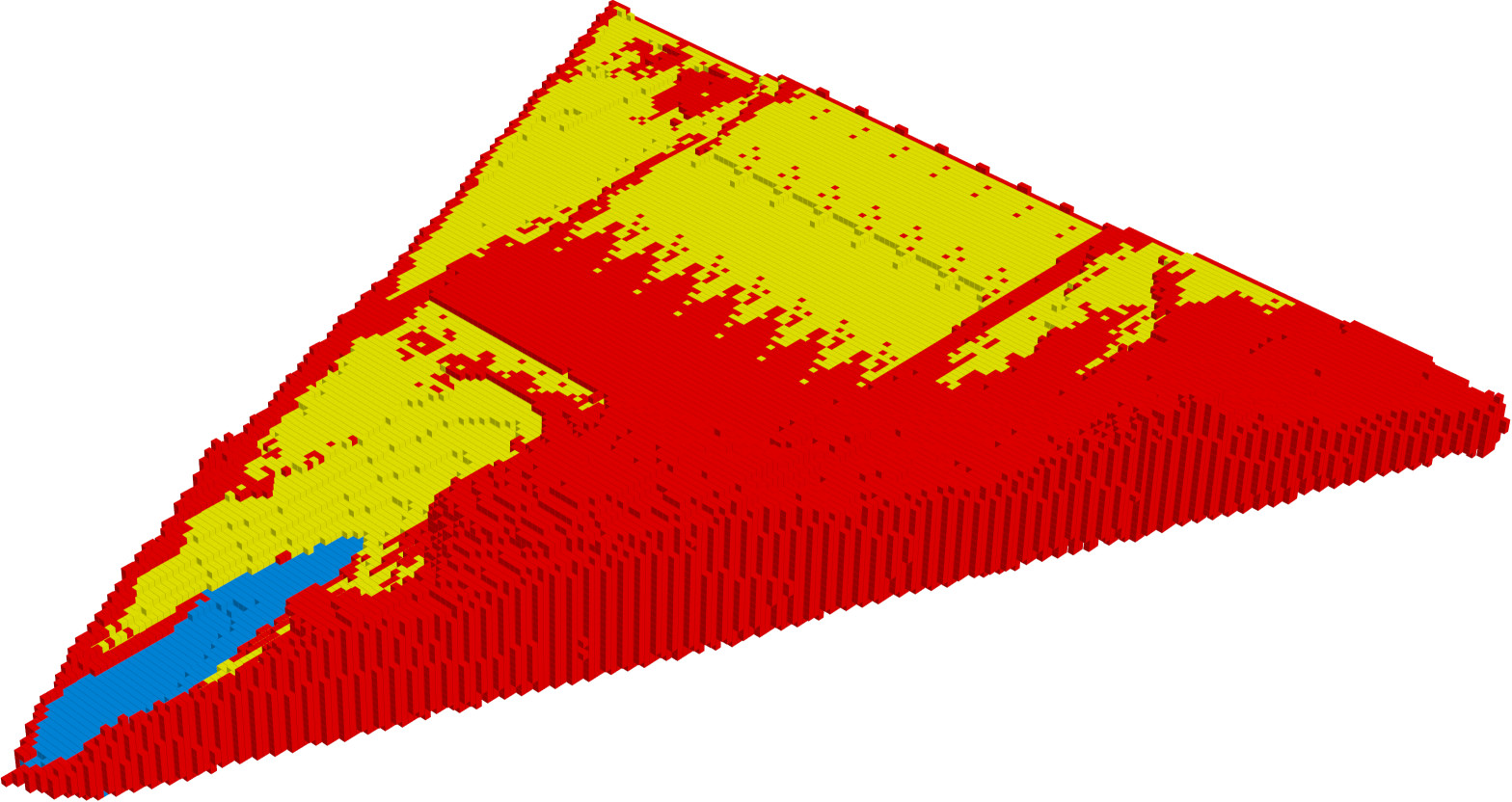}
    \end{minipage}%
    \begin{minipage}{0.16\textwidth}
        \centering
        \includegraphics[width=0.95\textwidth]{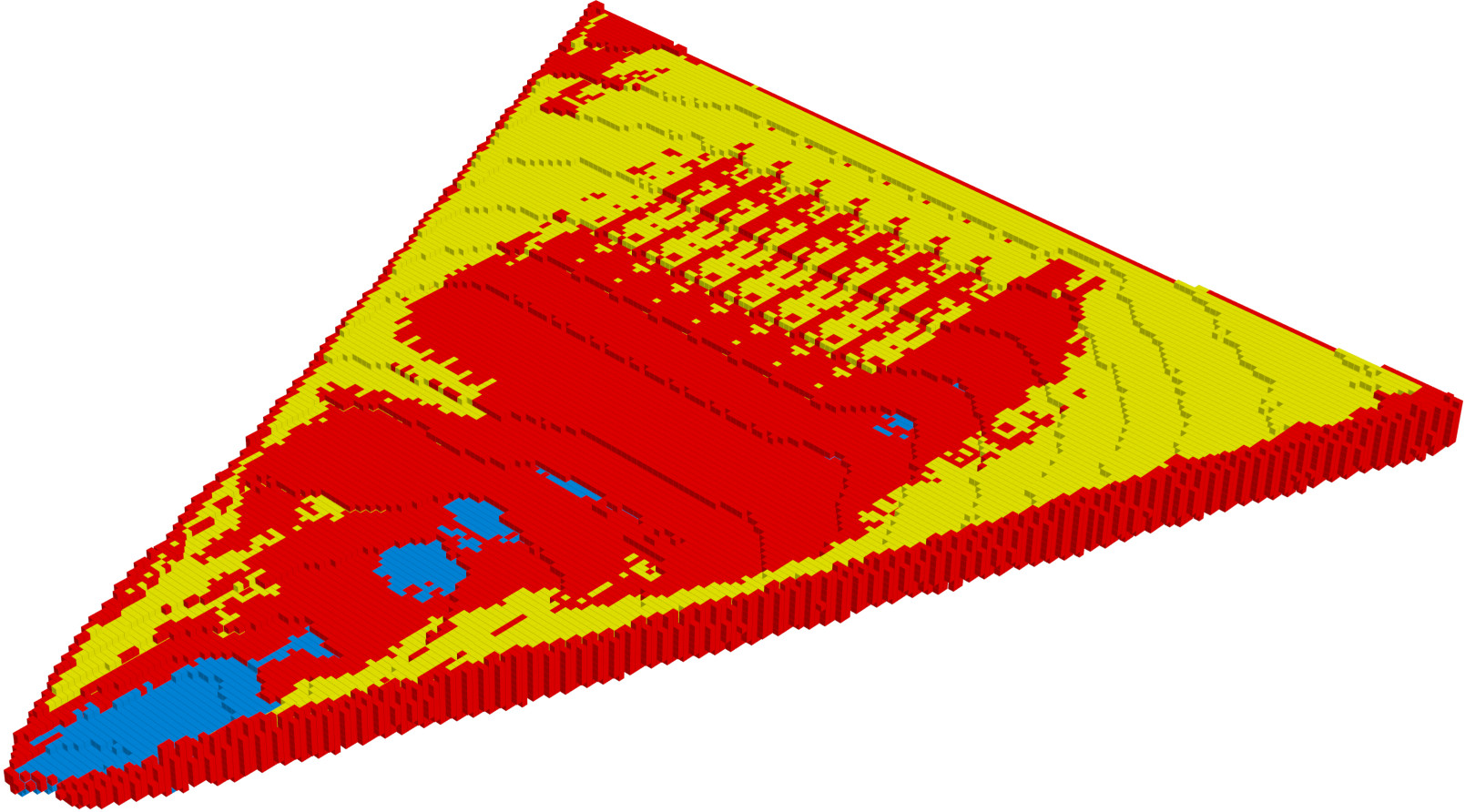}
    \end{minipage}%
    \begin{minipage}{0.16\textwidth}
        \centering
        \includegraphics[width=0.95\textwidth]{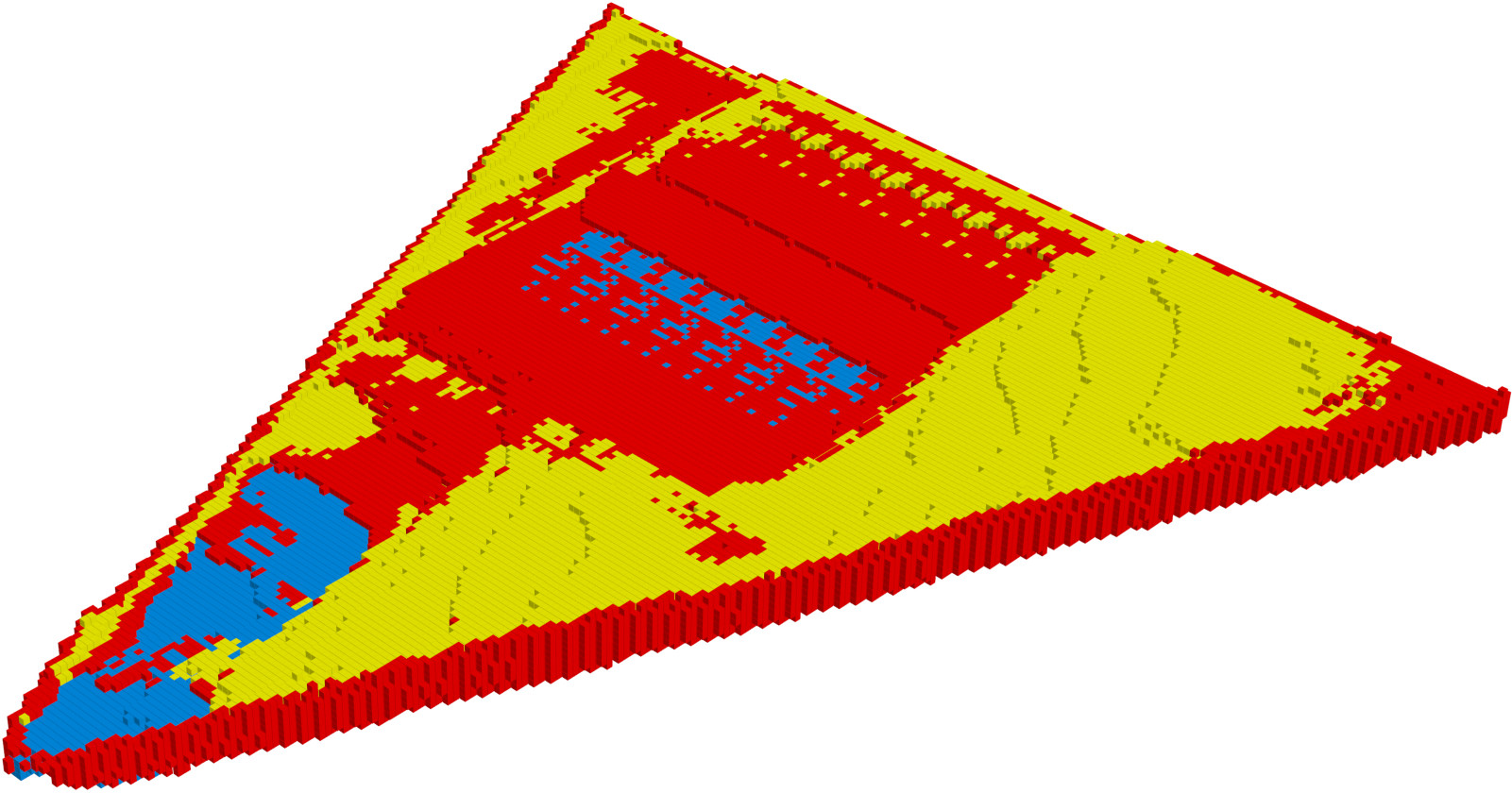}
    \end{minipage}
    \vskip\baselineskip

     % 5 row of images with titles on the left
    \begin{minipage}{0.16\textwidth}
        \centering
        {SSCNet}
    \end{minipage}%
    \begin{minipage}{0.16\textwidth}
        \centering
        \includegraphics[width=0.95\textwidth]{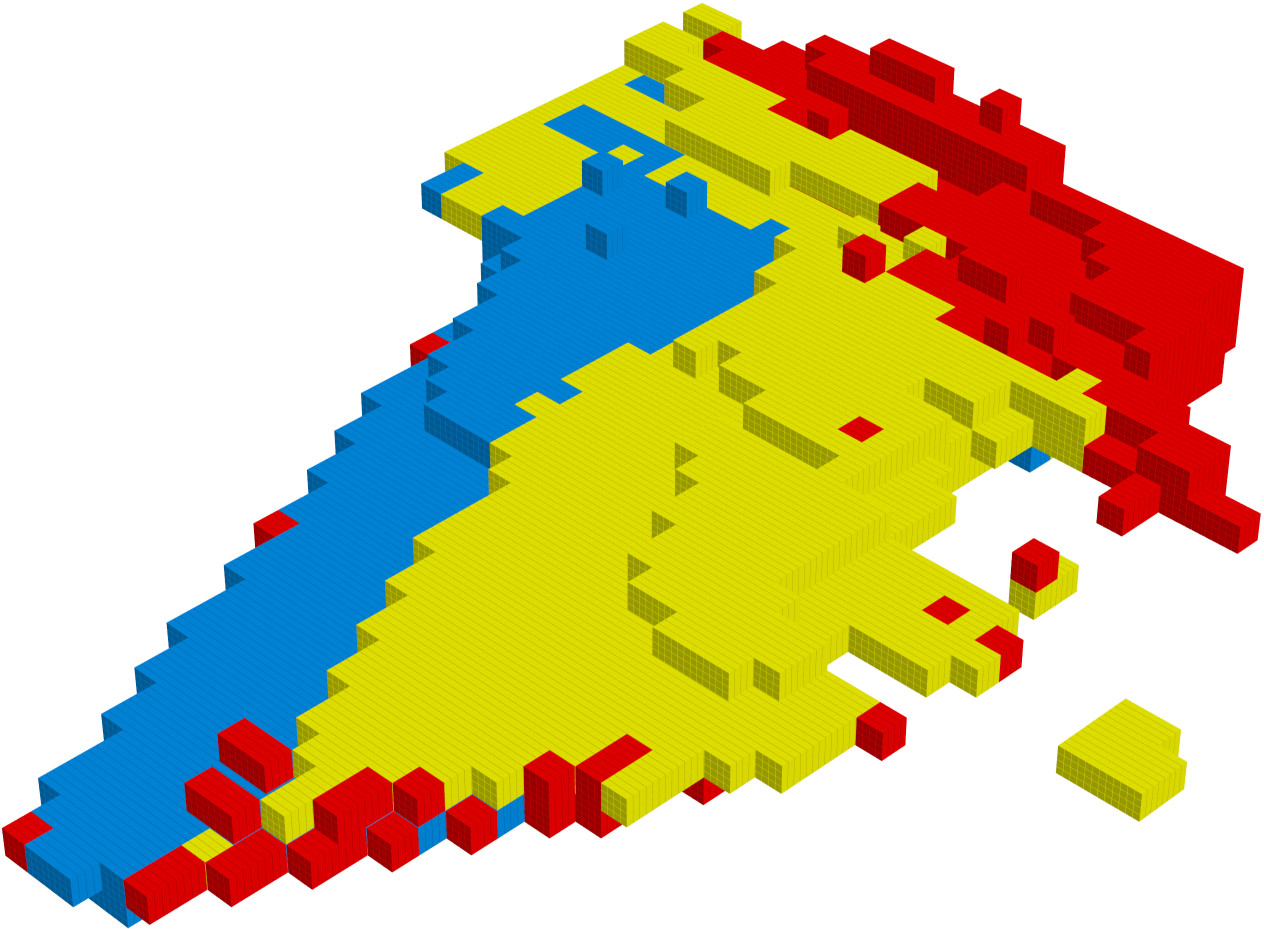}
    \end{minipage}%
    \begin{minipage}{0.16\textwidth}
        \centering
        \includegraphics[width=0.95\textwidth]{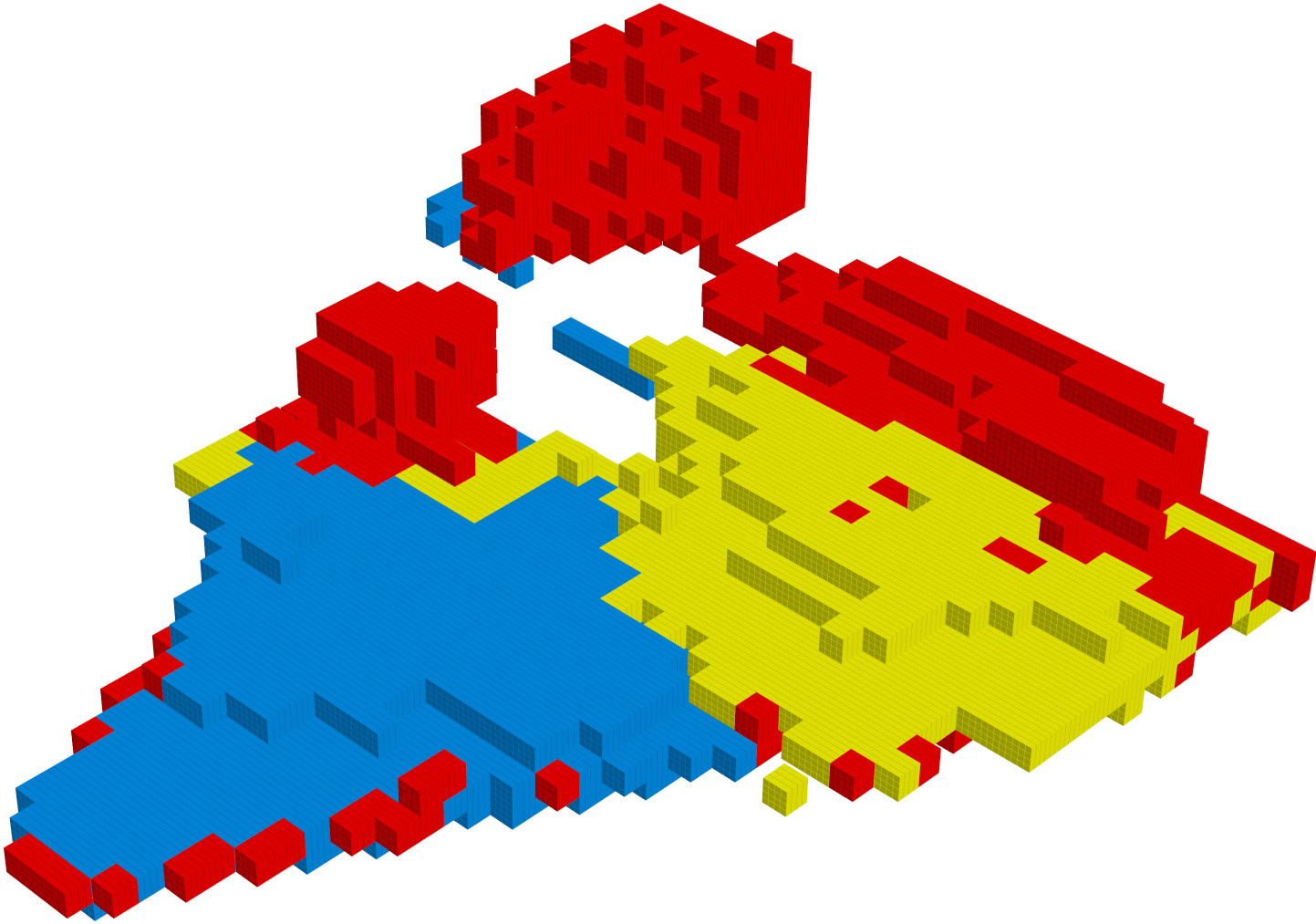}
    \end{minipage}%
    \begin{minipage}{0.16\textwidth}
        \centering
        \includegraphics[width=0.95\textwidth]{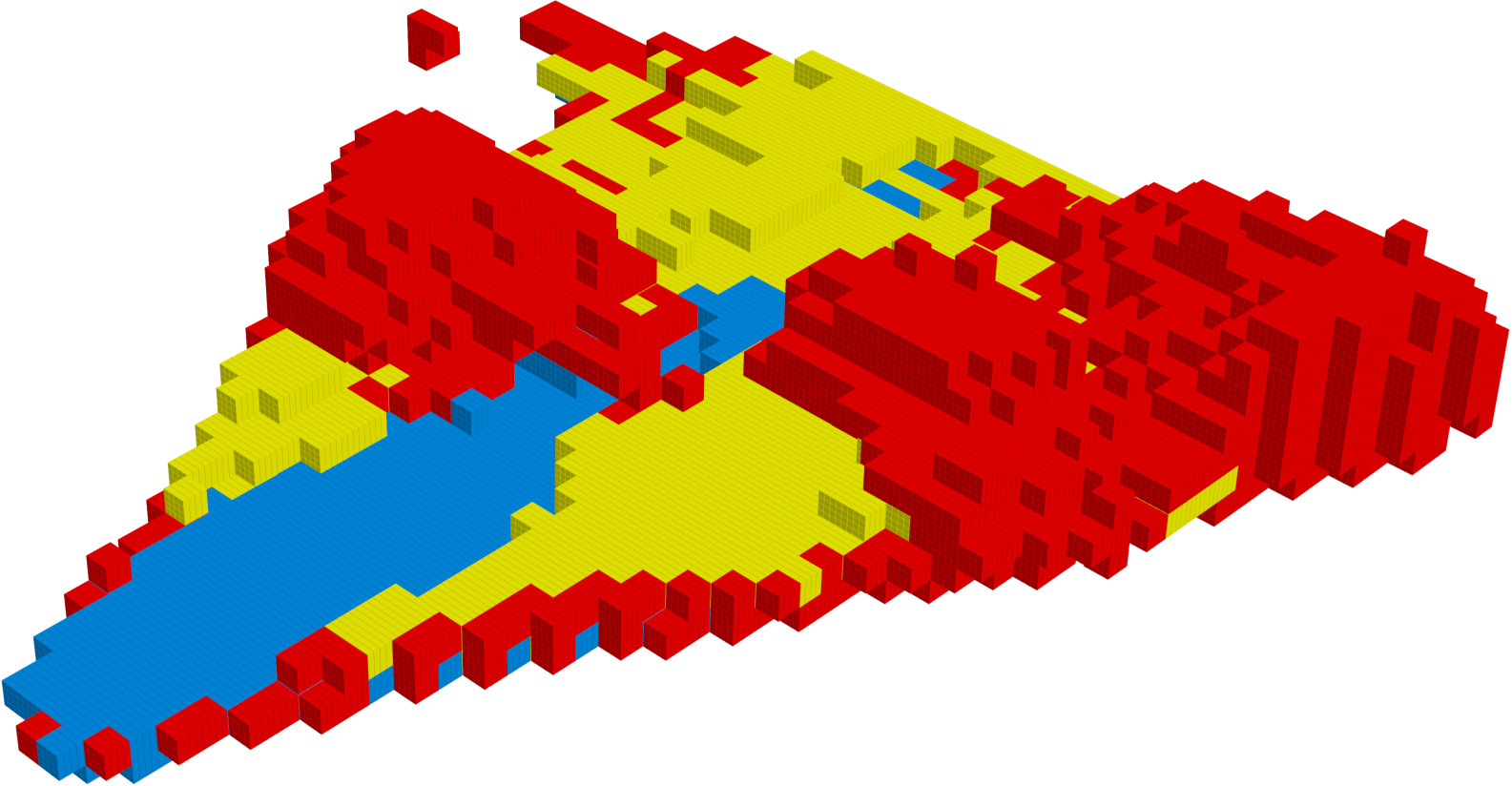}
    \end{minipage}%
    \begin{minipage}{0.16\textwidth}
        \centering
        \includegraphics[width=0.95\textwidth]{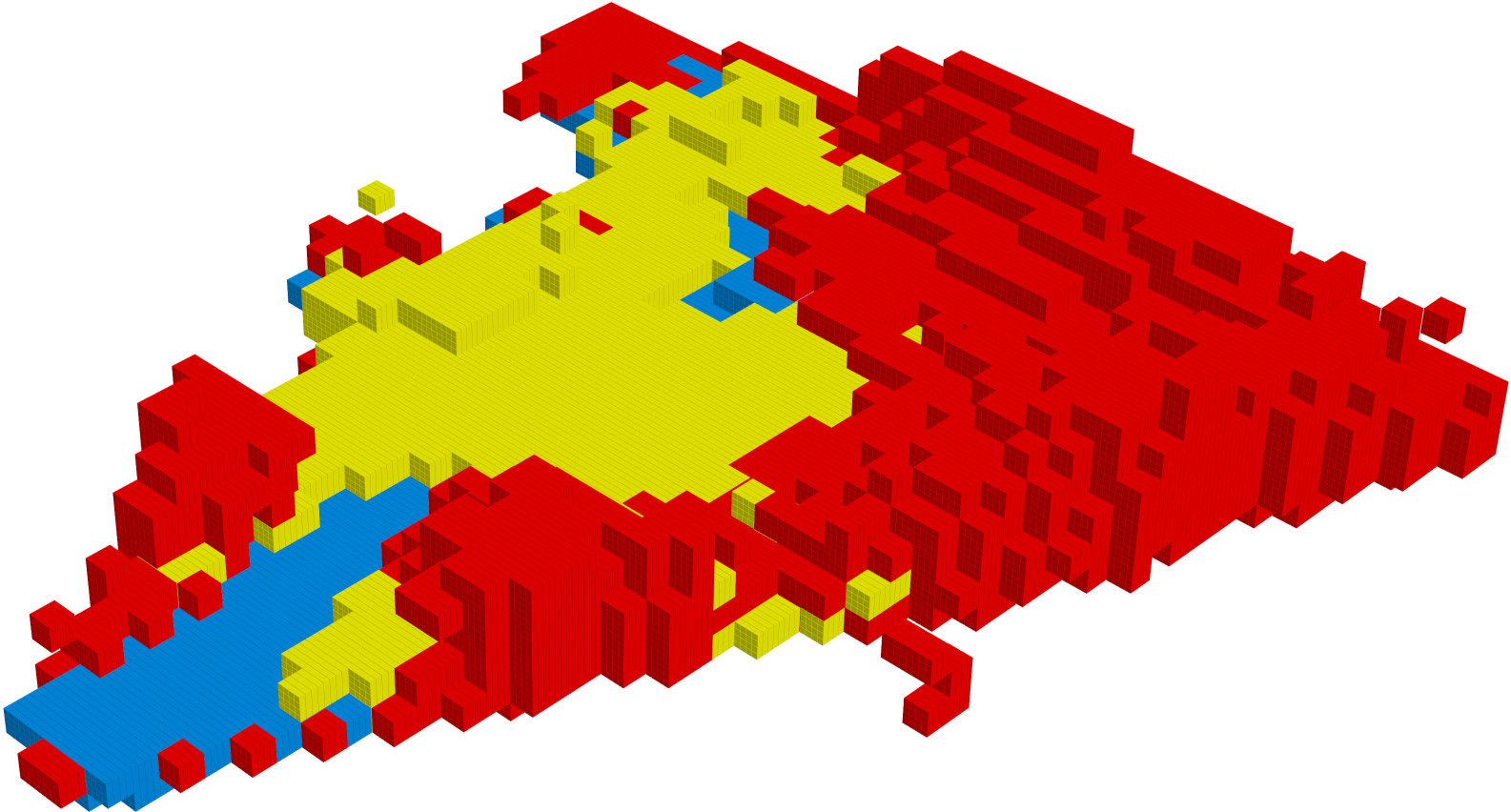}
    \end{minipage}%
    \begin{minipage}{0.16\textwidth}
        \centering
        \includegraphics[width=0.95\textwidth]{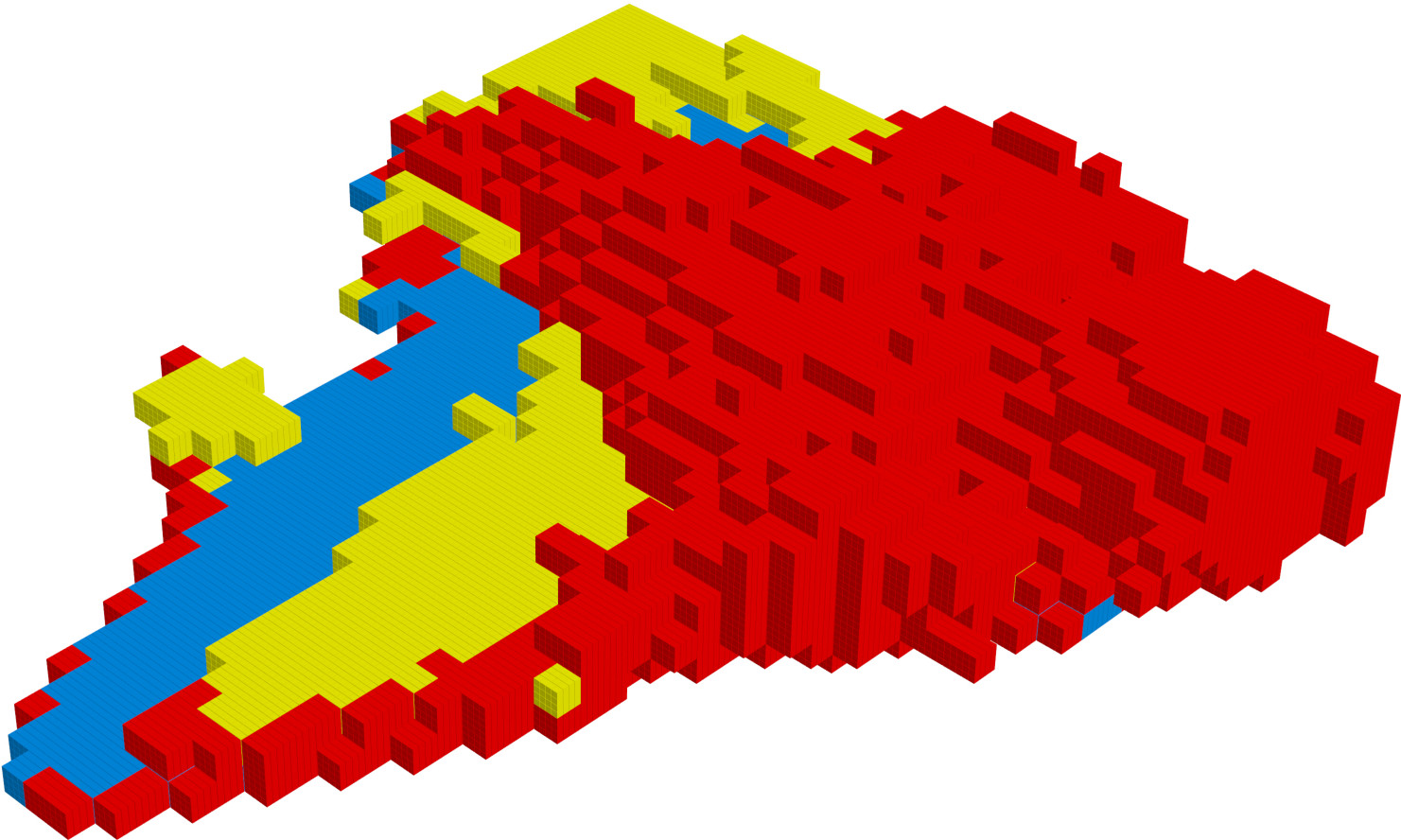}
    \end{minipage}
    \vskip\baselineskip

     % 6 row of images with titles on the left
    \begin{minipage}{0.16\textwidth}
        \centering
        {SSCNet-full}
    \end{minipage}%
    \begin{minipage}{0.16\textwidth}
        \centering
        \includegraphics[width=0.95\textwidth]{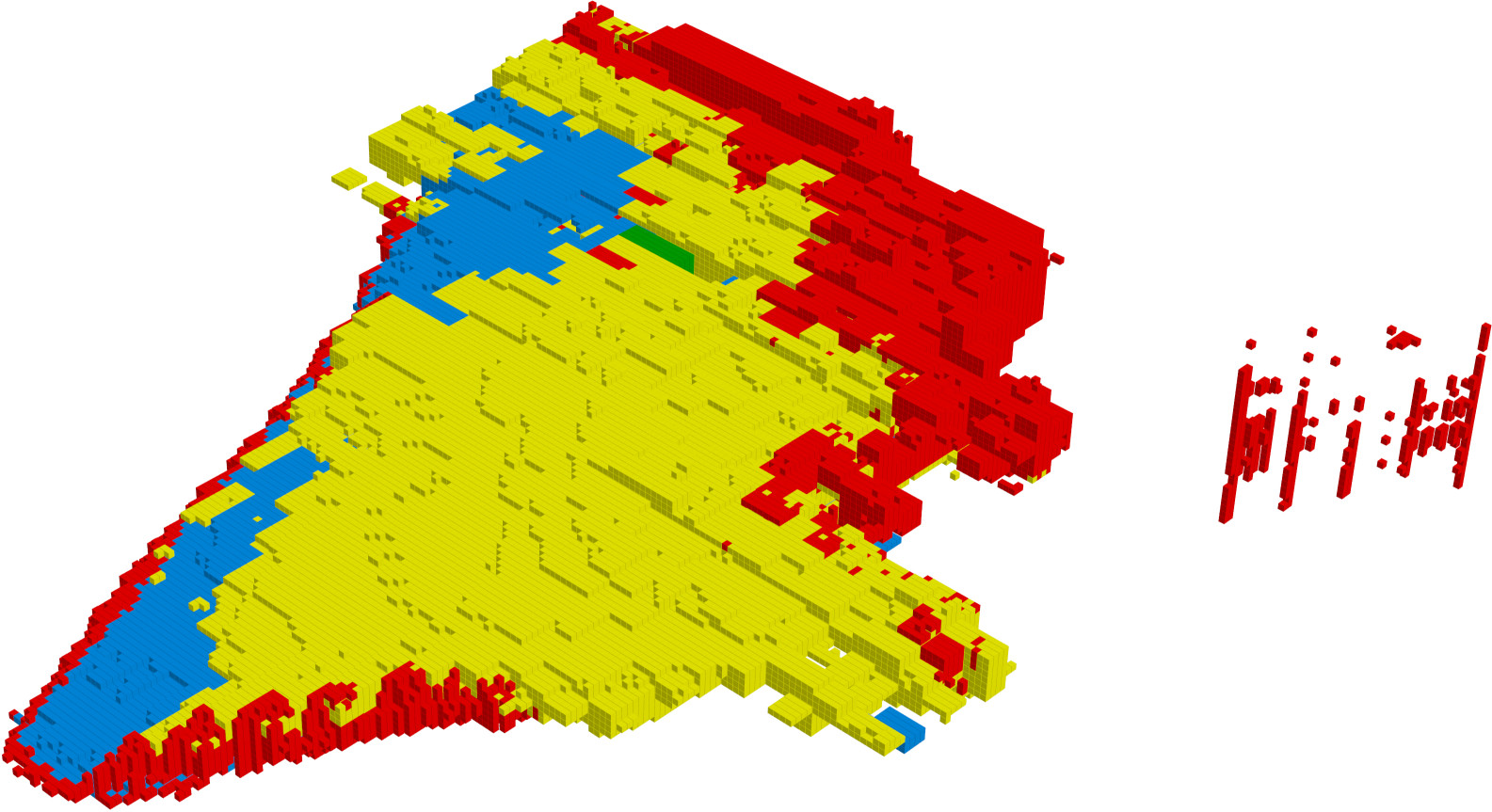}
    \end{minipage}%
    \begin{minipage}{0.16\textwidth}
        \centering
        \includegraphics[width=0.95\textwidth]{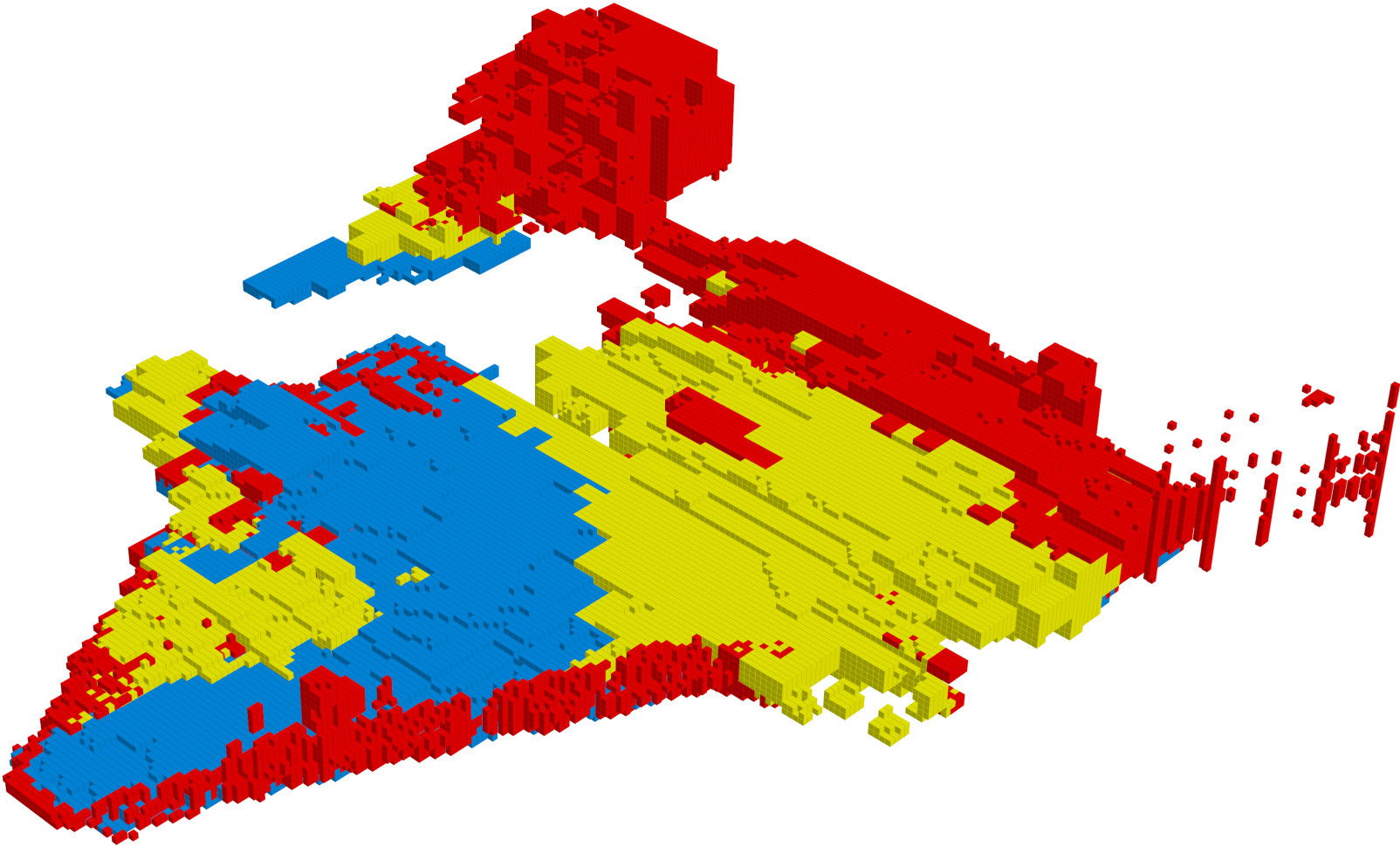}
    \end{minipage}%
    \begin{minipage}{0.16\textwidth}
        \centering
        \includegraphics[width=0.95\textwidth]{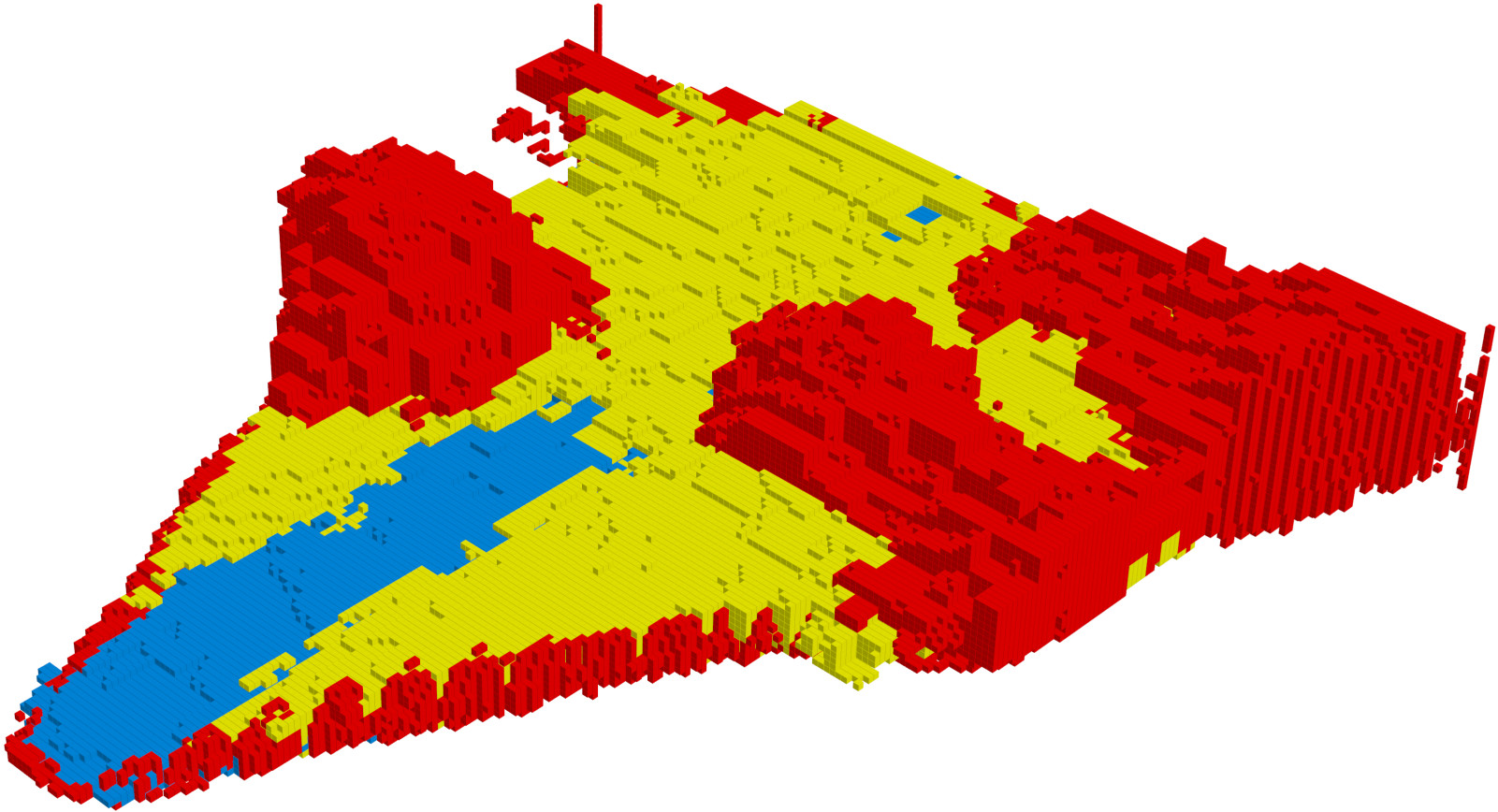}
    \end{minipage}%
    \begin{minipage}{0.16\textwidth}
        \centering
        \includegraphics[width=0.95\textwidth]{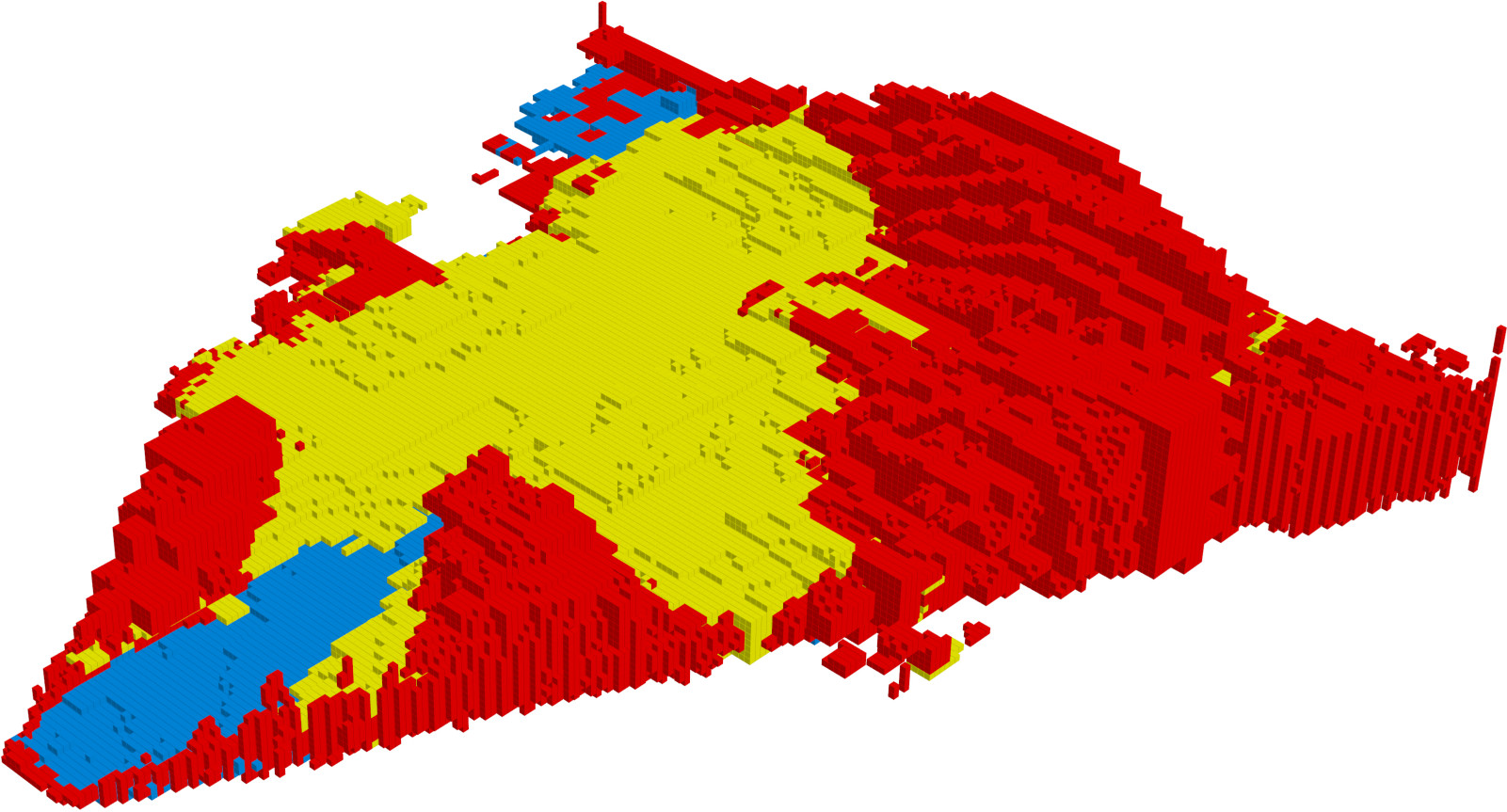}
    \end{minipage}%
    \begin{minipage}{0.16\textwidth}
        \centering
        \includegraphics[width=0.95\textwidth]{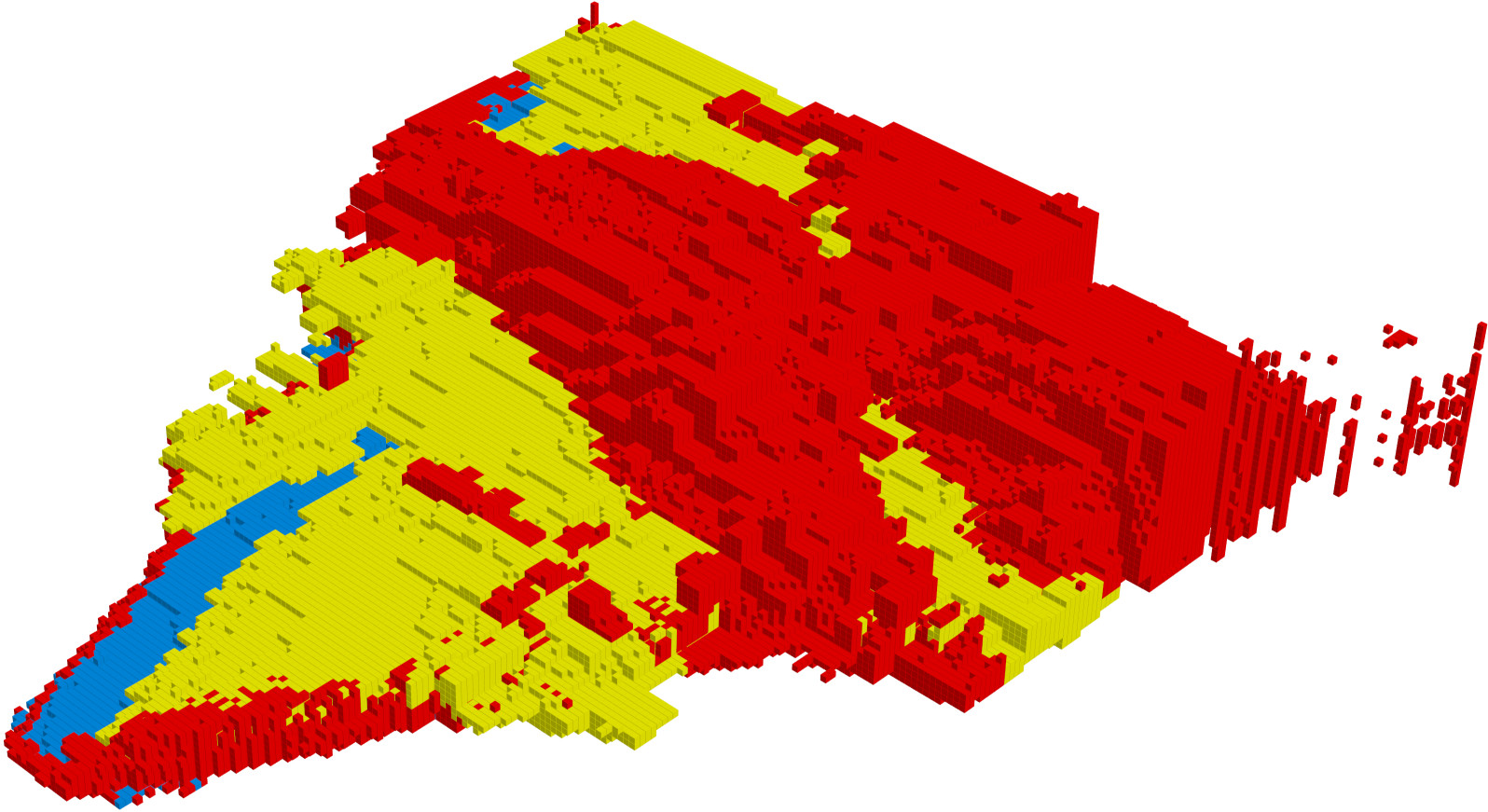}
    \end{minipage}
    \vskip\baselineskip

     % 7 row of images with titles on the left
    \begin{minipage}{0.16\textwidth}
        \centering
        {LMSCNet}
    \end{minipage}%
    \begin{minipage}{0.16\textwidth}
        \centering
        \includegraphics[width=0.95\textwidth]{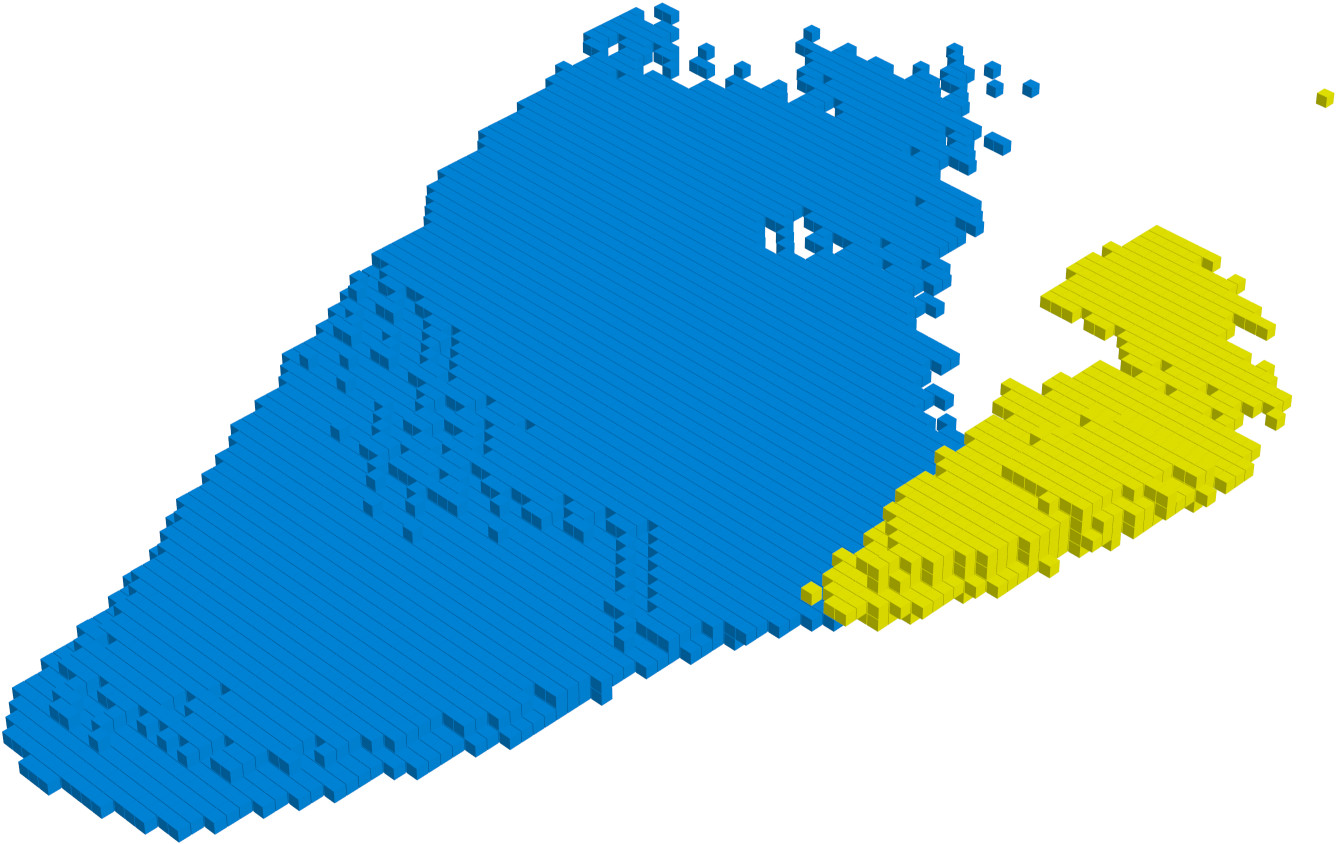}
    \end{minipage}%
    \begin{minipage}{0.16\textwidth}
        \centering
        \includegraphics[width=0.95\textwidth]{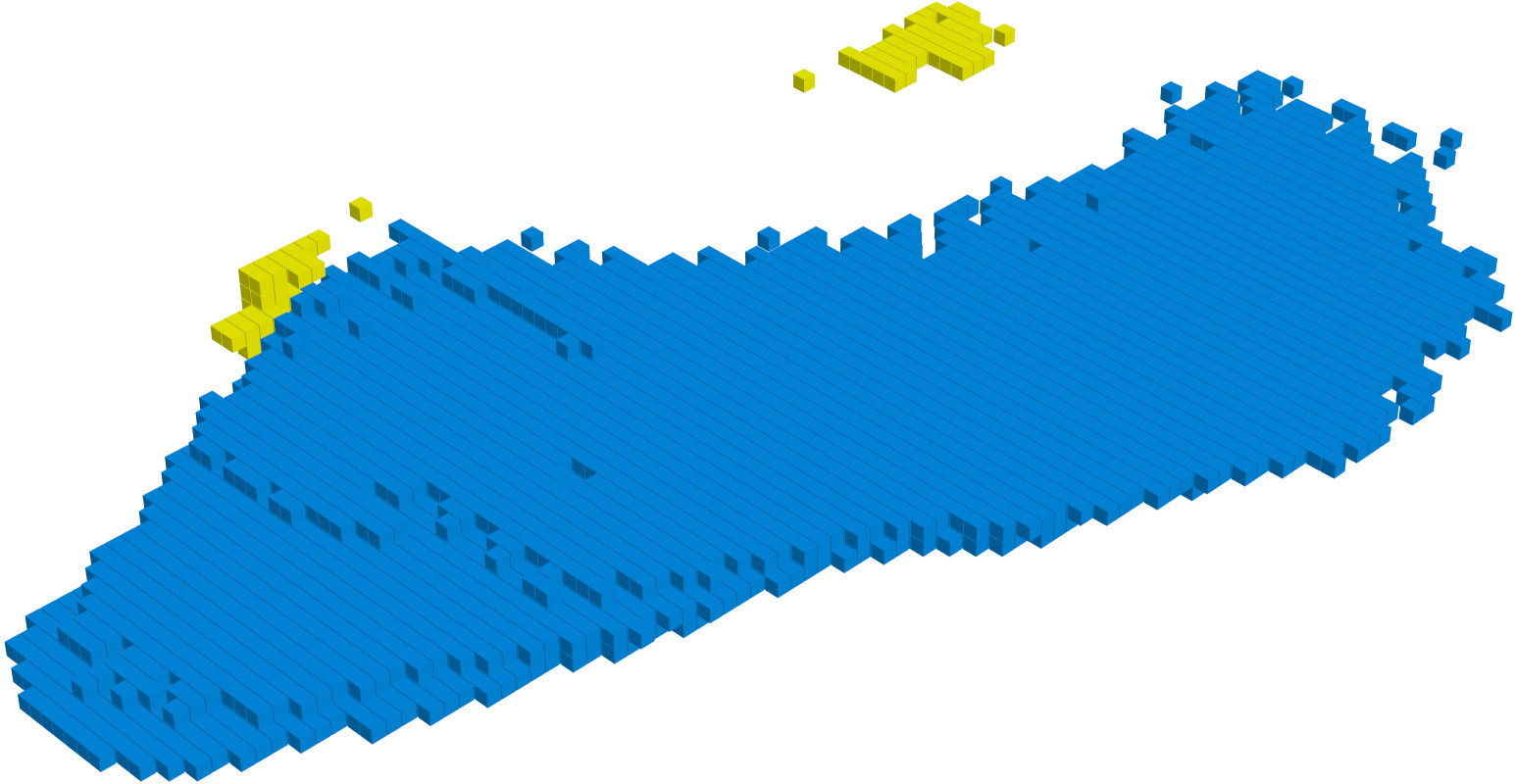}
    \end{minipage}%
    \begin{minipage}{0.16\textwidth}
        \centering
        \includegraphics[width=0.95\textwidth]{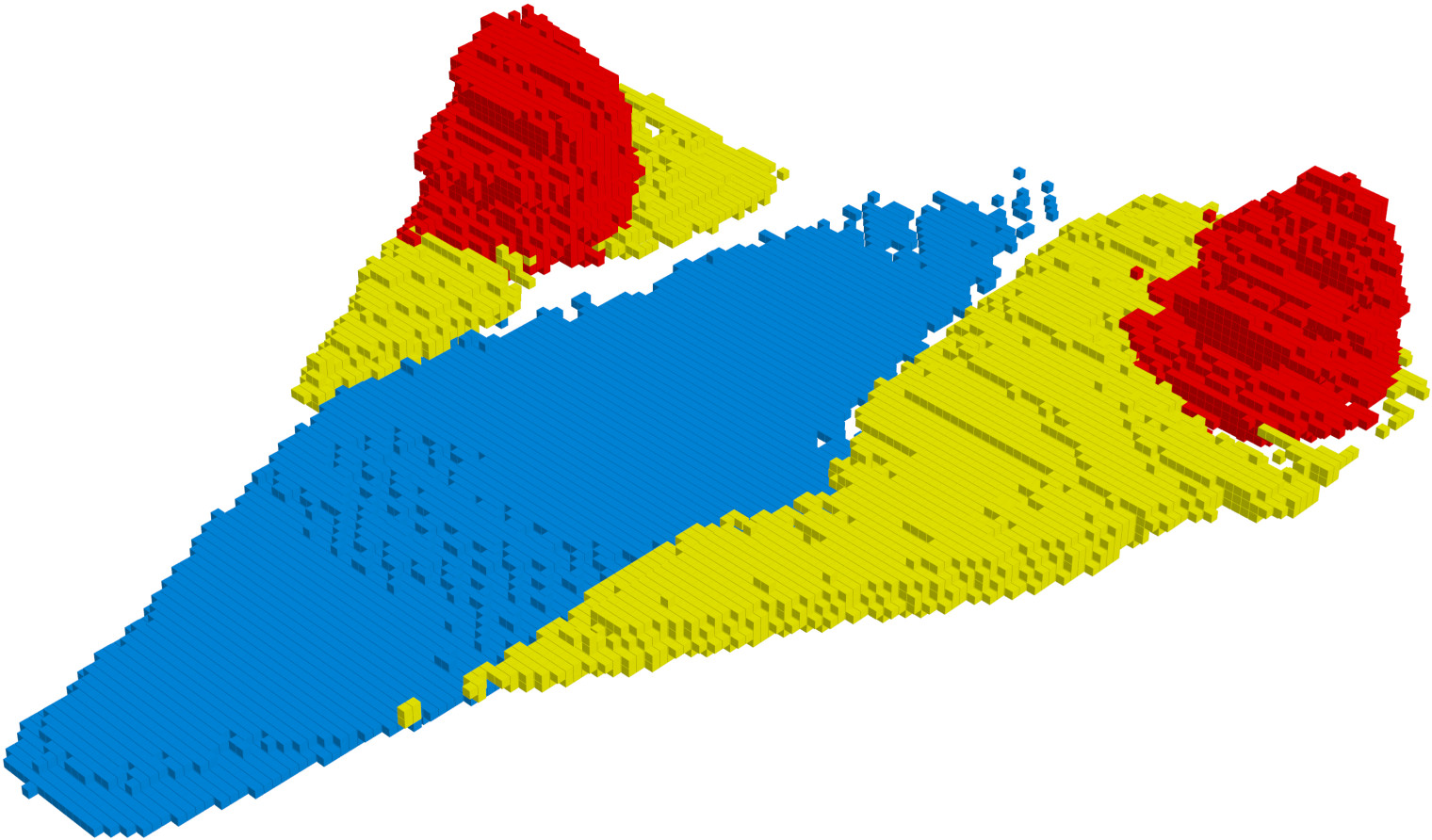}
    \end{minipage}%
    \begin{minipage}{0.16\textwidth}
        \centering
        \includegraphics[width=0.95\textwidth]{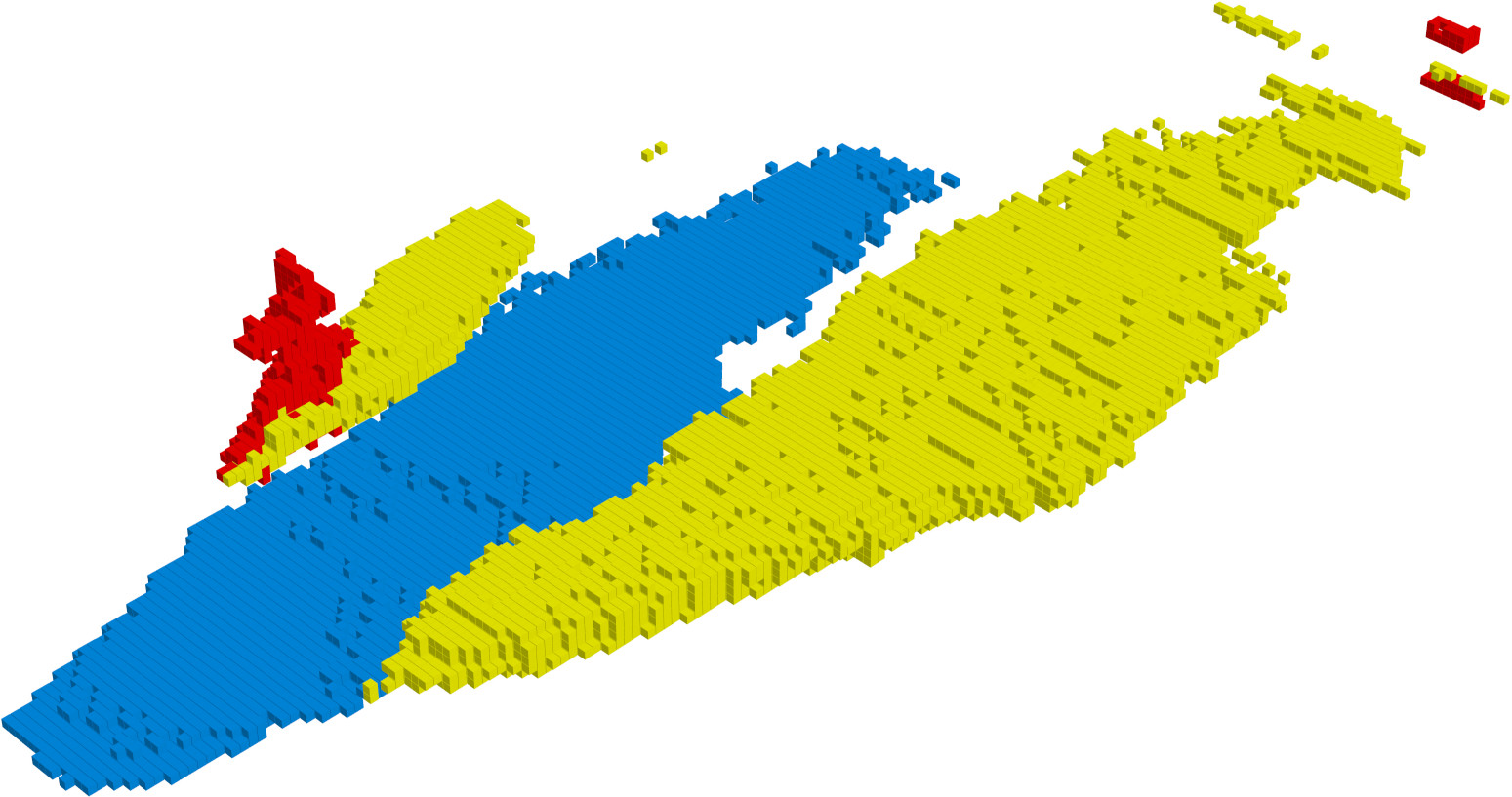}
    \end{minipage}%
    \begin{minipage}{0.16\textwidth}
        \centering
        \includegraphics[width=0.95\textwidth]{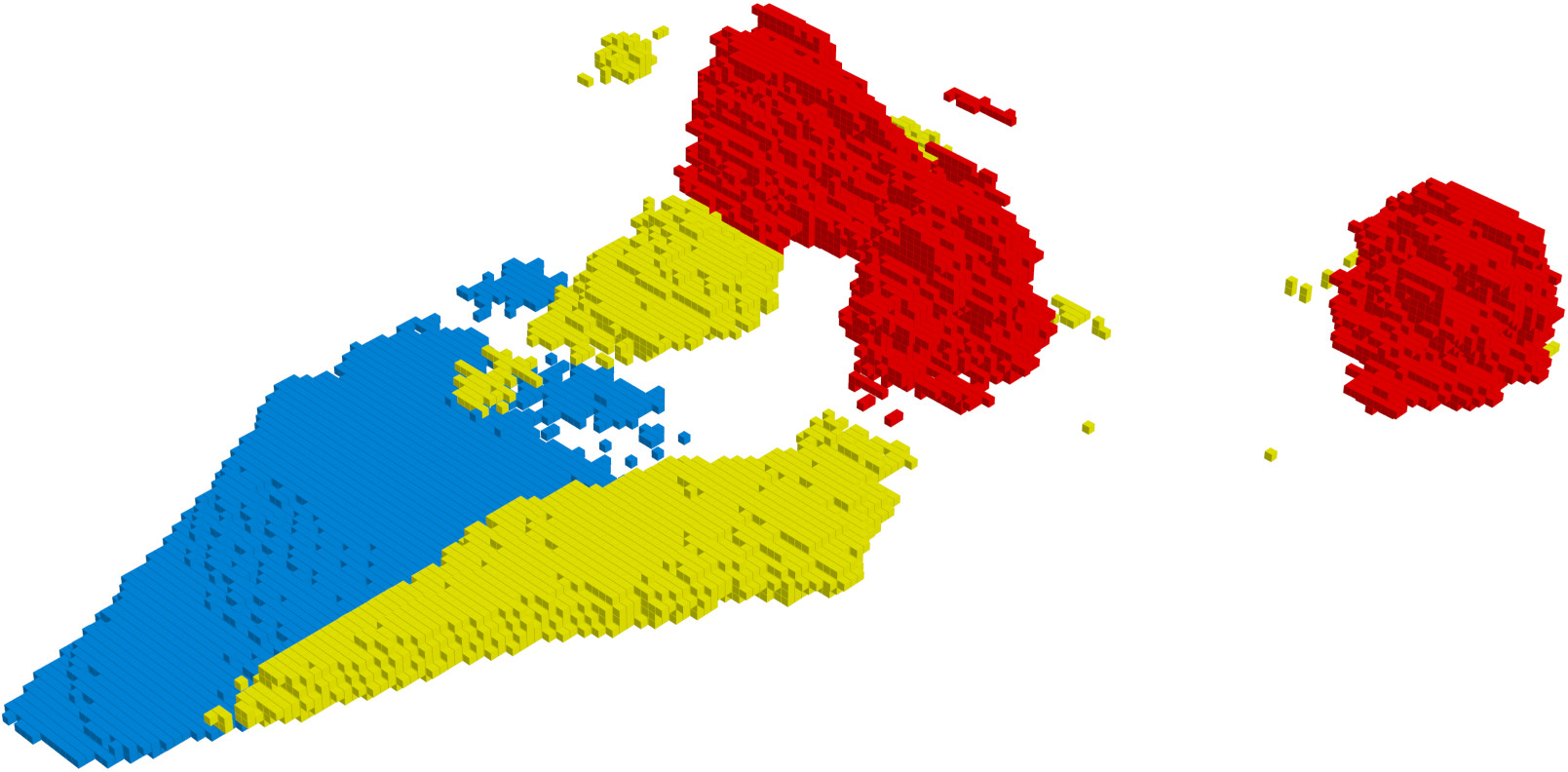}
    \end{minipage}
    \vskip\baselineskip

     % 8 row of images with titles on the left
    \begin{minipage}{0.16\textwidth}
        \centering
        {LMSCNet-SS}
    \end{minipage}%
    \begin{minipage}{0.16\textwidth}
        \centering
        \includegraphics[width=0.95\textwidth]{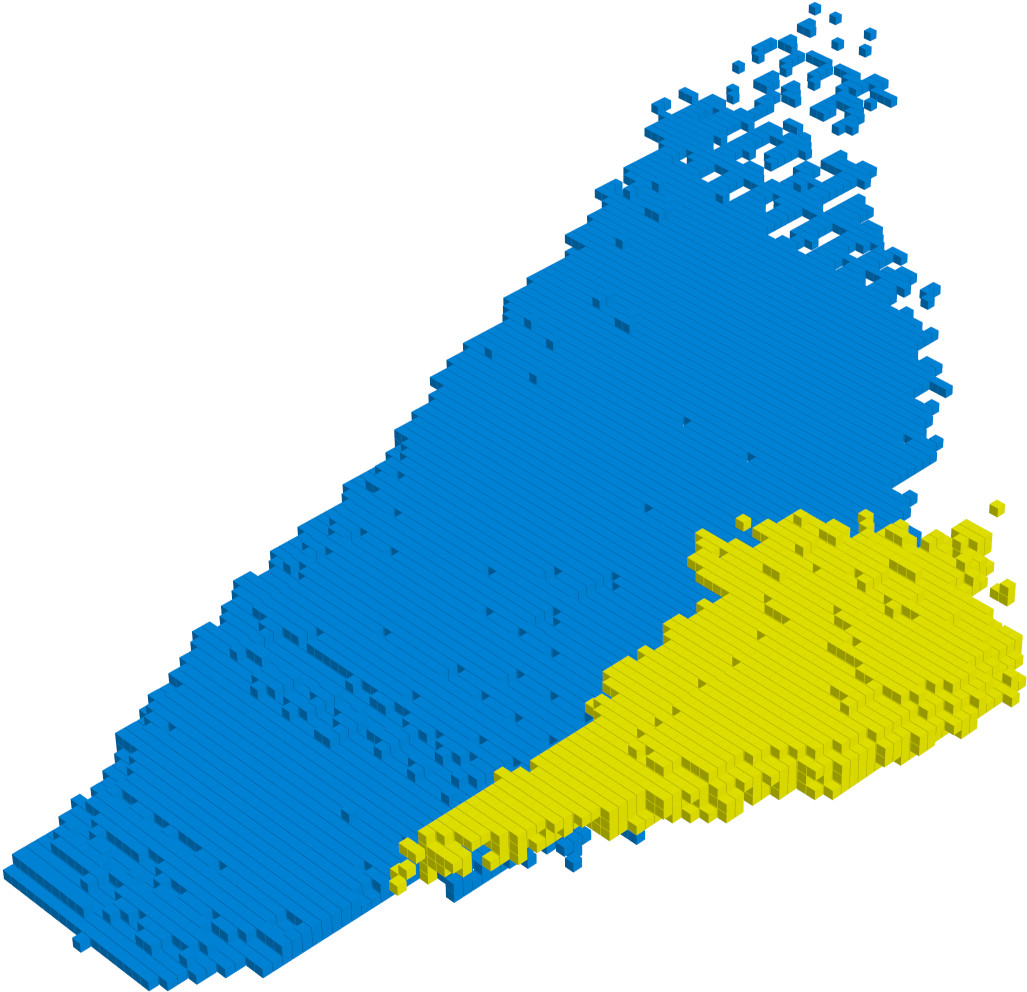}
    \end{minipage}%
    \begin{minipage}{0.16\textwidth}
        \centering
        \includegraphics[width=0.95\textwidth]{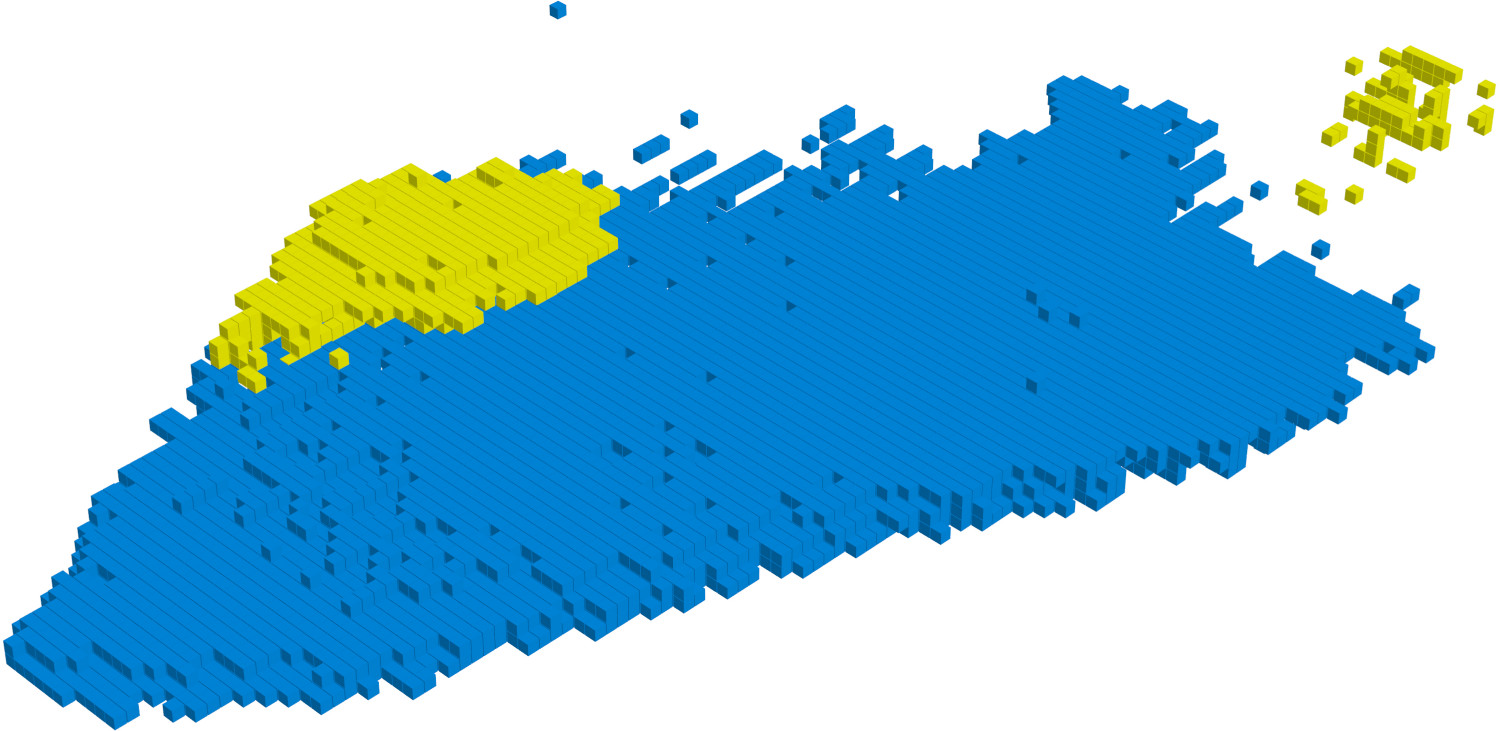}
    \end{minipage}%
    \begin{minipage}{0.16\textwidth}
        \centering
        \includegraphics[width=0.95\textwidth]{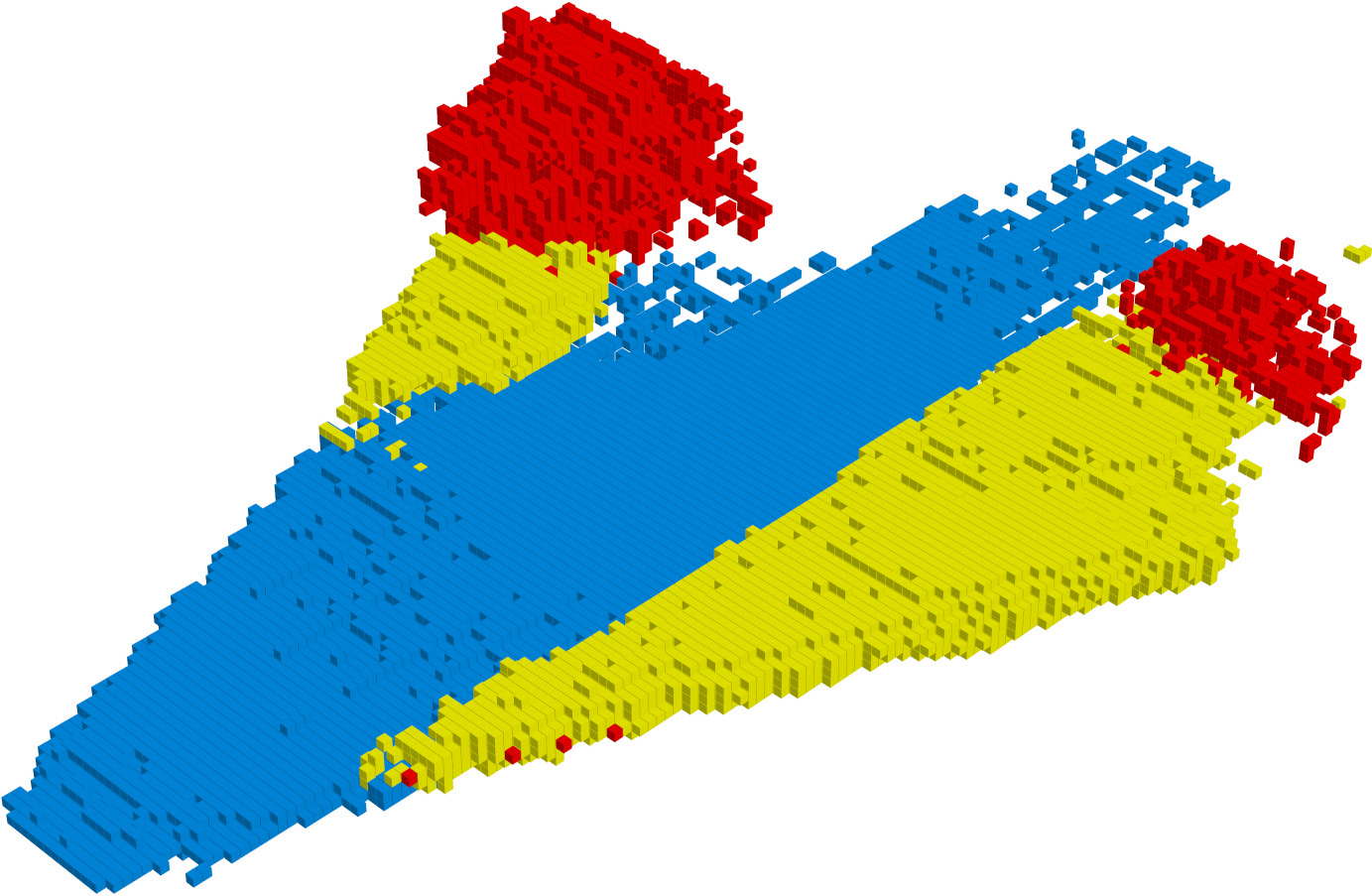}
    \end{minipage}%
    \begin{minipage}{0.16\textwidth}
        \centering
        \includegraphics[width=0.95\textwidth]{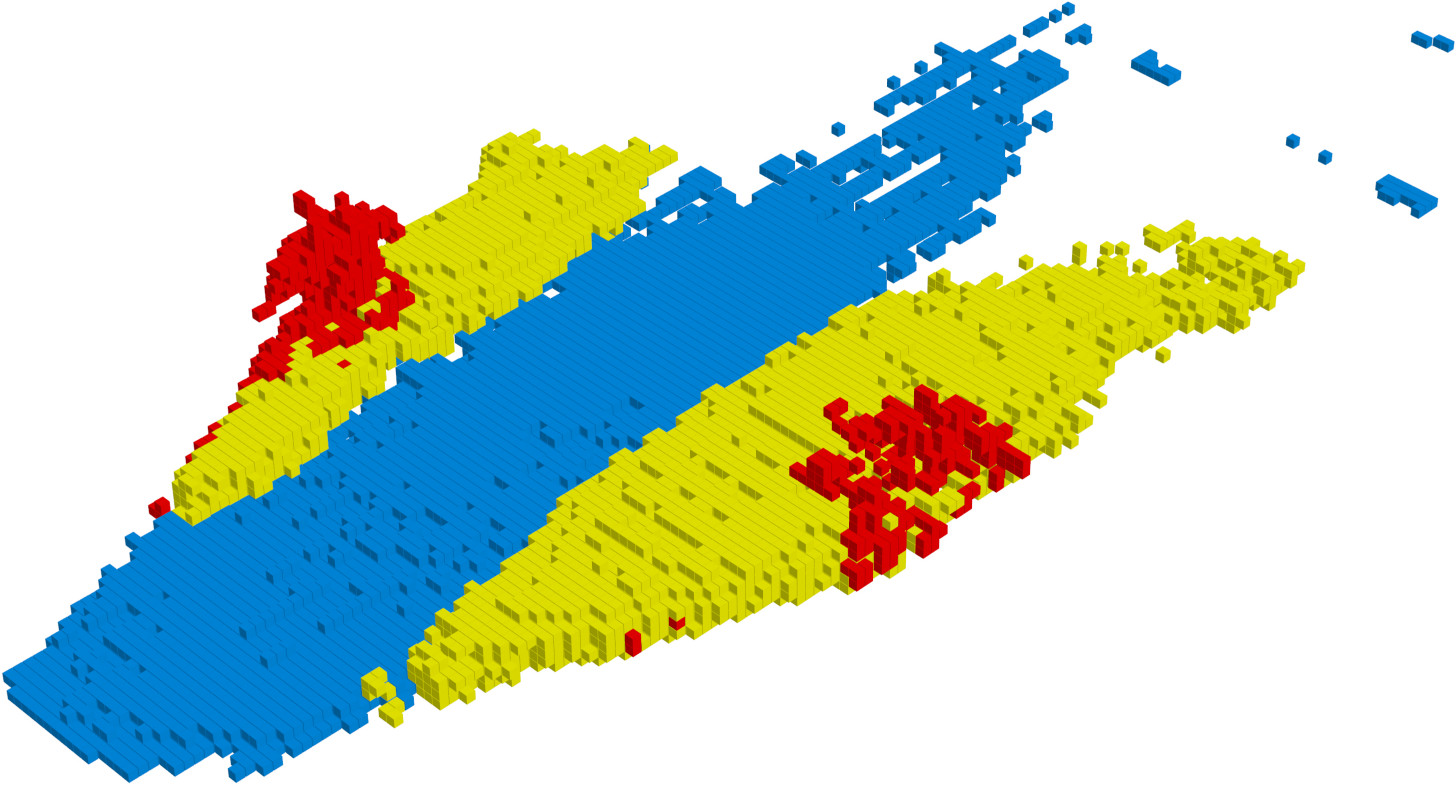}
    \end{minipage}%
    \begin{minipage}{0.16\textwidth}
        \centering
        \includegraphics[width=0.95\textwidth]{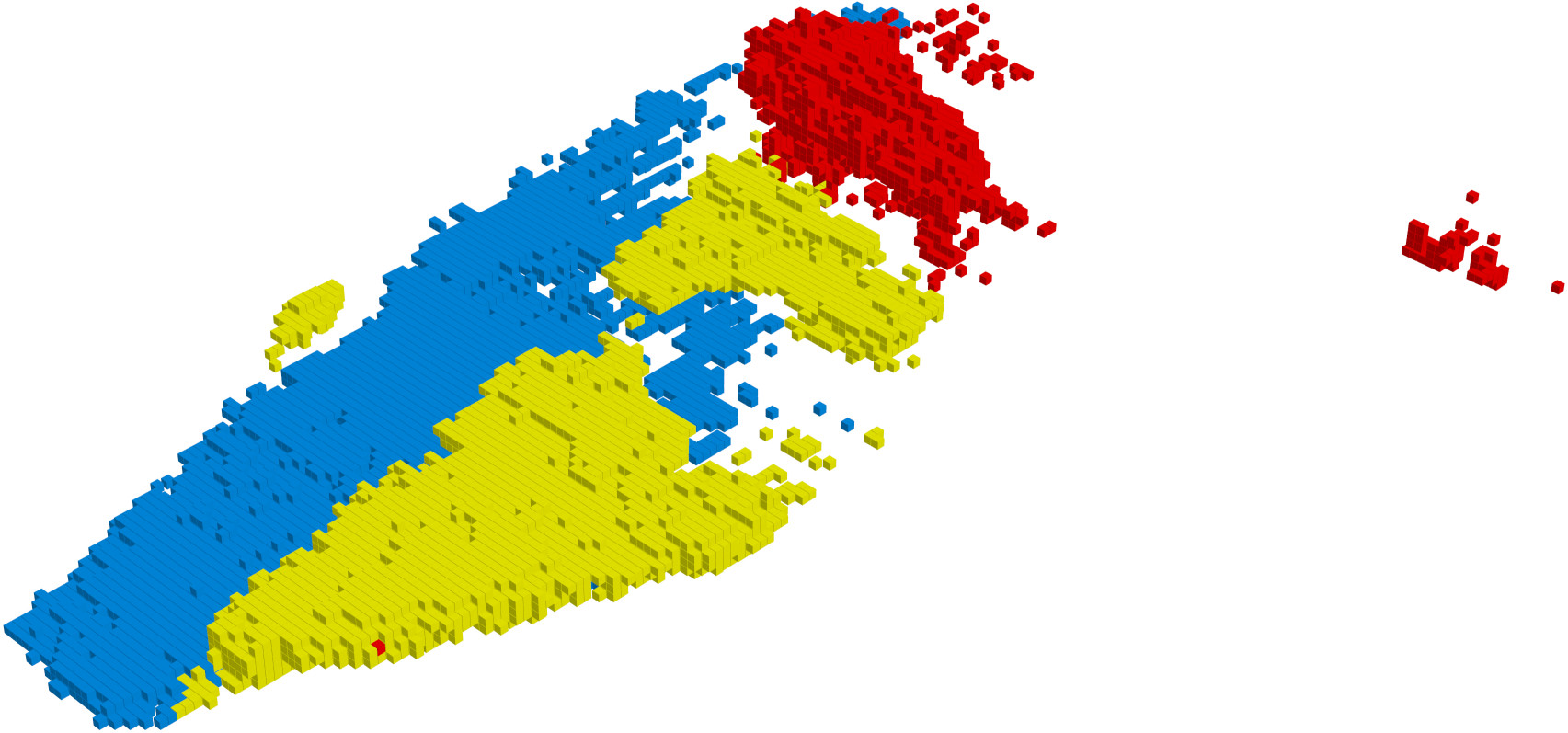}
    \end{minipage}
    \vskip\baselineskip

     % 9 row of images with titles on the left
    \begin{minipage}{0.16\textwidth}
        \centering
        {ORD-BKI}
    \end{minipage}%
    \begin{minipage}{0.16\textwidth}
        \centering
        \includegraphics[width=0.95\textwidth]{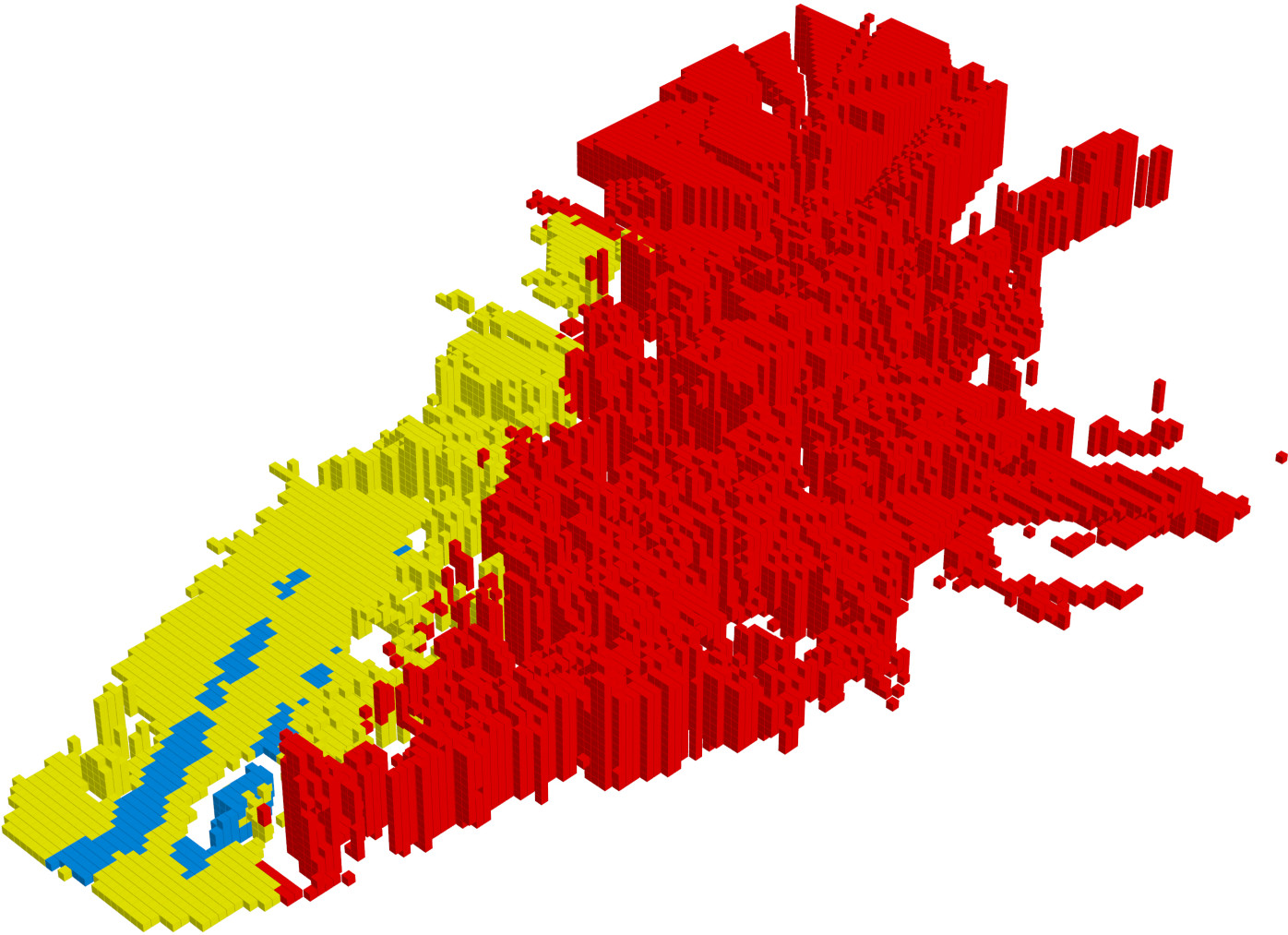}
    \end{minipage}%
    \begin{minipage}{0.16\textwidth}
        \centering
        \includegraphics[width=0.95\textwidth]{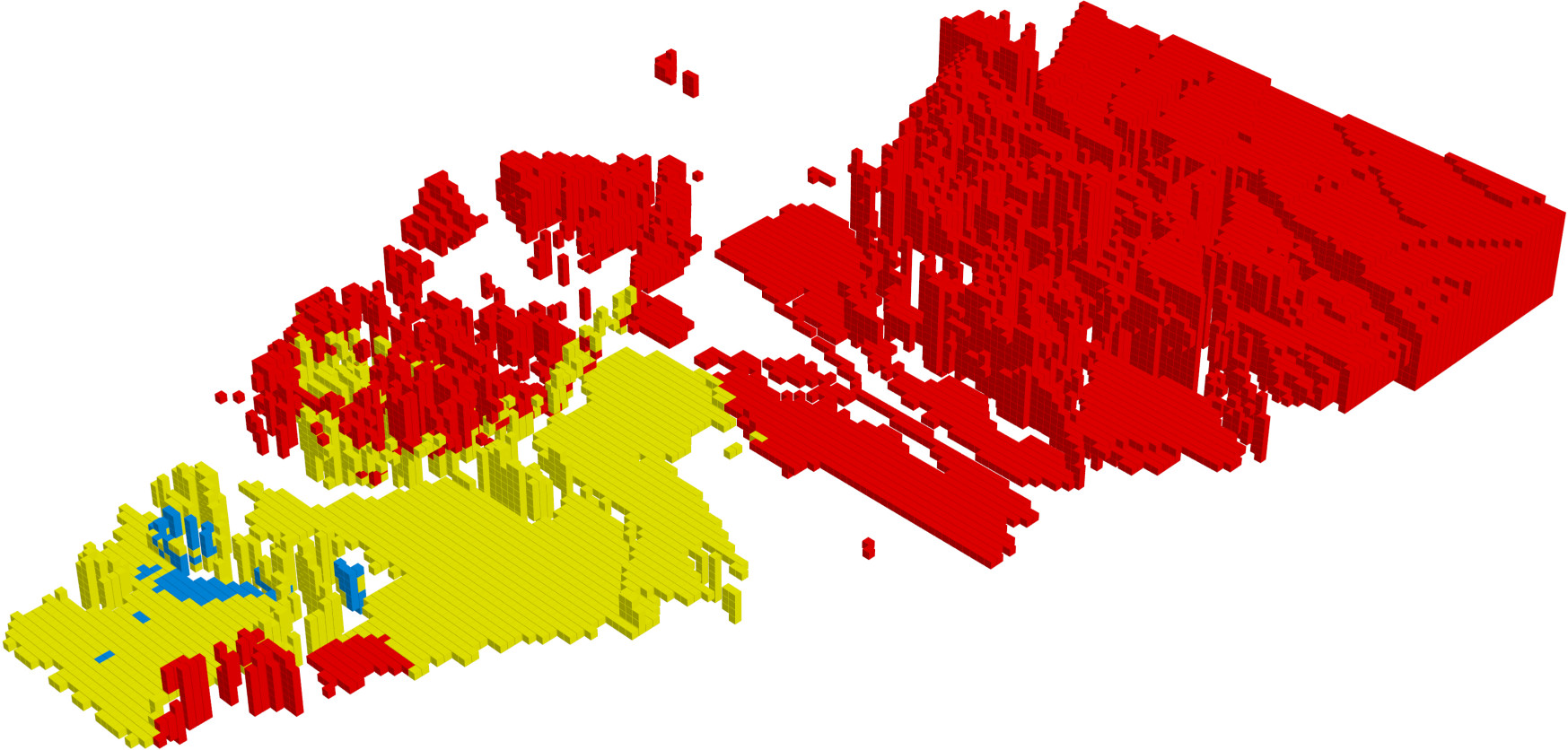}
    \end{minipage}%
    \begin{minipage}{0.16\textwidth}
        \centering
        \includegraphics[width=0.95\textwidth]{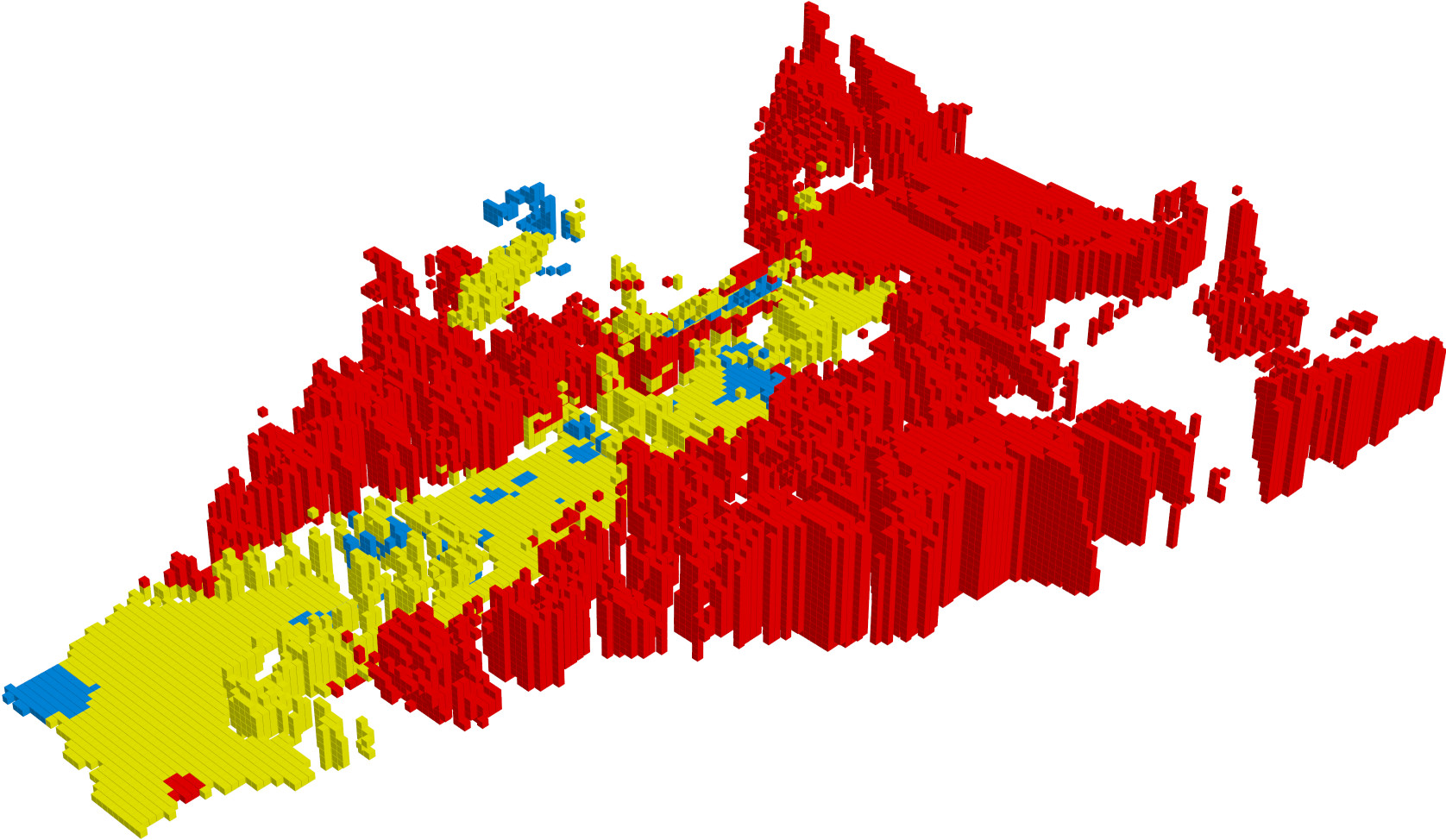}
    \end{minipage}%
    \begin{minipage}{0.16\textwidth}
        \centering
        \includegraphics[width=0.95\textwidth]{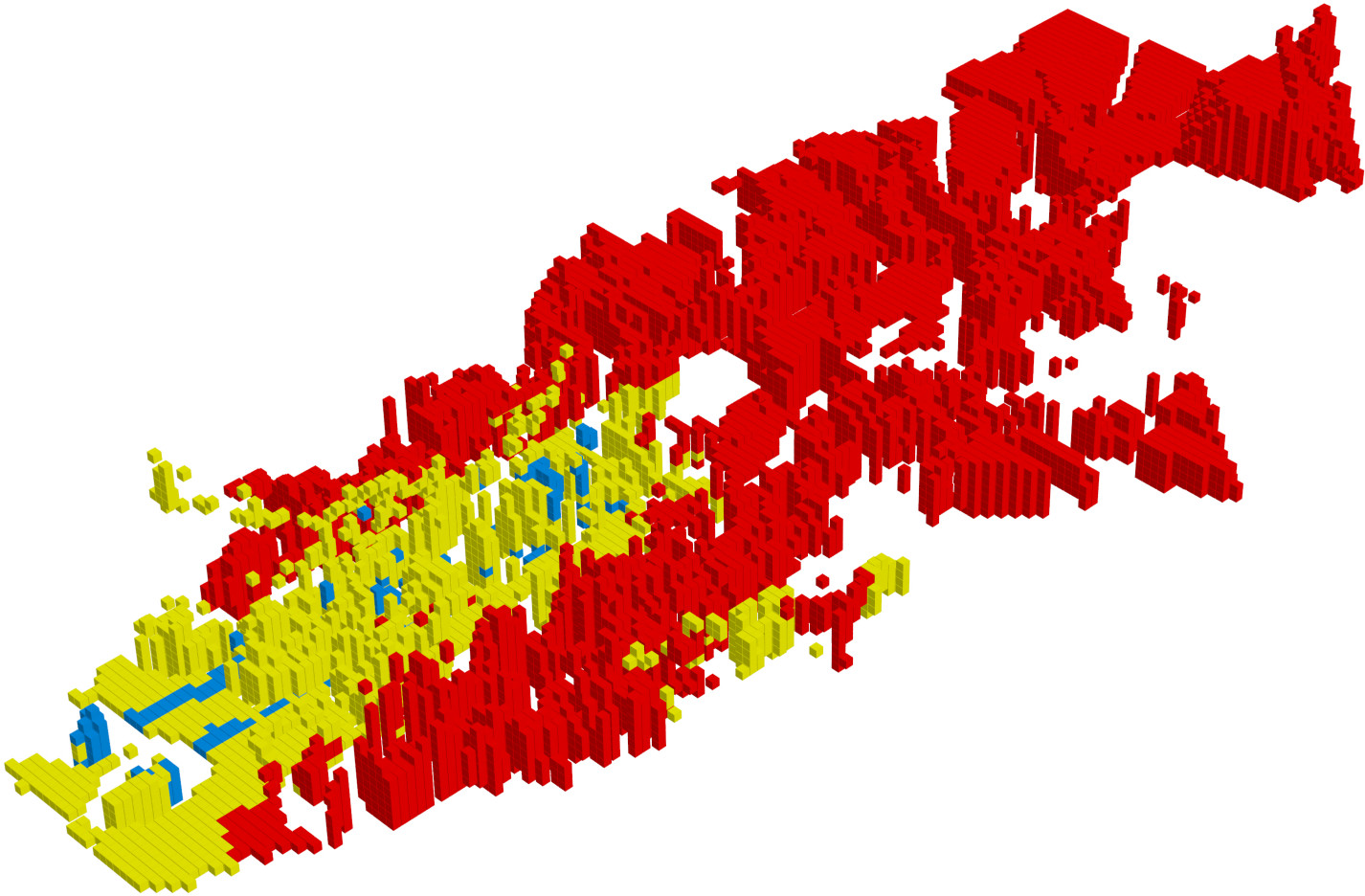}
    \end{minipage}%
    \begin{minipage}{0.16\textwidth}
        \centering
        \includegraphics[width=0.95\textwidth]{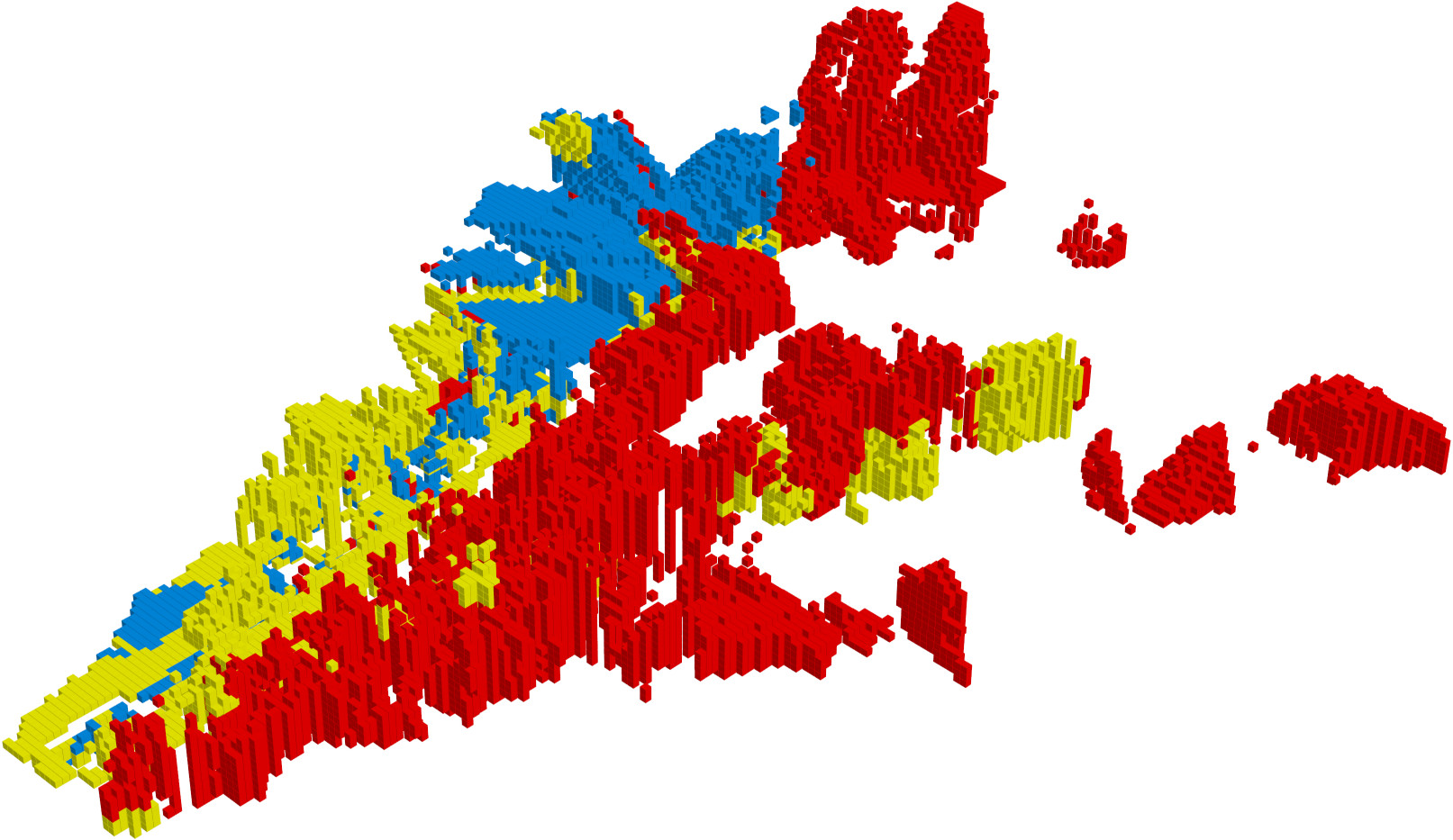}
    \end{minipage}
    % \vskip\baselineskip

    % Additional rows can follow in the same format
    \caption{Visualization of traversability prediction results from various methods. The rows present different views and predictions, from top to bottom: (row 1) Camera views depicting real-world scenes, (row 2) Semantic annotations of the environment, (row 3) Traversability annotations representing the ground truth, (row 4) Traversability estimations produced by 3DTTNet, and (row 5) Predictions from alternative algorithms (rows 5-10). Color-category mappings are defined in Fig.~\ref{fig_2}(a) for semantic annotations and Fig.~\ref{fig_2}(b) for traversability annotations respectively.}

\label{fig_11}
\end{figure*}

Visualization of traversable terrain modeling results from different models is presented in Fig.~\ref{fig_11}. By comparing these results with the ground truth, it is evident that 3DTTNet effectively identifies road traversability challenges in off-road environments. Additionally, the model accurately infers and completes regions covering with vegetation and other obstacles.

\begin{figure}[htbp]
\centering
\includegraphics[width=\columnwidth]{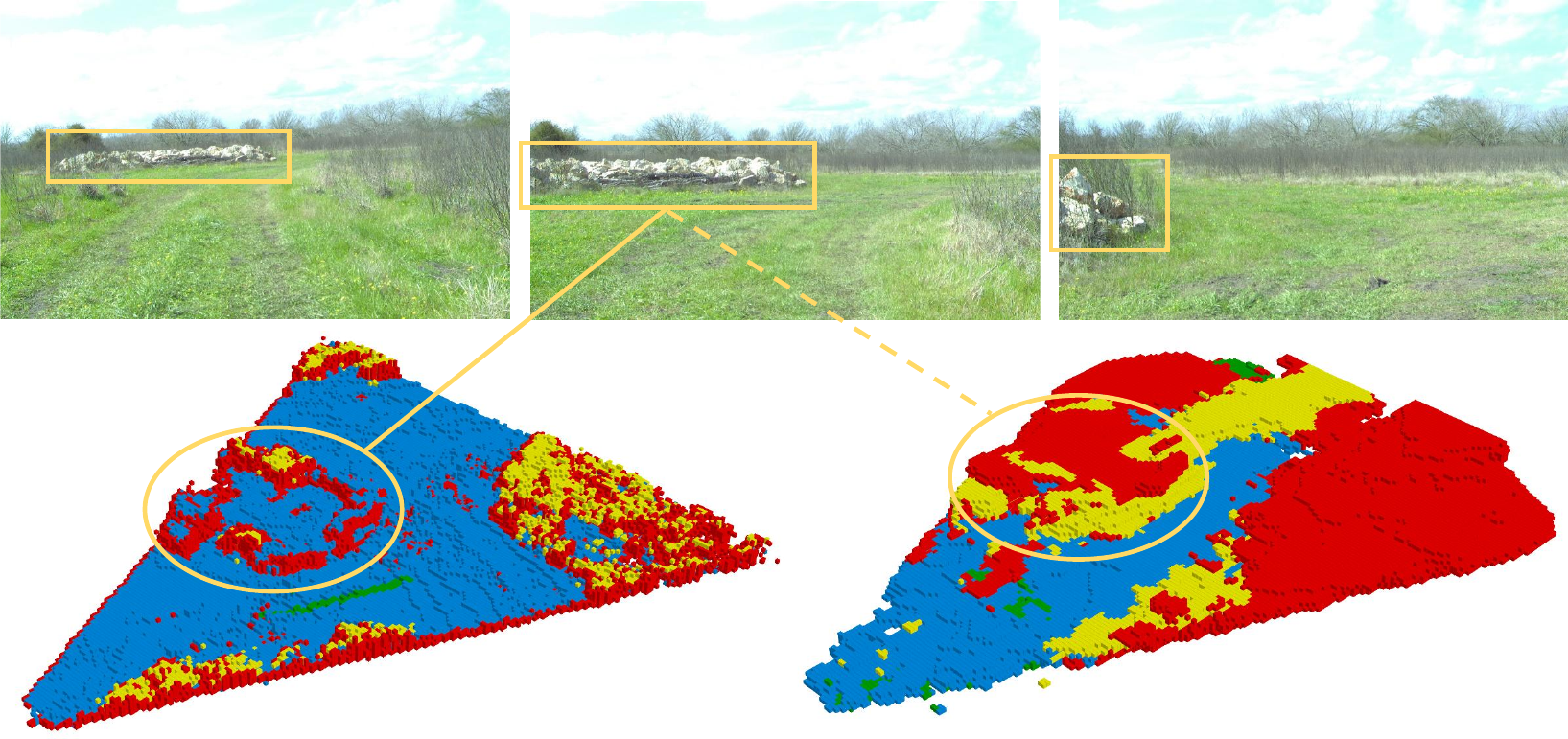}
\caption{Detection of rocky piles. Top images: camera views 10 seconds before (left), at the current time (center), and 10 seconds after (right). Bottom images: ground truth (left) and traversability estimations from 3DTTNet (right).}
\label{fig_12}
\end{figure}

As illustrated in Fig.~\ref{fig_12}, a pile of rocks appears in front of the platform. Remarkably, the model effectively recognizes the rocky pile as impassable and accurately assesses its spatial distribution. Although the ground truth data designates the edges of the rocky obstacles as lethal regions, it incorrectly assigns labels to the flat areas within the rocky pile. In contrast, 3DTTNet correctly infers the entire rocky pile as lethal, demonstrating its generalization capabilities.

\begin{figure}[htbp]
\centering
\includegraphics[width=\columnwidth]{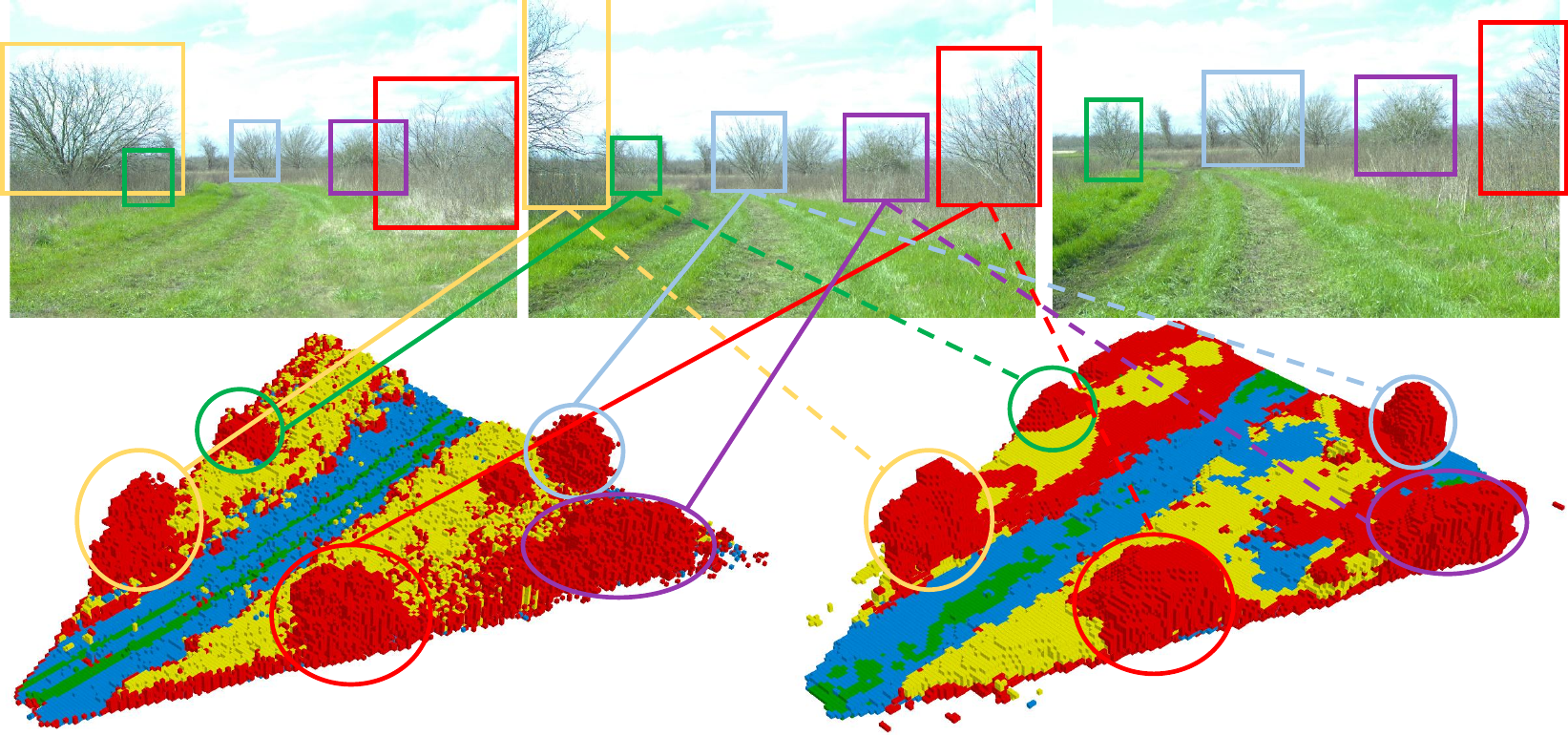}
\caption{Detection of multiple consecutive trees. Top images: camera views 10 seconds before (left), at the current time (center), and 10 seconds after (right). Bottom images: ground truth (left) and traversability estimations from 3DTTNet (right).}
\label{fig_13}
\end{figure}

\begin{figure}[htbp]
\centering
\includegraphics[width=\columnwidth]{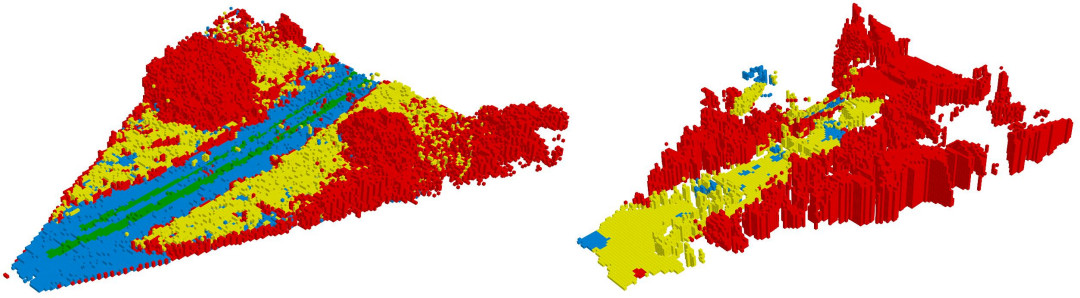}
\caption{Ground truth (left) and ORD-BKI inference result.}
\label{fig_14}
\end{figure}

In off-road scenarios, trees are among the most common obstacles that ground unmanned platforms must avoid during trajectory planning. Therefore, accurate detection of trees is essential. Fig.~\ref{fig_13} illustrates the detection results of 3DTTNet for multiple consecutively distributed trees. Different colored bounding boxes indicate the correspondence between various trees in images captured at different times, the ground truth, and traversability estimation results. By accurately identifying the spatial distribution of trees within the effective detection range, 3DTTNet enables ground unmanned platforms to effectively avoid these common obstacles in off-road environments, thereby facilitating safe trajectory planning.

The inference results of the ORD-BKI model are visualized, as shown in Fig.~\ref{fig_14}. The results demonstrate that although the ORD-BKI model effectively identifies traversable ground areas in the two-dimensional plane, its performance in representing the 3D environment is suboptimal. This observation suggests that semantic mapping approaches are significantly influenced by the density of LiDAR point clouds. Such approaches are more suitable for platforms equipped with multiple supplementary blind-spot LiDARs. In contrast, unmanned platforms equipped with only a single LiDAR sensor experience limitations in representational performance for 3D traversable area recognition tasks, due to the low point cloud density within the field of view of the camera.

Ground surfaces become partially exposed owing to vehicle passage, resulting in the formation of ruts. These rutted areas typically exhibit sparse weed distribution and lack clear boundaries with the surrounding densely weed-covered regions. Their characteristics lie between dirt and grass, which poses significant challenges for precise identification. Therefore, the recognition accuracy of 3DTTNet in these transitional areas requires further improvement.

\subsection{Real Vehicle Experiments}

\begin{figure}[htbp]
\centering
\includegraphics[height=1.5in]{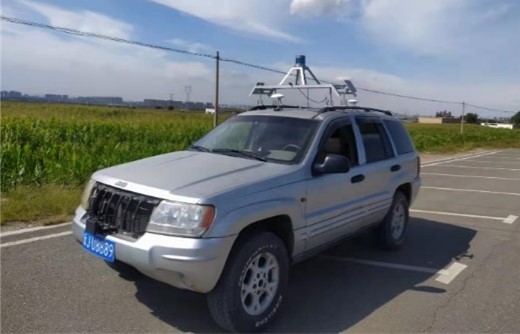}
\caption{Real vehicle test platform.}
\label{fig_15}
\end{figure}

3DTTNet is implemented on a real vehicle platform and validated through tests conducted at the Xiaotianshan Professional Outdoor Off-Road Site in Zhangjiakou City, Hebei Province, China. The test vehicle is a Jeep Grand Cherokee, equipped with a Suteng Innovation 80-line LiDAR and a Senyun Intelligent monocular camera (see Fig.~\ref{fig_15}).

\subsubsection{Test 1}
The experiments are involved with six typical scenarios: (a) Rural path: rural roads without ground markings; (b) Stepped barrier: stepped obstacles composed of scattered rocks and earth mounds; (c) Vehicle barrier: vehicle obstacles presented in the direction of travel; (d) Plain: wide dirt roads with scattered weeds and shrubs; (e) Forest: dirt roads flanked by trees on both sides; and (f) Mud ruts: muddy roads filled with deep vehicle ruts. These scenarios simulate diverse challenges in real off-road environments, aiming to evaluate the perception performance and robustness of the algorithm under varying complexities and dynamic conditions.

It is noteworthy that the training data is based on the RELLIS-3D dataset, which is collected using the unmanned ground platform Warthog. However, significant differences exist between the Warthog and the real-vehicle test platform concerning sensor models, vehicle dimensions, and sensor configurations. For instance, the Warthog platform and the Jeep Grand Cherokee differ in vehicle size, as well as in sensor installation heights and angles. These disparities pose challenges to the generalization capability of the algorithm.

\begin{figure*}[htbp]
    \begin{minipage}{0.143\textwidth}
        \centering
        {Results}
        % Top-left empty cell for title
    \end{minipage}%
    % 1 row titles
    \begin{minipage}{0.142\textwidth}
        \centering
        {(a) Rural \\ Path}
    \end{minipage}%
    \begin{minipage}{0.142\textwidth}
        \centering
        {(b) Stepped \\ Barrier}
    \end{minipage}%
    \begin{minipage}{0.142\textwidth}
        \centering
        {(c) Vehicle \\ Barrier}
    \end{minipage}%
    \begin{minipage}{0.142\textwidth}
        \centering
        {(d) Plain}
    \end{minipage}%
    \begin{minipage}{0.142\textwidth}
        \centering
        {(e) Forest}
    \end{minipage}%
    \begin{minipage}{0.142\textwidth}
        \centering
        {(f) Mud Ruts}
    \end{minipage}
    \vskip\baselineskip
    
    % 2 row of images with titles on the left
    \begin{minipage}{0.143\textwidth}
        \centering
        {Camera View}
    \end{minipage}%
    \begin{minipage}{0.142\textwidth}
        \centering
        \includegraphics[width=0.95\textwidth]{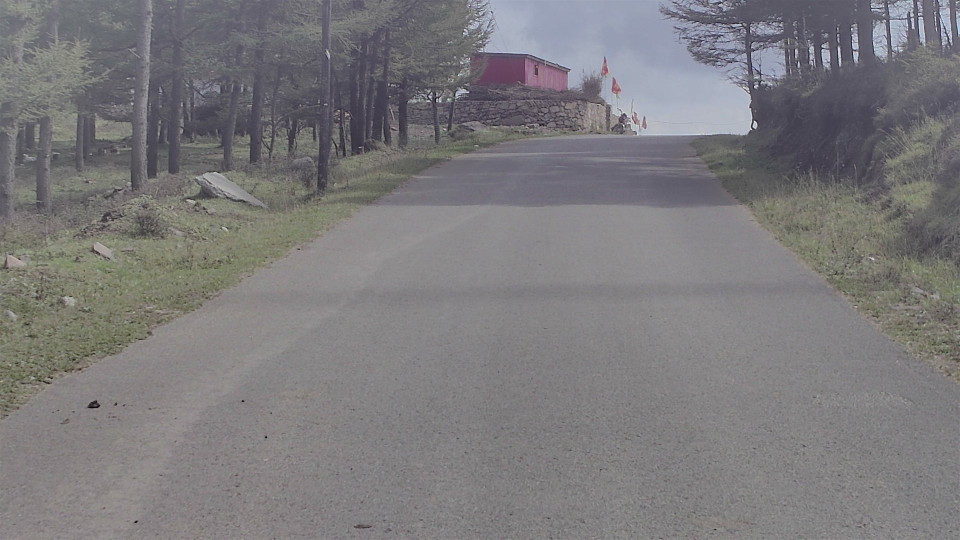}
    \end{minipage}%
    \begin{minipage}{0.142\textwidth}
        \centering
        \includegraphics[width=0.95\textwidth]{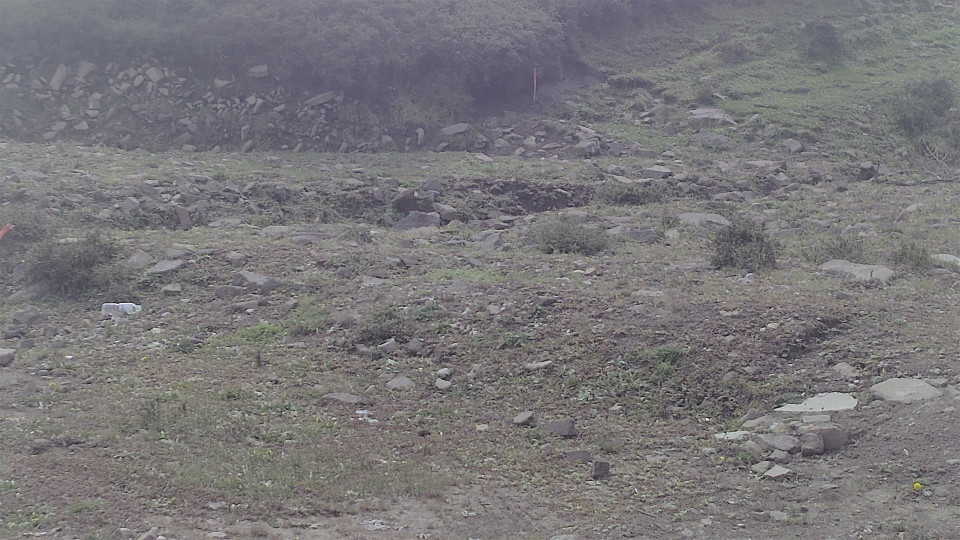}
    \end{minipage}%
    \begin{minipage}{0.142\textwidth}
        \centering
        \includegraphics[width=0.95\textwidth]{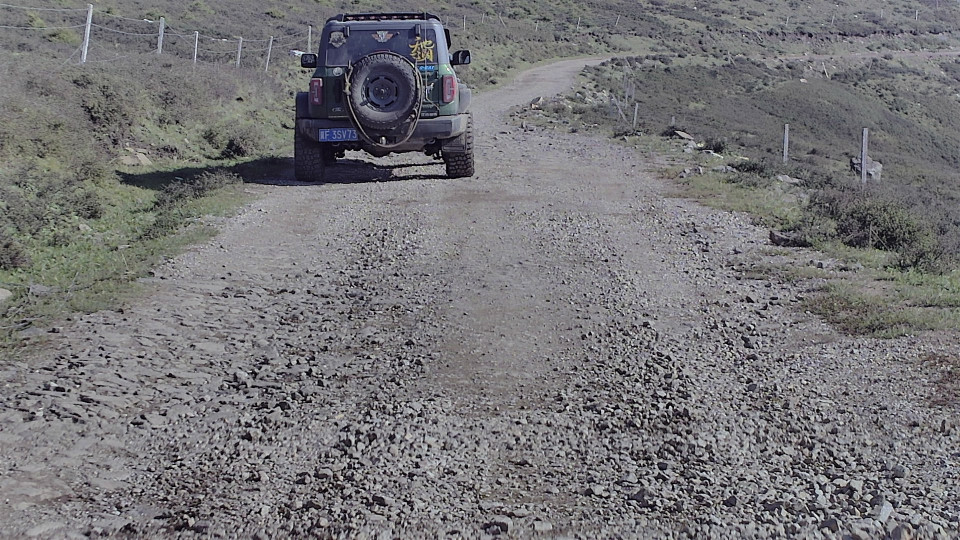}
    \end{minipage}%
    \begin{minipage}{0.142\textwidth}
        \centering
        \includegraphics[width=0.95\textwidth]{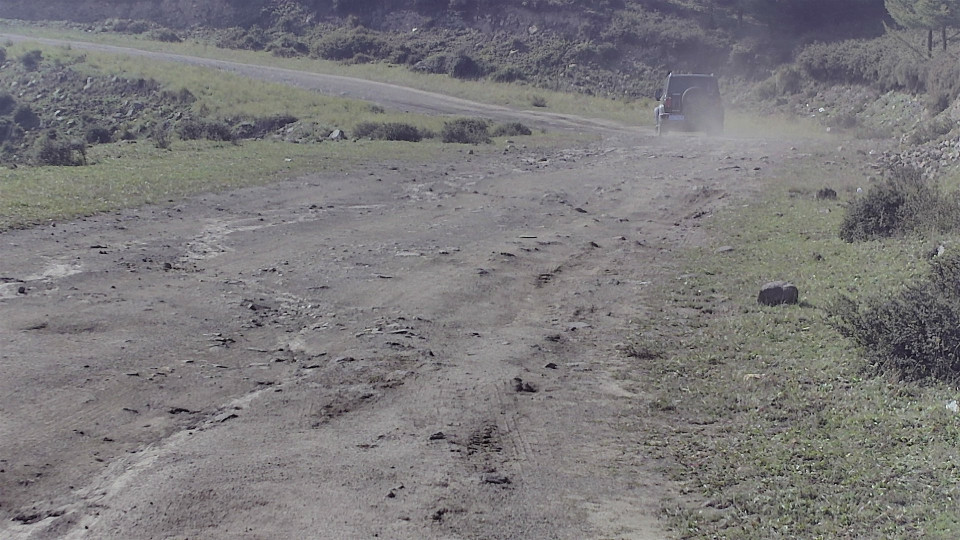}
    \end{minipage}%
    \begin{minipage}{0.142\textwidth}
        \centering
        \includegraphics[width=0.95\textwidth]{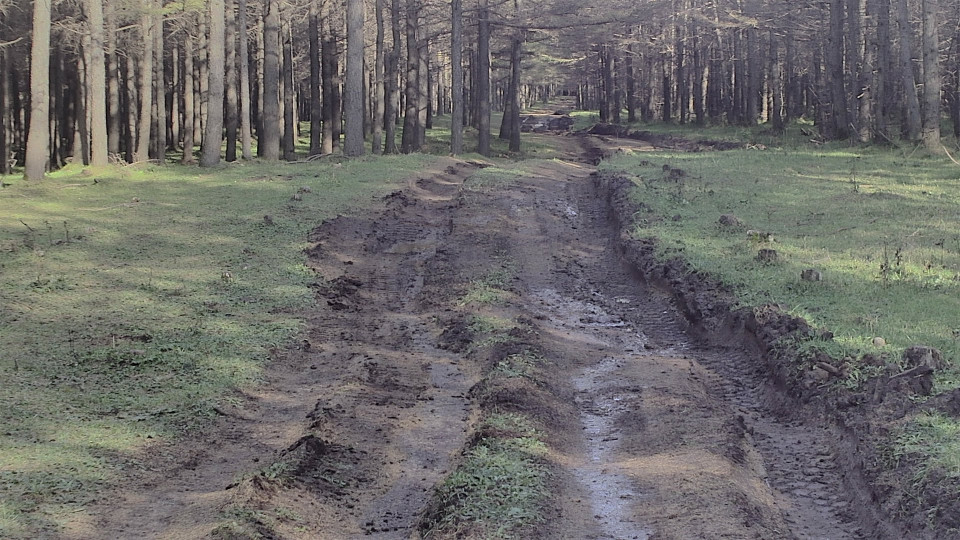}
    \end{minipage}
    \begin{minipage}{0.142\textwidth}
        % \centering
        \includegraphics[width=0.95\textwidth]{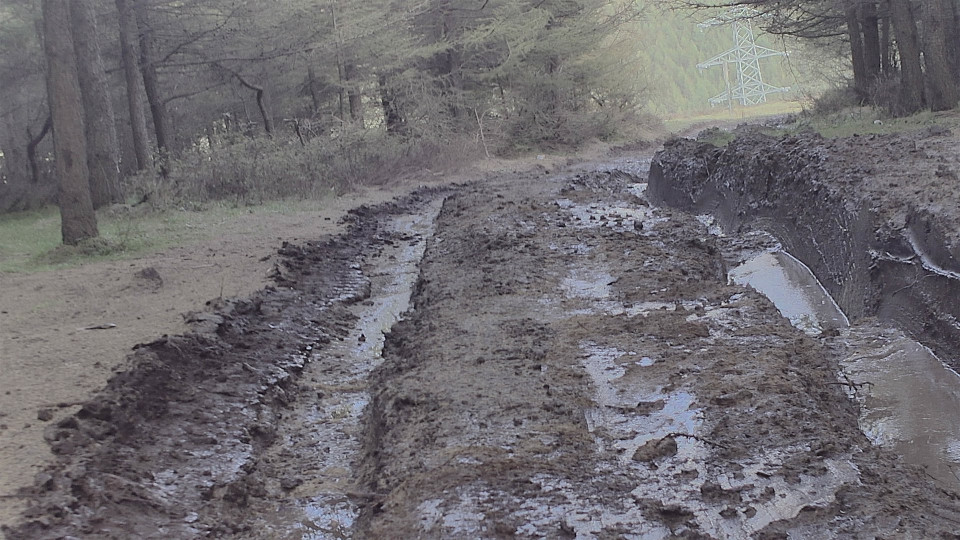}
    \end{minipage}
    \vskip\baselineskip
    
    % 3 row of images with titles on the left
    \begin{minipage}{0.143\textwidth}
        \centering
        {3DTTNet(Ours)}
    \end{minipage}%
    \begin{minipage}{0.142\textwidth}
        \centering
        \includegraphics[width=0.95\textwidth]{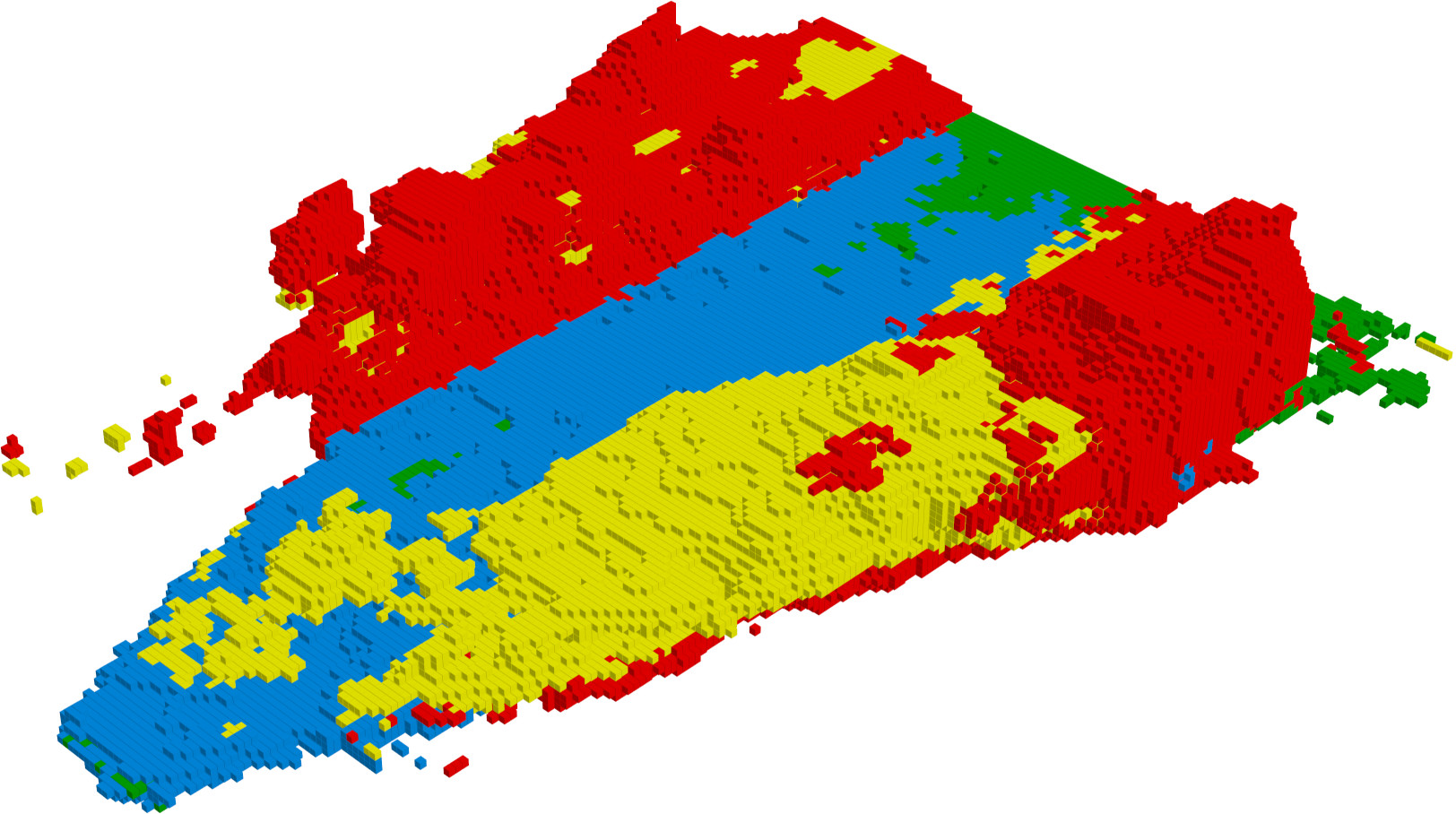}
    \end{minipage}%
    \begin{minipage}{0.142\textwidth}
        \centering
        \includegraphics[width=0.95\textwidth]{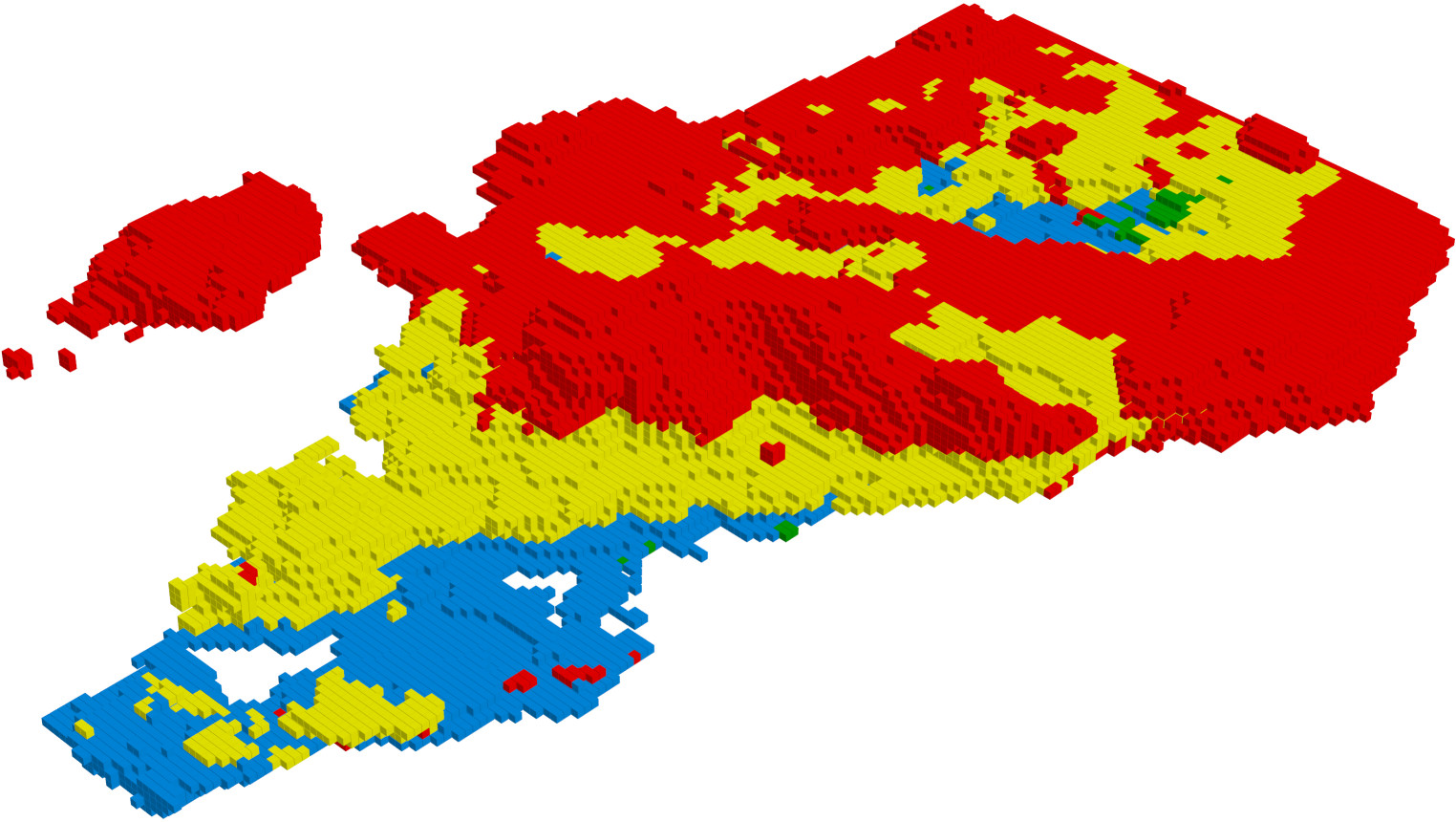}
    \end{minipage}%
    \begin{minipage}{0.142\textwidth}
        \centering
        \includegraphics[width=0.95\textwidth]{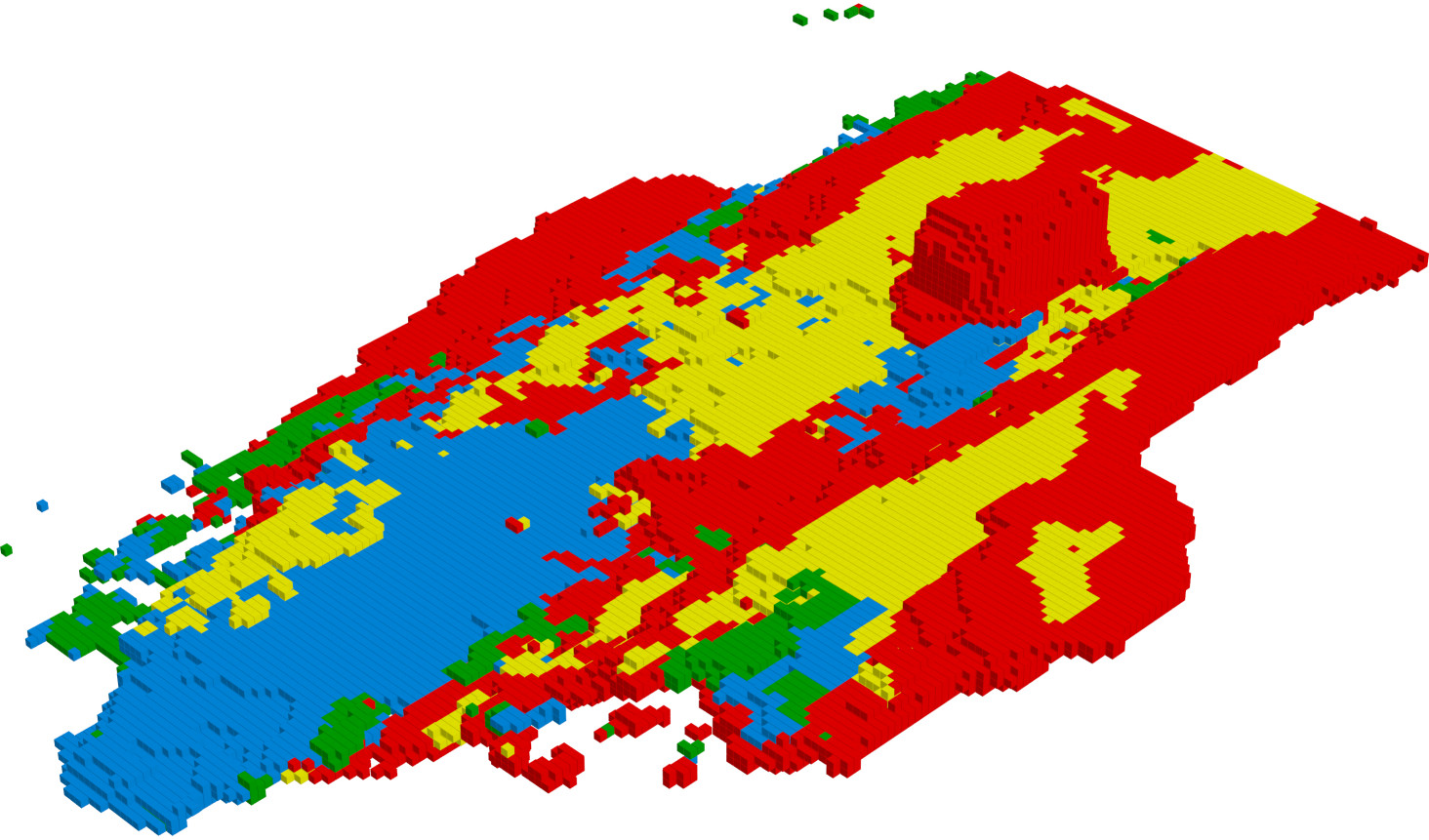}
    \end{minipage}%
    \begin{minipage}{0.142\textwidth}
        \centering
        \includegraphics[width=0.95\textwidth]{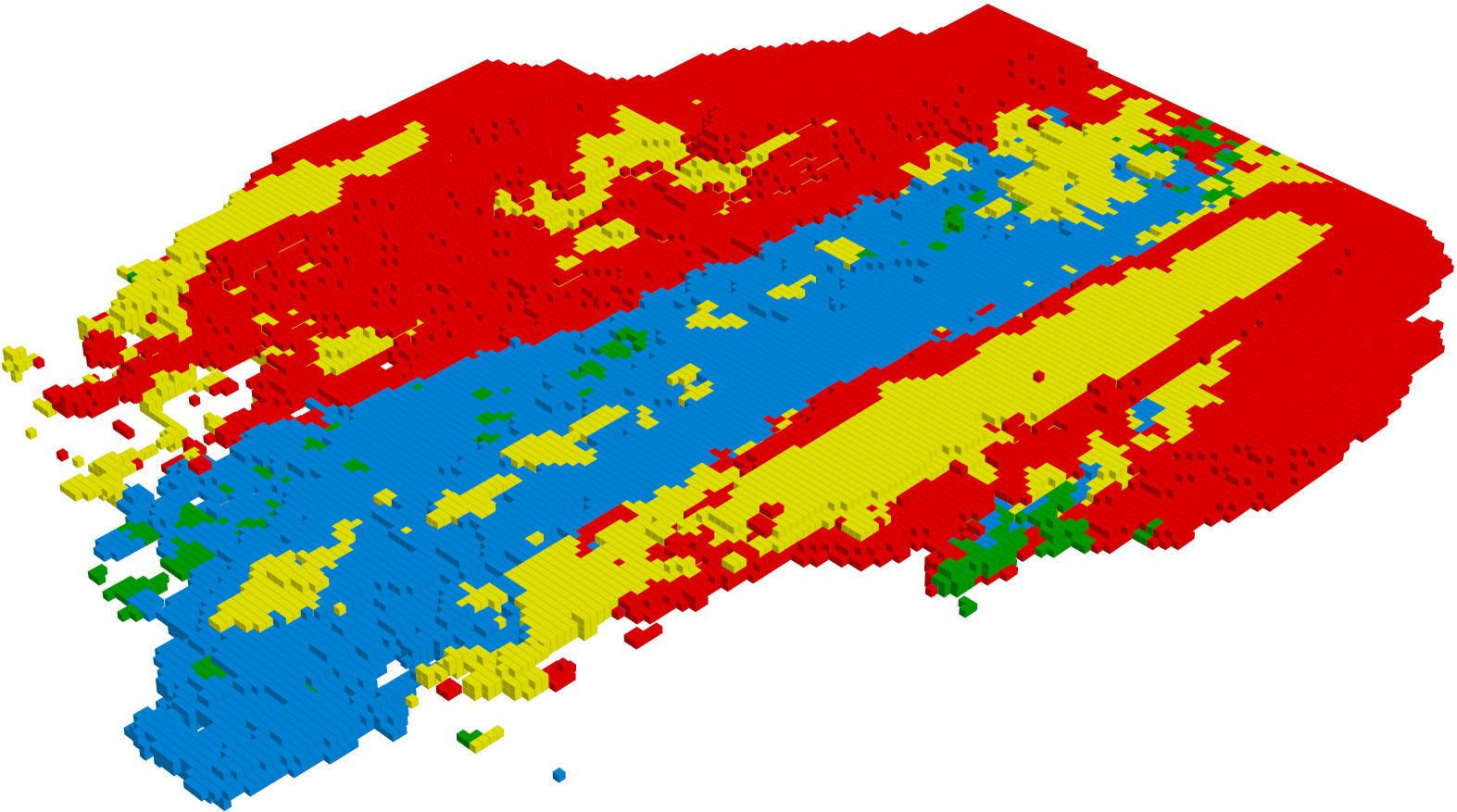}
    \end{minipage}%
    \begin{minipage}{0.142\textwidth}
        \centering
        \includegraphics[width=0.95\textwidth]{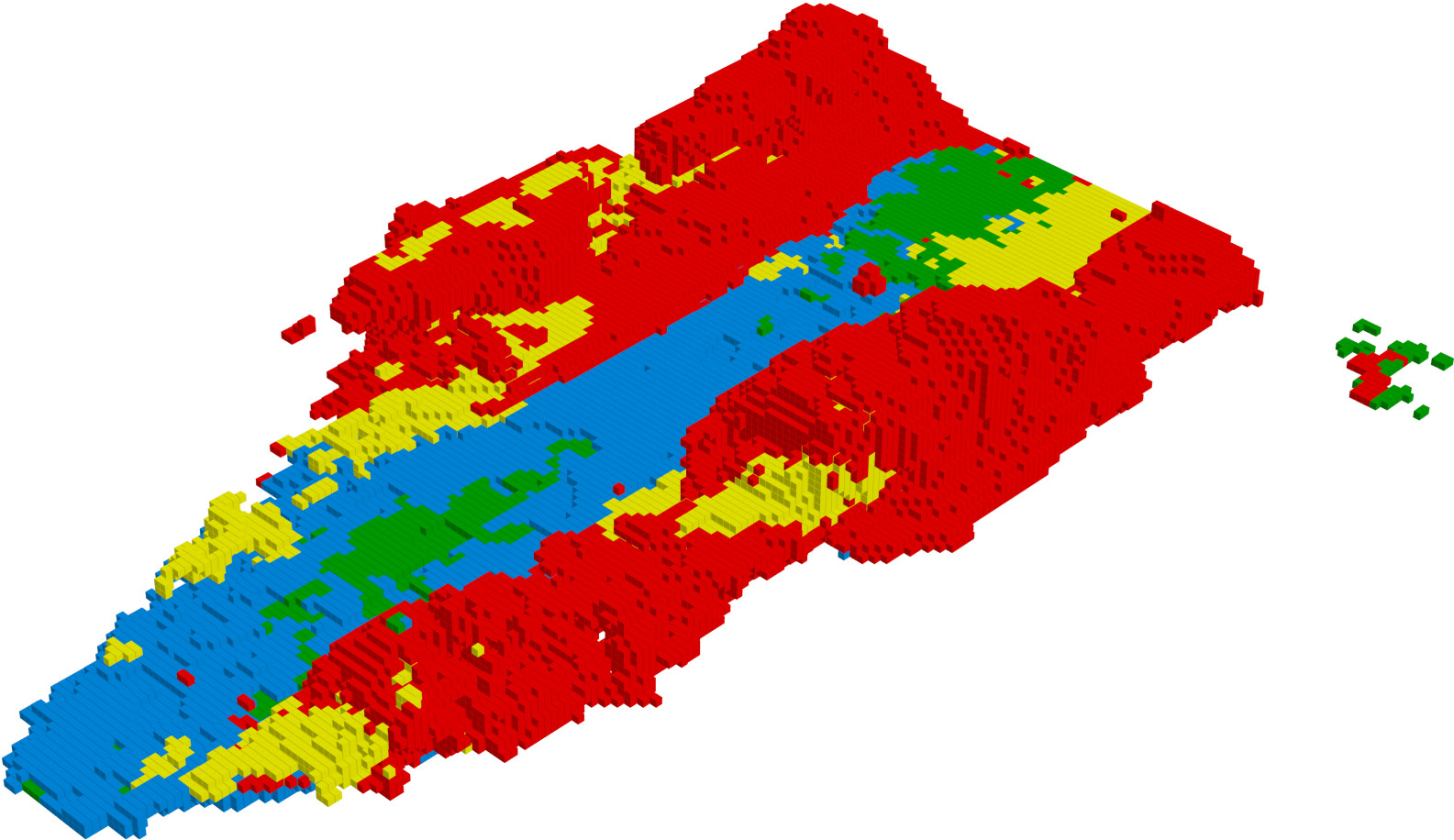}
    \end{minipage}
    \begin{minipage}{0.142\textwidth}
        \centering
        \includegraphics[width=0.95\textwidth]{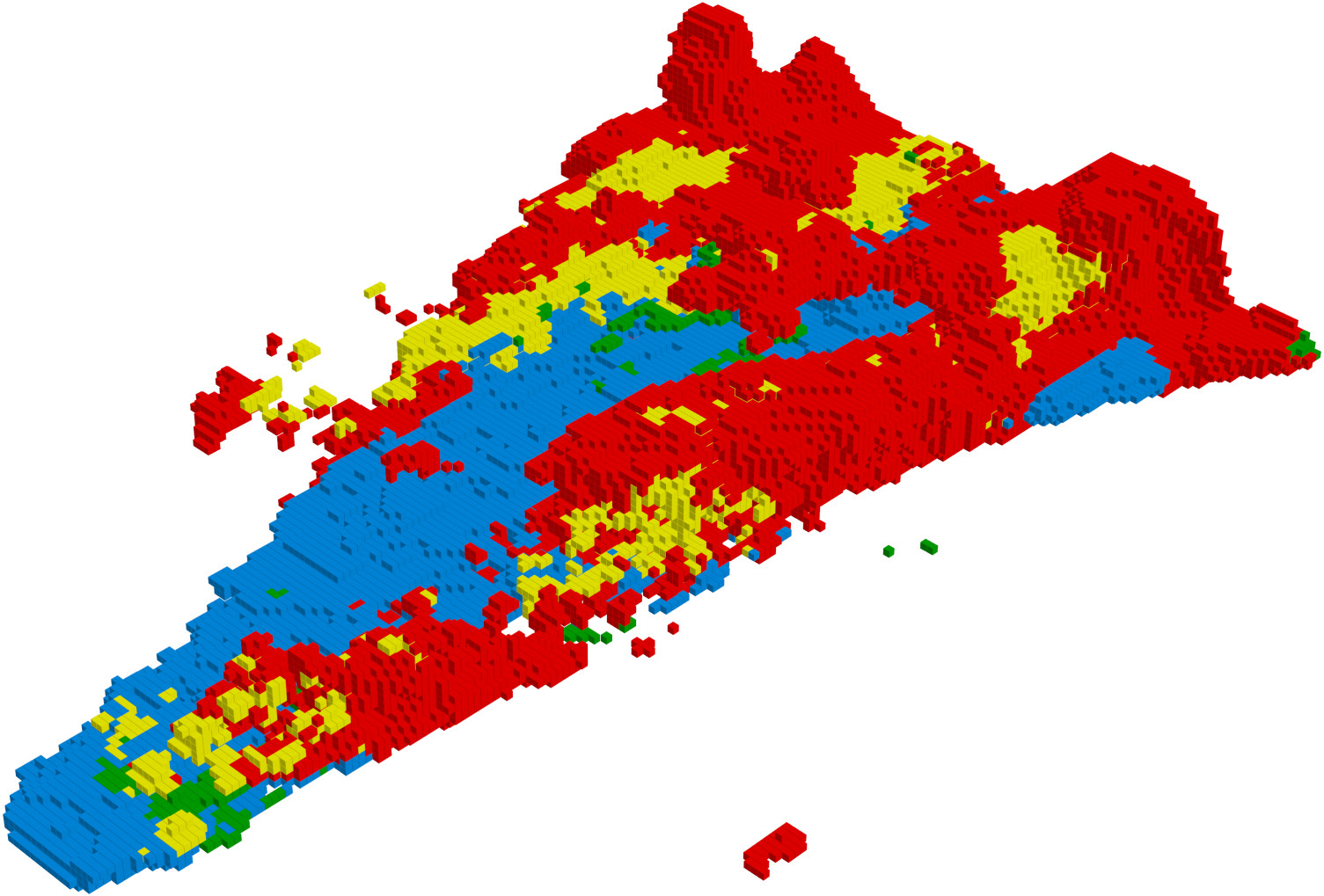}
    \end{minipage}
    \vskip\baselineskip

    % 4 row of images with titles on the left
    \begin{minipage}{0.143\textwidth}
        \centering
        {MonoScene}
    \end{minipage}%
    \begin{minipage}{0.142\textwidth}
        \centering
        \includegraphics[width=0.95\textwidth]{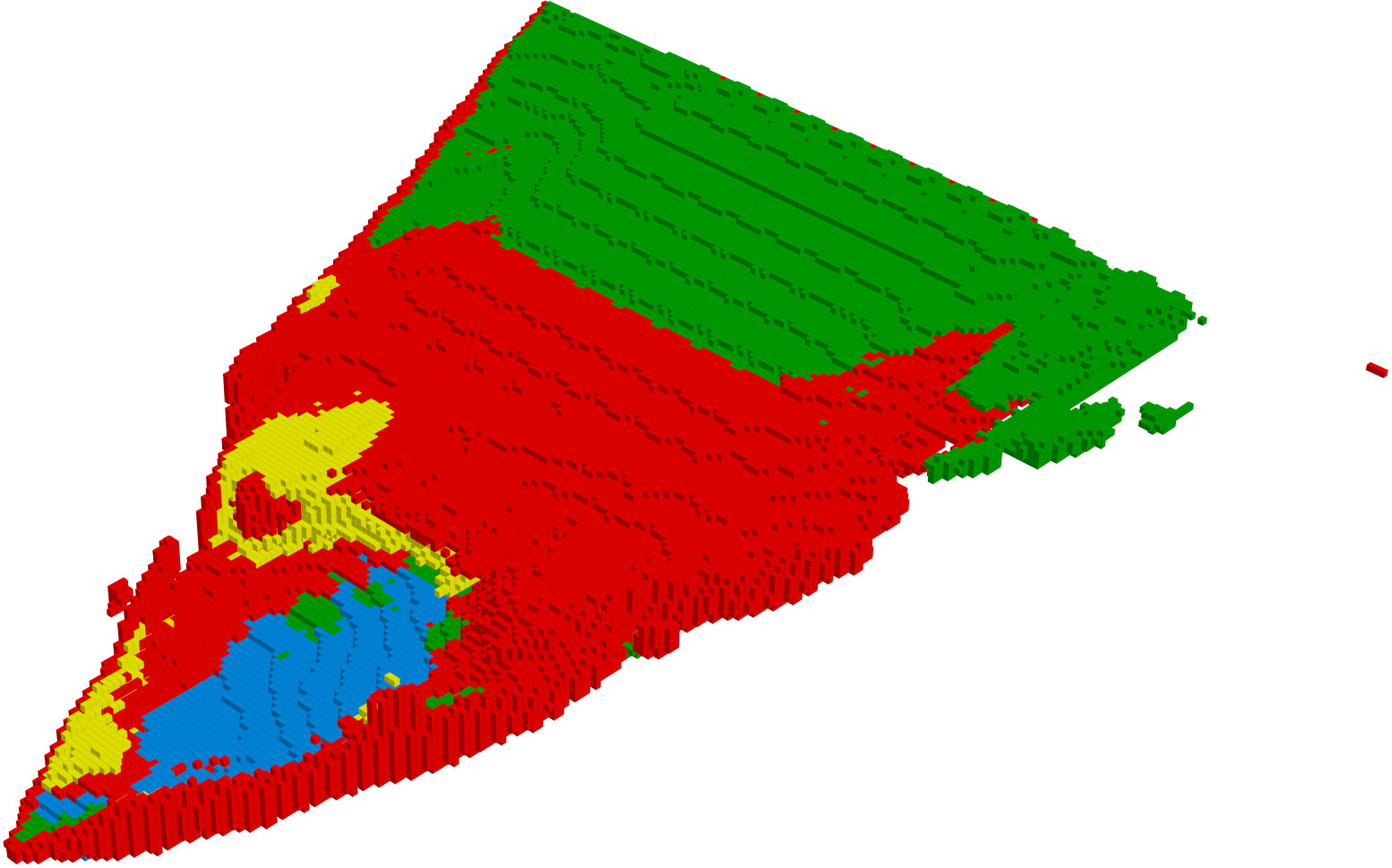}
    \end{minipage}%
    \begin{minipage}{0.142\textwidth}
        \centering
        \includegraphics[width=0.95\textwidth]{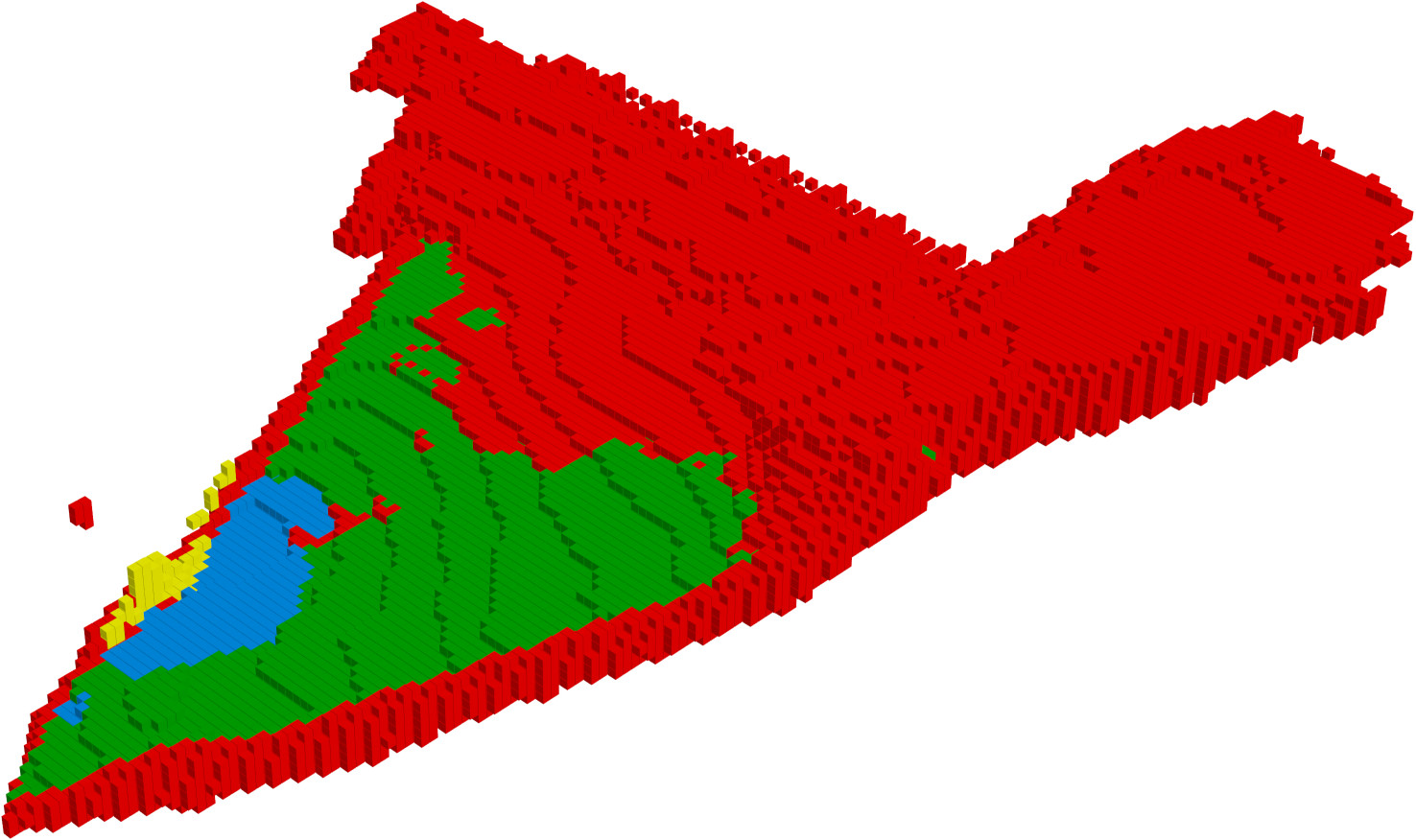}
    \end{minipage}%
    \begin{minipage}{0.142\textwidth}
        \centering
        \includegraphics[width=0.95\textwidth]{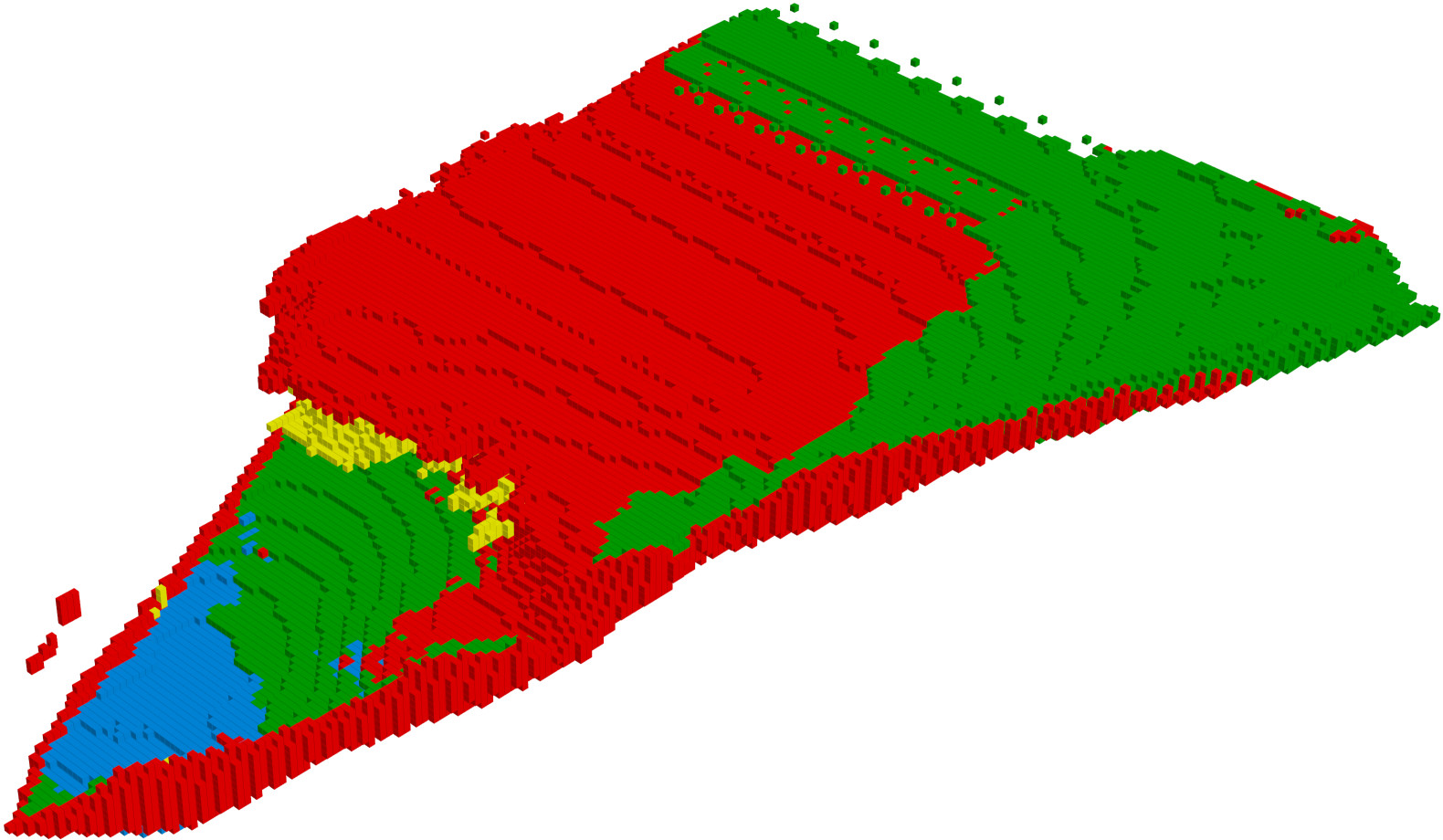}
    \end{minipage}%
    \begin{minipage}{0.142\textwidth}
        \centering
        \includegraphics[width=0.95\textwidth]{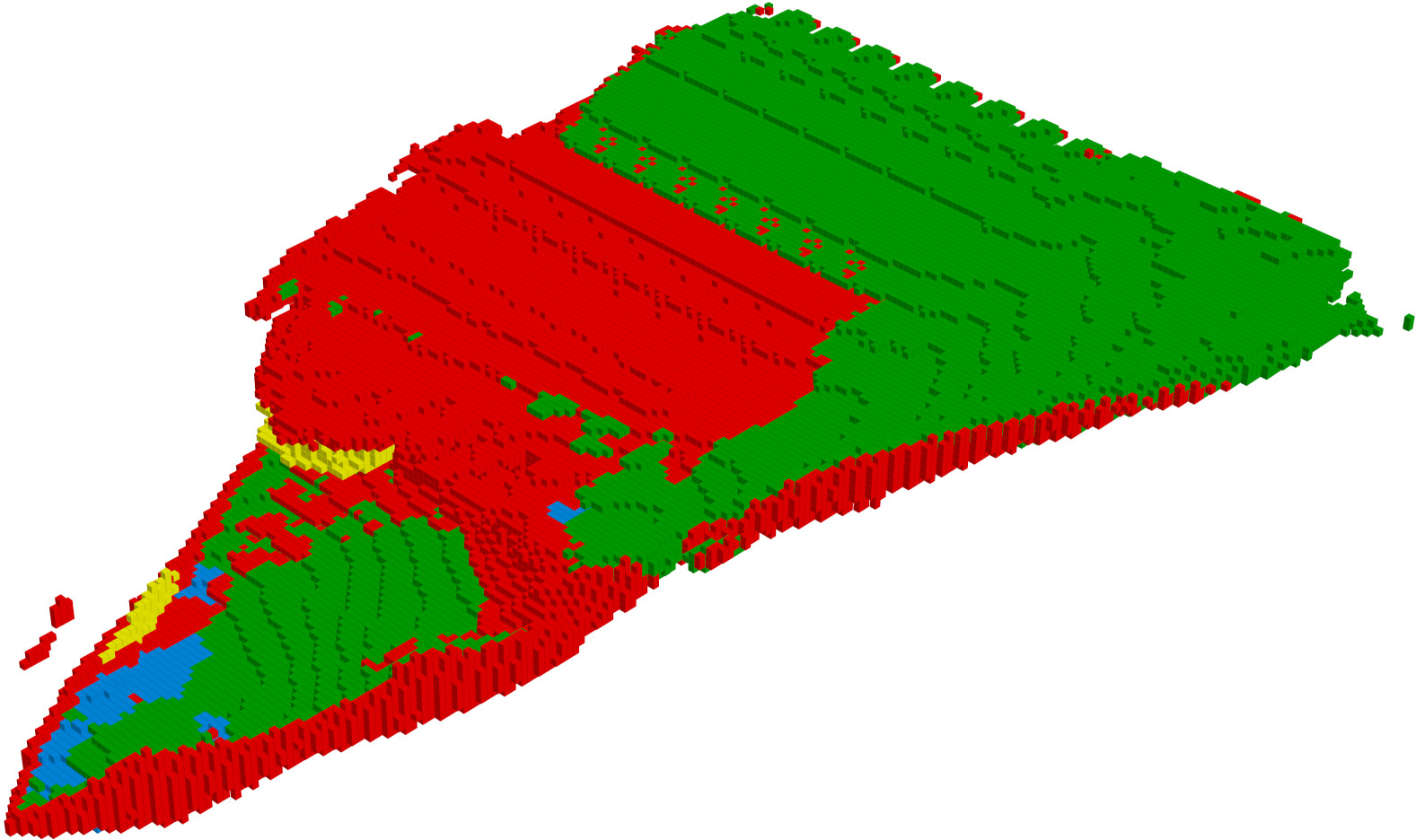}
    \end{minipage}%
    \begin{minipage}{0.142\textwidth}
        \centering
        \includegraphics[width=0.95\textwidth]{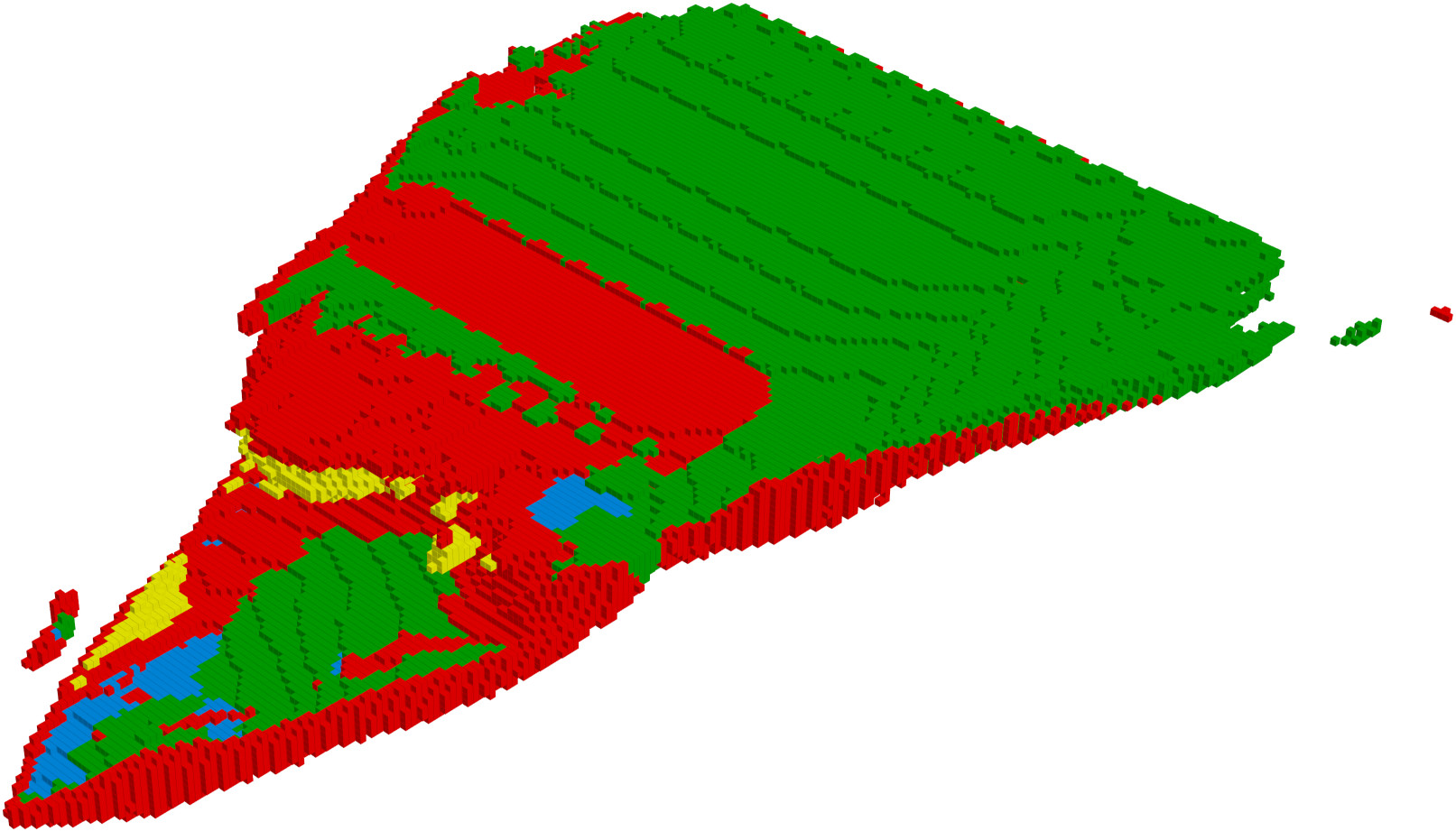}
    \end{minipage}
    \begin{minipage}{0.142\textwidth}
        \centering
        \includegraphics[width=0.95\textwidth]{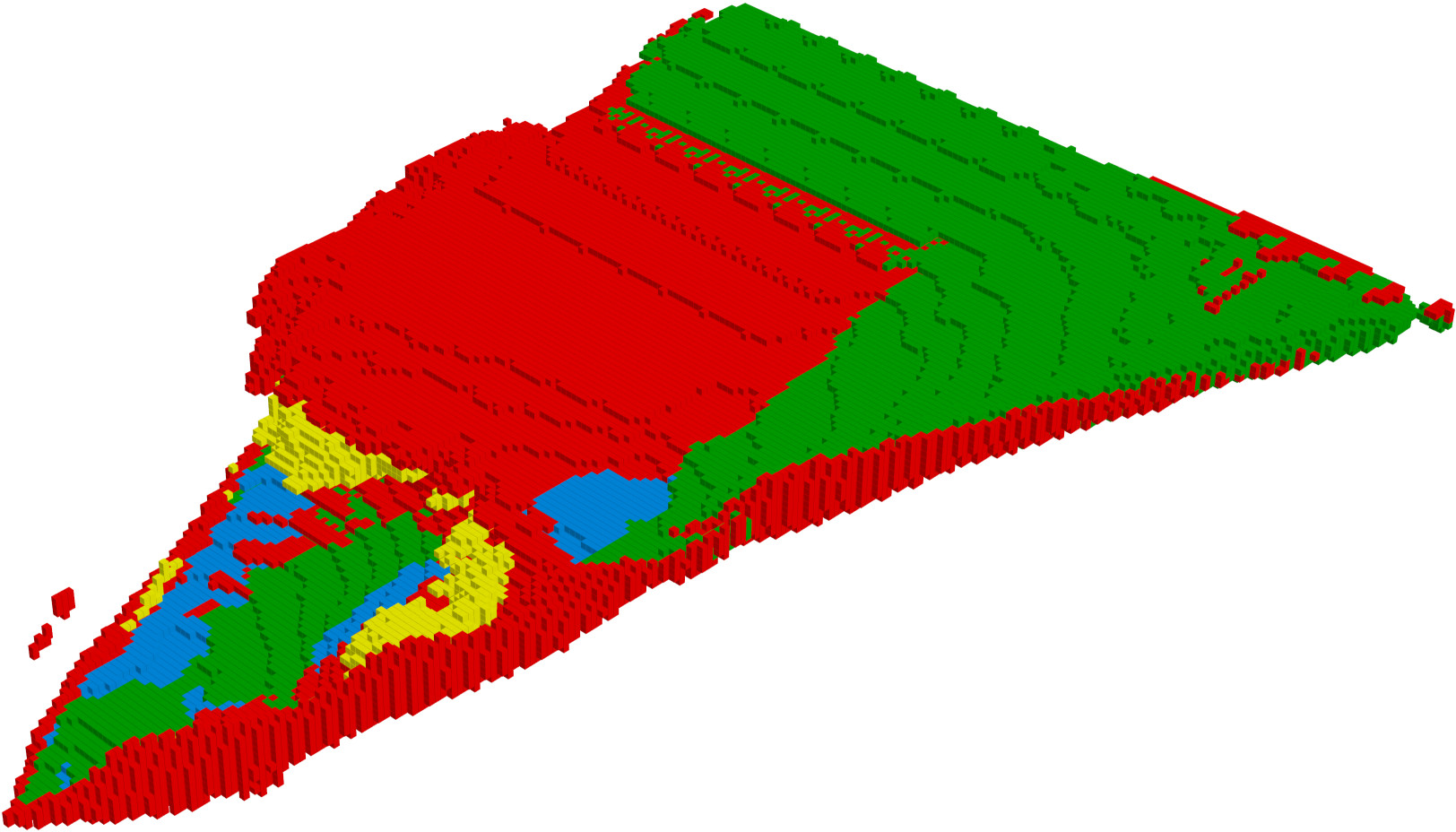}
    \end{minipage}
    \vskip\baselineskip

    % 5 row of images with titles on the left
    \begin{minipage}{0.143\textwidth}
        \centering
        {SSCNet}
    \end{minipage}%
    \begin{minipage}{0.142\textwidth}
        \centering
        \includegraphics[width=0.95\textwidth]{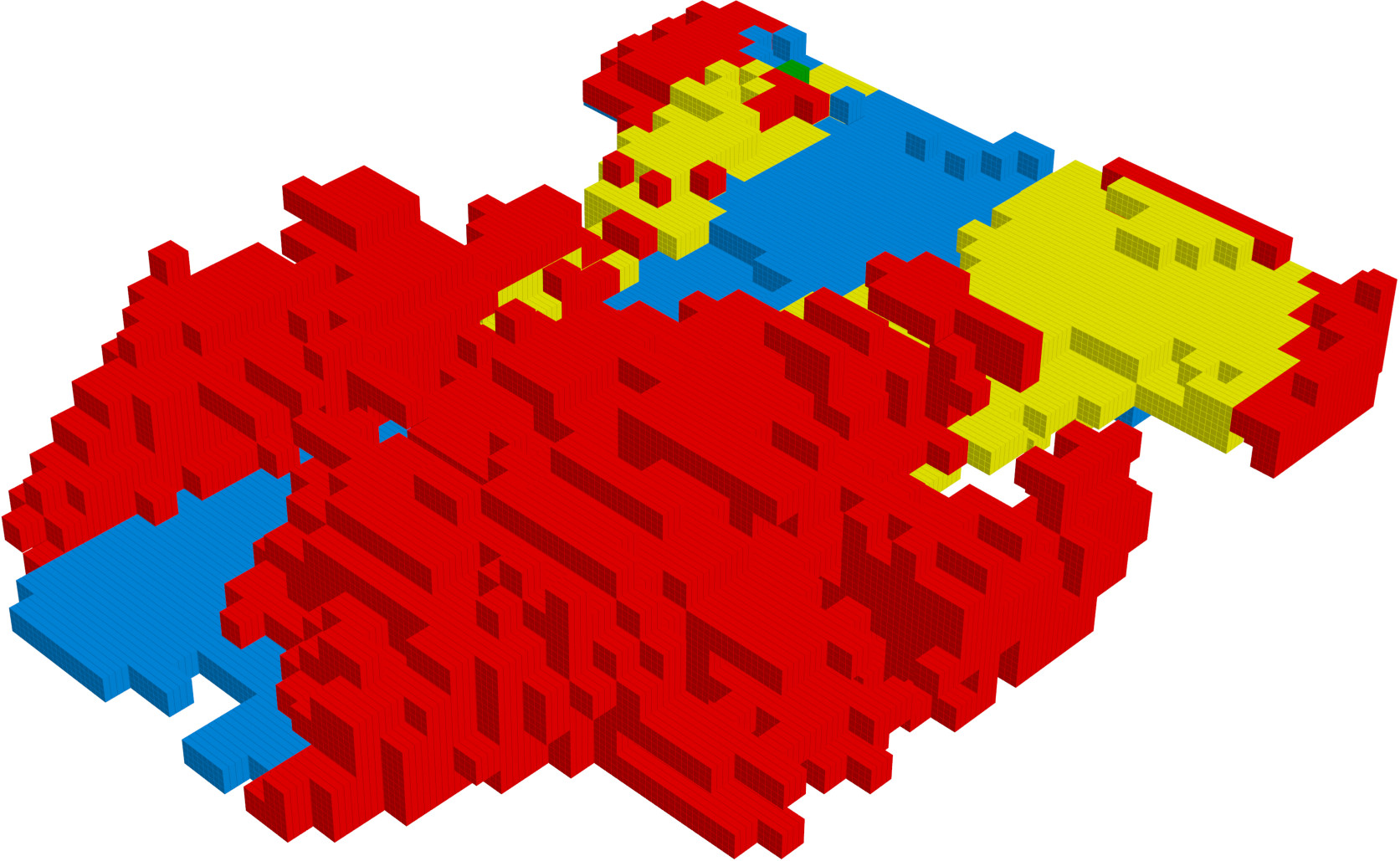}
    \end{minipage}%
    \begin{minipage}{0.142\textwidth}
        \centering
        \includegraphics[width=0.95\textwidth]{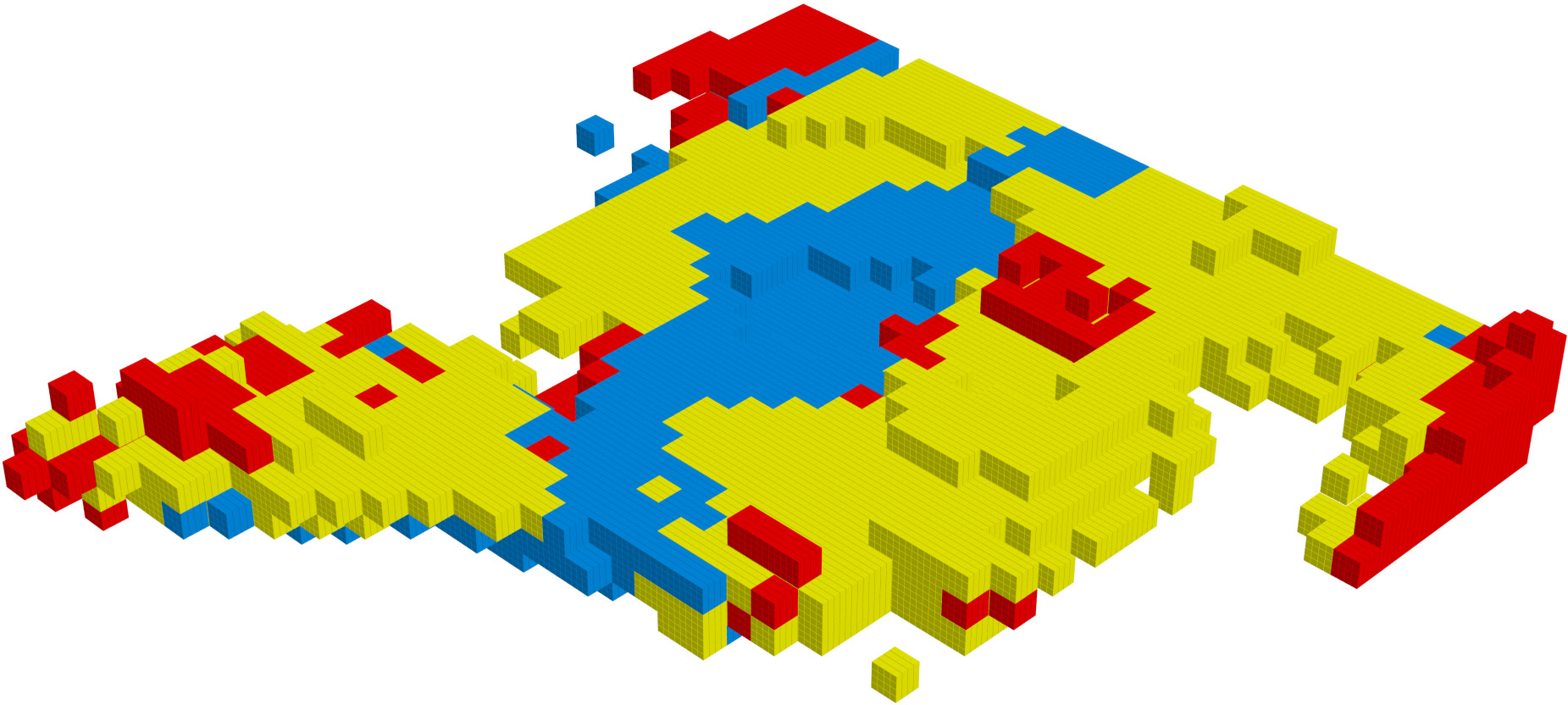}
    \end{minipage}%
    \begin{minipage}{0.142\textwidth}
        \centering
        \includegraphics[width=0.95\textwidth]{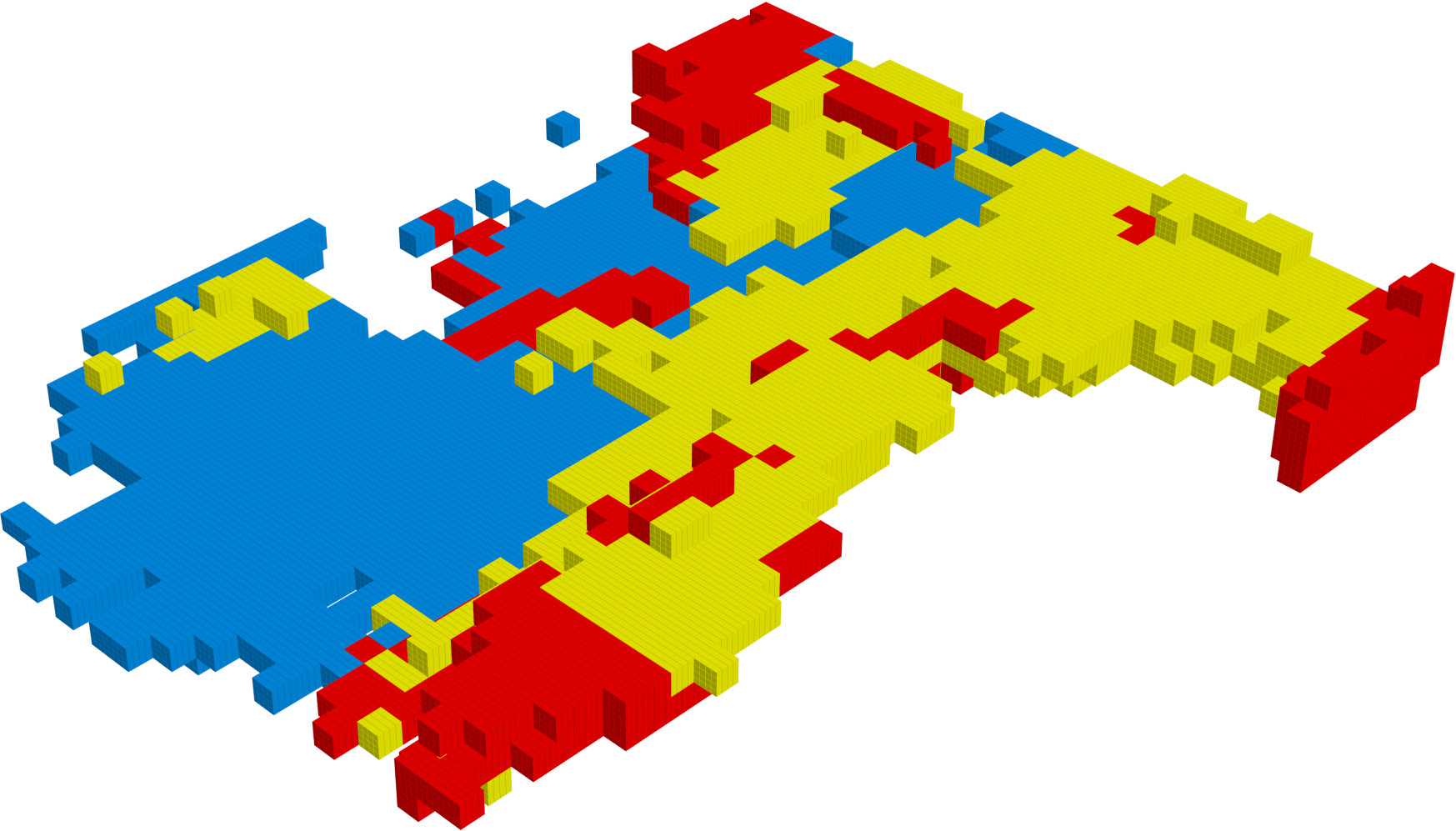}
    \end{minipage}%
    \begin{minipage}{0.142\textwidth}
        \centering
        \includegraphics[width=0.95\textwidth]{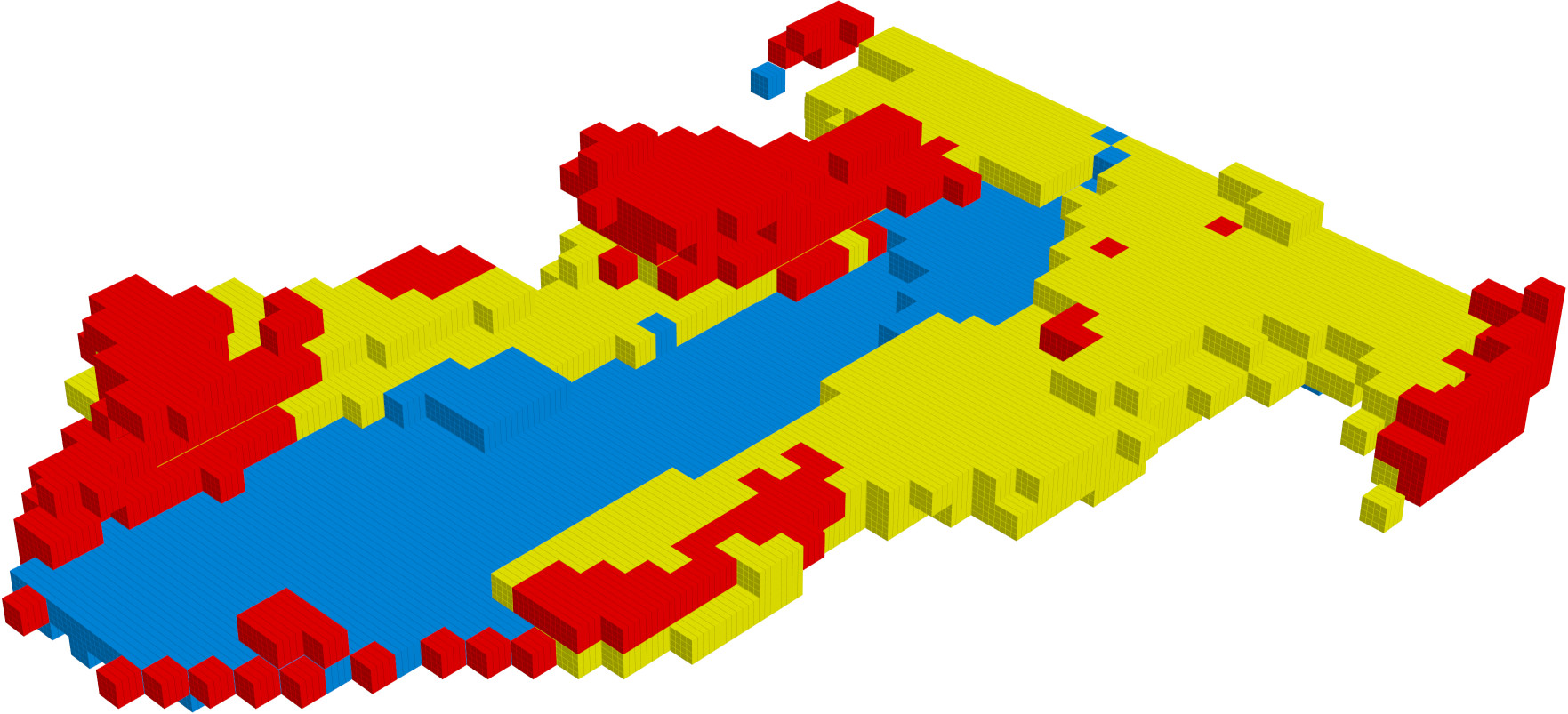}
    \end{minipage}%
    \begin{minipage}{0.142\textwidth}
        \centering
        \includegraphics[width=0.95\textwidth]{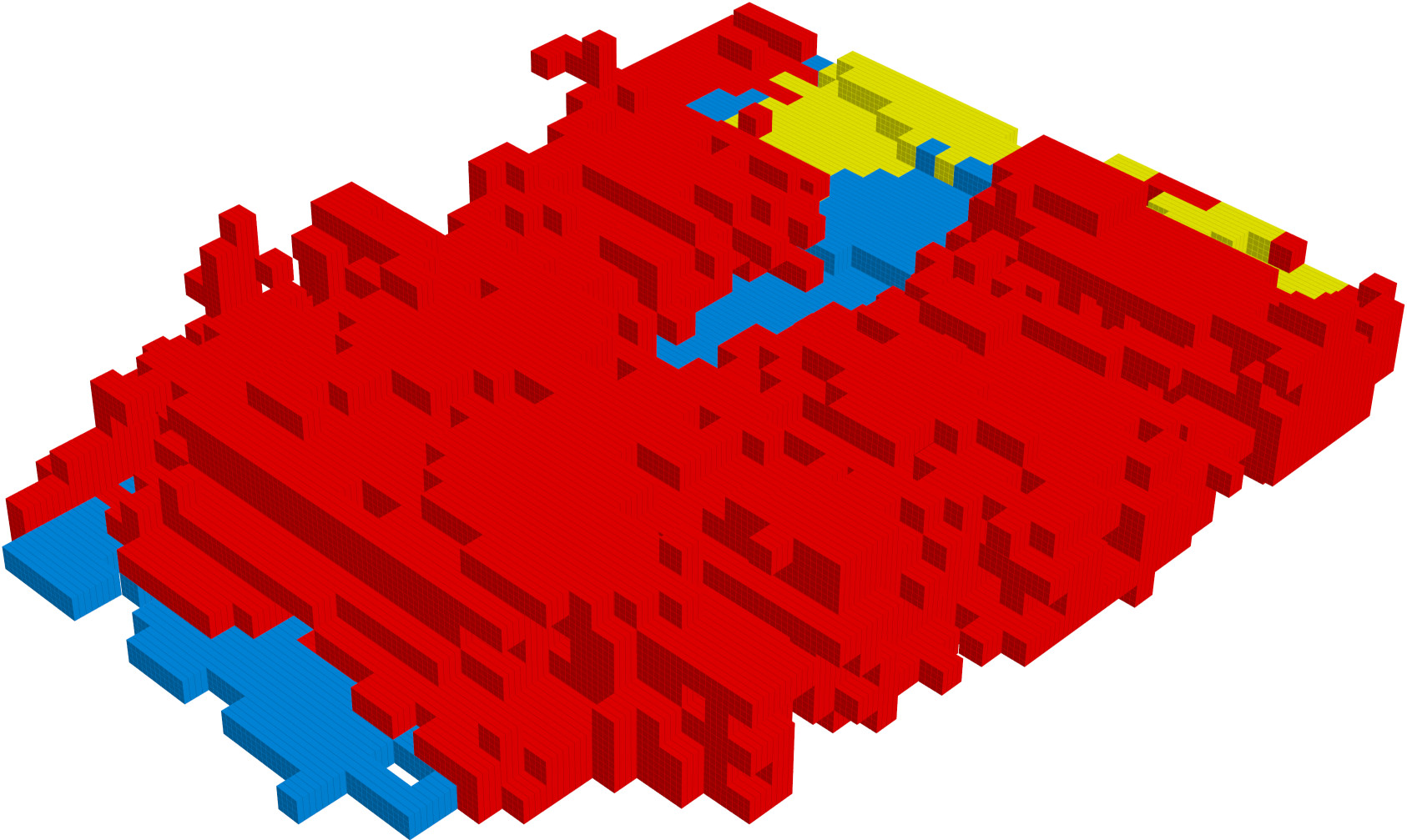}
    \end{minipage}
    \begin{minipage}{0.142\textwidth}
        \centering
        \includegraphics[width=0.95\textwidth]{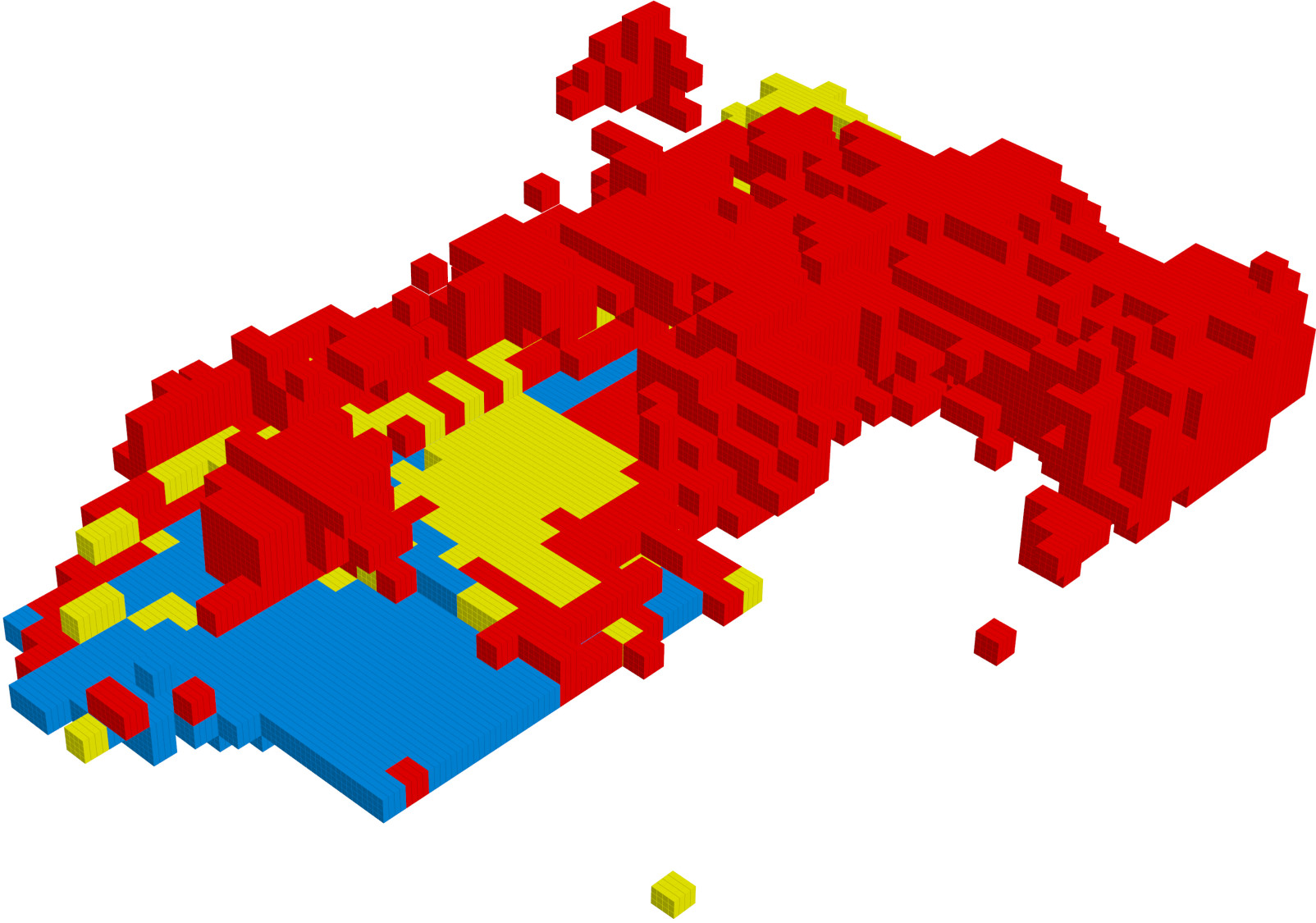}
    \end{minipage}
    \vskip\baselineskip

    % 6 row of images with titles on the left
    \begin{minipage}{0.143\textwidth}
        \centering
        {SSCNet-full}
    \end{minipage}%
    \begin{minipage}{0.142\textwidth}
        \centering
        \includegraphics[width=0.95\textwidth]{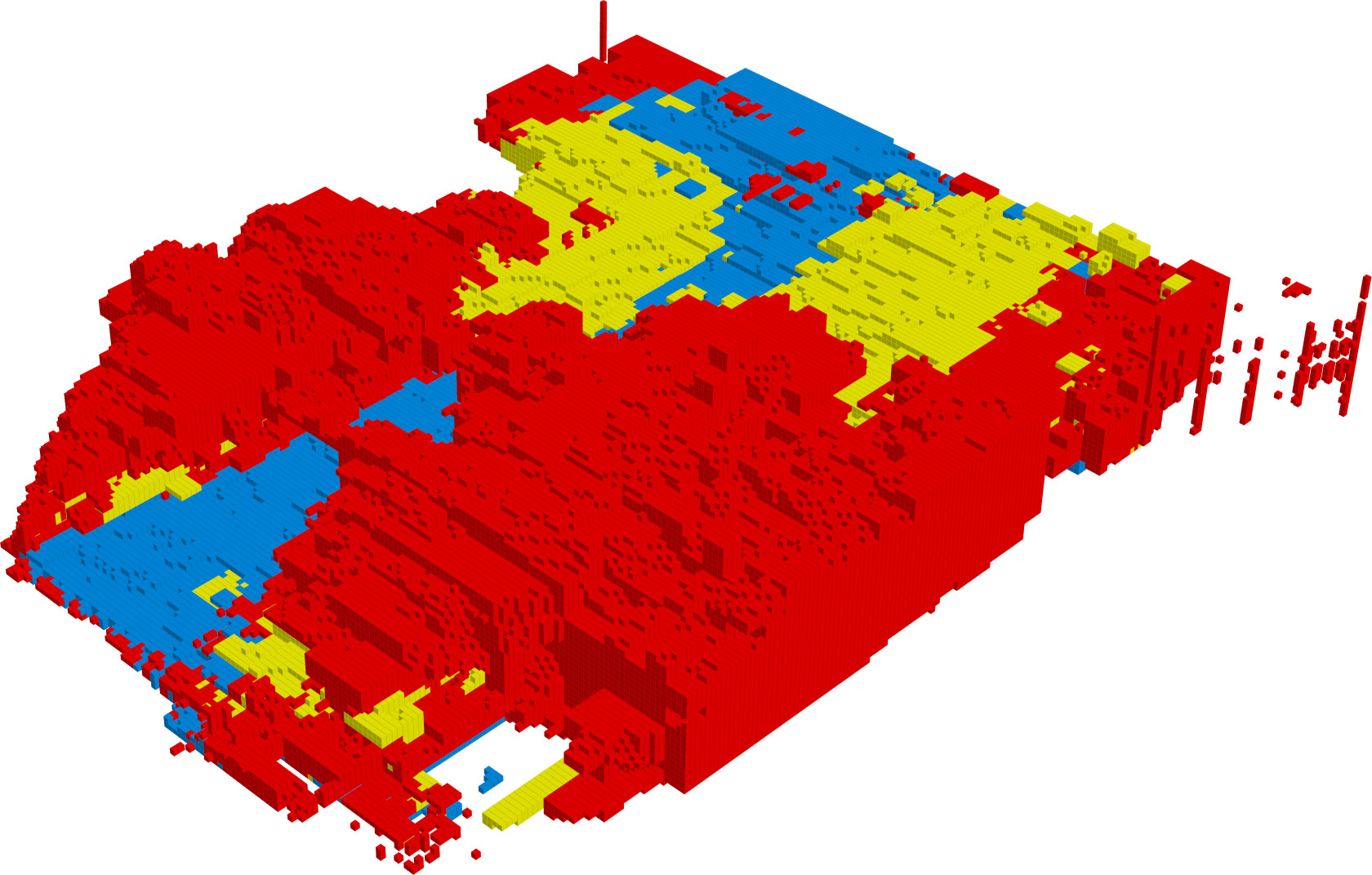}
    \end{minipage}%
    \begin{minipage}{0.142\textwidth}
        \centering
        \includegraphics[width=0.95\textwidth]{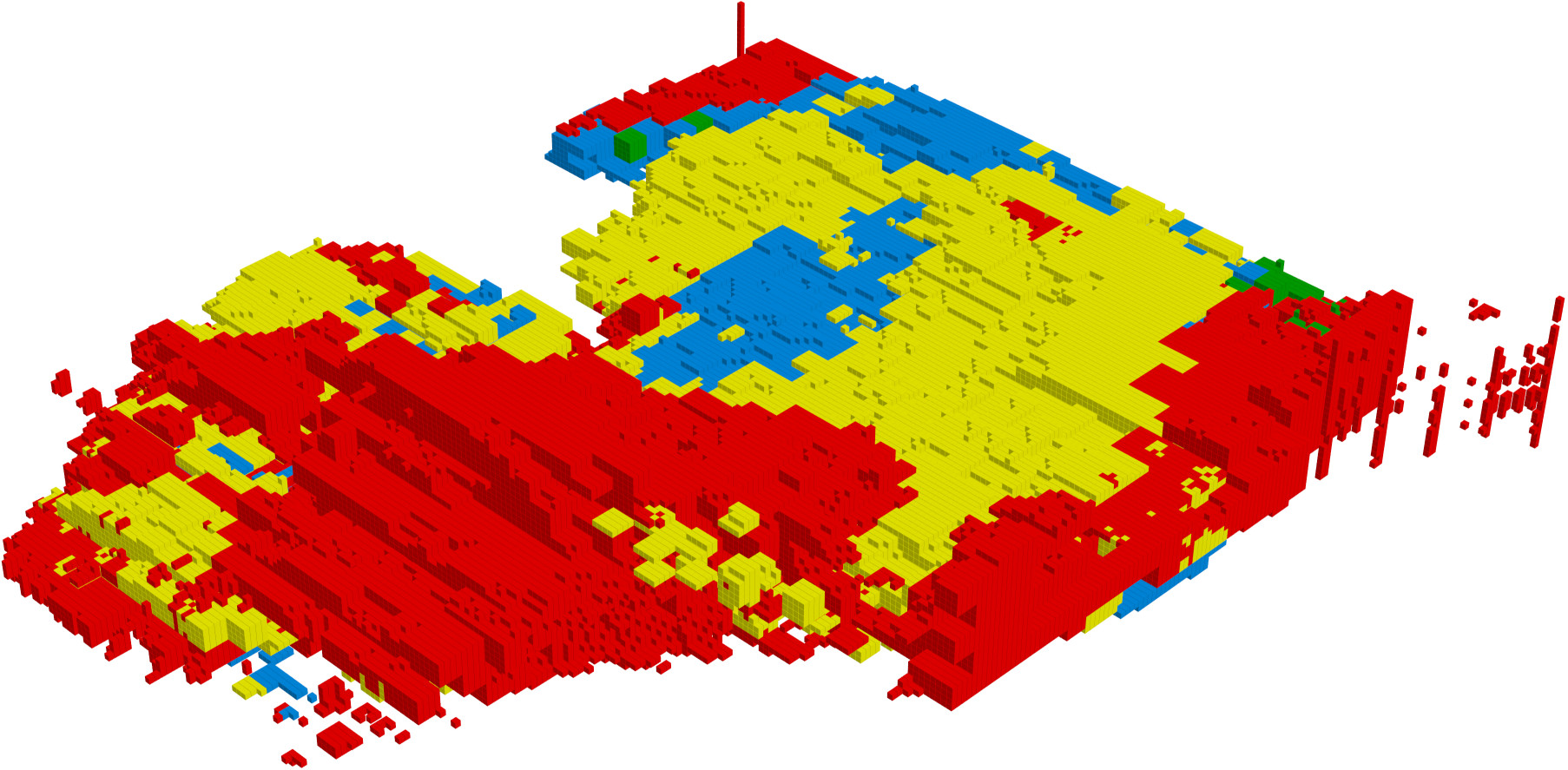}
    \end{minipage}%
    \begin{minipage}{0.142\textwidth}
        \centering
        \includegraphics[width=0.95\textwidth]{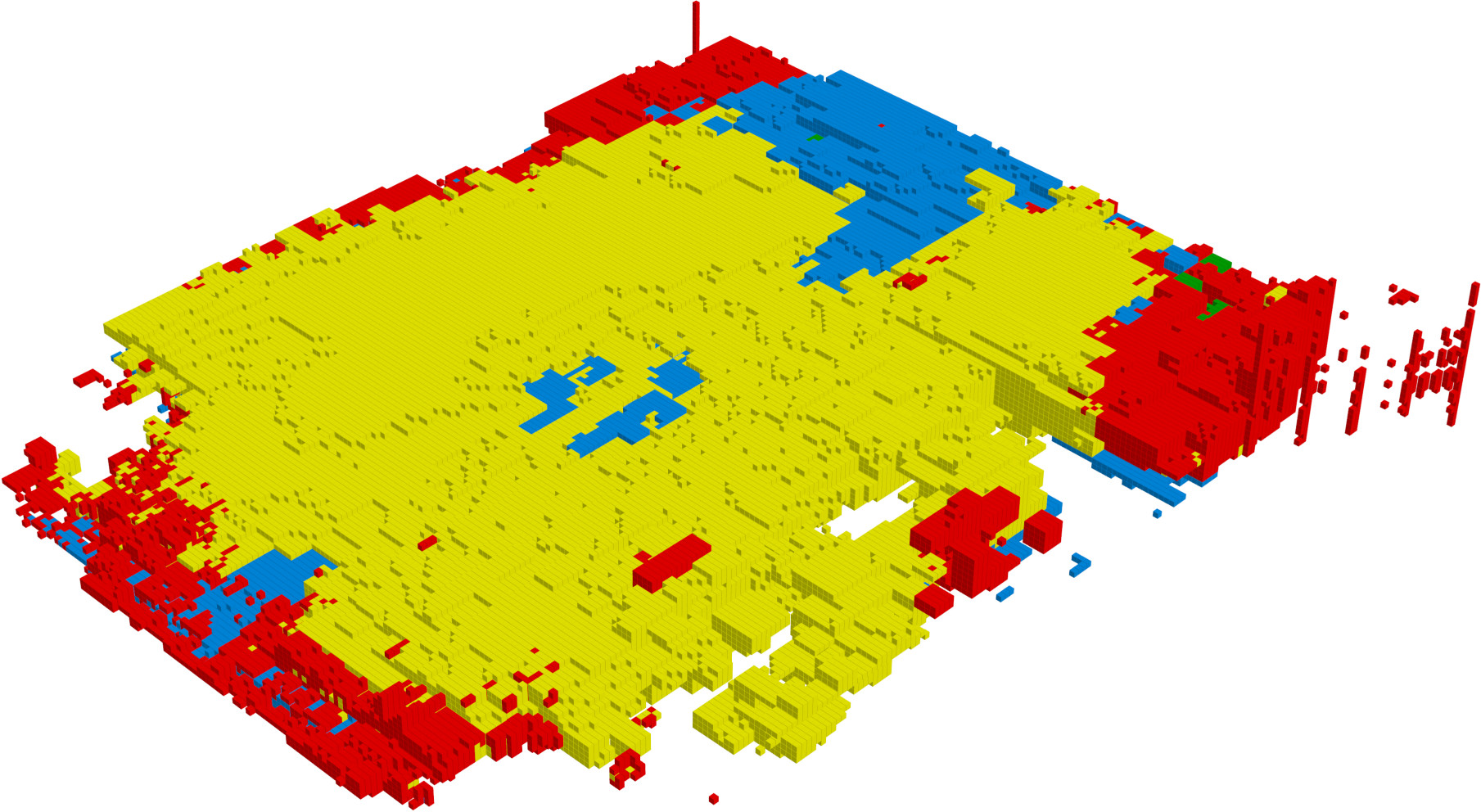}
    \end{minipage}%
    \begin{minipage}{0.142\textwidth}
        \centering
        \includegraphics[width=0.95\textwidth]{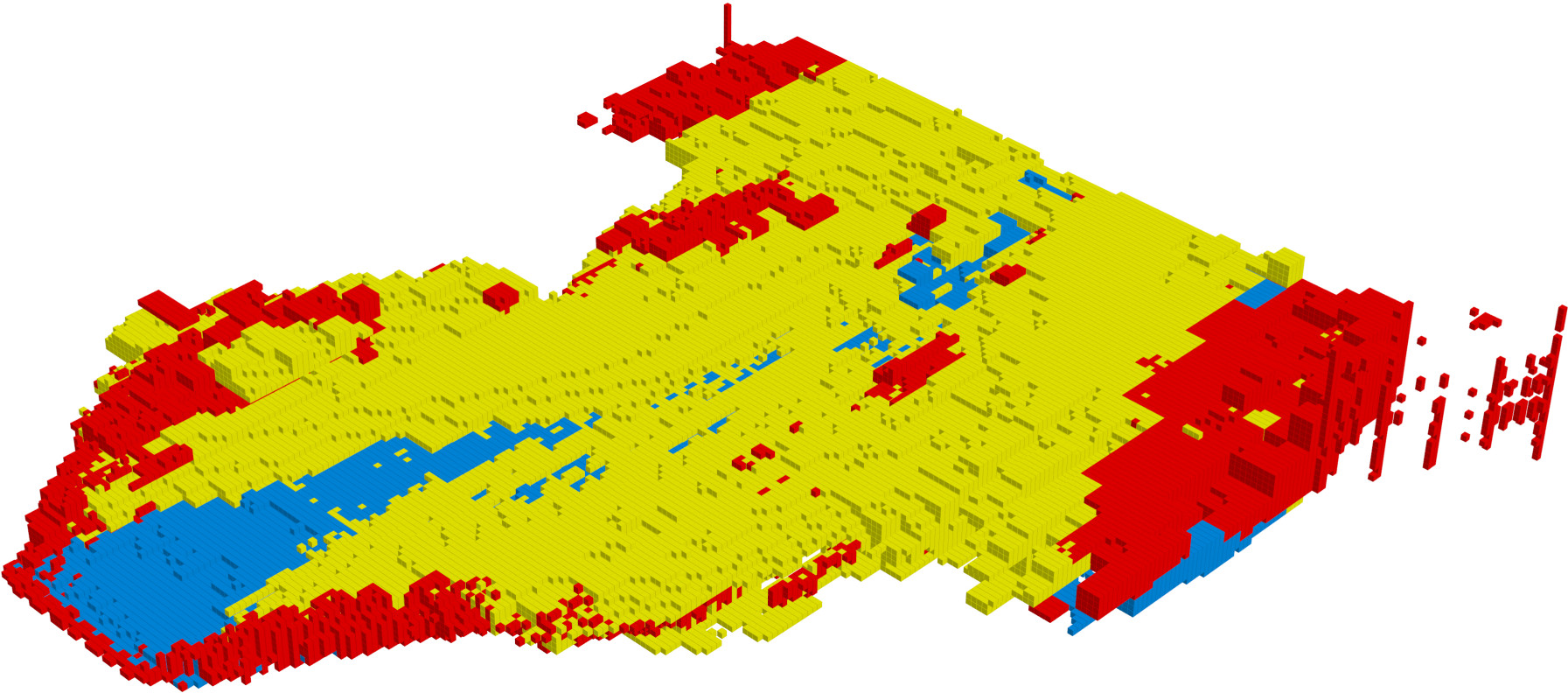}
    \end{minipage}%
    \begin{minipage}{0.142\textwidth}
        \centering
        \includegraphics[width=0.95\textwidth]{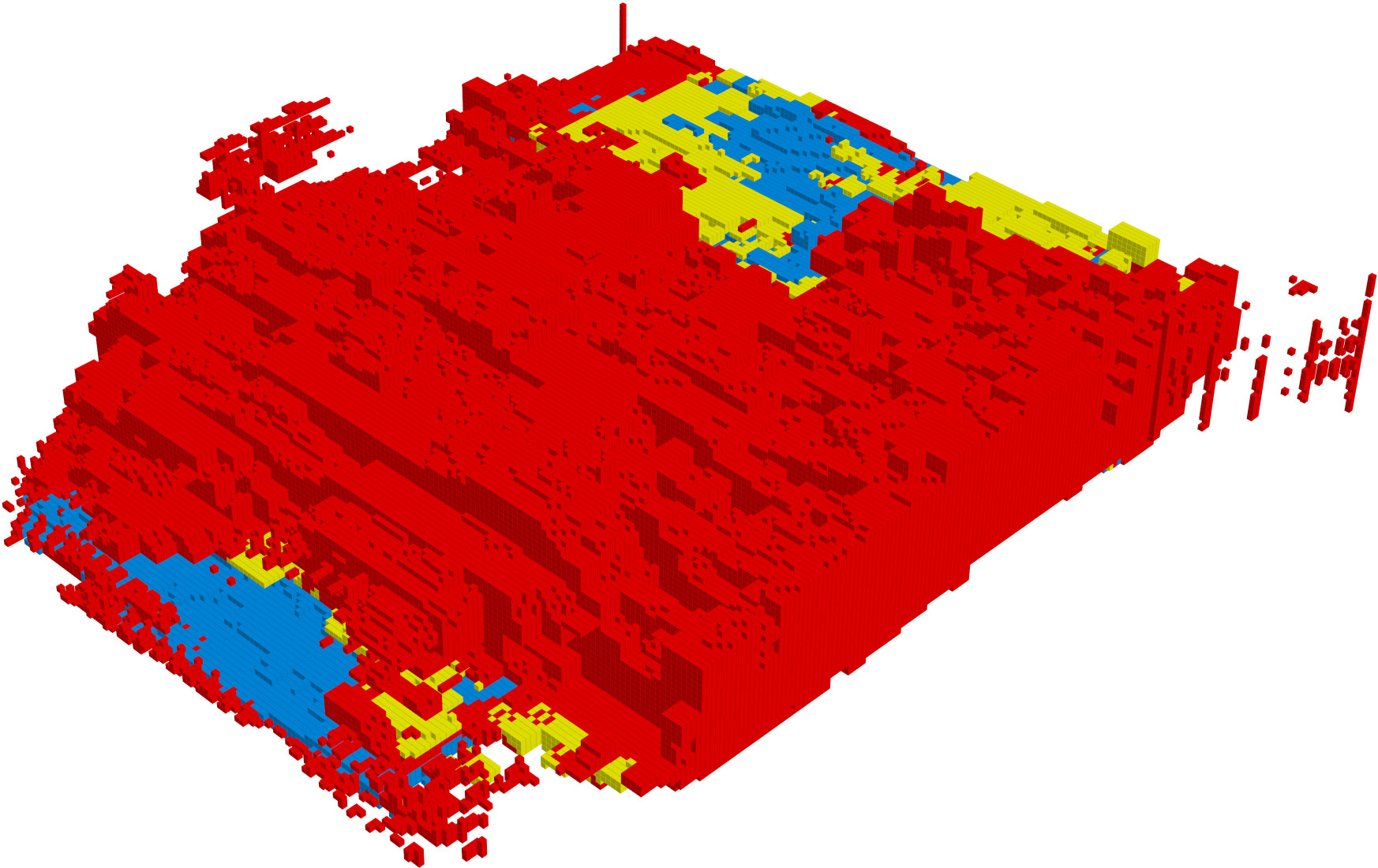}
    \end{minipage}
    \begin{minipage}{0.142\textwidth}
        \centering
        \includegraphics[width=0.95\textwidth]{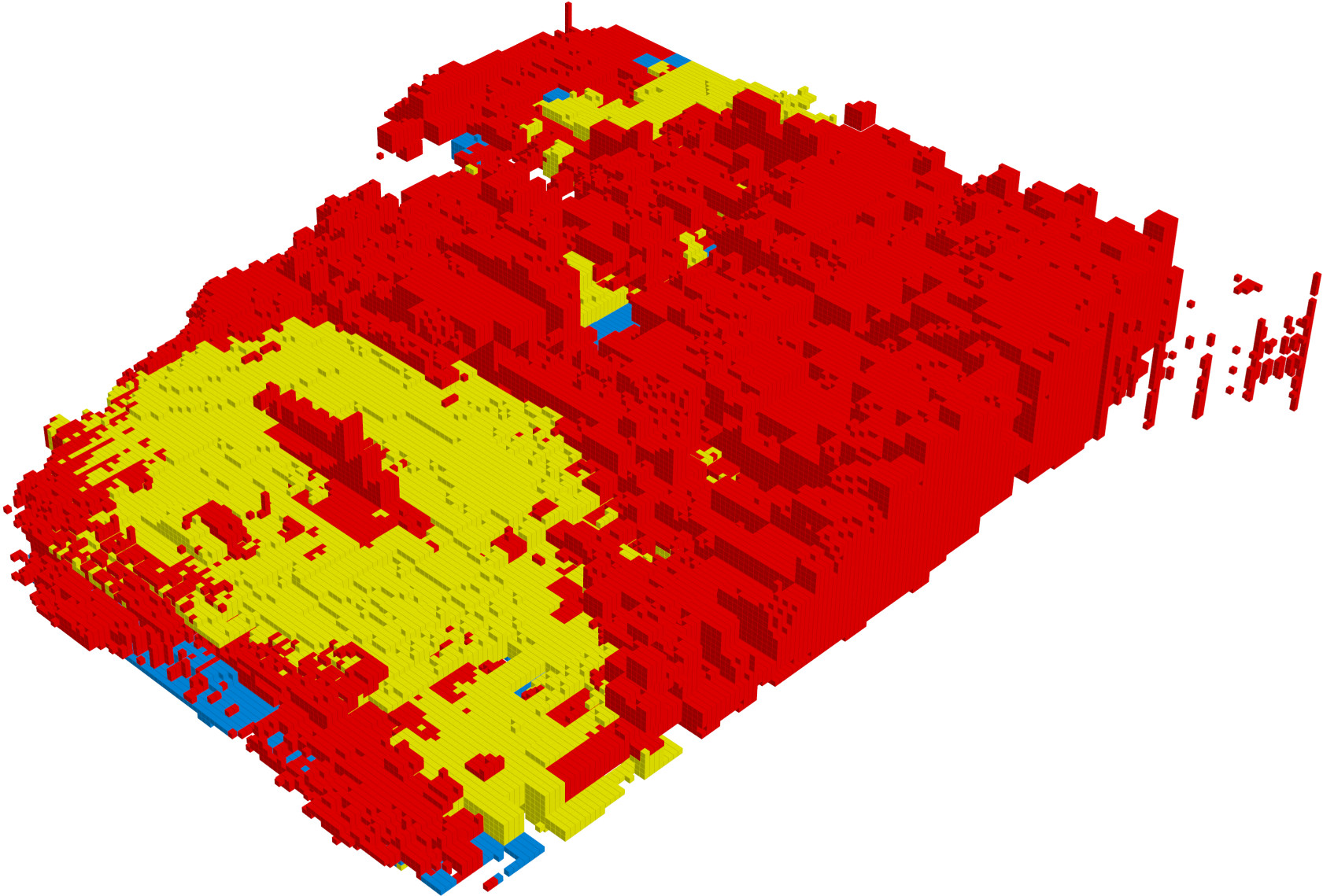}
    \end{minipage}
    \vskip\baselineskip

    % 7 row of images with titles on the left
    \begin{minipage}{0.143\textwidth}
        \centering
        {LMSCNet}
    \end{minipage}%
    \begin{minipage}{0.142\textwidth}
        \centering
        \includegraphics[width=0.95\textwidth]{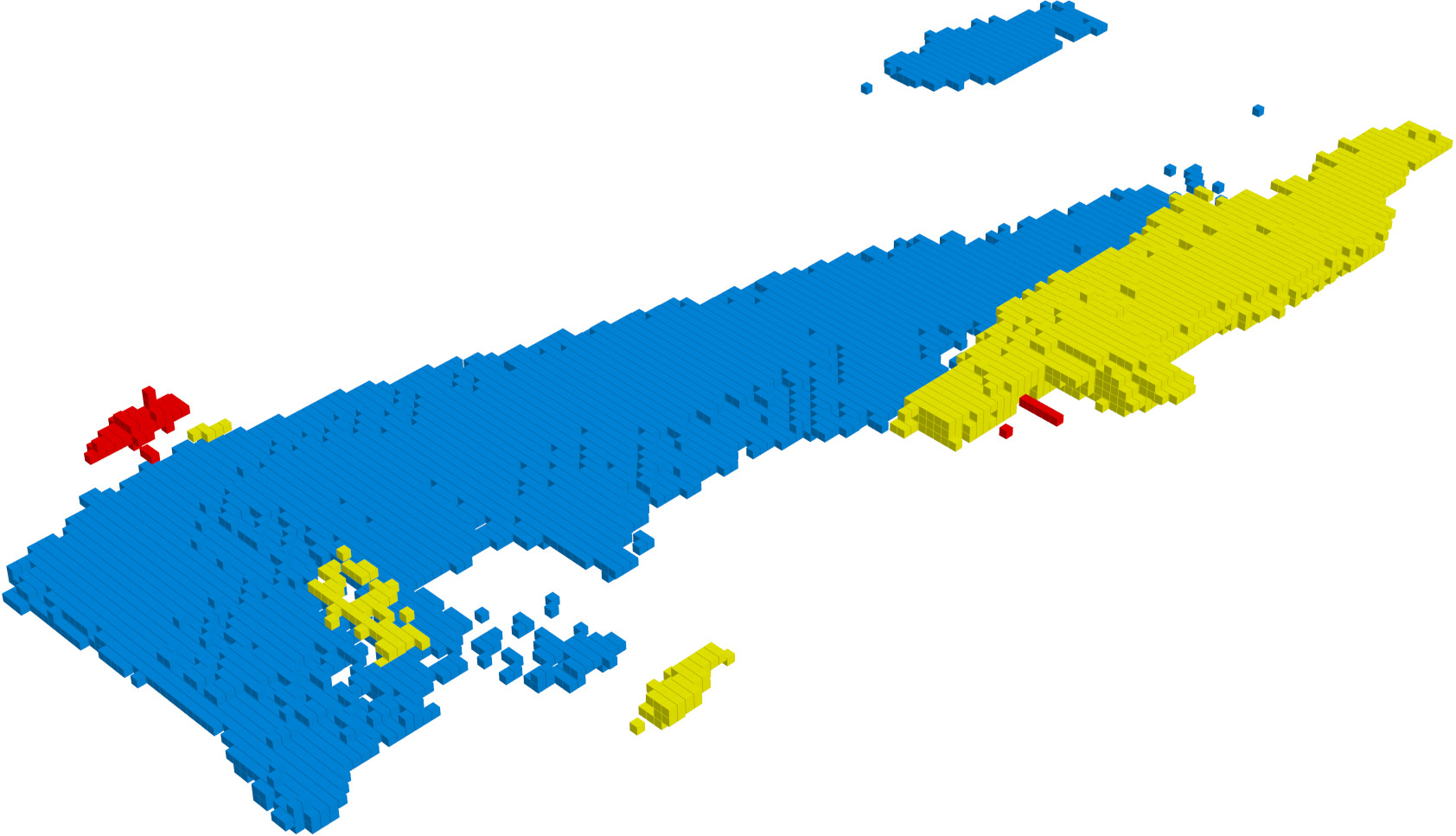}
    \end{minipage}%
    \begin{minipage}{0.142\textwidth}
        \centering
        \includegraphics[width=0.95\textwidth]{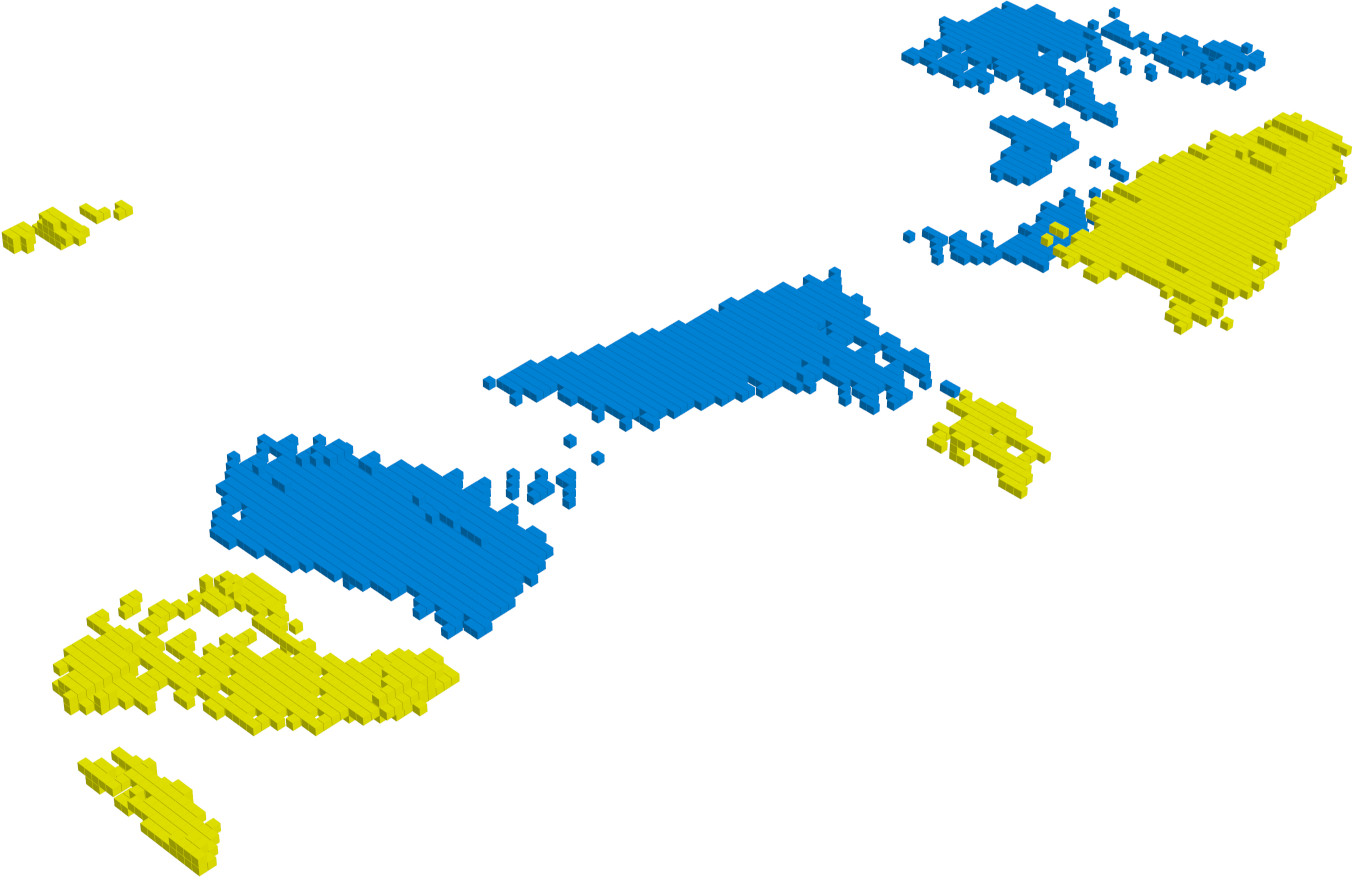}
    \end{minipage}%
    \begin{minipage}{0.142\textwidth}
        \centering
        \includegraphics[width=0.95\textwidth]{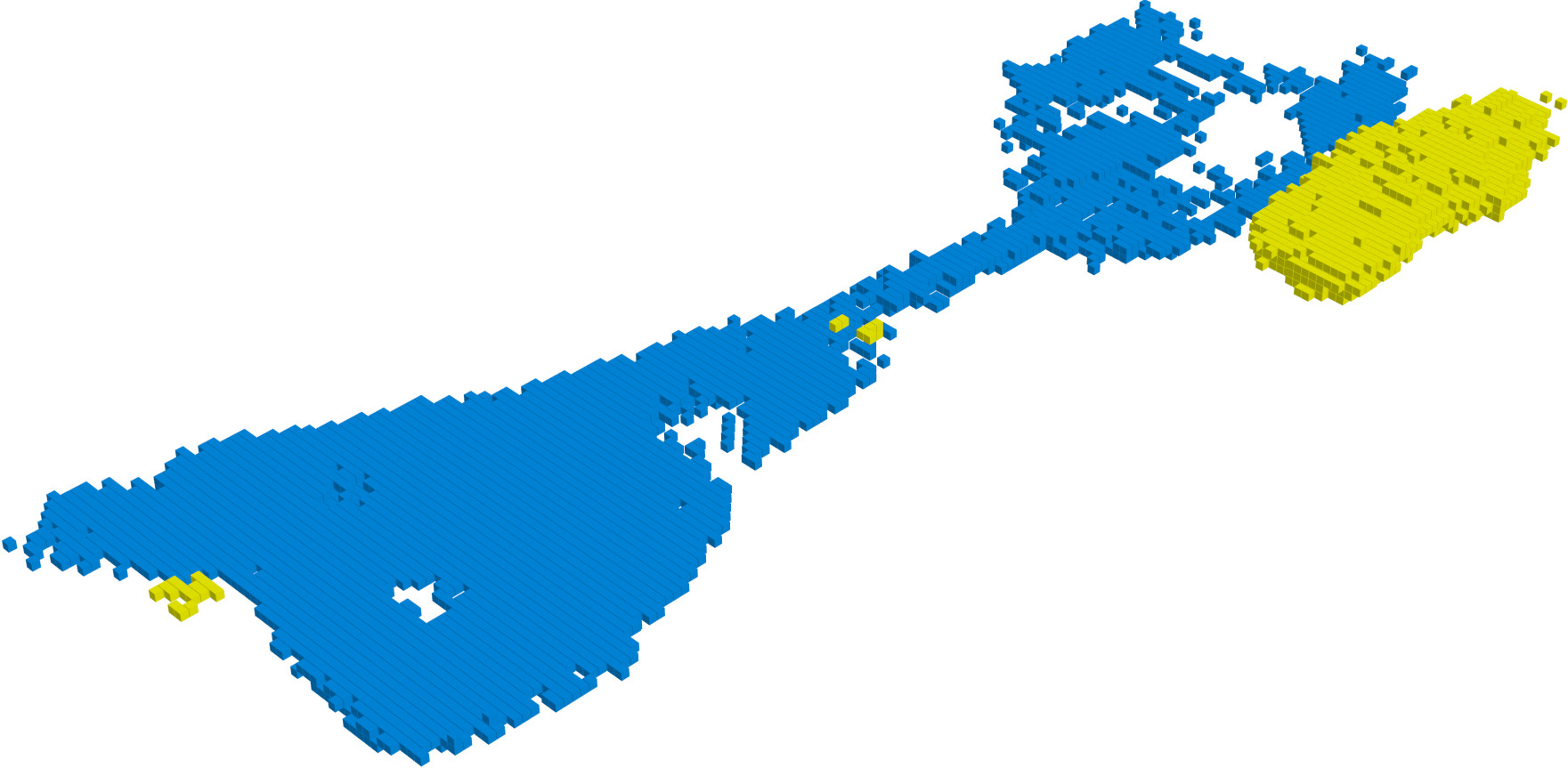}
    \end{minipage}%
    \begin{minipage}{0.142\textwidth}
        \centering
        \includegraphics[width=0.95\textwidth]{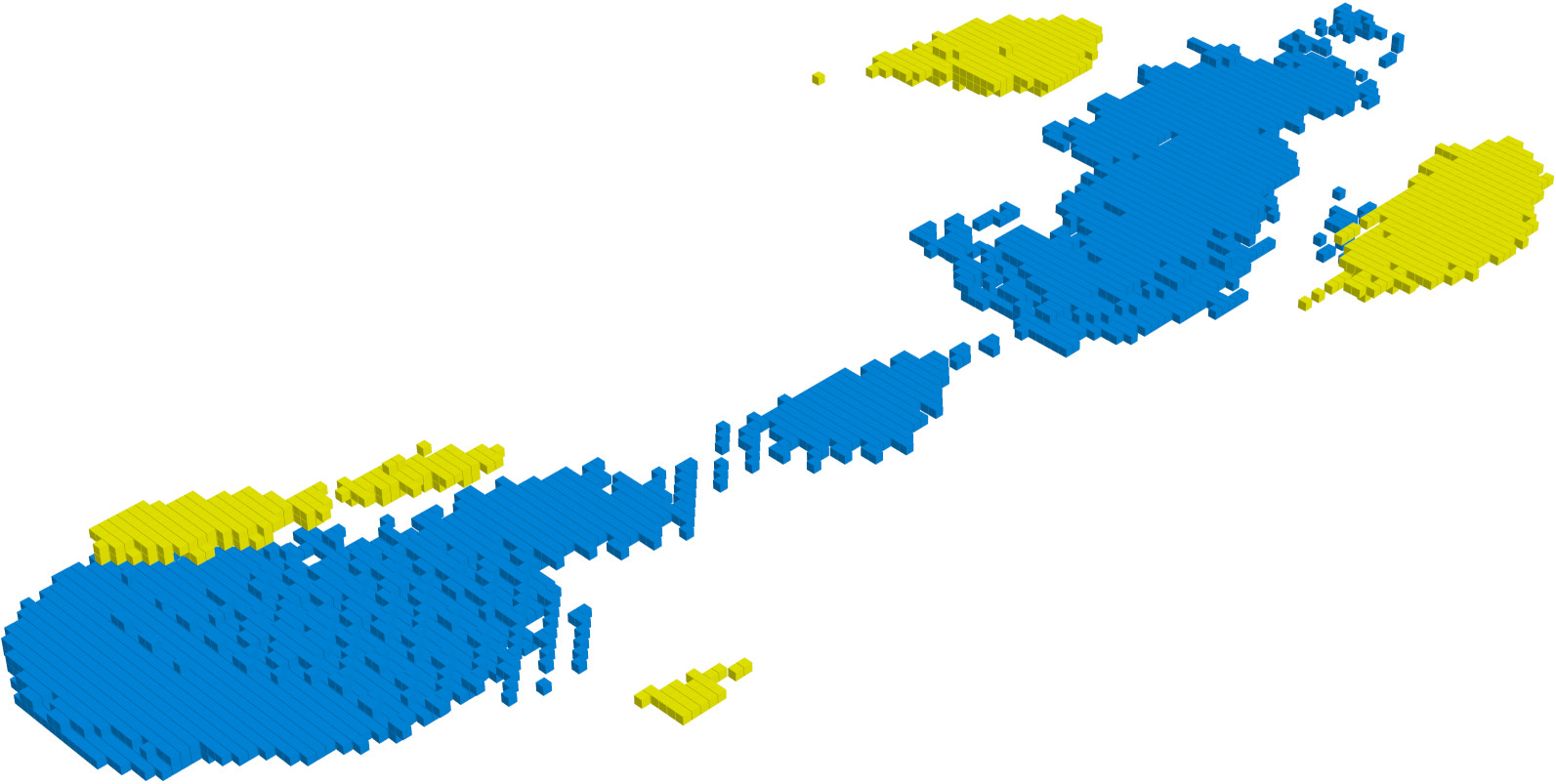}
    \end{minipage}%
    \begin{minipage}{0.142\textwidth}
        \centering
        \includegraphics[width=0.95\textwidth]{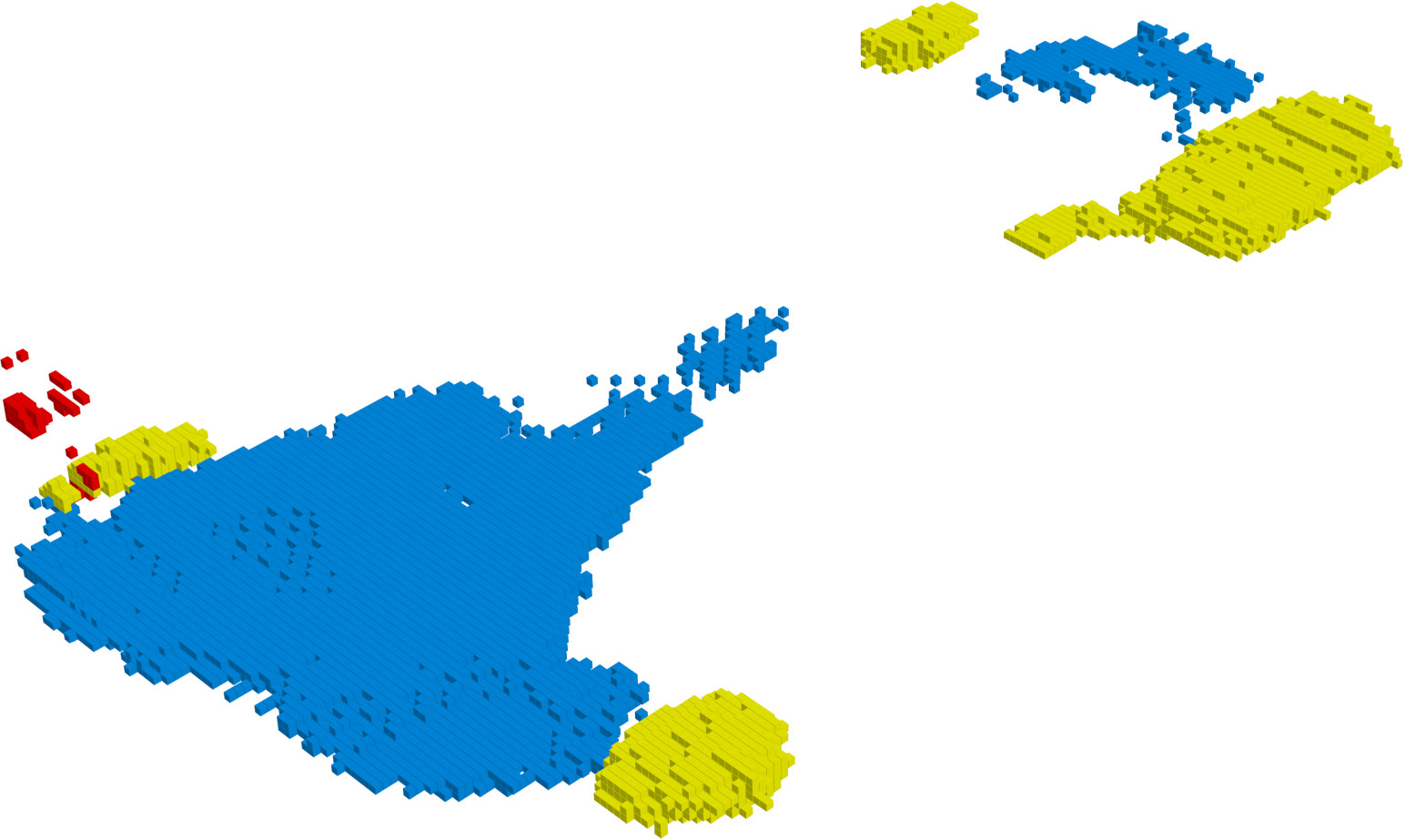}
    \end{minipage}
    \begin{minipage}{0.142\textwidth}
        \centering
        \includegraphics[width=0.95\textwidth]{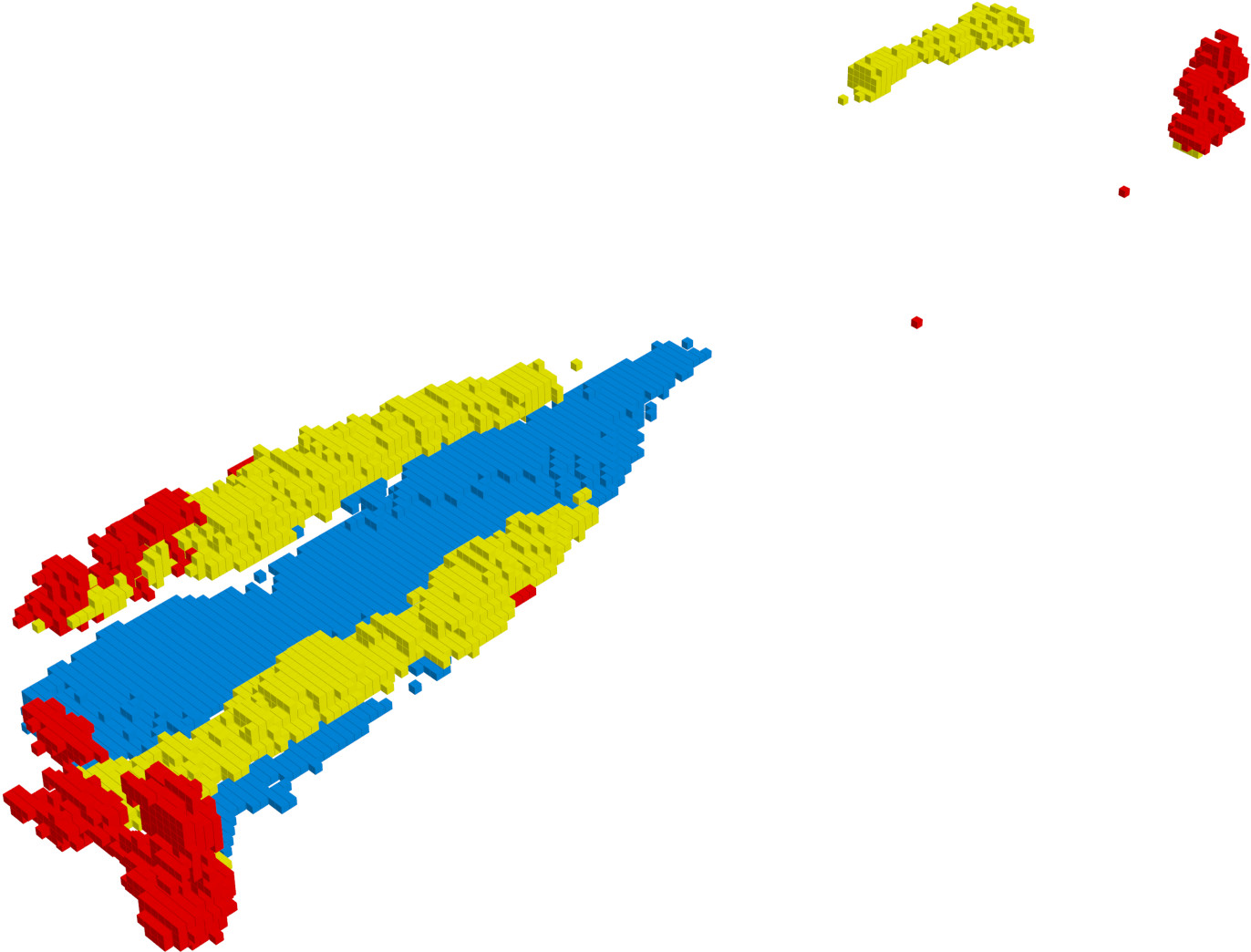}
    \end{minipage}
    \vskip\baselineskip

    % 8 row of images with titles on the left
    \begin{minipage}{0.143\textwidth}
        \centering
        {LMSCNet-SS}
    \end{minipage}%
    \begin{minipage}{0.142\textwidth}
        \centering
        \includegraphics[width=0.95\textwidth]{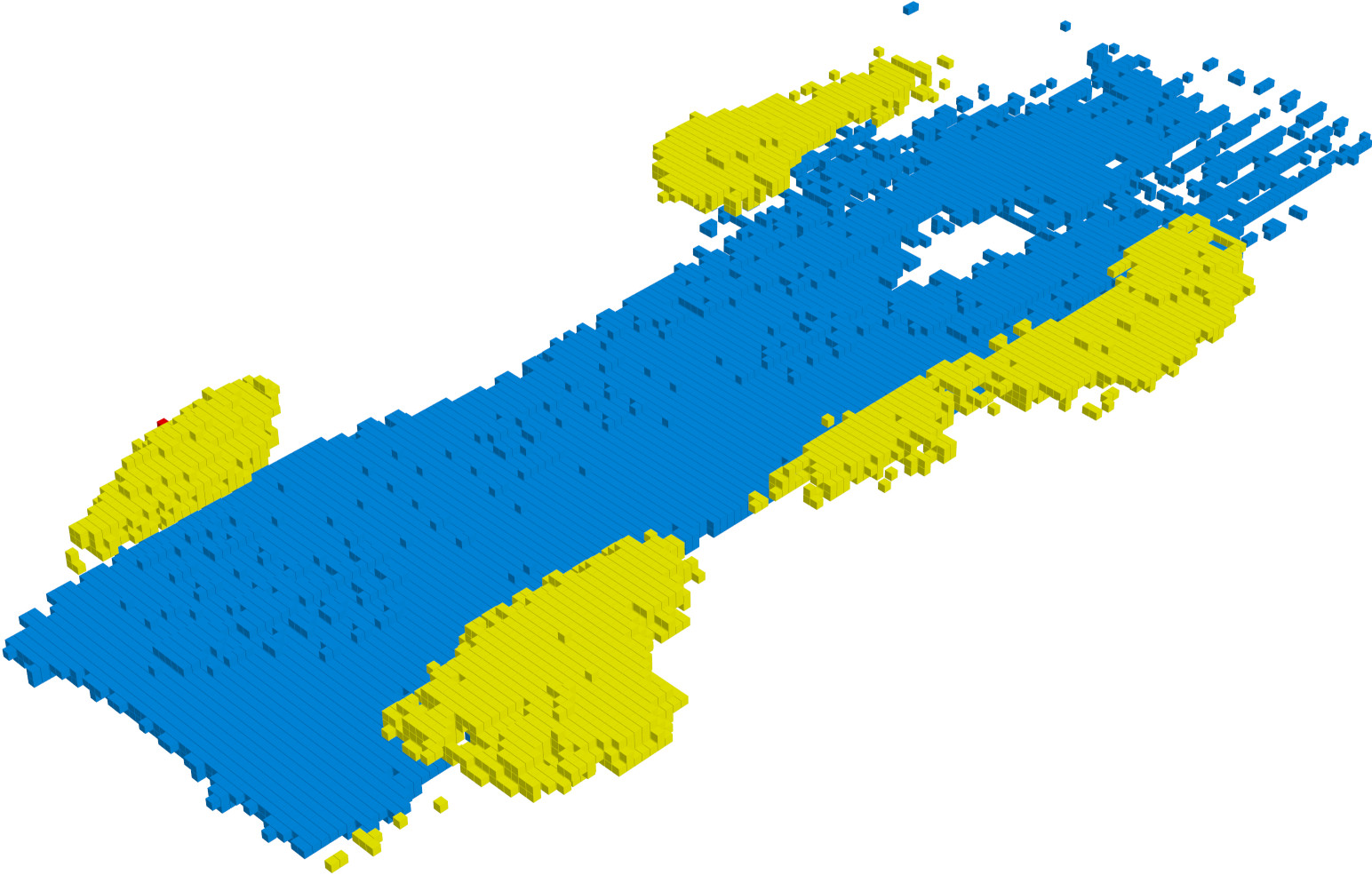}
    \end{minipage}%
    \begin{minipage}{0.142\textwidth}
        \centering
        \includegraphics[width=0.95\textwidth]{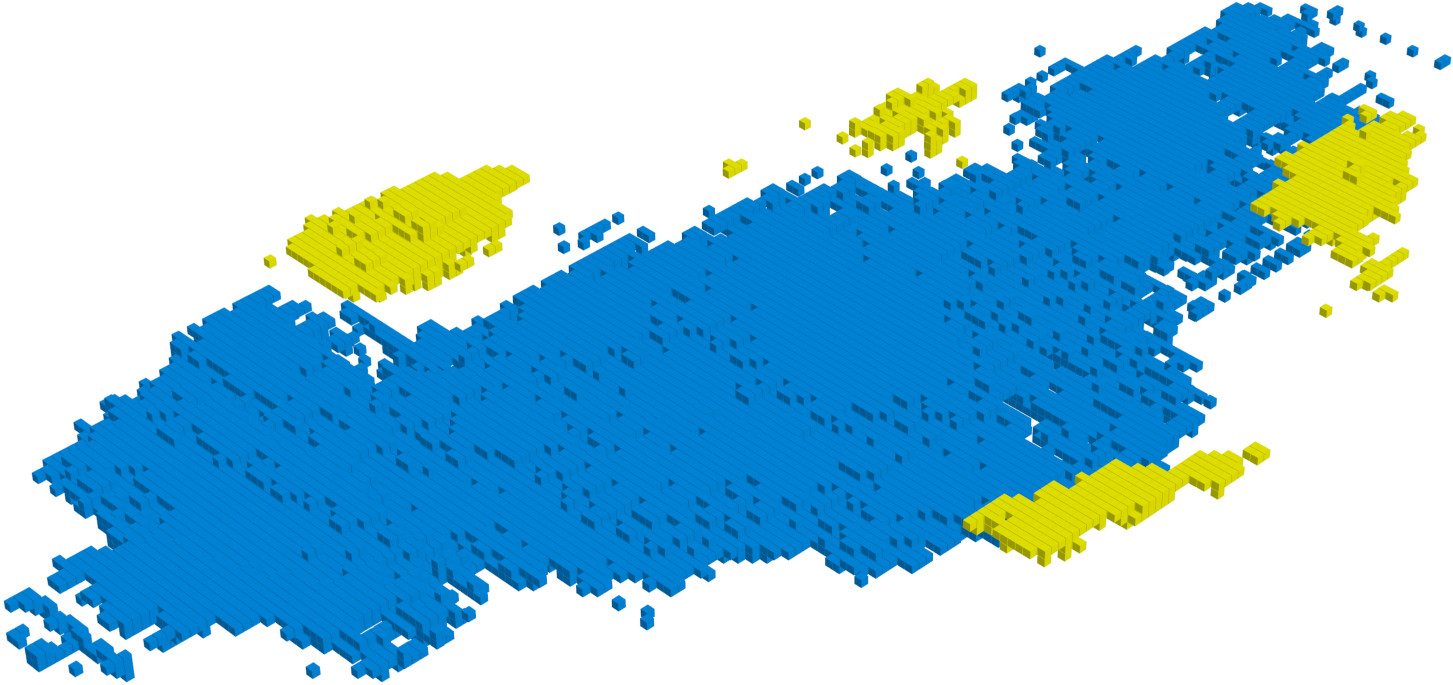}
    \end{minipage}%
    \begin{minipage}{0.142\textwidth}
        \centering
        \includegraphics[width=0.95\textwidth]{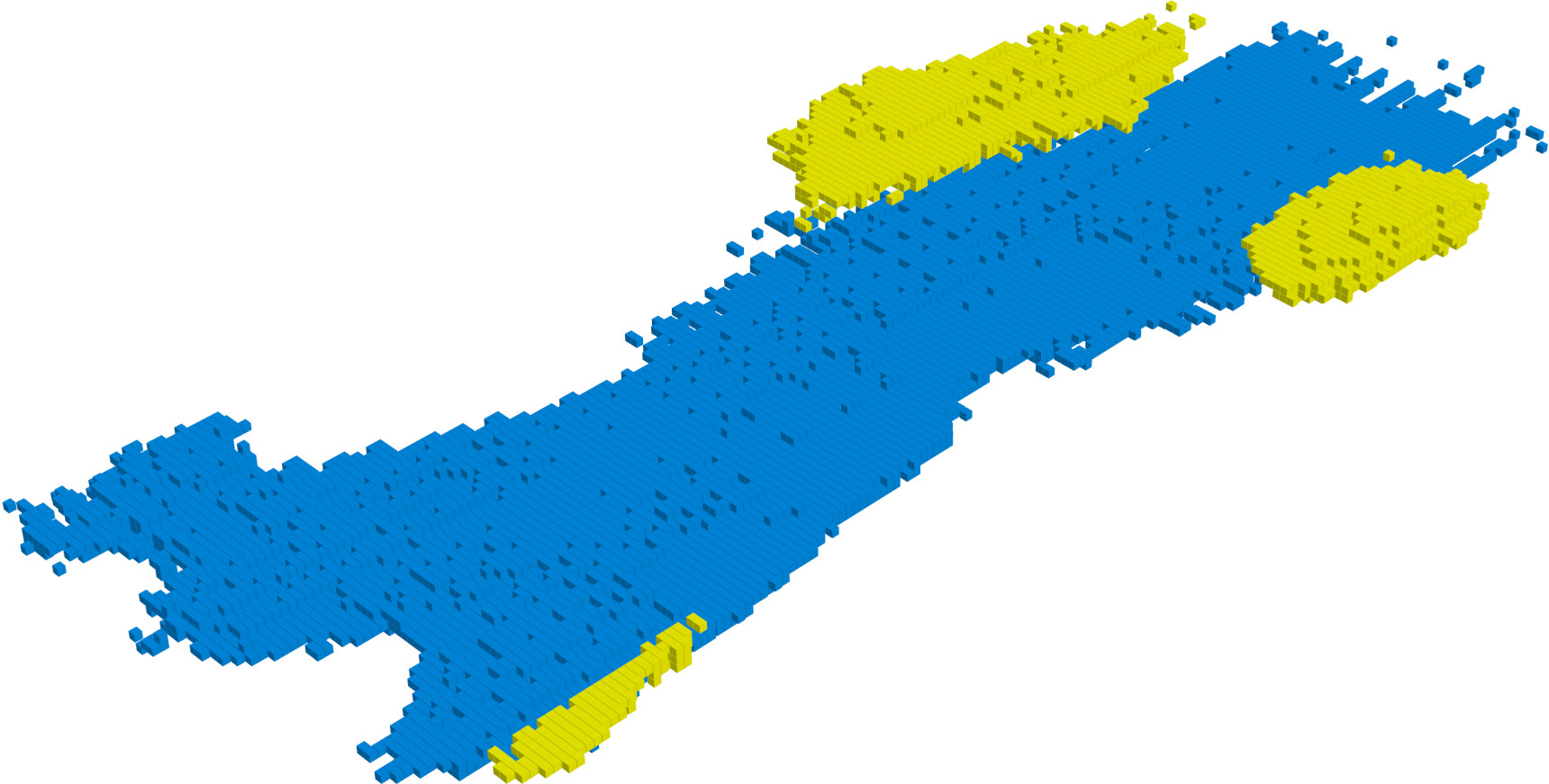}
    \end{minipage}%
    \begin{minipage}{0.142\textwidth}
        \centering
        \includegraphics[width=0.95\textwidth]{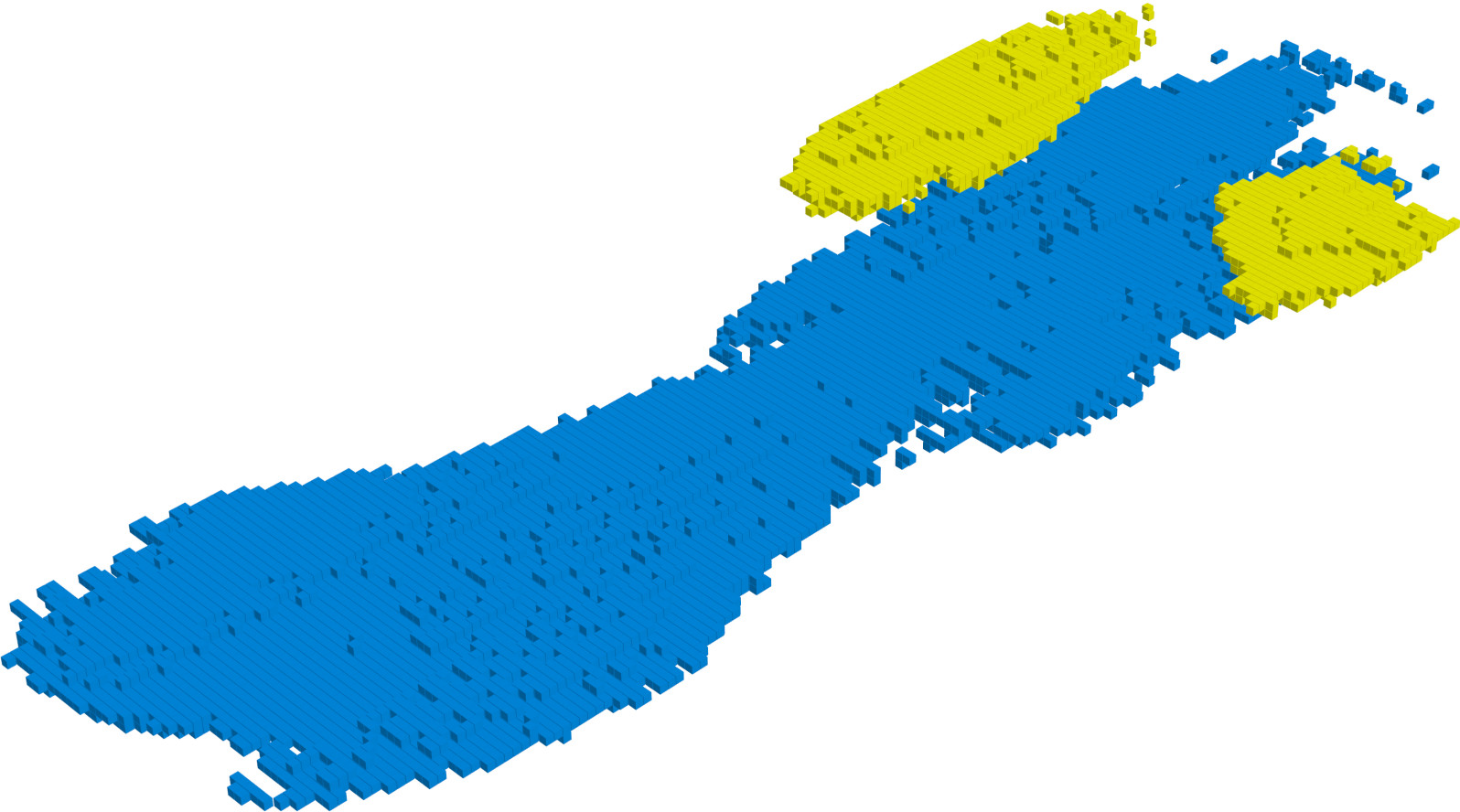}
    \end{minipage}%
    \begin{minipage}{0.142\textwidth}
        \centering
        \includegraphics[width=0.95\textwidth]{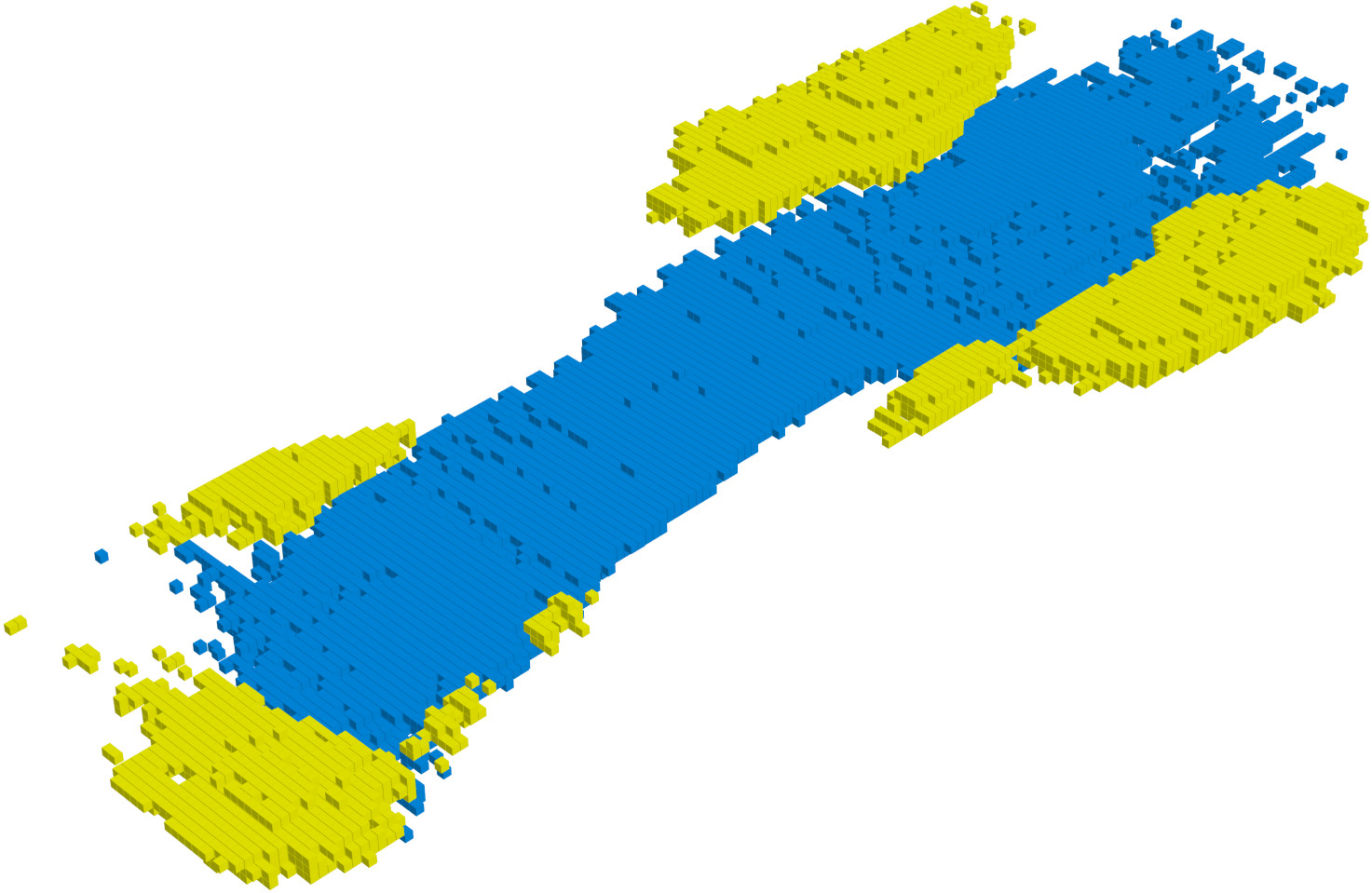}
    \end{minipage}
    \begin{minipage}{0.142\textwidth}
        \centering
        \includegraphics[width=0.95\textwidth]{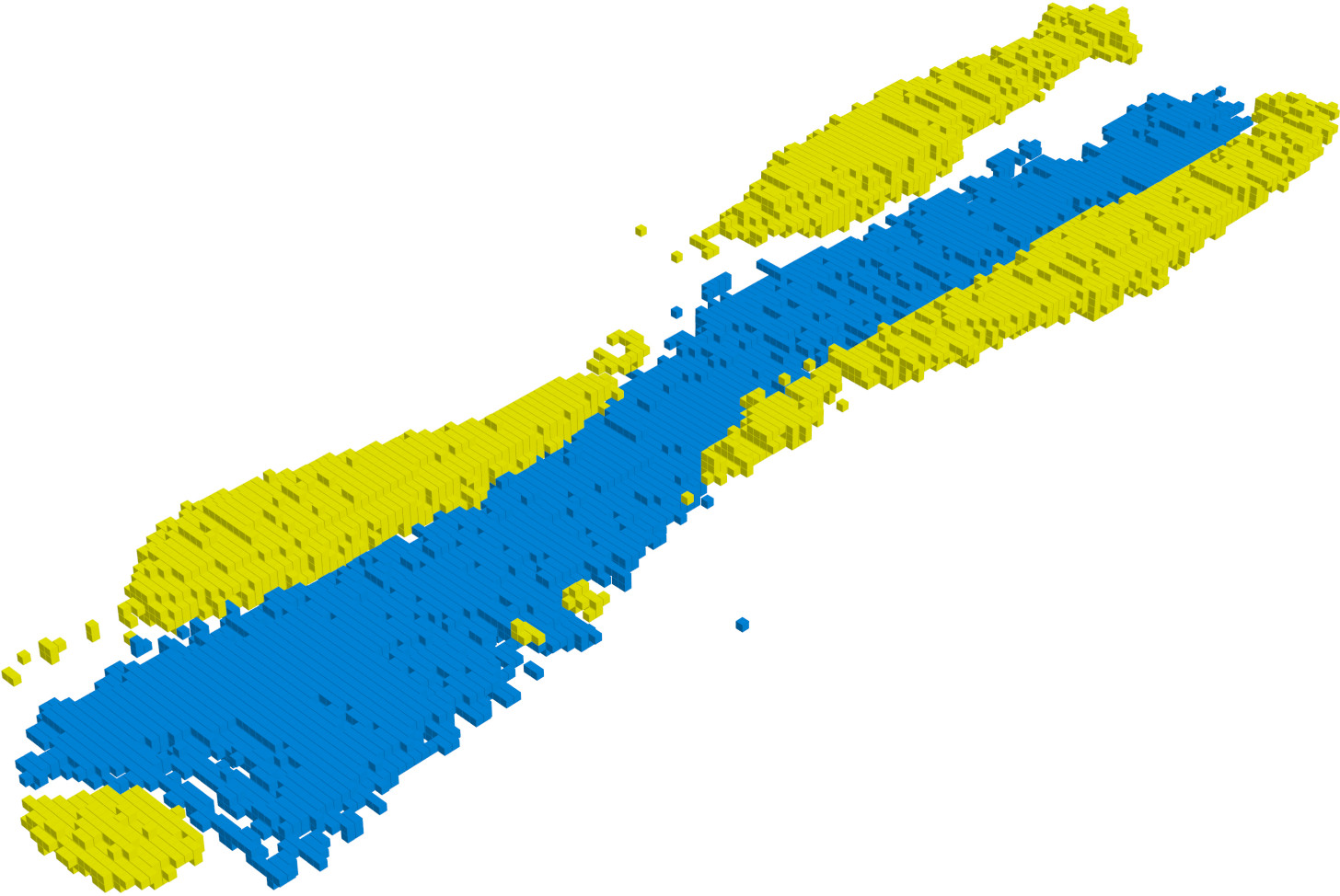}
    \end{minipage}
    \vskip\baselineskip

    % 9 row of images with titles on the left
    \begin{minipage}{0.143\textwidth}
        \centering
        {ORD-BKI}
    \end{minipage}%
    \begin{minipage}{0.142\textwidth}
        \centering
        \includegraphics[width=0.95\textwidth]{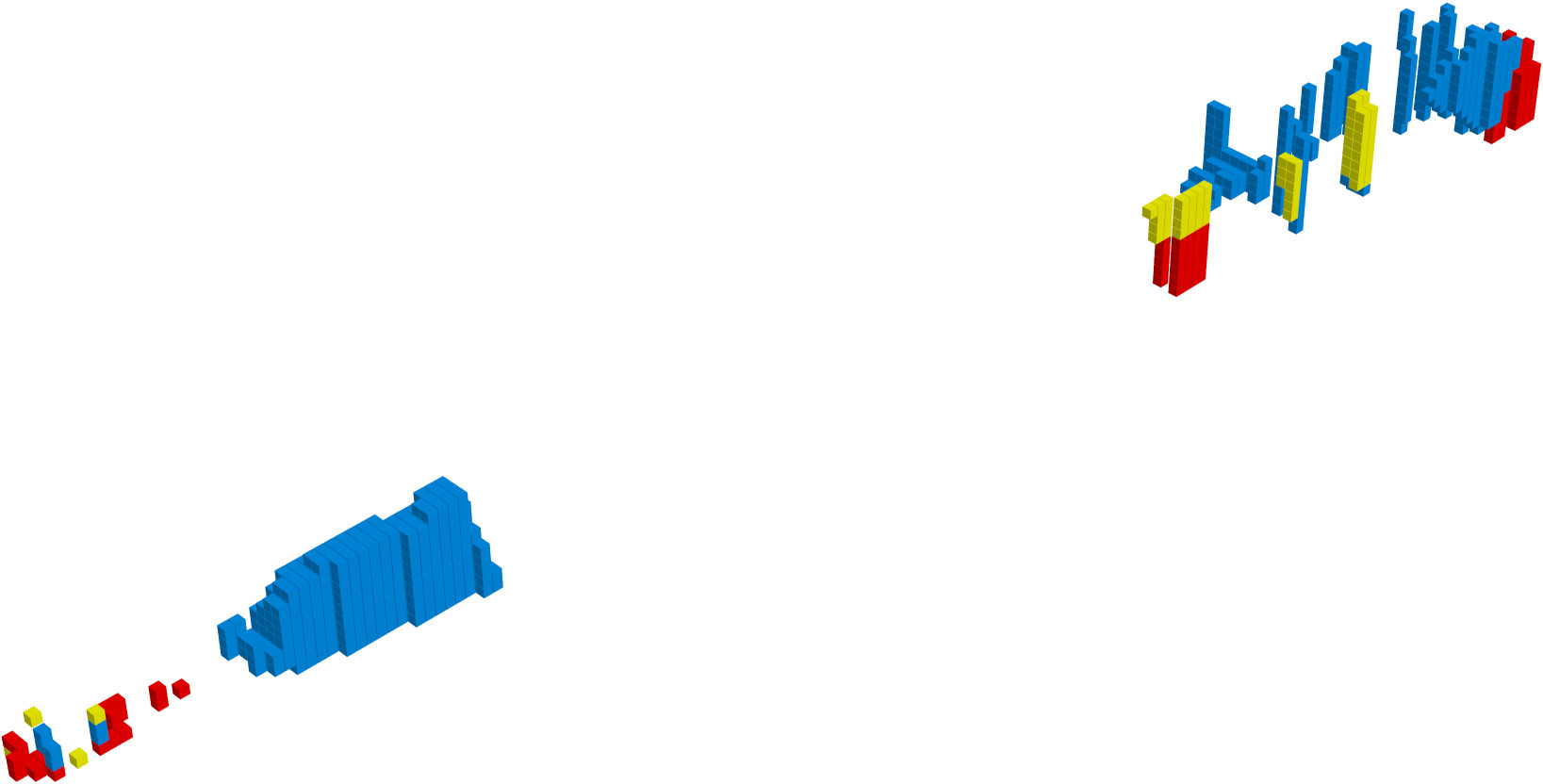}
    \end{minipage}%
    \begin{minipage}{0.142\textwidth}
        \centering
        \includegraphics[width=0.95\textwidth]{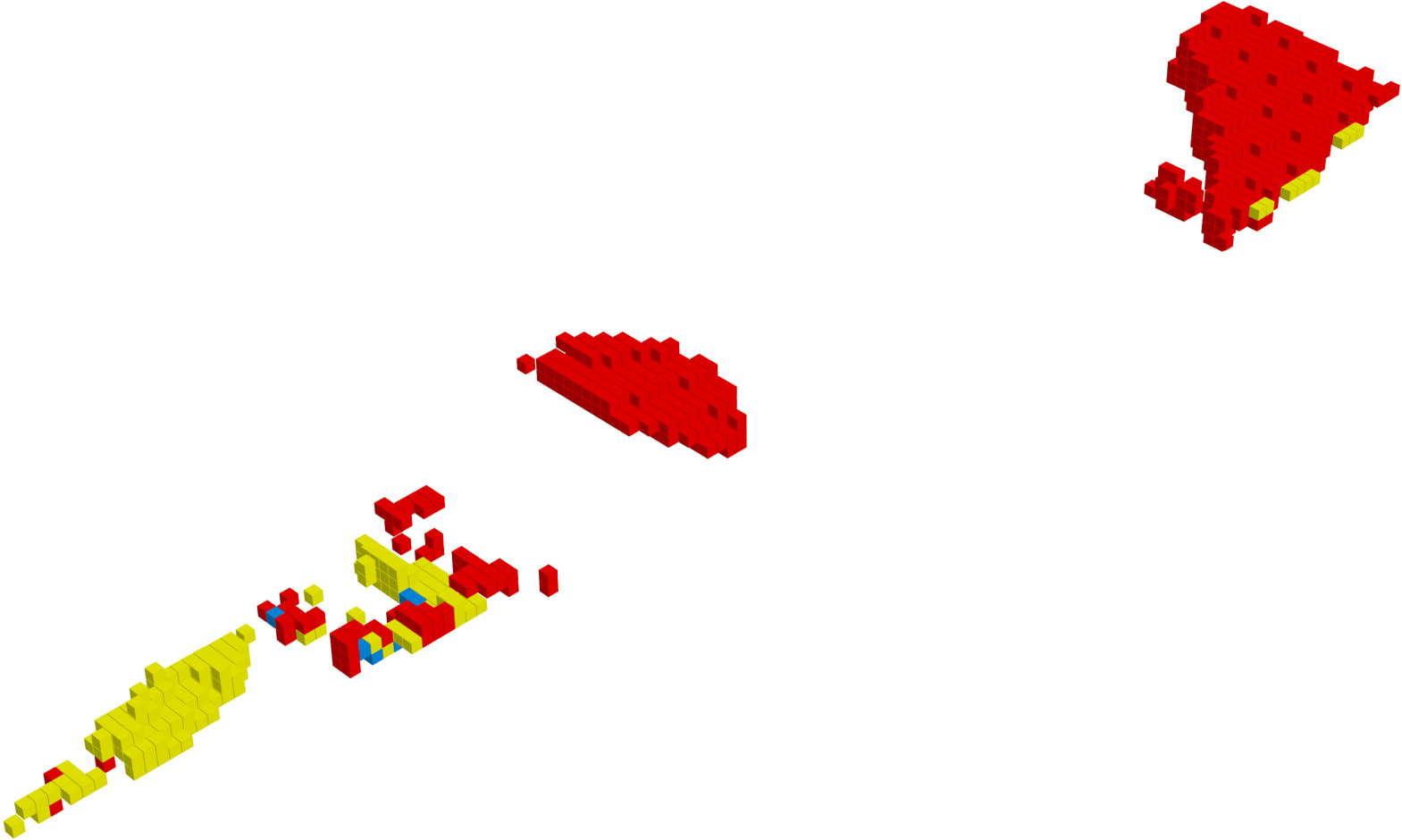}
    \end{minipage}%
    \begin{minipage}{0.142\textwidth}
        \centering
        \includegraphics[width=0.95\textwidth]{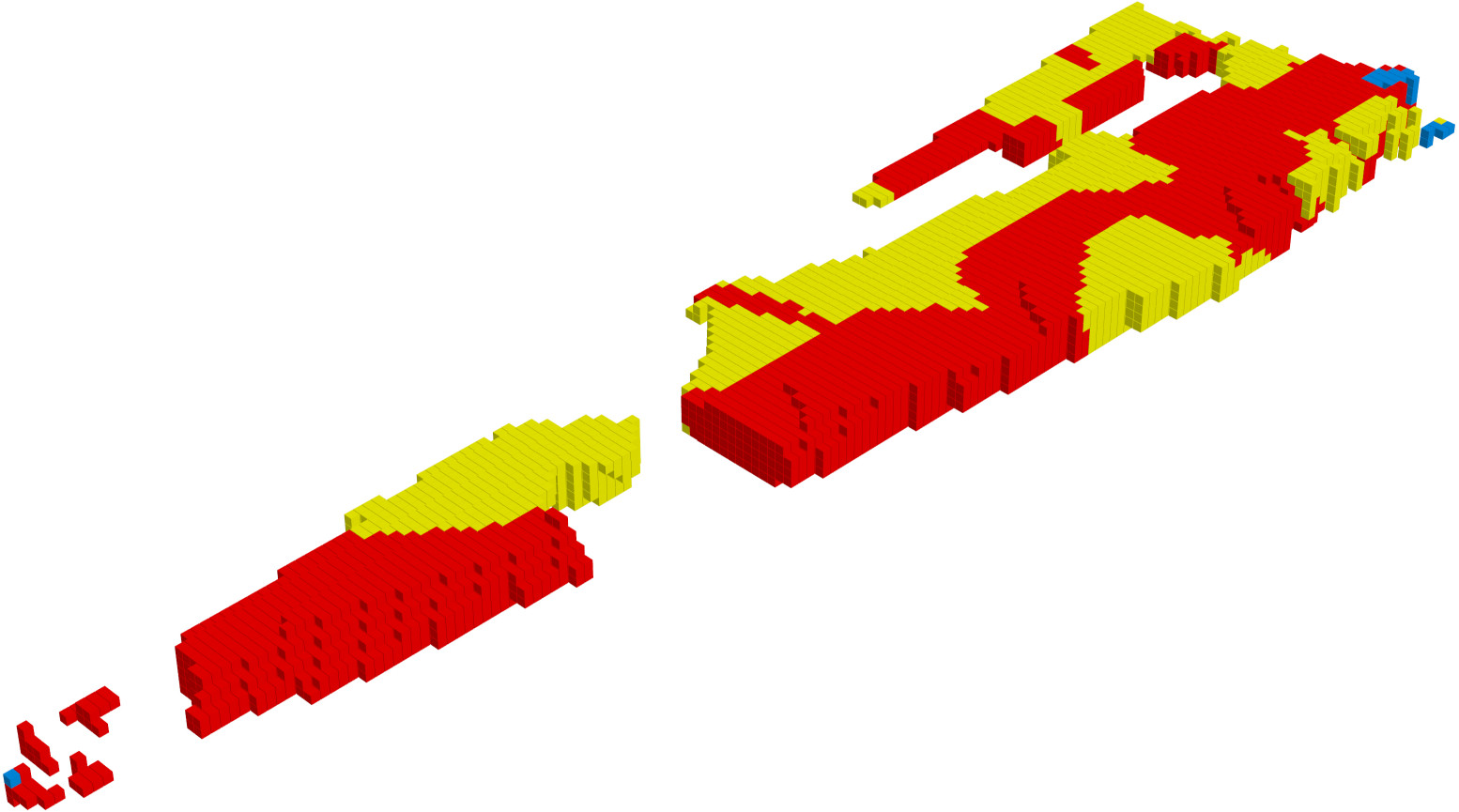}
    \end{minipage}%
    \begin{minipage}{0.142\textwidth}
        \centering
        \includegraphics[width=0.95\textwidth]{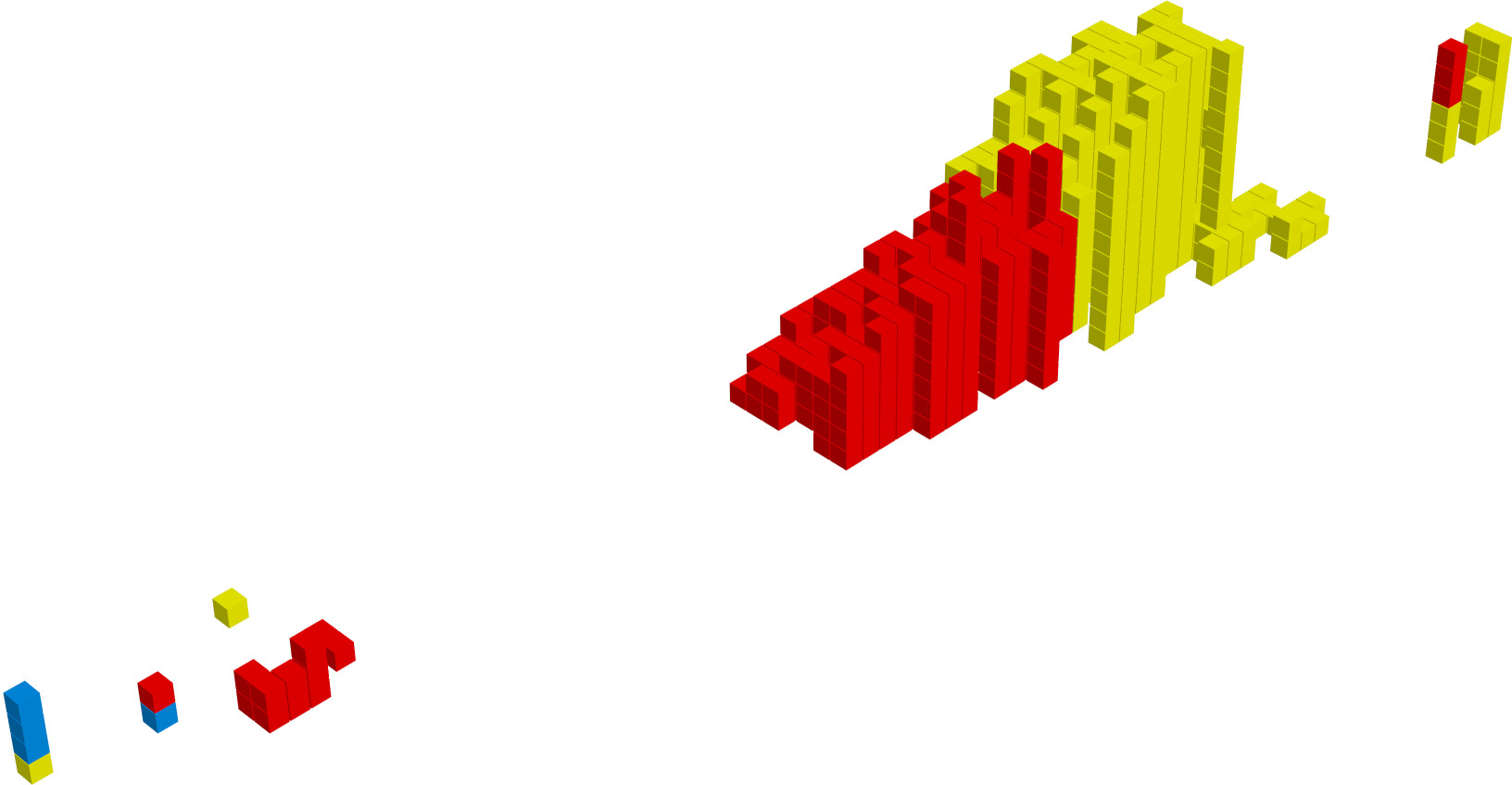}
    \end{minipage}%
    \begin{minipage}{0.142\textwidth}
        \centering
        \includegraphics[width=0.95\textwidth]{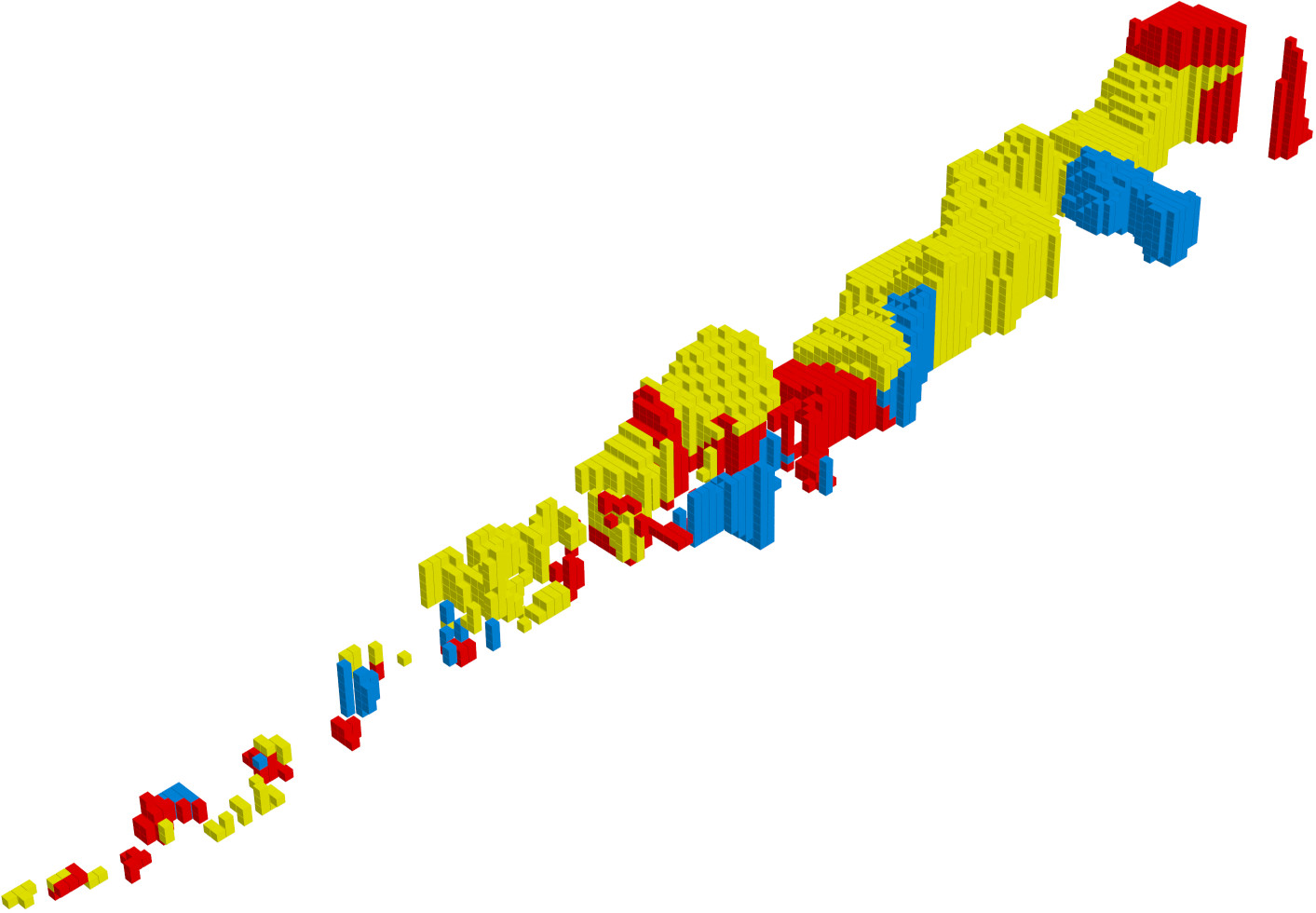}
    \end{minipage}
    \begin{minipage}{0.142\textwidth}
        \centering
        \includegraphics[width=0.95\textwidth]{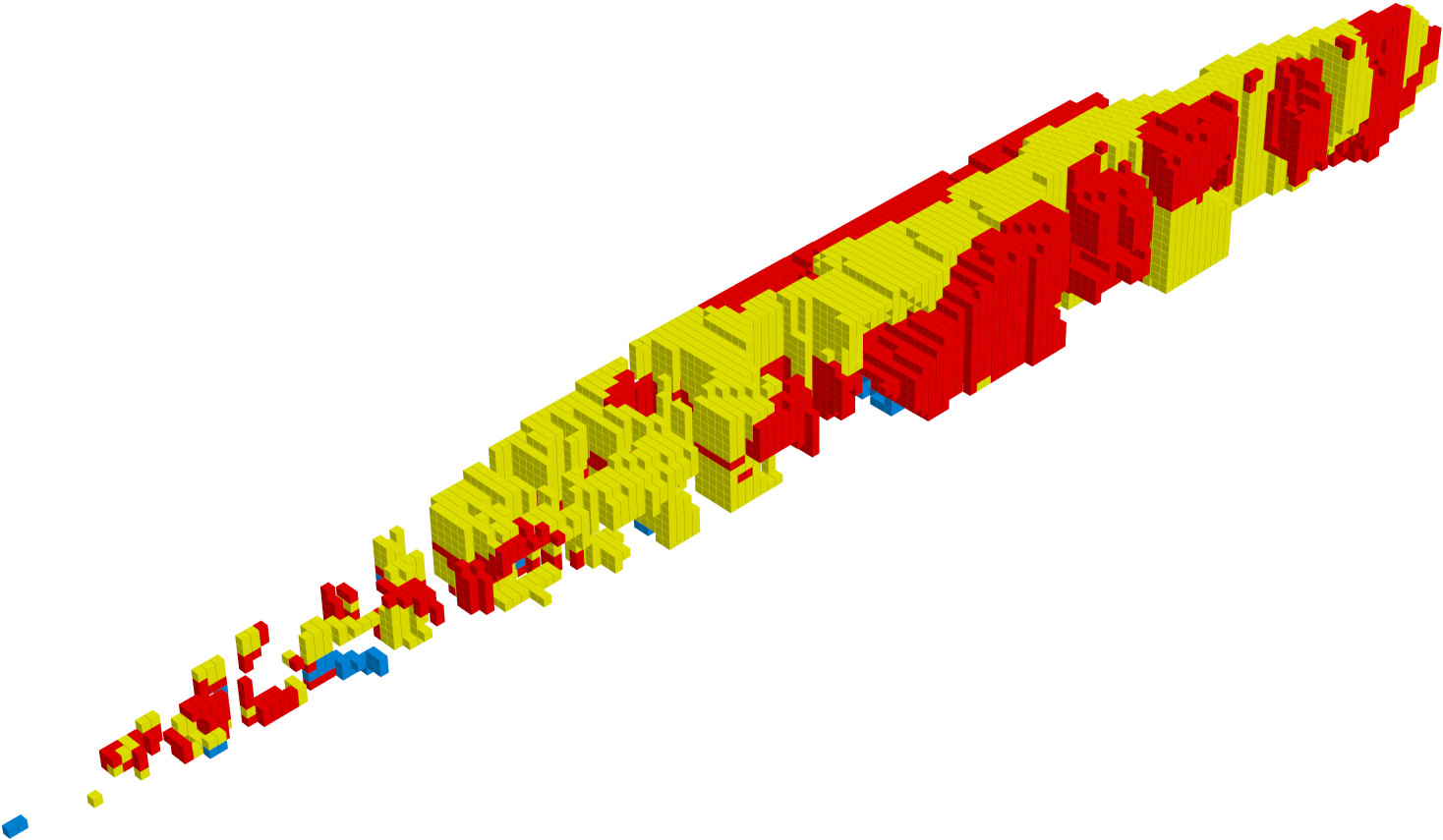}
    \end{minipage}
    \vskip\baselineskip
    % Additional rows can follow in the same format
    \caption{Real-vehicle test scenarios conducted at the Xiaotianshan Professional Outdoor Off-Road Site. The six test scenarios, presented from left to right, are: (a) Rural Path, (b) Stepped Barrier, (c) Vehicle Barrier, (d) Plain, (e) Forest, and (f) Mud Ruts, along with traversability estimation results from different algorithms. The first row displays the camera views, followed by the estimations from 3DTTNet (Ours), MonoScene, SSCNet, SSCNet-full, LMSCNet, LMSCNet-SS, and ORD-BKI. Different regions are labeled according to their traversability, with color-category mapping detailed in Fig.~\ref{fig_2}(b).}
    \label{fig_16}
\end{figure*}

\begin{table*}[htbp] 
    \centering
    \caption{Traversability estimate results from the real vehicle test. The table summarizes the inference times and performance across six test scenarios (labeled (a) to (f)) for various methods. Methods that successfully estimated traversability in each scenario are marked with a check ($\checkmark$), while methods that did not perform the estimation are marked with a dash (–).}

        \begin{tabular}{c|c|cccccc}
            \toprule
            {\textbf{Method}} &{\textbf{Inference Time}} &{\textbf{Scenario (a)}} &{\textbf{Scenario (b)}}  &{\textbf{Scenario (c)}} &{\textbf{Scenario (d)}} &{\textbf{Scenario (e)}} &{\textbf{Scenario (f)}}\\
            \midrule
                \textbf{MonoScene\cite{ref51}}& 4410ms& -&-&-&-&-&-\\
                \textbf{SSCNet\cite{ref52}}& 31ms&  $\checkmark$&-& -& $\checkmark$&-&-\\
                \textbf{SSCNet-full\cite{ref52}}& 24ms& $\checkmark$&-&-&-&-&-\\
                \textbf{LMSCNet\cite{ref53}}& 63ms& -&-&-&-&-&-\\
                \textbf{LMSCNet-SS\cite{ref53}}& 53ms& $\checkmark$&-&-&$\checkmark$&$\checkmark$&-\\
                \textbf{ORD-BKI\cite{ref48}}& 5243ms& -&-&-&-&-&-\\
            \midrule
                \textbf{3DTTNet(Ours)}& 340ms& $\checkmark$& $\checkmark$& $\checkmark$& $\checkmark$& $\checkmark$& $\checkmark$\\
            \bottomrule
        \end{tabular}
    
    \label{tab:table_4}
\end{table*}

Despite these differences, the real-vehicle test results (see Fig.~\ref{fig_16} and Table~\ref{tab:table_4}) demonstrate that the proposed algorithm accurately identifies the 3D traversability of terrains in off-road environments. This outcome indicates the robust adaptability and generalization capabilities of the algorithm, enabling effective operation across various sensor configurations and vehicle platforms.

\subsubsection{Test 2}

\begin{figure*}[htbp]
    % 1 and 2 scenarios
    % 1 row titles
    \begin{minipage}{0.5\textwidth}
        \centering
        {(a) Cliff Terrain}
    \end{minipage}%
    \begin{minipage}{0.5\textwidth}
        \centering
        {(b) Constrained Forest Road}
    \end{minipage}%
    
    % 2 row titles
    \begin{minipage}{0.25\textwidth}
        \centering
        {Camera View}
    \end{minipage}%
    \begin{minipage}{0.25\textwidth}
        \centering
        {Trajectory}
    \end{minipage}%
    \begin{minipage}{0.25\textwidth}
        \centering
        {Camera View}
    \end{minipage}%
    \begin{minipage}{0.25\textwidth}
        \centering
        {Trajectory}
    \end{minipage}%
    
    % 3 row of images with titles on the left
    \begin{minipage}{0.25\textwidth}
        \centering
        \includegraphics[width=0.95\textwidth]{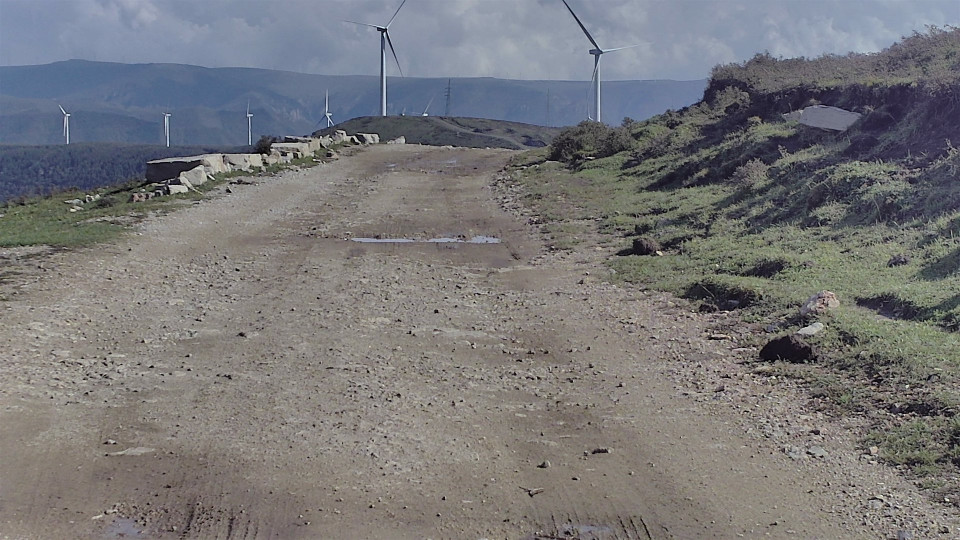}
    \end{minipage}%
    \begin{minipage}{0.25\textwidth}
        \centering
        \includegraphics[width=0.95\textwidth]{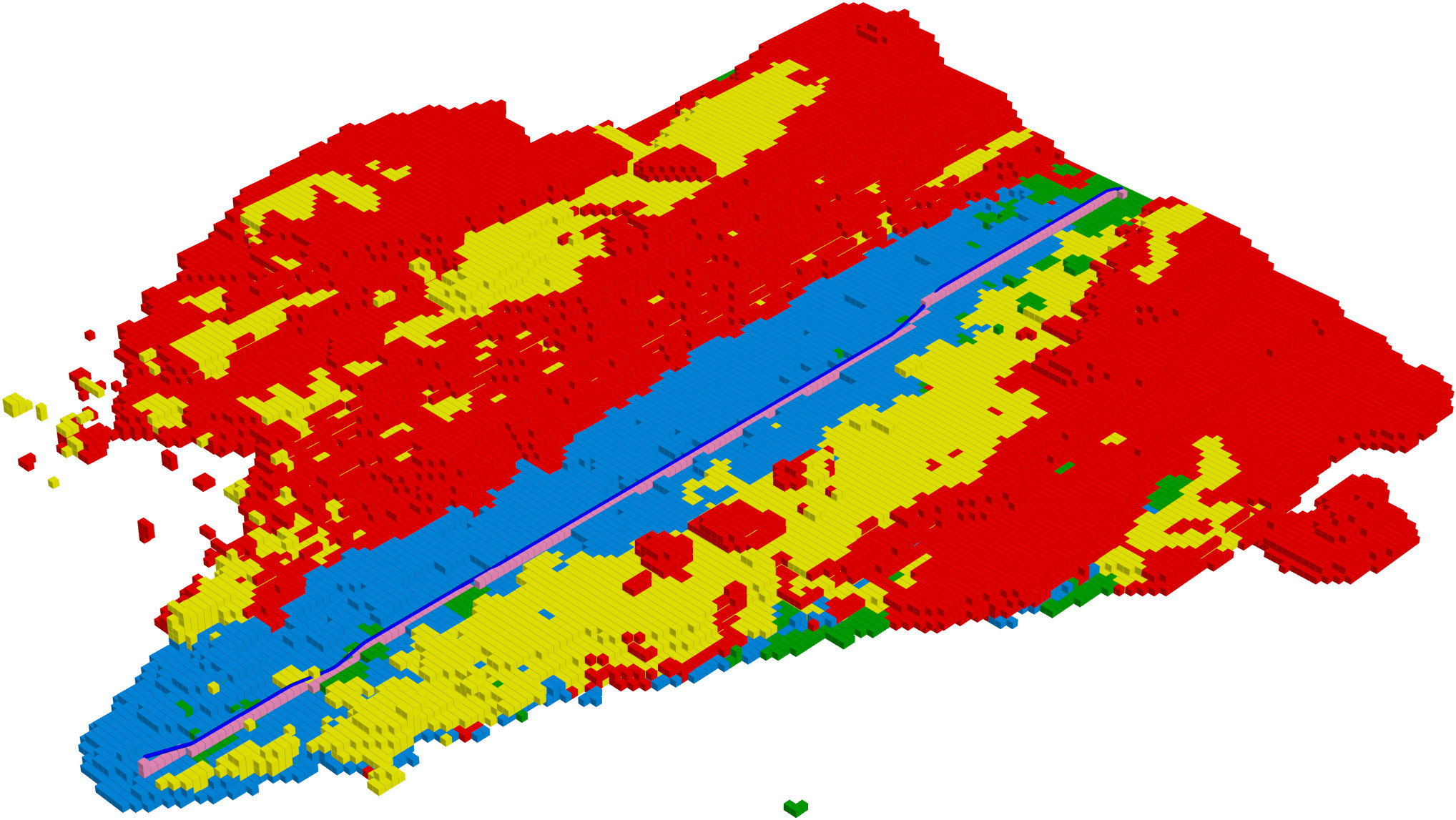}
    \end{minipage}%
    \begin{minipage}{0.25\textwidth}
        \centering
        \includegraphics[width=0.95\textwidth]{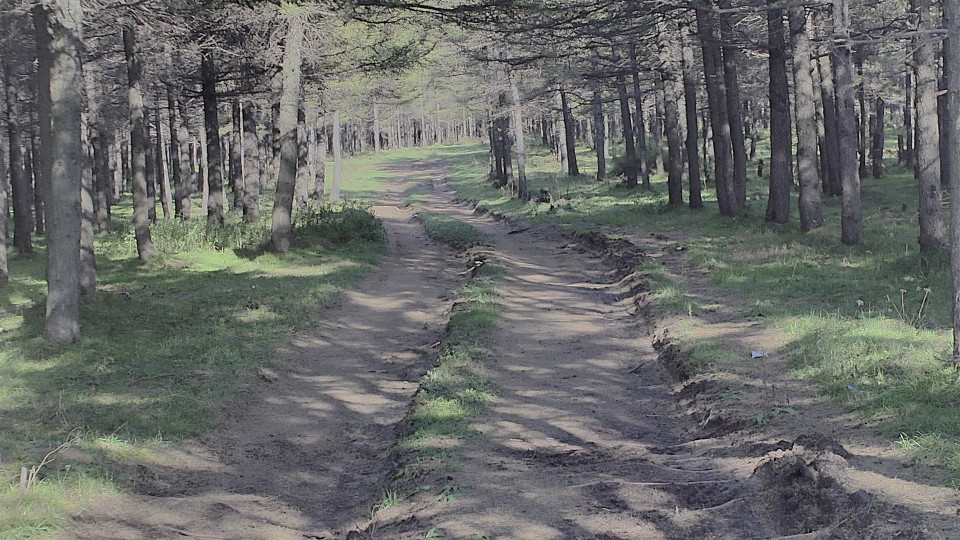}
    \end{minipage}%
    \begin{minipage}{0.25\textwidth}
        \centering
        \includegraphics[width=0.95\textwidth]{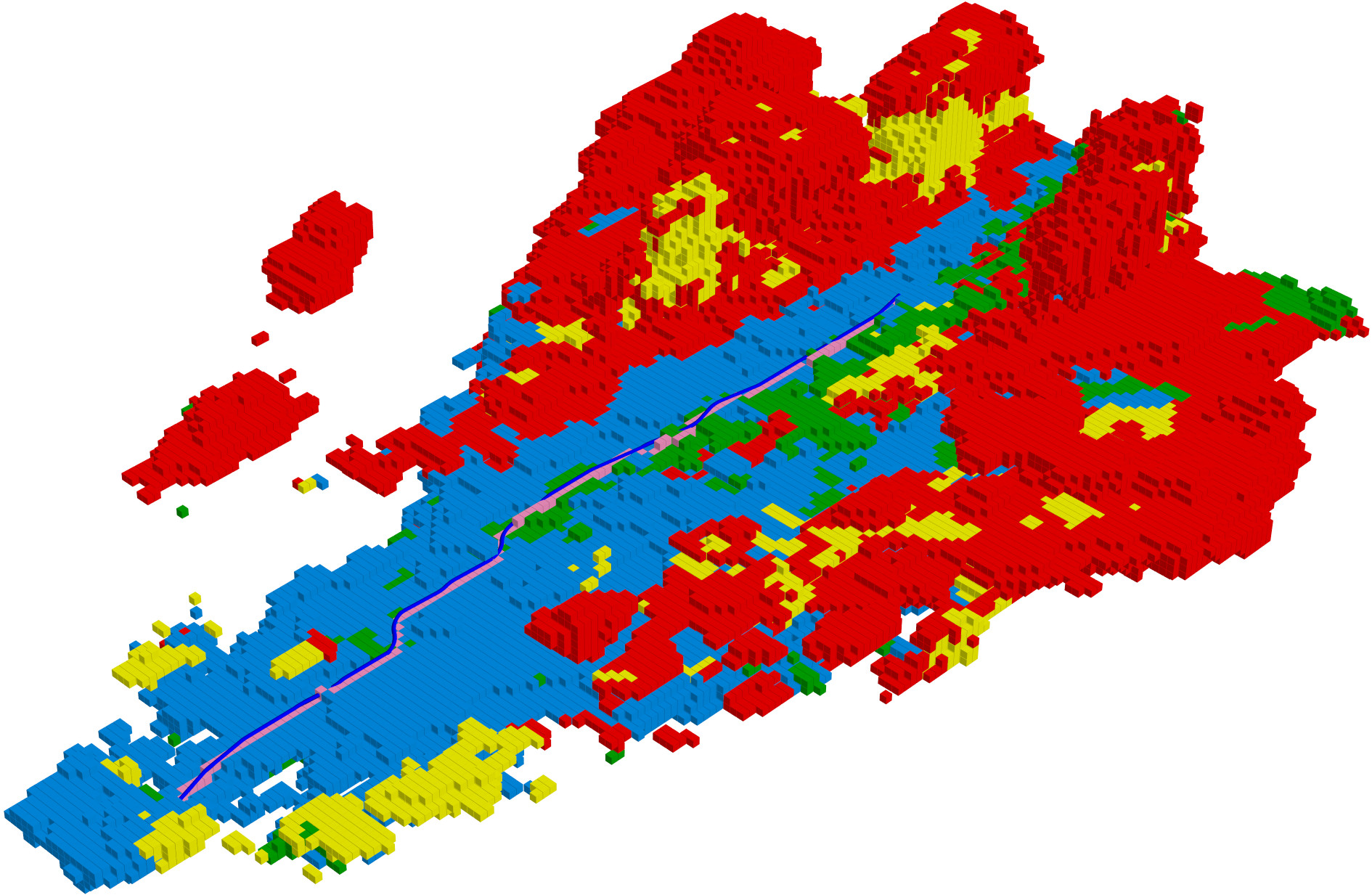}
    \end{minipage}%
    
    % 4 row of images with titles on the left
    \begin{minipage}{0.25\textwidth}
        \centering
        \includegraphics[width=0.95\textwidth]{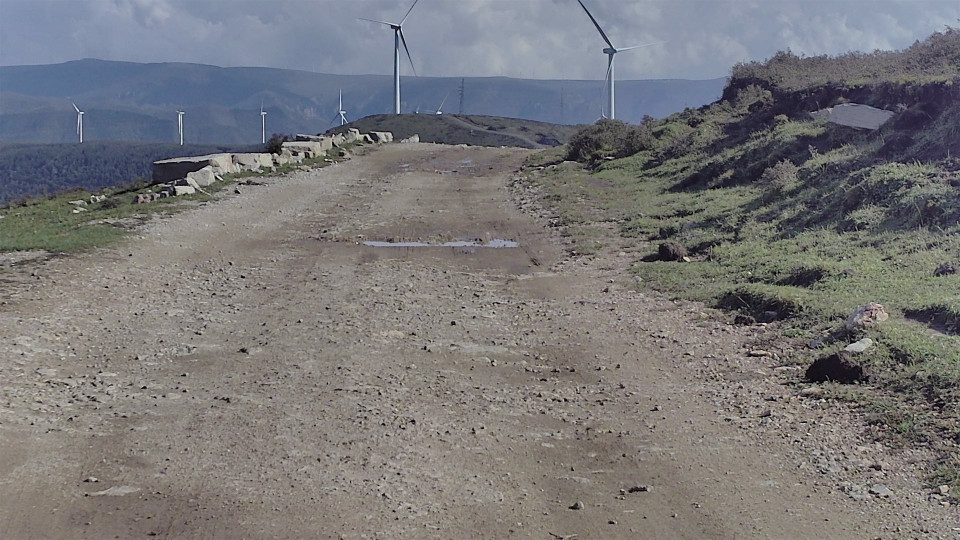}
    \end{minipage}%
    \begin{minipage}{0.25\textwidth}
        \centering
        \includegraphics[width=0.95\textwidth]{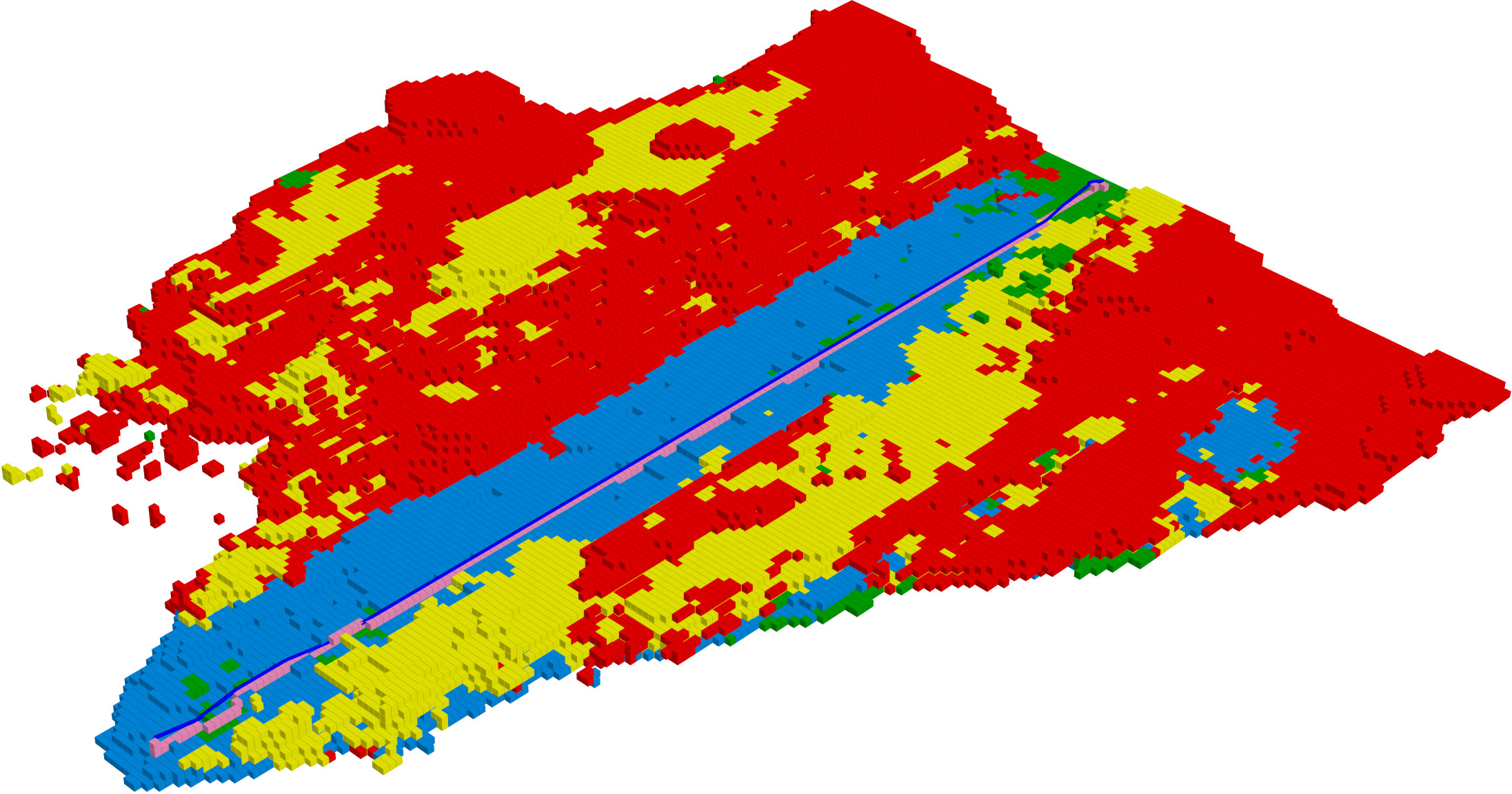}
    \end{minipage}%
    \begin{minipage}{0.25\textwidth}
        \centering
        \includegraphics[width=0.95\textwidth]{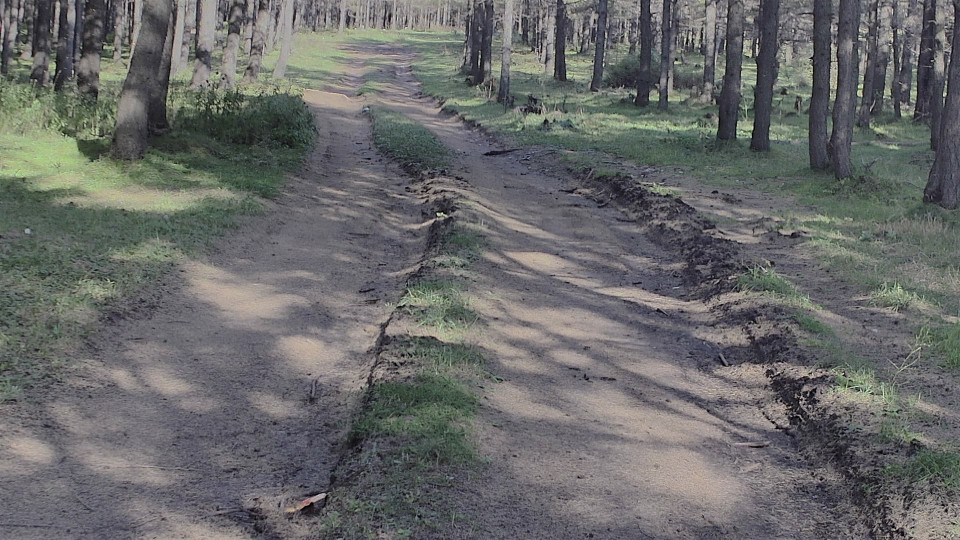}
    \end{minipage}%
    \begin{minipage}{0.25\textwidth}
        \centering
        \includegraphics[width=0.95\textwidth]{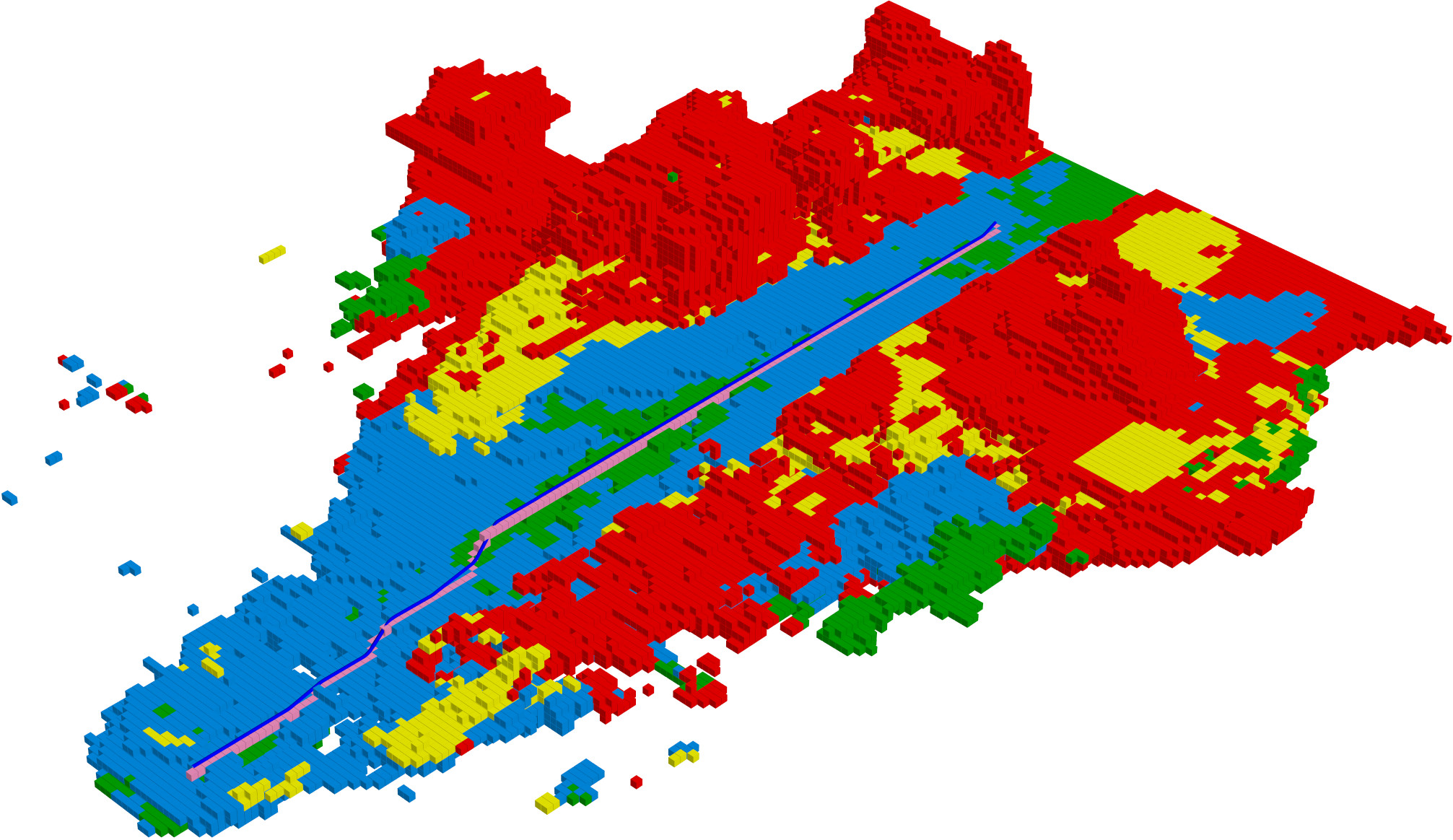}
    \end{minipage}%
    
    % 5 row of images with titles on the left
    \begin{minipage}{0.25\textwidth}
        \centering
        \includegraphics[width=0.95\textwidth]{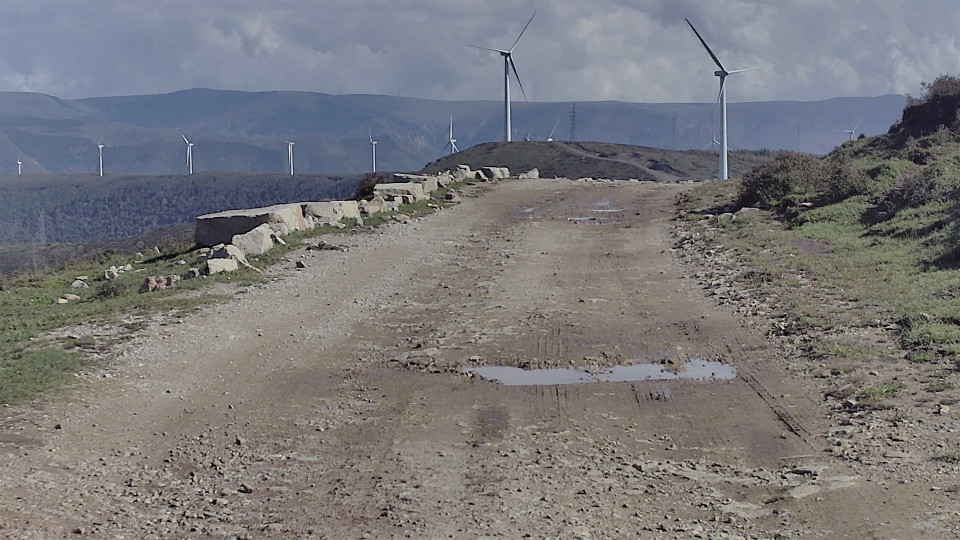}
    \end{minipage}%
    \begin{minipage}{0.25\textwidth}
        \centering
        \includegraphics[width=0.95\textwidth]{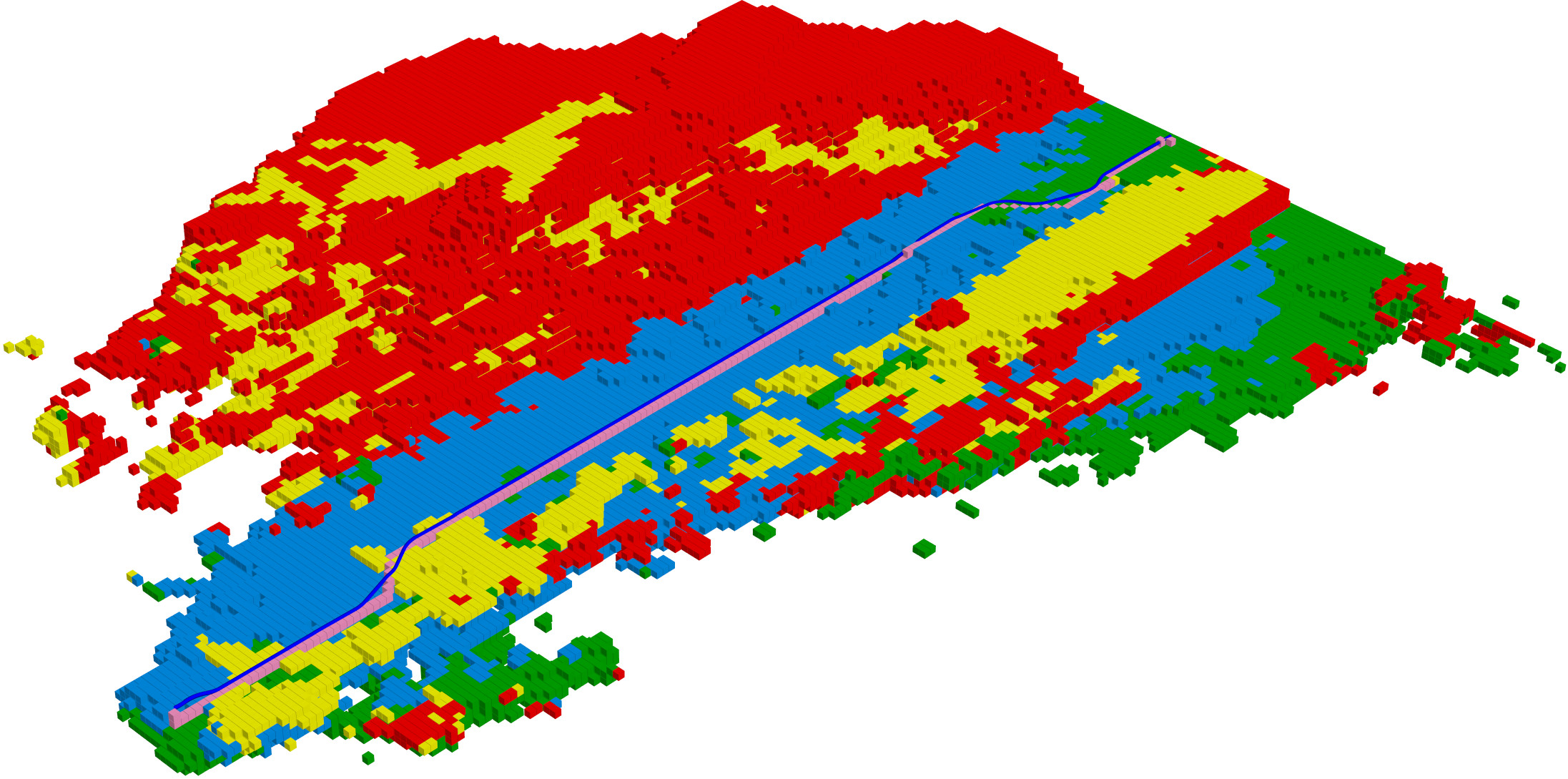}
    \end{minipage}%
    \begin{minipage}{0.25\textwidth}
        \centering
        \includegraphics[width=0.95\textwidth]{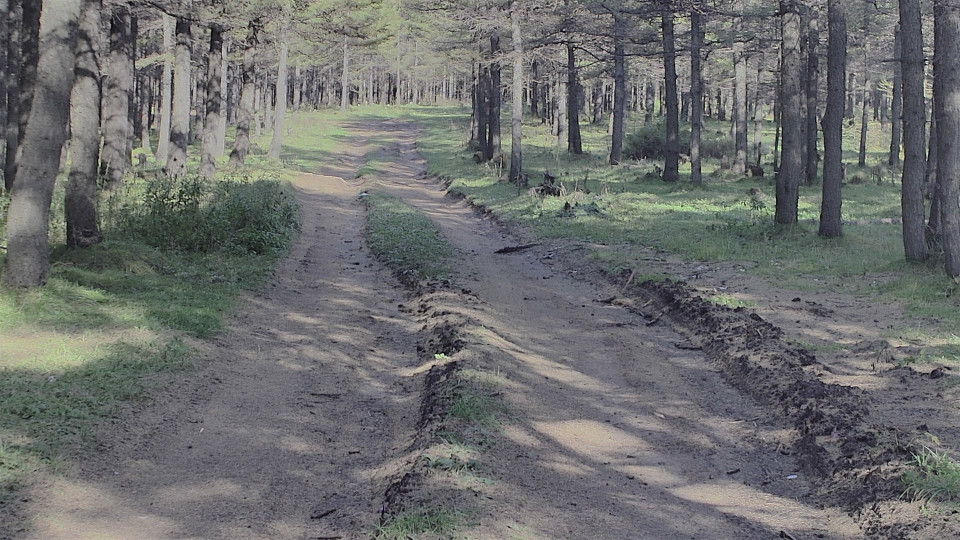}
    \end{minipage}%
    \begin{minipage}{0.25\textwidth}
        \centering
        \includegraphics[width=0.95\textwidth]{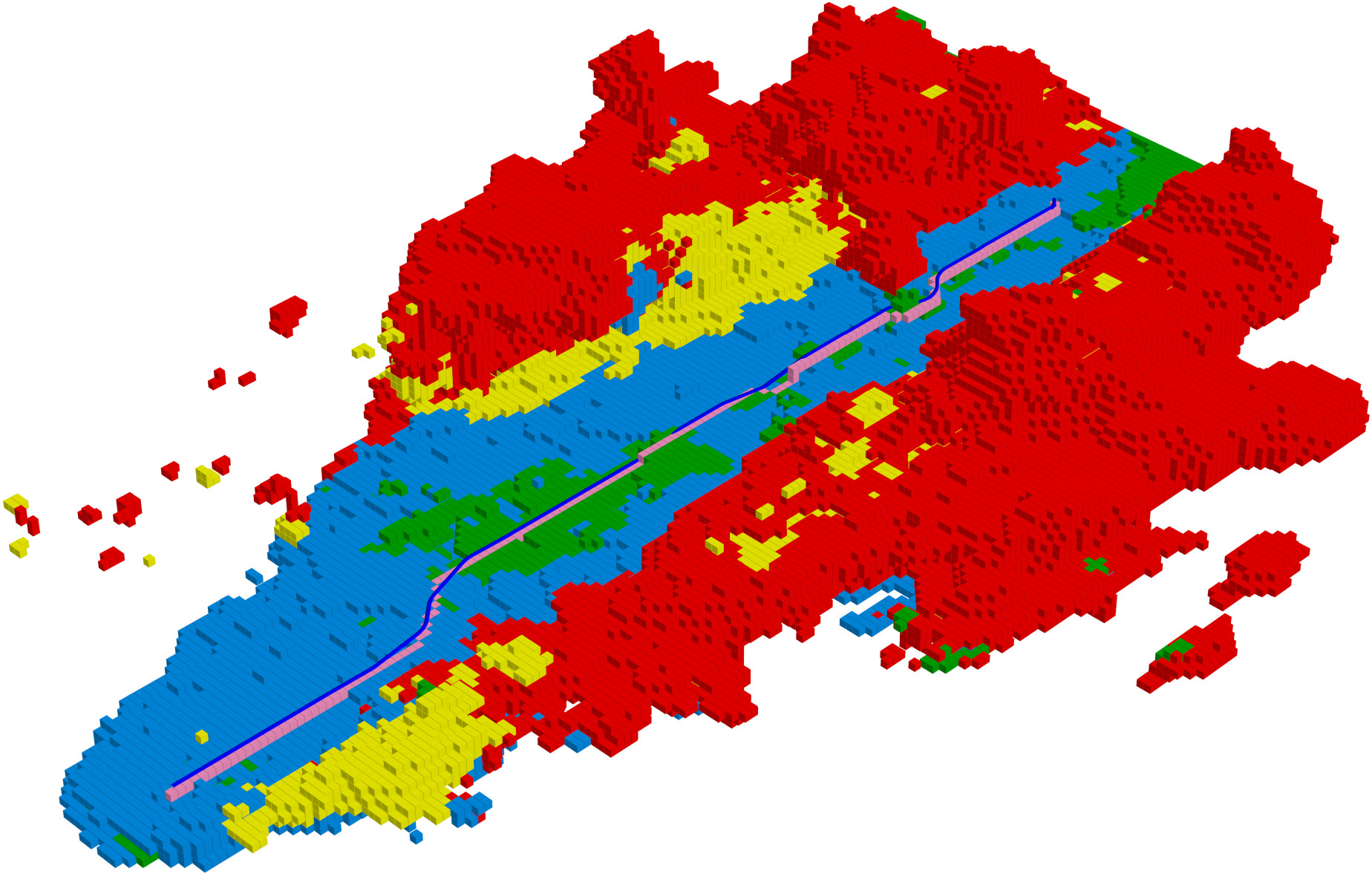}
    \end{minipage}%
    % \vskip\baselineskip
    
    % 3 and 4 scenarios
    % 1 row titles
    \begin{minipage}{0.5\textwidth}
        \centering
        {(c) Continuous Irregular Pits}
    \end{minipage}%
    \begin{minipage}{0.5\textwidth}
        \centering
        {(d) Water-filled Ruts}
    \end{minipage}%
    
    % 2 row titles
    \begin{minipage}{0.25\textwidth}
        \centering
        {Camera View}
    \end{minipage}%
    \begin{minipage}{0.25\textwidth}
        \centering
        {Trajectory}
    \end{minipage}%
    \begin{minipage}{0.25\textwidth}
        \centering
        {Camera View}
    \end{minipage}%
    \begin{minipage}{0.25\textwidth}
        \centering
        {Trajectory}
    \end{minipage}%
    
    % 3 row of images with titles on the left
    \begin{minipage}{0.25\textwidth}
        \centering
        \includegraphics[width=0.95\textwidth]{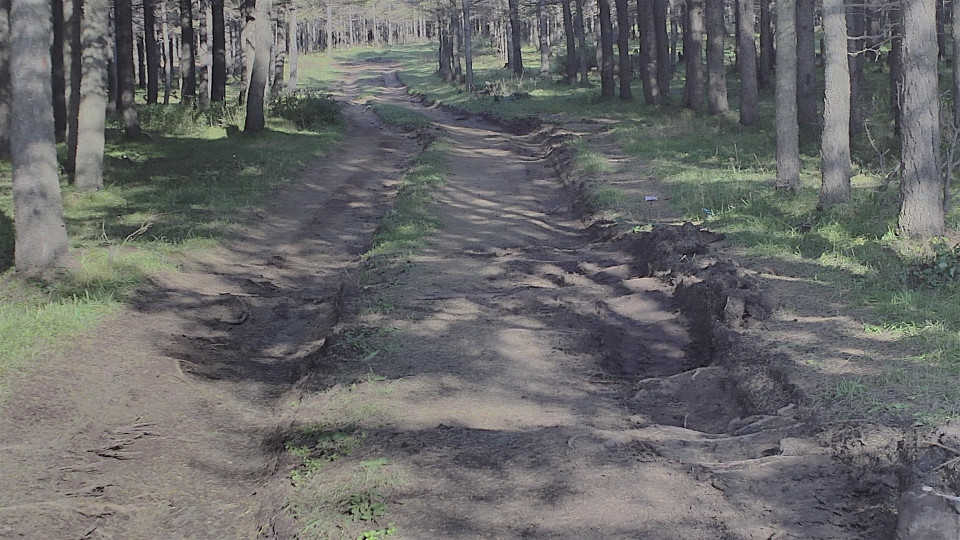}
    \end{minipage}%
    \begin{minipage}{0.25\textwidth}
        \centering
        \includegraphics[width=0.95\textwidth]{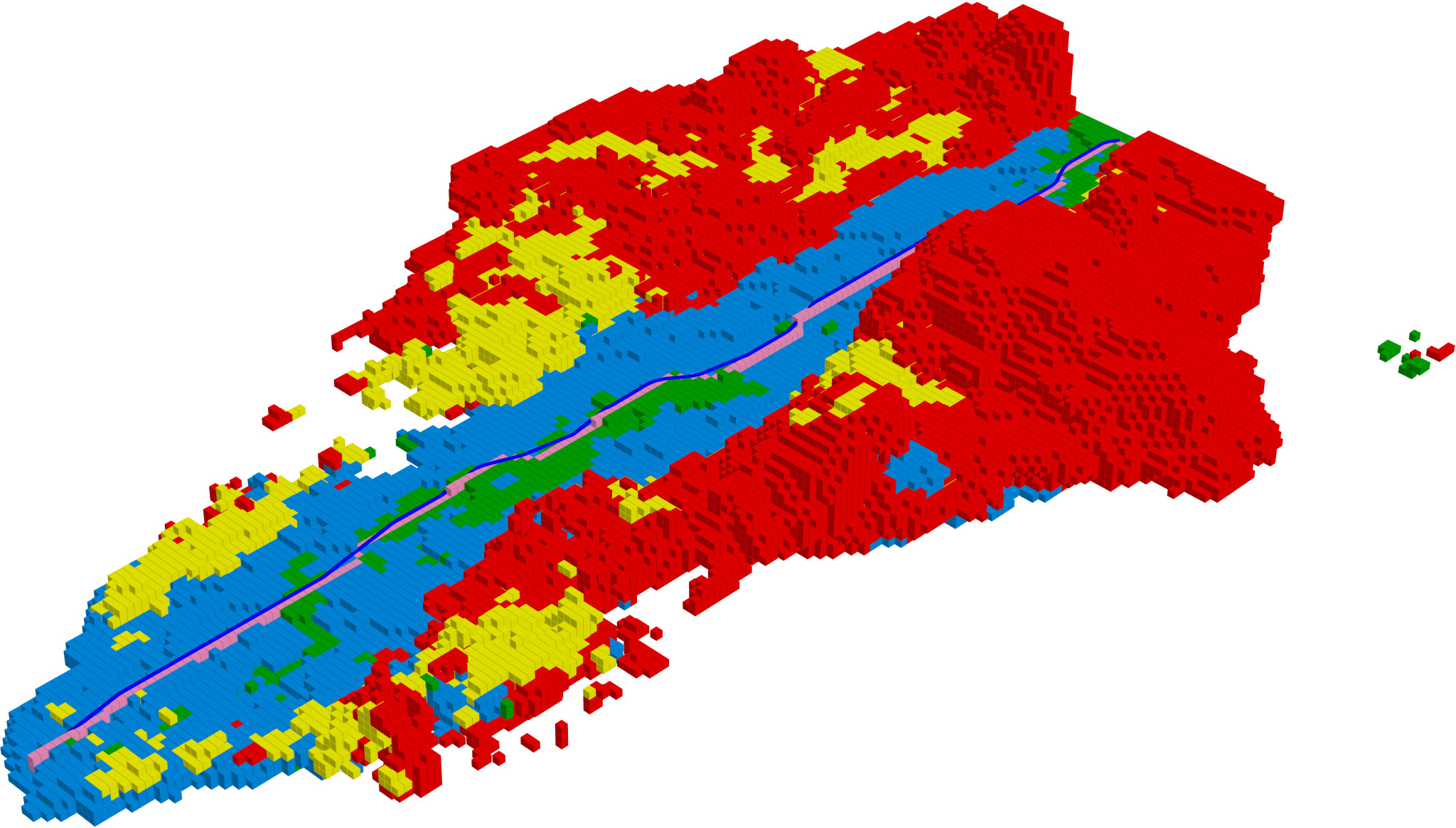}
    \end{minipage}%
    \begin{minipage}{0.25\textwidth}
        \centering
        \includegraphics[width=0.95\textwidth]{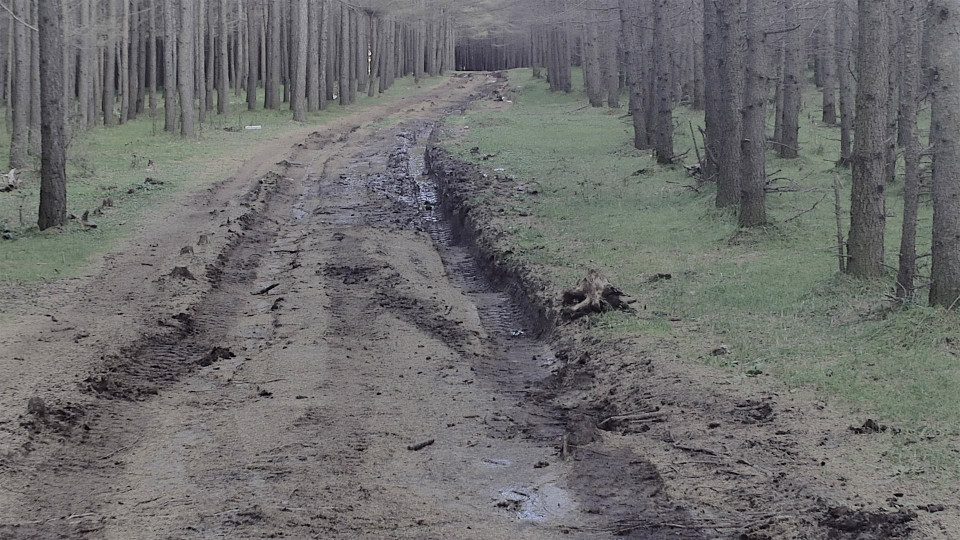}
    \end{minipage}%
    \begin{minipage}{0.25\textwidth}
        \centering
        \includegraphics[width=0.95\textwidth]{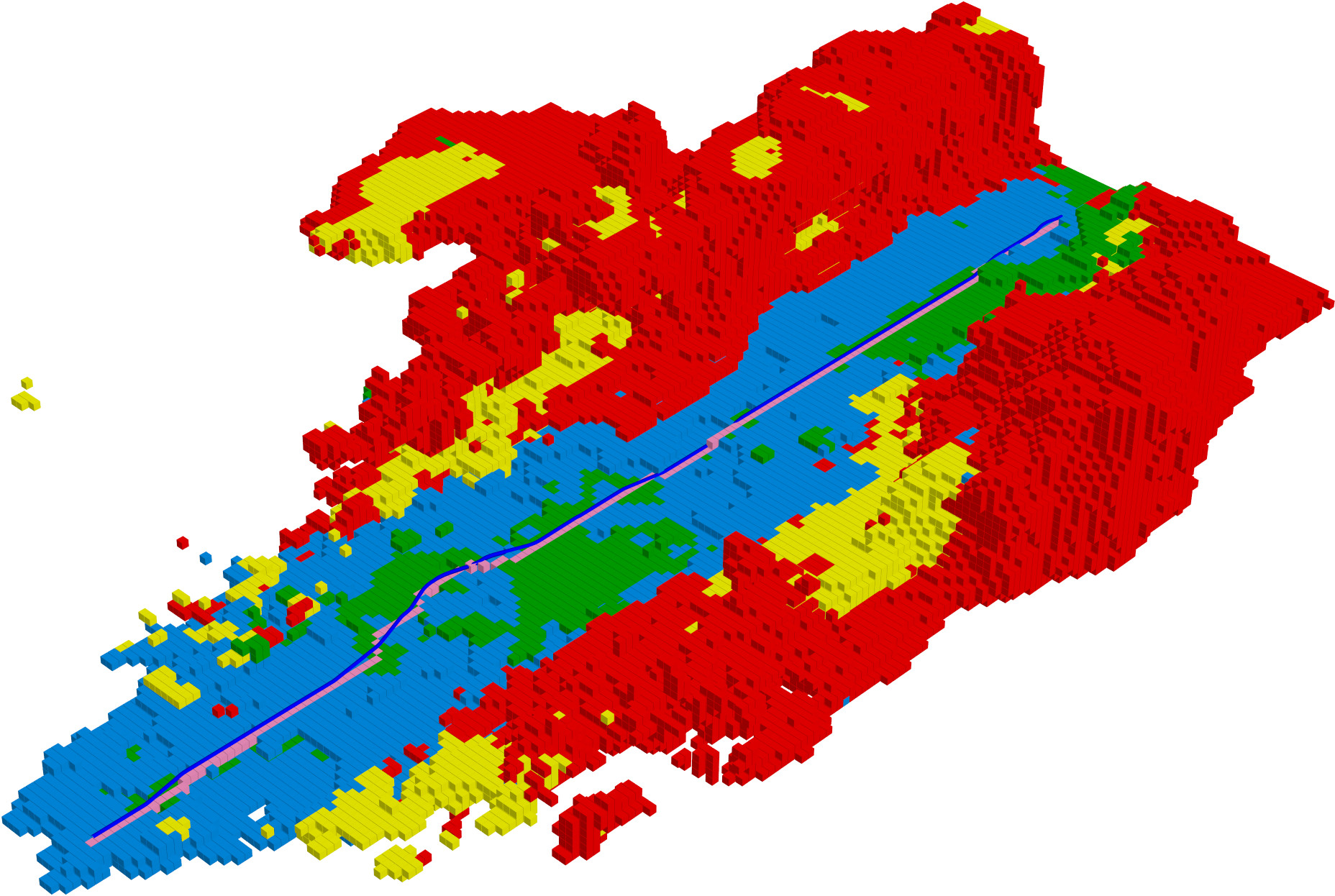}
    \end{minipage}%
    
    % 4 row of images with titles on the left
    \begin{minipage}{0.25\textwidth}
        \centering
        \includegraphics[width=0.95\textwidth]{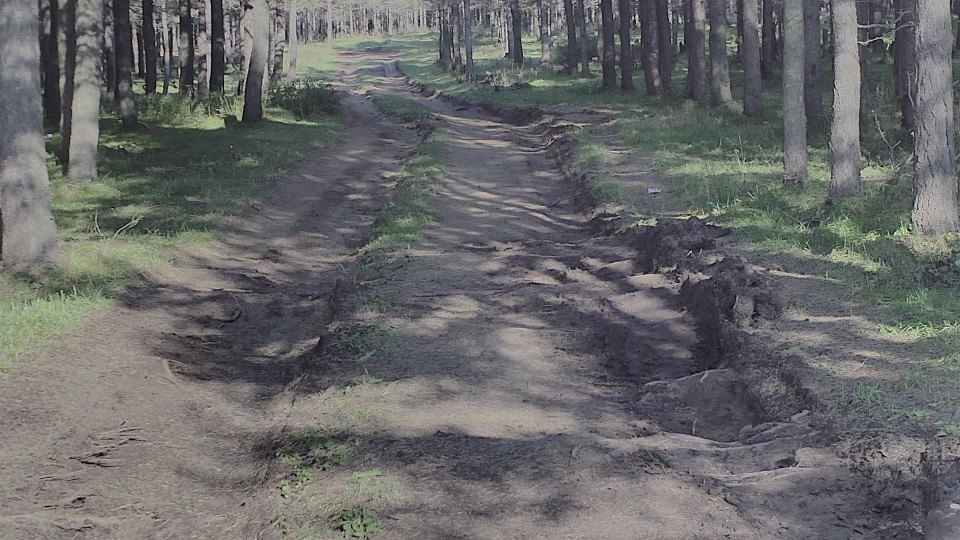}
    \end{minipage}%
    \begin{minipage}{0.25\textwidth}
        \centering
        \includegraphics[width=0.95\textwidth]{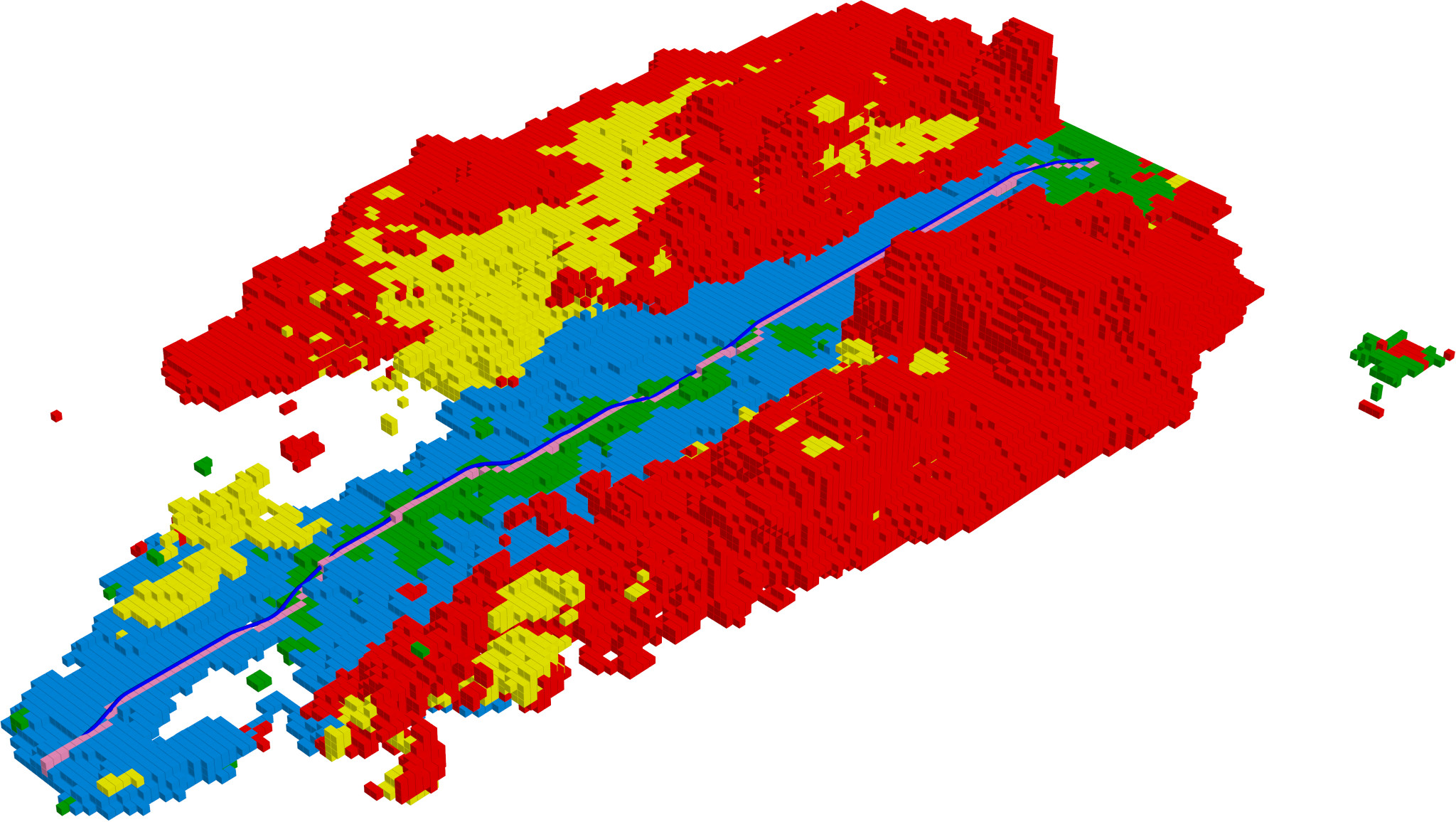}
    \end{minipage}%
    \begin{minipage}{0.25\textwidth}
        \centering
        \includegraphics[width=0.95\textwidth]{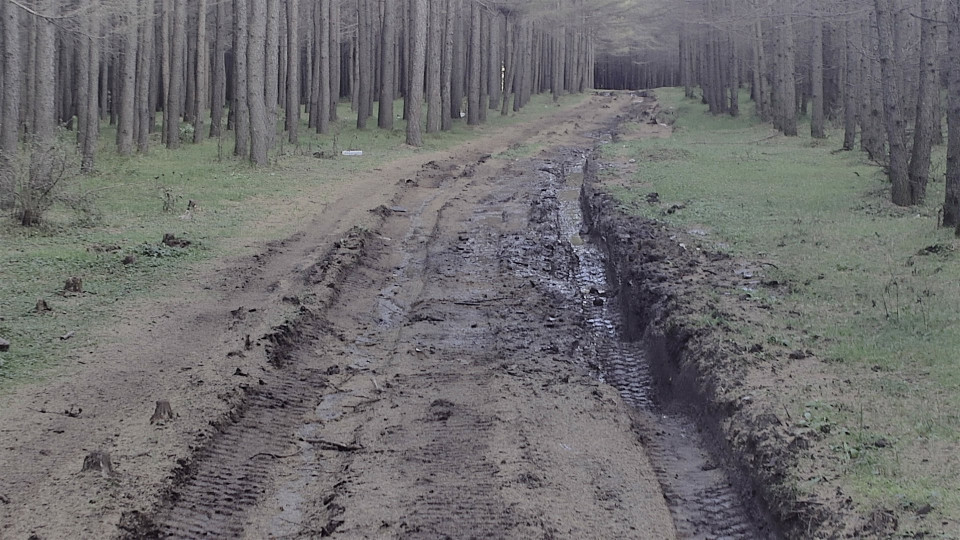}
    \end{minipage}%
    \begin{minipage}{0.25\textwidth}
        \centering
        \includegraphics[width=0.95\textwidth]{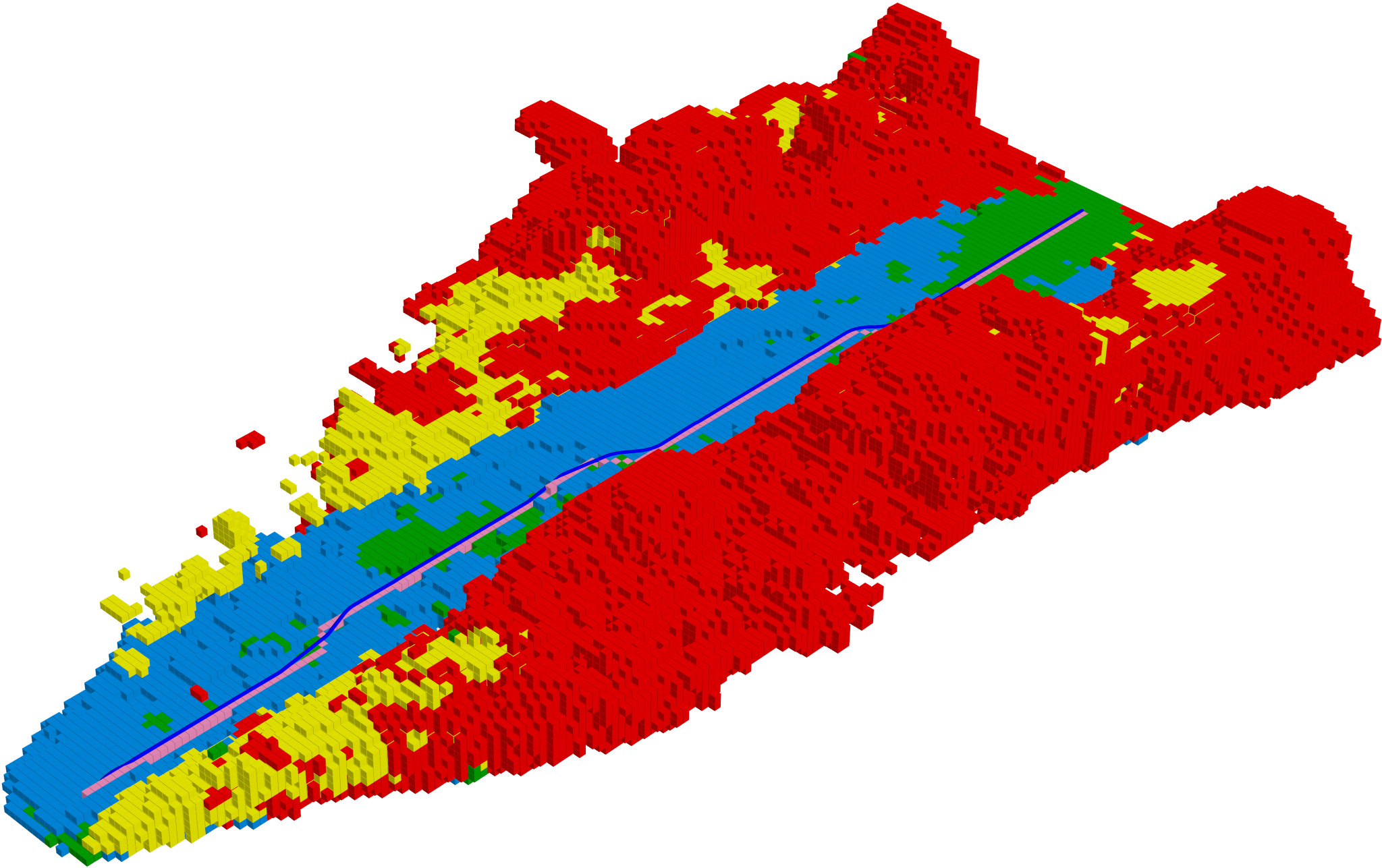}
    \end{minipage}%
    
    % 5 row of images with titles on the left
    \begin{minipage}{0.25\textwidth}
        \centering
        \includegraphics[width=0.95\textwidth]{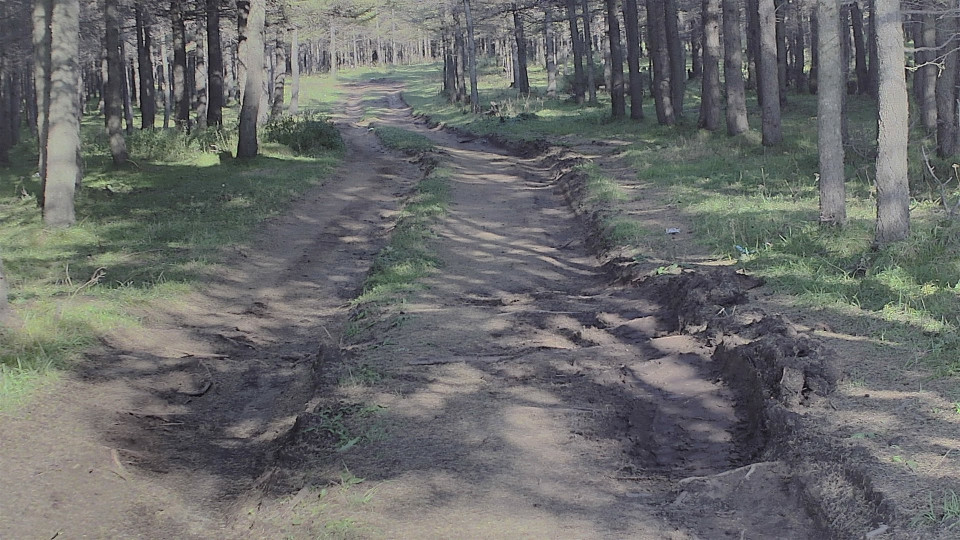}
    \end{minipage}%
    \begin{minipage}{0.25\textwidth}
        \centering
        \includegraphics[width=0.95\textwidth]{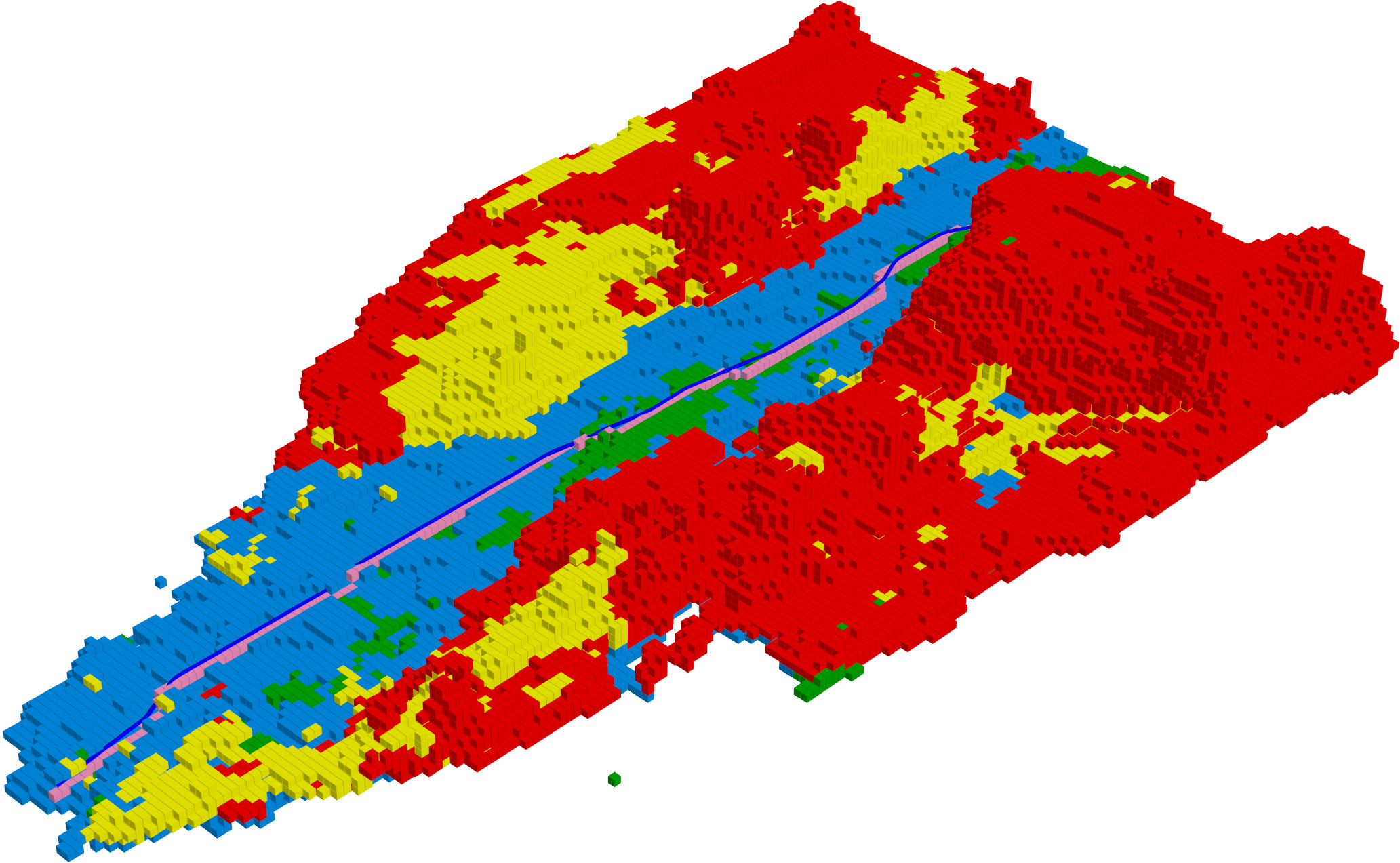}
    \end{minipage}%
    \begin{minipage}{0.25\textwidth}
        \centering
        \includegraphics[width=0.95\textwidth]{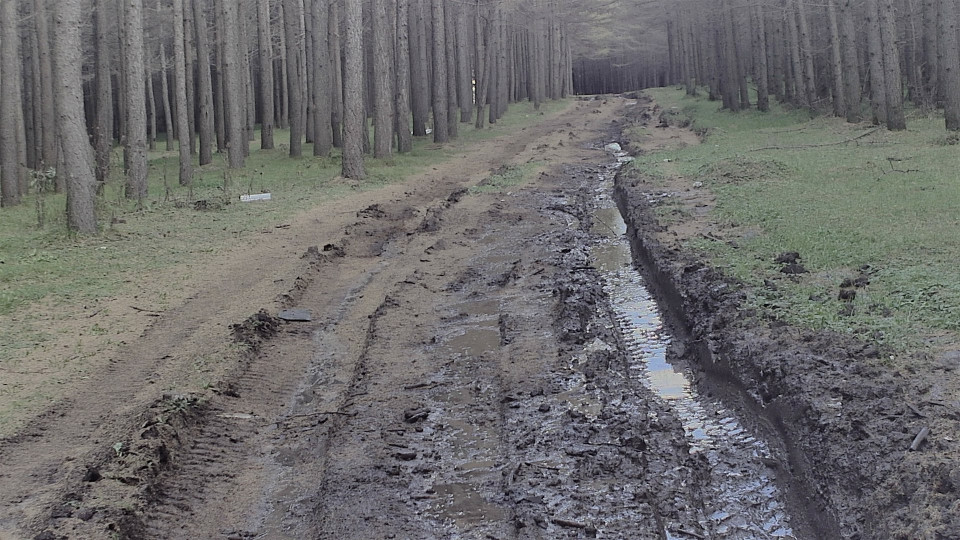}
    \end{minipage}%
    \begin{minipage}{0.25\textwidth}
        \centering
        \includegraphics[width=0.95\textwidth]{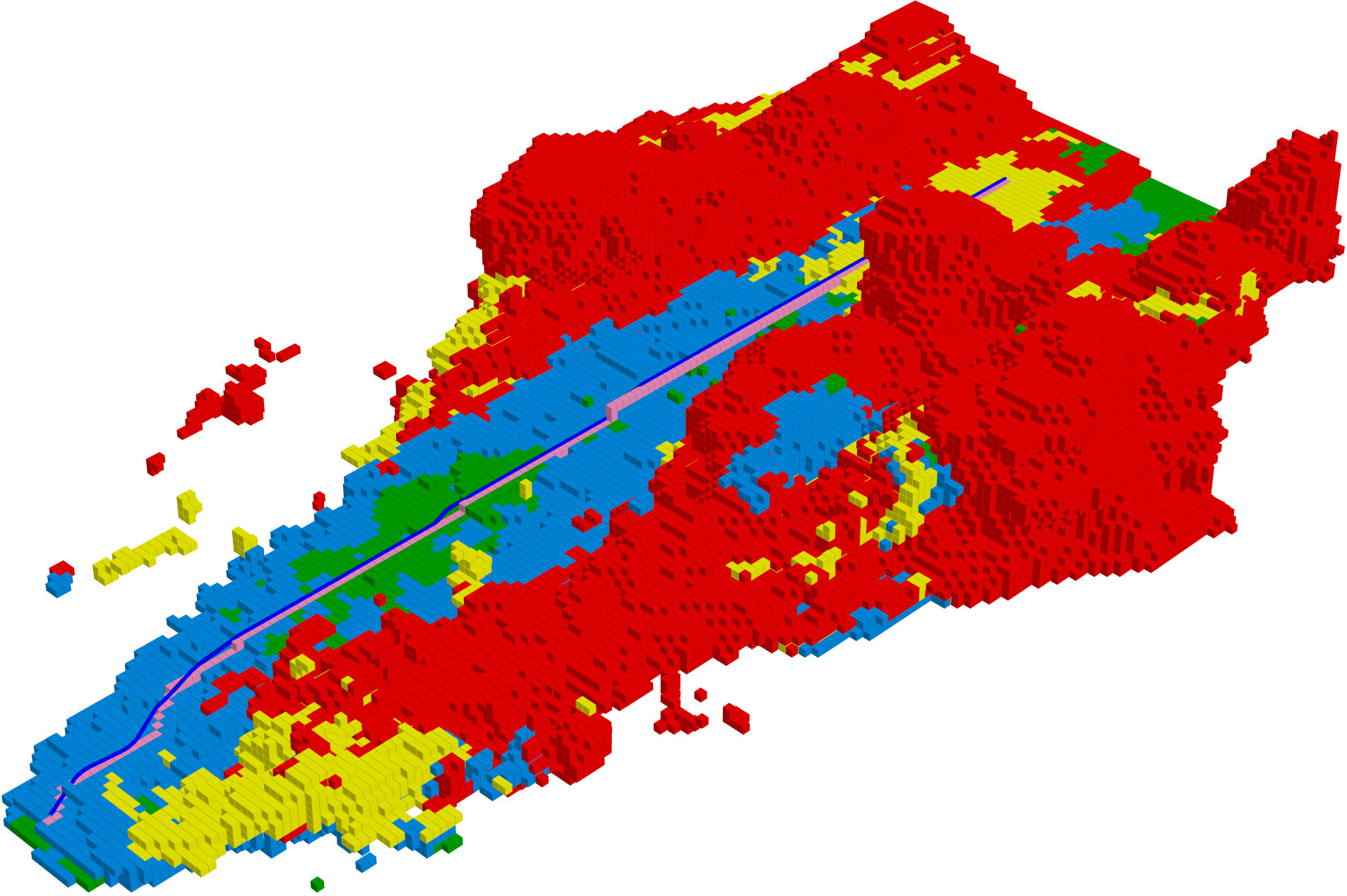}
    \end{minipage}%
    % \vskip\baselineskip
    % Additional rows can follow in the same format
    \caption{Multi-scenario trajectory planning visualization in off-road environments. Experimental results for four representative scenarios: (a) Cliff terrain, (b) Constrained forest road, (c) Continuous irregular pits, and (d) Water-filled ruts. Each case demonstrates three consecutive input frames (left panels, arranged chronologically from top to bottom) and corresponding planning outputs (right panels). The Hybrid A* algorithm generates initial trajectory nodes (light pink grids), while the Reeds-Shepp algorithm produces optimized smooth trajectories (dark blue curves).}
    \label{fig_18}
\end{figure*}

We further validated the practicality of the proposed framework through trajectory planning experiments. Leveraging the traversability cost map generated by 3DTTNet, we integrated vehicle kinematic constraints into the Hybrid A* algorithm to search for collision-free initial trajectories in complex off-road environments. To ensure dynamic feasibility, the Reeds-Shepp algorithm was subsequently applied to refine trajectory smoothness and adherence to motion constraints. The experiments encompass representative scenarios, including: (a) Cliff terrain: A highly unstructured environment with a cliff on the road’s left side, loose rock piles along the edge, and a steep slope beneath, challenging both perception and trajectory safety; (b) Constrained forest road: A narrow, winding path flanked by dense vegetation, requiring precise navigation in confined spaces with limited visibility; (c) Continuous irregular pits: A forest road with densely distributed irregular pits on the right side, severely degrading vehicle passability and demanding adaptive obstacle avoidance; (d) Water-filled ruts: Deep ruts formed by preceding vehicles, compounded by rainwater accumulation, introducing slippery surfaces and unpredictable terrain deformation.

These scenarios simulate extreme yet realistic challenges in off-road environments, including non-uniform obstacles, dynamic terrain conditions, and constrained mobility. As demonstrated in Fig.~\ref{fig_18}, our framework generates real-time trajectories that balance safety and kinematic feasibility across these scenarios. The results highlight the system’s robustness in heterogeneous environments and its potential to address critical gaps in autonomous off-road navigation, such as handling irregular geometries and partial occlusions.

\begin{figure}[t]
    \centering
    \subfloat[{\scriptsize (a)}]{%
        \includegraphics[width=0.48\columnwidth]{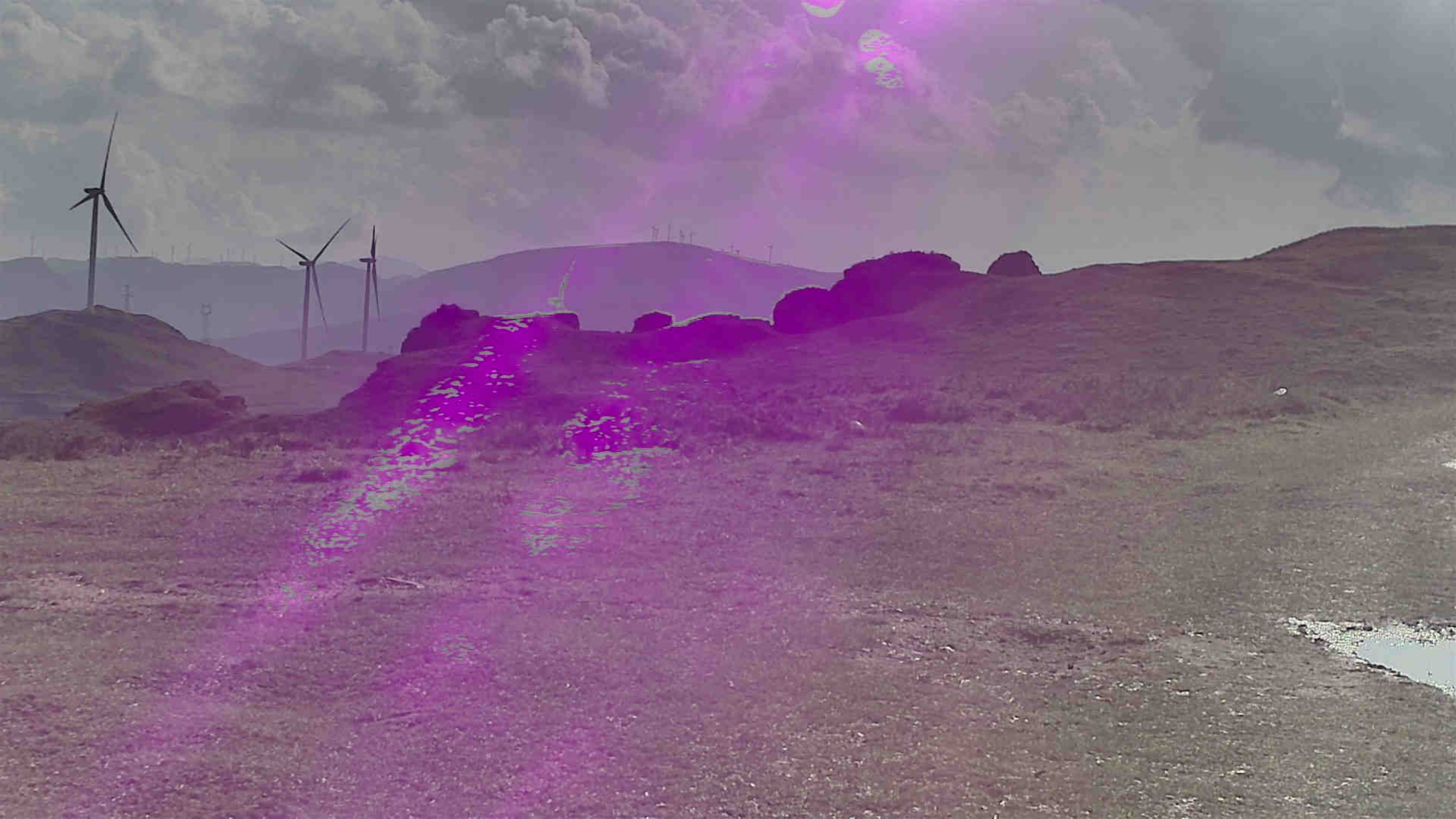}%
        \label{fig:sunlight_1}
    }
    \hfill
    \subfloat[{\scriptsize (b)}]{%
        \includegraphics[width=0.48\columnwidth]{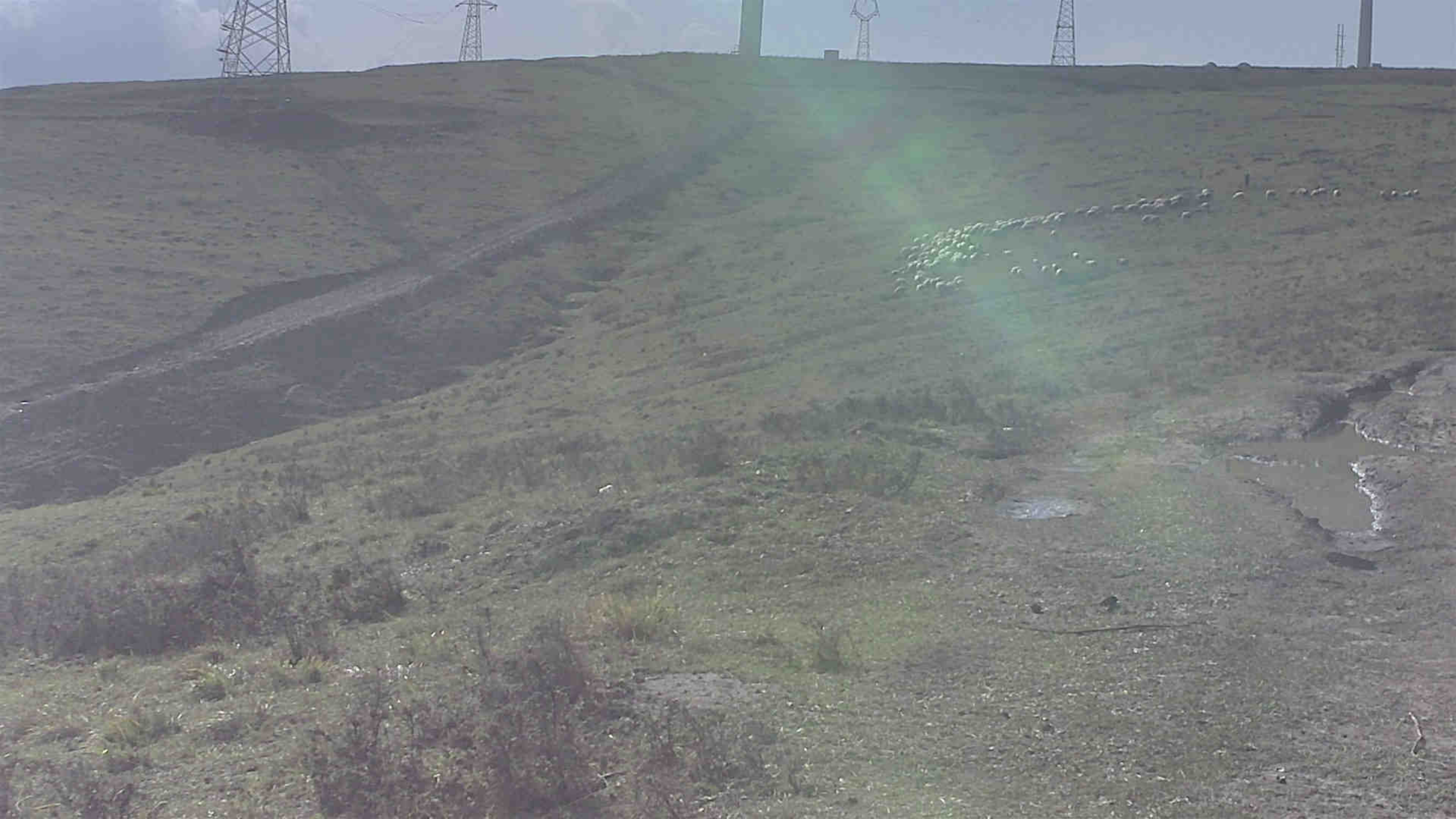}%
        \label{fig:sunlight_2}
    }
    \caption{Effects of intense sunlight on input images: (a) Purple Halos, and (b) Localized Overexposure.}
    \label{fig_17}
\end{figure}

From the experiments, it is observed that under intense sunlight conditions (see Fig.~\ref{fig_17}), the input images exhibit purple halos and localized highlight overexposure effects, leading to incorrect recognitions of the model. Accordingly, future work will involve enhancing the robustness of the model to drastic changes in environmental illumination to improve recognition accuracy under challenging lighting scenarios.

\section{Conclusion}
\label{sec:conc}
In this paper, we propose 3DTTNet, a multimodal method for off-road traversable terrain modeling that utilizes LiDAR point clouds and monocular images to generate 3D traversable predictions. On basis of combining geometric and semantic information, 3DTTNet enhances environmental feature extraction, enabling more accurate estimation of traversable regions in complex, occluded terrains. Furthermore, we introduce the RELLIS-OCC dataset with 3D traversable annotations, enabling detailed terrain assessment in off-road environments. Extensive experiments validate the superior performance of 3DTTNet compared to baseline approaches, particularly in handling obstacles with diverse geometries and partial occlusion. 3DTTNet holds significant potential for real-world deployment, as it can be easily adapted to different platforms and a range of off-road environments, providing a robust solution for autonomous navigation in challenging conditions.

For future work, the proposed traversability estimation framework will be united with 3D motion planning algorithms to enable autonomous navigation in off-road conditions. Additionally, recognizing the limitations of existing public datasets, we plan to collect an expanded off-road dataset containing complex obstacle information and evaluate the performance of 3DTTNet in detecting negative and overhanging obstacles.

\bibliographystyle{ieeetr} 
\bibliography{myref}

\begin{thebibliography}{10}

\bibitem{ref_add1}
A.~Geiger, P.~Lenz, C.~Stiller, and R.~Urtasun, ``Vision meets robotics: The kitti dataset,'' {\em The International Journal of Robotics Research}, vol.~32, no.~11, pp.~1231--1237, 2013.

\bibitem{ref_add2}
H.~Caesar, V.~Bankiti, A.~H. Lang, S.~Vora, V.~E. Liong, Q.~Xu, A.~Krishnan, Y.~Pan, G.~Baldan, and O.~Beijbom, ``nuscenes: A multimodal dataset for autonomous driving,'' in {\em Proceedings of the IEEE/CVF conference on computer vision and pattern recognition}, pp.~11621--11631, 2020.

\bibitem{ref_add3}
P.~Sun, H.~Kretzschmar, X.~Dotiwalla, A.~Chouard, V.~Patnaik, P.~Tsui, J.~Guo, Y.~Zhou, Y.~Chai, B.~Caine, {\em et~al.}, ``Scalability in perception for autonomous driving: Waymo open dataset,'' in {\em Proceedings of the IEEE/CVF conference on computer vision and pattern recognition}, pp.~2446--2454, 2020.

\bibitem{ref_add5}
C.~R. Qi, H.~Su, K.~Mo, and L.~J. Guibas, ``Pointnet: Deep learning on point sets for 3d classification and segmentation,'' in {\em Proceedings of the IEEE conference on computer vision and pattern recognition}, pp.~652--660, 2017.

\bibitem{ref_add6}
C.~R. Qi, L.~Yi, H.~Su, and L.~J. Guibas, ``Pointnet++: Deep hierarchical feature learning on point sets in a metric space,'' {\em Advances in neural information processing systems}, vol.~30, 2017.

\bibitem{ref_add8}
Y.~Zhou and O.~Tuzel, ``Voxelnet: End-to-end learning for point cloud based 3d object detection,'' in {\em Proceedings of the IEEE conference on computer vision and pattern recognition}, pp.~4490--4499, 2018.

\bibitem{ref_add9}
Y.~Huang, W.~Zheng, Y.~Zhang, J.~Zhou, and J.~Lu, ``Tri-perspective view for vision-based 3d semantic occupancy prediction,'' in {\em Proceedings of the IEEE/CVF conference on computer vision and pattern recognition}, pp.~9223--9232, 2023.

\bibitem{ref_add10}
X.~Wang, Z.~Zhu, W.~Xu, Y.~Zhang, Y.~Wei, X.~Chi, Y.~Ye, D.~Du, J.~Lu, and X.~Wang, ``Openoccupancy: A large scale benchmark for surrounding semantic occupancy perception,'' in {\em Proceedings of the IEEE/CVF International Conference on Computer Vision}, pp.~17850--17859, 2023.

\bibitem{ref_add12}
Y.~Wei, L.~Zhao, W.~Zheng, Z.~Zhu, J.~Zhou, and J.~Lu, ``Surroundocc: Multi-camera 3d occupancy prediction for autonomous driving,'' in {\em Proceedings of the IEEE/CVF International Conference on Computer Vision}, pp.~21729--21740, 2023.

\bibitem{ref1}
C.~H. Tong, D.~Gingras, K.~Larose, T.~D. Barfoot, and {\'E}.~Dupuis, ``The canadian planetary emulation terrain 3d mapping dataset,'' {\em The International Journal of Robotics Research}, vol.~32, no.~4, pp.~389--395, 2013.

\bibitem{ref2}
A.~Valada, G.~L. Oliveira, T.~Brox, and W.~Burgard, ``Deep multispectral semantic scene understanding of forested environments using multimodal fusion,'' in {\em 2016 International Symposium on Experimental Robotics}, pp.~465--477, Springer, 2017.

\bibitem{ref3}
D.~Maturana, P.-W. Chou, M.~Uenoyama, and S.~Scherer, ``Real-time semantic mapping for autonomous off-road navigation,'' in {\em Field and Service Robotics: Results of the 11th International Conference}, pp.~335--350, Springer, 2018.

\bibitem{ref4}
M.~Wigness, S.~Eum, J.~G. Rogers, D.~Han, and H.~Kwon, ``A rugd dataset for autonomous navigation and visual perception in unstructured outdoor environments,'' in {\em 2019 IEEE/RSJ International Conference on Intelligent Robots and Systems (IROS)}, pp.~5000--5007, IEEE, 2019.

\bibitem{ref5}
G.~Gresenz, J.~White, and D.~C. Schmidt, ``An off-road terrain dataset including images labeled with measures of terrain roughness,'' in {\em 2021 IEEE International Conference on Autonomous Systems (ICAS)}, pp.~1--5, IEEE, 2021.

\bibitem{ref6}
P.~Jiang, P.~Osteen, M.~Wigness, and S.~Saripalli, ``Rellis-3d dataset: Data, benchmarks and analysis,'' in {\em 2021 IEEE international conference on robotics and automation (ICRA)}, pp.~1110--1116, IEEE, 2021.

\bibitem{ref7}
G.~Chustz and S.~Saripalli, ``Rooad: Rellis off-road odometry analysis dataset,'' in {\em 2022 IEEE Intelligent Vehicles Symposium (IV)}, pp.~1504--1510, IEEE, 2022.

\bibitem{ref8}
S.~Sharma, L.~Dabbiru, T.~Hannis, G.~Mason, D.~W. Carruth, M.~Doude, C.~Goodin, C.~Hudson, S.~Ozier, J.~E. Ball, {\em et~al.}, ``Cat: Cavs traversability dataset for off-road autonomous driving,'' {\em IEEE Access}, vol.~10, pp.~24759--24768, 2022.

\bibitem{ref9}
C.~Min, W.~Jiang, D.~Zhao, J.~Xu, L.~Xiao, Y.~Nie, and B.~Dai, ``Orfd: A dataset and benchmark for off-road freespace detection,'' in {\em 2022 international conference on robotics and automation (ICRA)}, pp.~2532--2538, IEEE, 2022.

\bibitem{ref10}
T.~Yan, X.~Zheng, W.~Liu, B.~Liang, and Z.~Chen, ``The synthetic off-road trail dataset for unmanned motorcycle,'' in {\em 2022 IEEE 95th Vehicular Technology Conference:(VTC2022-Spring)}, pp.~1--7, IEEE, 2022.

\bibitem{ref11}
S.~Triest, M.~Sivaprakasam, S.~J. Wang, W.~Wang, A.~M. Johnson, and S.~Scherer, ``Tartandrive: A large-scale dataset for learning off-road dynamics models,'' in {\em 2022 International Conference on Robotics and Automation (ICRA)}, pp.~2546--2552, IEEE, 2022.

\bibitem{ref13}
S.~Yao, R.~Guan, Z.~Wu, Y.~Ni, Z.~Huang, R.~W. Liu, Y.~Yue, W.~Ding, E.~G. Lim, H.~Seo, {\em et~al.}, ``Waterscenes: A multi-task 4d radar-camera fusion dataset and benchmarks for autonomous driving on water surfaces,'' {\em IEEE Transactions on Intelligent Transportation Systems}, 2024.

\bibitem{ref14}
J.~Knights, K.~Vidanapathirana, M.~Ramezani, S.~Sridharan, C.~Fookes, and P.~Moghadam, ``Wild-places: A large-scale dataset for lidar place recognition in unstructured natural environments,'' in {\em 2023 IEEE international conference on robotics and automation (ICRA)}, pp.~11322--11328, IEEE, 2023.

\bibitem{ref15}
P.~Mortimer, R.~Hagmanns, M.~Granero, T.~Luettel, J.~Petereit, and H.-J. Wuensche, ``The goose dataset for perception in unstructured environments,'' in {\em 2024 IEEE International Conference on Robotics and Automation (ICRA)}, pp.~14838--14844, IEEE, 2024.

\bibitem{ref12}
M.~Sivaprakasam, P.~Maheshwari, M.~G. Castro, S.~Triest, M.~Nye, S.~Willits, A.~Saba, W.~Wang, and S.~Scherer, ``Tartandrive 2.0: More modalities and better infrastructure to further self-supervised learning research in off-road driving tasks,'' {\em arXiv preprint arXiv:2402.01913}, 2024.

\bibitem{ref16}
C.~S. Dima, N.~Vandapel, and M.~Hebert, ``Classifier fusion for outdoor obstacle detection,'' in {\em IEEE International Conference on Robotics and Automation, 2004. Proceedings. ICRA'04. 2004}, vol.~1, pp.~665--671, IEEE, 2004.

\bibitem{ref17}
D.~Bradley, S.~Thayer, A.~Stentz, and P.~Rander, ``Vegetation detection for mobile robot navigation,'' {\em Robotics Institute, Carnegie Mellon University, Pittsburgh, PA, Tech. Rep. CMU-RI-TR-04-12}, 2004.

\bibitem{ref21}
H.~Dahlkamp, A.~Kaehler, D.~Stavens, S.~Thrun, and G.~R. Bradski, ``Self-supervised monocular road detection in desert terrain.,'' in {\em Robotics: science and systems}, vol.~38, Philadelphia, 2006.

\bibitem{ref22}
J.~Sock, J.~Kim, J.~Min, and K.~Kwak, ``Probabilistic traversability map generation using 3d-lidar and camera,'' in {\em 2016 IEEE international conference on robotics and automation (ICRA)}, pp.~5631--5637, IEEE, 2016.

\bibitem{ref23}
J.~Larson and M.~Trivedi, ``Lidar based off-road negative obstacle detection and analysis,'' in {\em 2011 14th International IEEE Conference on Intelligent Transportation Systems (ITSC)}, pp.~192--197, IEEE, 2011.

\bibitem{ref24}
Z.~W.-J. YANG Jian-Hua and W.~Zhao-Hui, ``Cooperative multi-robot observation of multiple moving targets based on contribution model,'' {\em Pattern Recognition and Artificial Intelligence}, vol.~28, no.~04, pp.~335--343, 2015.

\bibitem{ref25}
C.~J. Holder and T.~P. Breckon, ``Learning to drive: End-to-end off-road path prediction,'' {\em IEEE Intelligent Transportation Systems Magazine}, vol.~13, no.~2, pp.~217--221, 2019.

\bibitem{ref26}
L.~Dabbiru, S.~Sharma, C.~Goodin, S.~Ozier, C.~Hudson, D.~Carruth, M.~Doude, G.~Mason, and J.~Ball, ``Traversability mapping in off-road environment using semantic segmentation,'' in {\em Autonomous Systems: Sensors, Processing, and Security for Vehicles and Infrastructure 2021}, vol.~11748, pp.~78--83, SPIE, 2021.

\bibitem{ref29}
Z.-l. DING, Y.-h. HU, J.-w. GONG, G.-m. XIONG, and C.~L{\"U}, ``Adaptive road extraction method in different scene based on deep learning,'' {\em Transactions of Beijing institute of Technology}, vol.~39, no.~11, pp.~1133--1137, 2019.

\bibitem{ref30}
L.~Dabbiru, C.~Goodin, N.~Scherrer, and D.~Carruth, ``Lidar data segmentation in off-road environment using convolutional neural networks (cnn),'' {\em SAE International Journal of Advances and Current Practices in Mobility}, vol.~2, no.~2020-01-0696, pp.~3288--3292, 2020.

\bibitem{ref31}
A.~Shaban, X.~Meng, J.~Lee, B.~Boots, and D.~Fox, ``Semantic terrain classification for off-road autonomous driving,'' in {\em Conference on Robot Learning}, pp.~619--629, PMLR, 2022.

\bibitem{ref33}
S.~Huang, G.~Xiong, B.~Zhu, J.~Gong, and H.~Chen, ``Lidar-camera fusion based high-resolution network for efficient road segmentation,'' in {\em 2020 3rd International Conference on Unmanned Systems (ICUS)}, pp.~830--835, IEEE, 2020.

\bibitem{ref35}
T.~Homberger, L.~Wellhausen, P.~Fankhauser, and M.~Hutter, ``Support surface estimation for legged robots,'' in {\em 2019 International Conference on Robotics and Automation (ICRA)}, pp.~8470--8476, IEEE, 2019.

\bibitem{ref36}
B.~Gao, A.~Xu, Y.~Pan, X.~Zhao, W.~Yao, and H.~Zhao, ``Off-road drivable area extraction using 3d lidar data,'' in {\em 2019 IEEE Intelligent Vehicles Symposium (IV)}, pp.~1505--1511, IEEE, 2019.

\bibitem{ref37}
J.~Mei, Y.~Yu, H.~Zhao, and H.~Zha, ``Scene-adaptive off-road detection using a monocular camera,'' {\em IEEE Transactions on Intelligent Transportation Systems}, vol.~19, no.~1, pp.~242--253, 2017.

\bibitem{ref38}
G.~Reina, A.~Milella, and R.~Worst, ``Lidar and stereo combination for traversability assessment of off-road robotic vehicles,'' {\em Robotica}, vol.~34, no.~12, pp.~2823--2841, 2016.

\bibitem{ref32}
D.~Maturana, P.-W. Chou, M.~Uenoyama, and S.~Scherer, ``Real-time semantic mapping for autonomous off-road navigation,'' in {\em Field and Service Robotics: Results of the 11th International Conference}, pp.~335--350, Springer, 2018.

\bibitem{ref42}
C.~H. ZHOU~Mengru, G.~H. XIONG~Guangming, and L.~Qingxiao, ``Road traversability analysis of unmanned tracked platform in off-road environment,'' {\em Acta Armamentarii}, vol.~43, no.~10, p.~2485, 2022.

\bibitem{ref55}
E.~Pairet, J.~D. Hern{\'a}ndez, M.~Carreras, Y.~Petillot, and M.~Lahijanian, ``Online mapping and motion planning under uncertainty for safe navigation in unknown environments,'' {\em IEEE Transactions on Automation Science and Engineering}, vol.~19, no.~4, pp.~3356--3378, 2022.

\bibitem{ref39}
C.~Goodin, L.~Dabbiru, C.~Hudson, G.~Mason, D.~Carruth, and M.~Doude, ``Fast terrain traversability estimation with terrestrial lidar in off-road autonomous navigation,'' in {\em Unmanned Systems Technology XXIII}, vol.~11758, pp.~189--199, SPIE, 2021.

\bibitem{ref40}
T.~Overbye and S.~Saripalli, ``G-vom: A gpu accelerated voxel off-road mapping system,'' in {\em 2022 IEEE Intelligent Vehicles Symposium (IV)}, pp.~1480--1486, IEEE, 2022.

\bibitem{ref54}
C.~Jiang, Z.~Hu, Z.~P. Mourelatos, D.~Gorsich, P.~Jayakumar, Y.~Fu, and M.~Majcher, ``R2-rrt*: Reliability-based robust mission planning of off-road autonomous ground vehicle under uncertain terrain environment,'' {\em IEEE Transactions on Automation Science and Engineering}, vol.~19, no.~2, pp.~1030--1046, 2022.

\bibitem{ref41}
Y.~Zhao, P.~Liu, W.~Xue, R.~Miao, Z.~Gong, and R.~Ying, ``Semantic probabilistic traversable map generation for robot path planning,'' in {\em 2019 IEEE international conference on robotics and biomimetics (ROBIO)}, pp.~2576--2582, IEEE, 2019.

\bibitem{ref56}
Y.~Gao, J.~Wu, X.~Yang, and Z.~Ji, ``Efficient hierarchical reinforcement learning for mapless navigation with predictive neighbouring space scoring,'' {\em IEEE Transactions on Automation Science and Engineering}, vol.~21, no.~4, pp.~5457--5472, 2024.

\bibitem{ref43}
T.~H.~Y. Leung, D.~Ignatyev, and A.~Zolotas, ``Hybrid terrain traversability analysis in off-road environments,'' in {\em 2022 8th International Conference on Automation, Robotics and Applications (ICARA)}, pp.~50--56, IEEE, 2022.

\bibitem{ref45}
Y.~Li, Z.~Yu, C.~Choy, C.~Xiao, J.~M. Alvarez, S.~Fidler, C.~Feng, and A.~Anandkumar, ``Voxformer: Sparse voxel transformer for camera-based 3d semantic scene completion,'' in {\em Proceedings of the IEEE/CVF conference on computer vision and pattern recognition}, pp.~9087--9098, 2023.

\bibitem{ref50}
X.~Zhu, W.~Su, L.~Lu, B.~Li, X.~Wang, and J.~Dai, ``Deformable detr: Deformable transformers for end-to-end object detection,'' in {\em International Conference on Learning Representations}, 2021.

\bibitem{ref44}
J.~Y. Wong, {\em Theory of Ground Vehicles}.
\newblock Wiley, 4th~ed., 2008.

\bibitem{ref57}
N.~Ganganath, C.-T. Cheng, and K.~T. Chi, ``A constraint-aware heuristic path planner for finding energy-efficient paths on uneven terrains,'' {\em IEEE transactions on industrial informatics}, vol.~11, no.~3, pp.~601--611, 2015.

\bibitem{ref48}
L.~Gan, R.~Zhang, J.~W. Grizzle, R.~M. Eustice, and M.~Ghaffari, ``Bayesian spatial kernel smoothing for scalable dense semantic mapping,'' {\em IEEE Robotics and Automation Letters}, vol.~5, no.~2, pp.~790--797, 2020.

\bibitem{ref46}
A.~Kendall, V.~Badrinarayanan, and R.~Cipolla, ``Bayesian segnet: Model uncertainty in deep convolutional encoder-decoder architectures for scene understanding,'' {\em arXiv preprint arXiv:1511.02680}, 2015.

\bibitem{ref47}
W.~R. Vega-Brown, M.~Doniec, and N.~G. Roy, ``Nonparametric bayesian inference on multivariate exponential families,'' {\em Advances in Neural Information Processing Systems}, vol.~27, 2014.

\bibitem{ref51}
A.-Q. Cao and R.~De~Charette, ``Monoscene: Monocular 3d semantic scene completion,'' in {\em Proceedings of the IEEE/CVF Conference on Computer Vision and Pattern Recognition}, pp.~3991--4001, 2022.

\bibitem{ref52}
S.~Song, F.~Yu, A.~Zeng, A.~X. Chang, M.~Savva, and T.~Funkhouser, ``Semantic scene completion from a single depth image,'' in {\em Proceedings of the IEEE conference on computer vision and pattern recognition}, pp.~1746--1754, 2017.

\bibitem{ref53}
L.~Roldao, R.~de~Charette, and A.~Verroust-Blondet, ``Lmscnet: Lightweight multiscale 3d semantic completion,'' in {\em 2020 International Conference on 3D Vision (3DV)}, pp.~111--119, IEEE, 2020.

\end{thebibliography}

\begin{IEEEbiography}[{\includegraphics[width=1in,height=1.25in,clip,keepaspectratio]{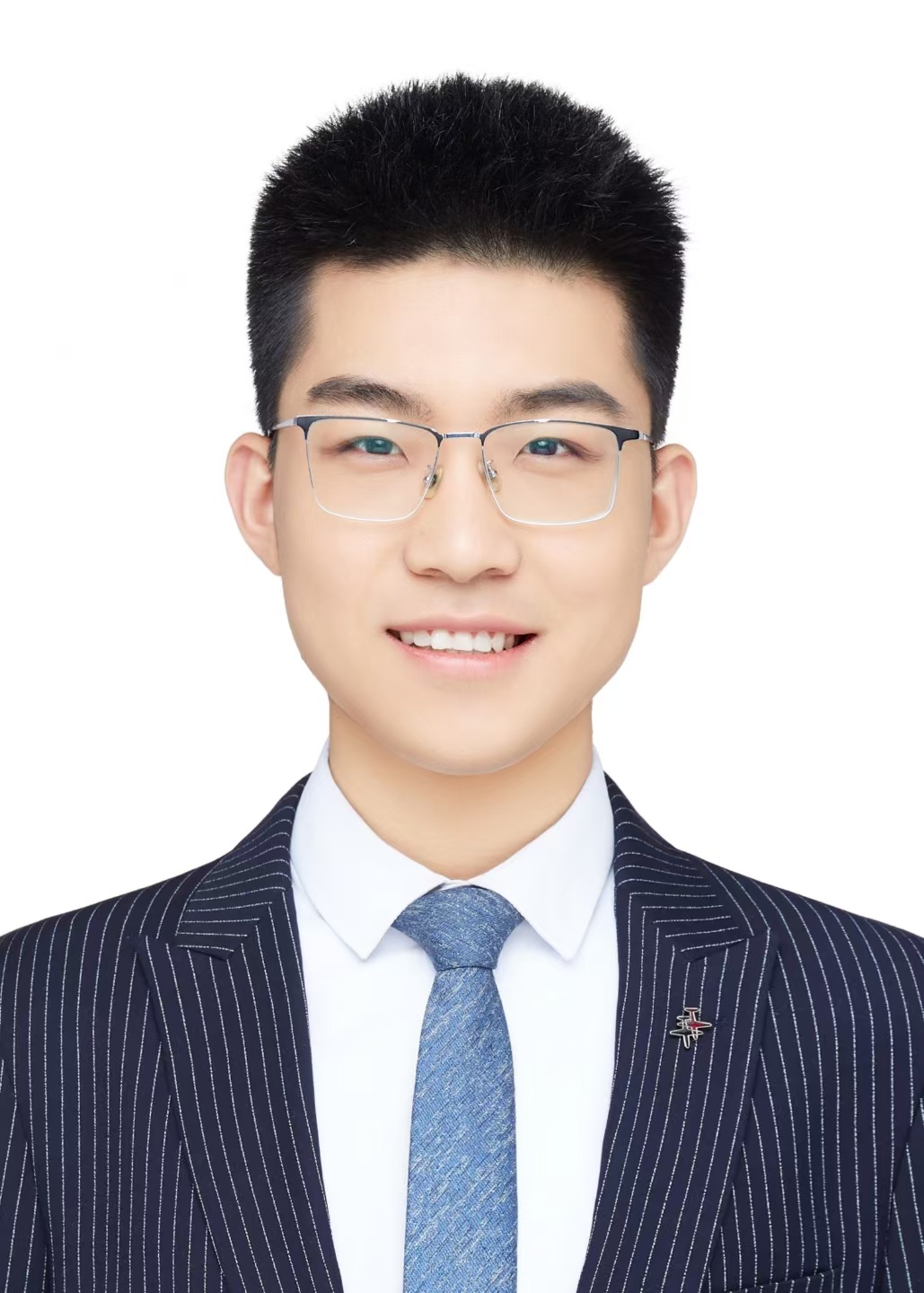}}]{Zitong Chen}
     received the B.S. degree in vehicle engineering from Chang'an University in 2022. He is currently pursuing his Master degree in mechanical engineering at the National Engineering Research Center for Electric Vehicles, Beijing Institute of Technology. His research interests focus on developing embodied agents for unmanned ground vehicles operating in off-road conditions or unstructured environments.
\end{IEEEbiography}

\vspace{-22pt}
\begin{IEEEbiography}[{\includegraphics[width=1in,height=1.25in,clip,keepaspectratio]{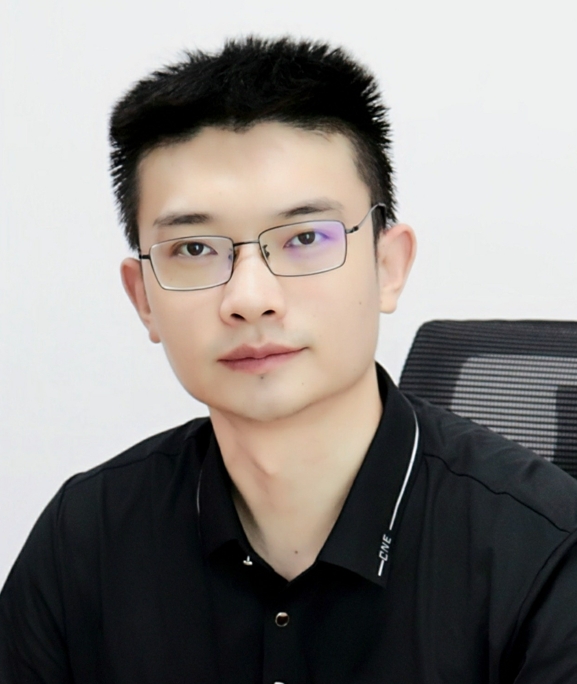}}]{Chao Sun}
    received the B.S. and Ph.D. degree in Mechanical Engineering from Beihang University and Beijing Institute of Technology in 2010 and 2016, respectively. He was a postdoctoral researcher at the Energy, Controls, and Applications Lab in University of California, Berkeley, CA, USA. Currently, he is an Associate Professor at Beijing Institute of Technology, studying on automated and connected vehicles and hybrid electric vehicles.
\end{IEEEbiography}

\vspace{-22pt}
\begin{IEEEbiography}[{\includegraphics[width=1in,height=1.25in,clip,keepaspectratio]{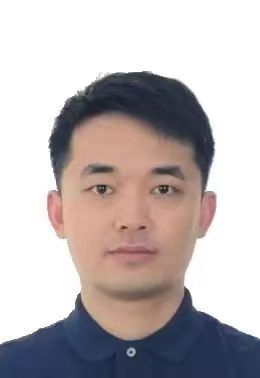}}]{Shida Nie}
    received the Ph.D. degrees in vehicle engineering from the Jilin University, Changchun, China. He is currently an associate research fellow of vehicle engineering with the School of Mechanical Engineering, Beijing Institute of Technology. His research interests include autonomous driving, vehicle dynamics control, and vehicle simulation technology. 
\end{IEEEbiography}

\vspace{-22pt}
\begin{IEEEbiography}[{\includegraphics[width=1in,height=1.25in,clip,keepaspectratio]{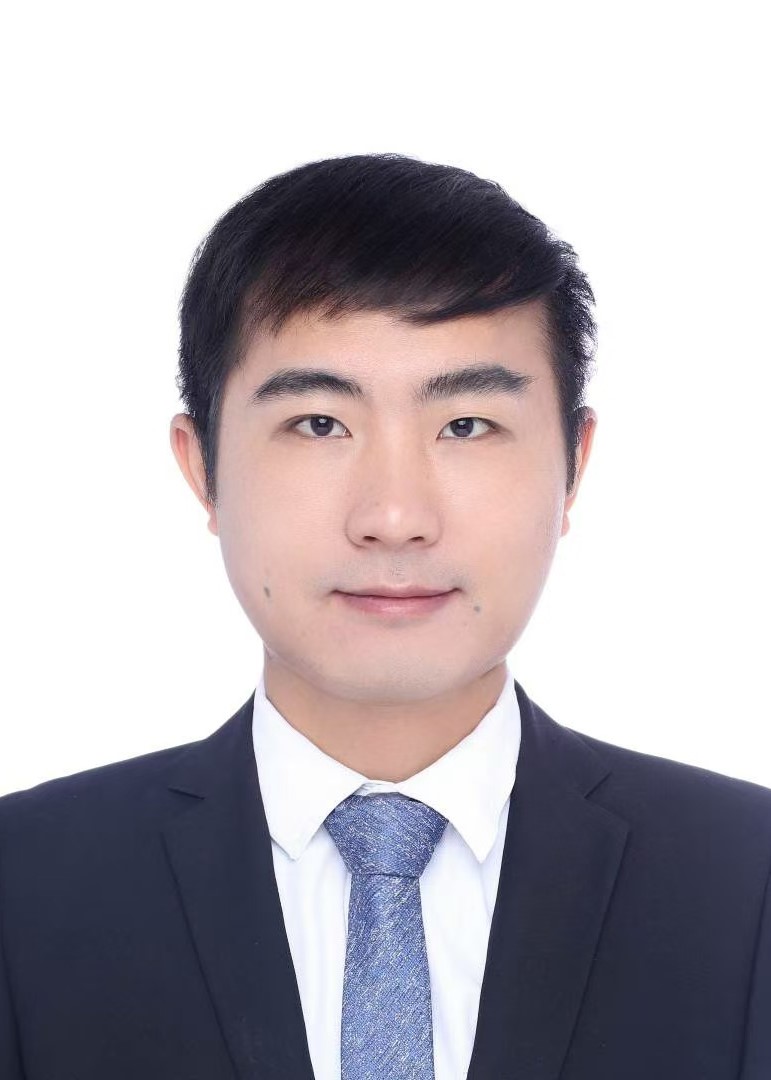}}]{Chen Min}
     received the Ph.D. degree in school of computer science from Peking University, Beijing, China, in 2024. He is currently an associate research fellow with the Research Center for Intelligent Computing Systems, Institute of Computing Technology, Chinese Academy of Sciences, Beijing, China. His research interest is in 3D computer vision, with a particular interest in 3D perception for autonomous driving and embodied intelligence.
\end{IEEEbiography}

\vspace{-22pt}
\begin{IEEEbiography}[{\includegraphics[width=1in,height=1.25in,clip,keepaspectratio]{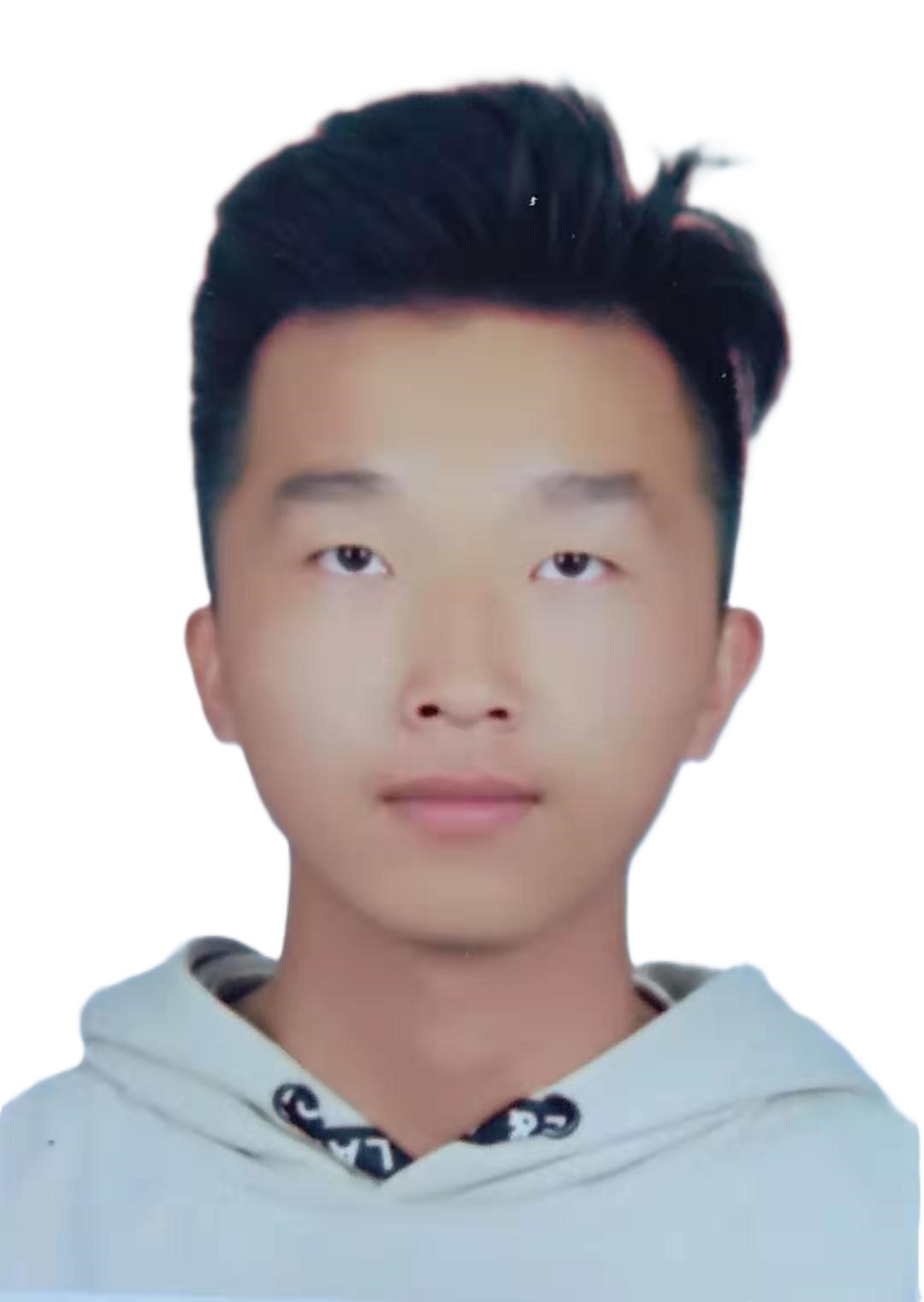}}]{Changjiu Ning}
    received the B.S. degree in vehicle engineering from Beijing Institute of Technology, Beijing, China, in 2021. He is currently pursuing the Ph.D. degree in mechanical engineering at the National Engineering Research Center for Electric Vehicles, Beijing Institute of Technology. His research interests include the planning and control of autonomous vehicles, the application of manifolds in unmanned vehicles, and deep learning.
\end{IEEEbiography}

\vspace{-22pt}
\begin{IEEEbiography}
[{\includegraphics[width=1in,height=1.25in,clip,keepaspectratio]{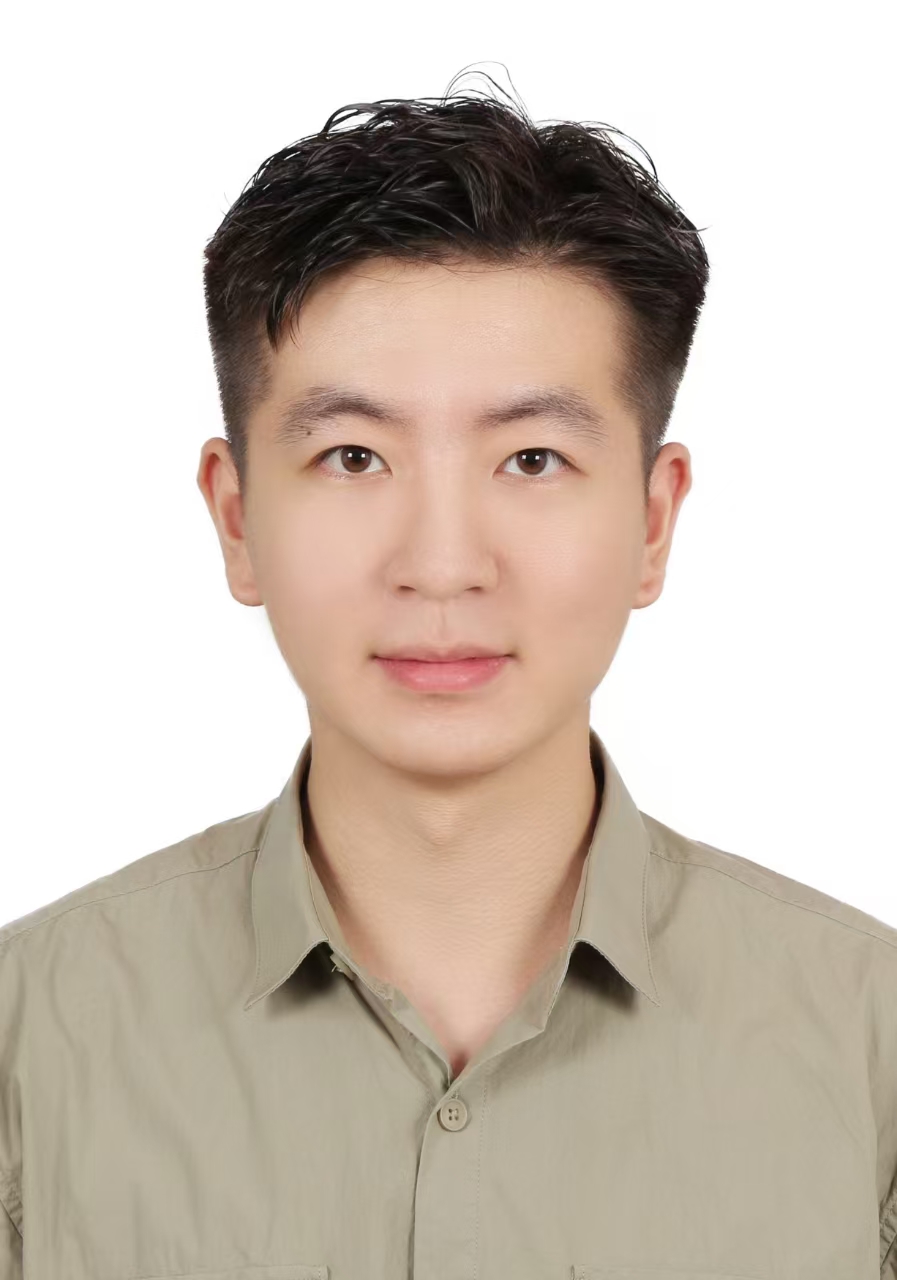}}]{Haoyu Li}
    received the B.S. degree in vehicle engineering from Hefei University of Technology, China, in 2020. He is currently pursuing the Ph.D. degree at the National Engineering Research Center for Electric Vehicles, Beijing Institute of Technology, China. His research interests include computer vision, deep learning, vehicle-infrastructure cooperative systems and their application in autonomous and electric vehicles.
\end{IEEEbiography}

\vspace{-22pt}
\begin{IEEEbiography}
[{\includegraphics[width=1in,height=1.25in,clip,keepaspectratio]{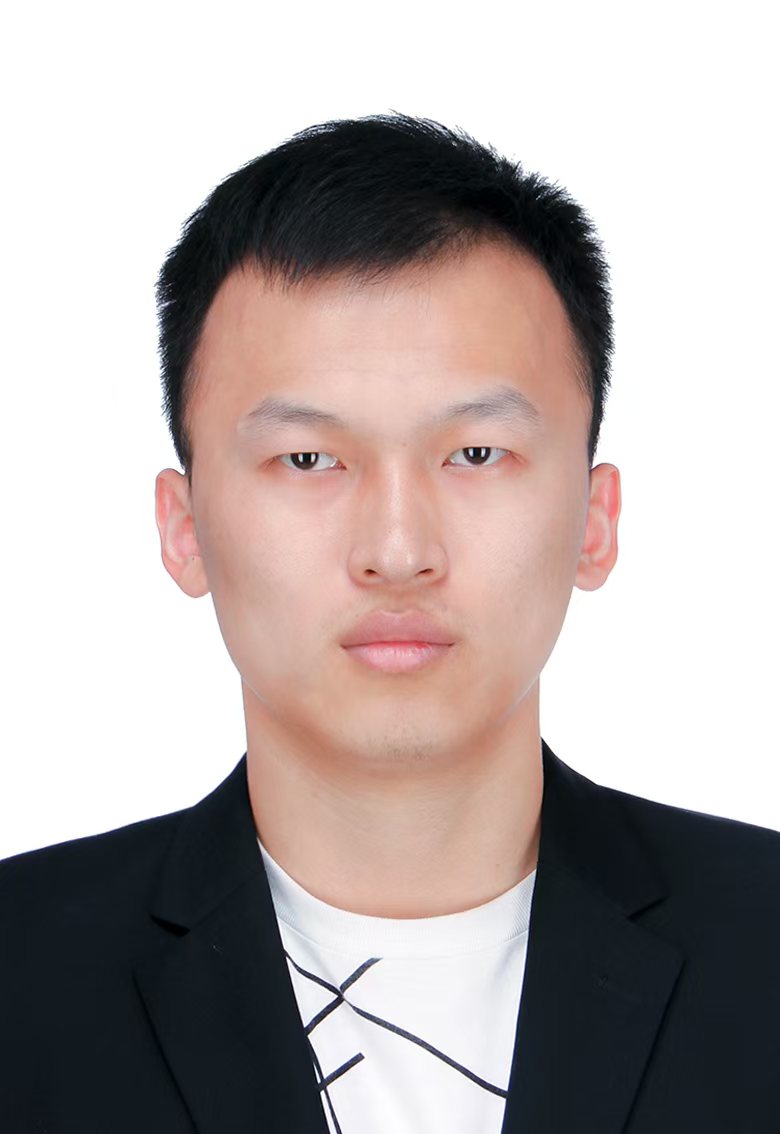}}]{Bo Wang}
    received the B.S. degree in vehicular technology from Tsinghua University, Beijing, China, in 2020. He is currently pursuing the Ph.D. degree in mechanical engineering at the National Engineering Research Center for Electric Vehicles, Beijing Institute of Technology. His research interests include pedestrian trajectory prediction and motion planning for automated and connected vehicles.
\end{IEEEbiography}

\vfill
\end{document}